\documentclass[acmtog, nonacm]{acmart}
\usepackage{booktabs} % For formal tables

\usepackage[ruled]{algorithm2e} % For algorithms

\SetAlFnt{\small}
\SetAlCapFnt{\small}
\SetAlCapNameFnt{\small}
\SetAlCapHSkip{0pt}

\usepackage{subcaption}
\usepackage[T1]{fontenc}
\usepackage{cleveref}
\usepackage{array} 
\usepackage{float} 
\usepackage{stfloats}
\usepackage{tcolorbox}
\usepackage{xcolor}
\usepackage{caption}
\usepackage{tikz}
\usepackage{placeins}
\usepackage{xspace}
\usepackage{xr}
\usepackage[normalem]{ulem}

\SetAlFnt{\small}
\SetAlCapFnt{\small}
\SetAlCapNameFnt{\small}
\SetAlCapHSkip{0pt}
\definecolor{darkred}{rgb}{0.55, 0.0, 0.0}

\newcommand{\method}{MaterialPicker\xspace}

\begin{document}
\setlength{\fboxsep}{1pt}
\title{MaterialPicker: Multi-Modal DiT-Based Material Generation}
\author{Xiaohe Ma}
\affiliation{
  \institution{State Key Lab of CAD\&CG, Zhejiang University}
  \city{Hangzhou}
  % \postcode{310058}
  \country{China}
}
\email{xiaohema1998@gmail.com}
\author{Valentin Deschaintre}
\affiliation{
  \institution{Adobe Research}
  \city{London}
  % \postcode{310058}
  \country{United Kingdom}
}
\email{deschain@adobe.com}
\author{Milo\v{s} Ha\v{s}an}
\affiliation{
  \institution{Adobe Research}
  \city{San Jose}
  % \postcode{95112}
  \country{USA}
}
\email{mihasan@adobe.com}
\author{Fujun Luan}
\affiliation{
  \institution{Adobe Research}
  \city{San Jose}
  % \postcode{310058}
  \country{USA}
}
\email{fluan@adobe.com}
\author{Kun Zhou}
\affiliation{
  \institution{State Key Lab of CAD\&CG, Zhejiang University and ZJU-FaceUnity Joint Lab of Intelligent Graphics}
  \city{Hangzhou}
  % \postcode{310058}
  \country{China}
}
\email{kunzhou@acm.org}
\author{Hongzhi Wu}
\affiliation{
  \institution{State Key Lab of CAD\&CG, Zhejiang University}
  \city{Hangzhou}
  % \postcode{310058}
  \country{China}
}
\email{hwu@acm.org}
\author{Yiwei Hu}
\affiliation{
  \institution{Adobe Research}
  \city{San Jose}
  % \postcode{310058}
  \country{USA}
}
\email{yiwhu@adobe.com}
\thanks{*Corresponding authors: Xiaohe Ma (\url{xiaohema1998@gmail.com}).}

\begin{abstract}
High-quality material generation is key for virtual environment authoring and inverse rendering. We propose \method, a multi-modal material generator leveraging a Diffusion Transformer (DiT) architecture, improving and simplifying the creation of high-quality materials from text prompts and/or photographs. Our method can generate a material based on an image crop of a material sample, even if the captured surface is distorted, viewed at an angle or partially occluded, as is often the case in photographs of natural scenes. We further allow the user to specify a text prompt to provide additional guidance for the generation. We finetune a pre-trained DiT-based video generator into a material generator, where each material map is treated as a frame in a video sequence. We evaluate our approach both quantitatively and qualitatively and show that it enables more diverse material generation and better distortion correction than previous work.
\end{abstract}


%     <concept>
%         <concept_id>10010147.10010178.10010224.10010240.10010243</concept_id>
%         <concept_desc>Computing methodologies~Appearance and texture representations</concept_desc>
%         <concept_significance>500</concept_significance>
%         </concept>
%   </ccs2012>

% \ccsdesc[500]{Computing methodologies~Appearance and texture representations}


\begin{teaserfigure}
    \centering
    \begin{minipage}{\linewidth}
        \includegraphics[width=0.995\linewidth]{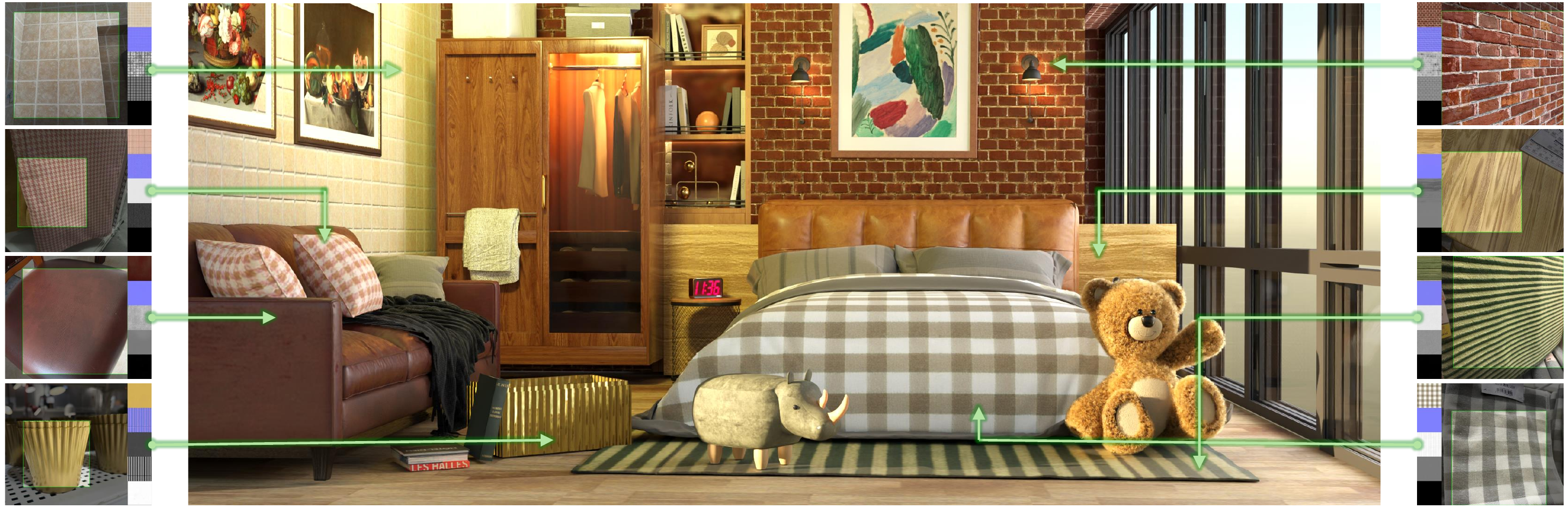}
    \end{minipage}
    \caption{We introduce \method, a DiT-based model that generates high-quality materials, conditioned on image crops and/or text prompts. Our model accurately captures textures and material properties even from photographs of distorted or partially obscured surfaces. We demonstrate \method by extracting material properties (albedo, normal, roughness, height and metallicity, shown in a column next to the input crops) from smartphone-captured photos, then applying these materials (as indicated by green arrows) in a 3D scene for photo-realistic rendering results.}
  \label{fig:teaser}
\end{teaserfigure}

\maketitle
   
\section{Introduction}
\label{sec:intro}
High-quality materials are a core requirement for photorealistic image synthesis. We present a multi-modal material generator, conditioned on a text prompt and/or an image. The image can be a photograph containing a material sample captured at any angle, potentially distorted or partially occluded. Our model lets users ``pick'' materials from any photograph just by outlining a rough crop square around the material sample of interest. 

Traditional material acquisition often requires tens or hundreds of photo samples under known light conditions and camera poses. Even with recent advances in material acquisition allowing single or few image(s) capture~\cite{Deschaintre2018, deschaintre2019flexible, Martin2022, Shi2020, Guo2021, zhou2022look, vecchio2024controlmat}, restrictions on the capture conditions are imposed. These methods typically require a camera flash as the only light source, and/or a fronto-parallel view of a flat sample. Even methods designed for capture using non-planar photographs \cite{lopes2024material} cannot handle significant texture distortion in the input photographs. Many recent material generation methods are trained from scratch on synthetic materials, limiting the generation diversity due to limited datasets \cite{vecchio2024controlmat, SubstanceAsset}, as compared to general-purpose text-to-image diffusion models \cite{nichol2021glide, ramesh2022hierarchical, rombach2022high}.

We propose to tackle these challenges with two new ideas. First, we create a dataset which contains 800K crops of synthetic scene renderings, textured with randomly assigned materials, with each crop paired with its ground truth material.  Using this data, we train our model for the ``material picking" task. We additionally use a text-to-material dataset~\cite{vecchio2024controlmat} containing 800K pairs of text descriptions and associated ground truth material maps, encouraging material generation diversity and resulting in a multi-modal generator that can accept images, text or both.

Second, we re-purpose a text-to-video generation model to generate material maps instead. We use a Diffusion-Transformer (DiT) based architecture, which has been shown to be effective for high-quality video synthesis \cite{videoworldsimulators2024}. However, our target domain is materials, which we represent as a set of 2D maps (albedo, normal, roughness, height, metallicity). To adapt our base DiT model, trained on videos, to materials, we finetune it by considering each material map as a ``frame'' in a video sequence. This approach preserves the strong prior information in the video model, improving our method's generalization and robustness.

We evaluate our model on both real and synthetic input images and compare it against the state-of-the-art methods for texture rectification \cite{hao2023diffusion}, material picking \cite{lopes2024material}, material acquisition \cite{vecchio2024controlmat} and text-to-material generation \cite{vecchio2024controlmat,vecchio2024matfuse}. We show that our approach generates materials that follow the input text prompt and/or match the appearance of the material sample in the input image, while correcting its distortions. 
In summary, we make the following contributions:
\begin{itemize}
    \item We propose a material generation model which uses text and/or image prompts as inputs, while being robust to distortion, occlusion and perspective in the input image. 
    \item We design a large-scale dataset of crops of material samples paired with the corresponding ground truth material maps, enabling our model to handle a range of viewing angles and distortions.
    \item We adapt a Diffusion Transformer text-to-video model for material generation by treating material maps as video frames, preserving the original prior knowledge embedded in the model to generate diverse materials.
\end{itemize}
\section{Related Work}
\label{sec:related_work}
\subsection{Material Acquisition and Generation}
While material acquisition has been a long standing challenge \cite{Guarnera16}, lightweight material acquisition and generation have seen significant progress using machine learning. Various methods were proposed to infer PBR material maps from only a single or few photographs \cite{Deschaintre2018, deschaintre2019flexible, DDB20, Shi2020, Guo2021, zhou2022look}. However, these methods rely on specific capture condition using a flash light co-located with the camera location. \citet{Martin2022} propose to use natural illumination but doesn't support direct metallic and roughness map estimations. Further, these methods rely on the camera being fronto-parallel or very close to it. This kind of photographs require specific captures, making the use of in the wild photos for material creation challenging.

As an alternative to create materials, generative model for materials were proposed. GAN-based approaches \cite{Guo2020, zhou2022tilegen} show that unconditional generation of materials is possible and can be used for material acquisition via optimization of their noise and latent spaces. Recent progress in generative model, and more specifically diffusion models \cite{rombach2022high}, enabled more stable, diffusion based, material generators \cite{vecchio2024matfuse, xue2024reflectancefusion}. Such diffusion models can also be used to support material acquisition tasks \cite{vecchio2024controlmat}, for example when paired with ControlNet \cite{zhang2023adding}. All these diffusion-based approaches either attempt to train the model from scratch, using solely synthetic material data \cite{vecchio2024matfuse, vecchio2024controlmat} or significantly alter the architecture of the original text-to-image model \cite{xue2024reflectancefusion}, preventing the use of the pre-existing priors in large scale image generation models \cite{rombach2022high}, limiting their generalization and diversity. Further, image prompts are limited to fronto-parallel photographs, which requires a specific capture.

Other methods leveraged transformers as a model for material generation \cite{guerrero2022matformer, hu2023generating} but focused on procedural material, which relies on generating functional graph generation, a very different modality. These procedural representations have resolution and editability benefits, but cannot easily model materials with complex texture patterns in the wild. In contrast, our model supports generating materials from any image or text prompt and produces varied, high-quality material samples.

\subsection{Material Extraction and Rectification}
Different methods were proposed to rectify textures or generally enable non-fronto-parallel textures as input. Some approaches~\cite{yeh2022photoscene, psdrroom} aim to evaluate the materials in an image through a retrieval and optimization method. Given an image, they retrieve the geometries and procedural materials in databases to optimize their position and appearance via differentiable rendering~\cite{match, Zhang:2020:PSDR}. Closest to our work is Material Palette \cite{lopes2024material}, targeting material extraction from a single photo, not restricted to fronto parallel images. The method leverages Dreambooth~\cite{ruiz2023dreambooth} optimized through a LoRA ~\cite{hu2021lora} on Stable Diffusion~\cite{rombach2022high} to learn a ``concept'' for each material. This lets them generate a texture with a similar appearance to the target material and  use a separate material estimation network to decompose the texture into material maps. However, this LoRA optimization step takes up to 3 minutes for each image, and we find that our approach reproduces better the target appearance. 

A related field is that of texture synthesis from real-world images. \citet{wu2020deep} present an automatic texture exemplar extraction based on Trimmed Texture CNN. VQGAN~\cite{esser2021taming} achieves high resolution image-to-image synthesis with a transformer-based architecture. These methods however do not support the common occlusions and deformations that occur in natural images. To tackle this limitation, \citet{hao2023diffusion} propose to rectify occlusions and distortions in texture images via a conditional denoising U-Net with an occlusion-aware latent transformer. We show that our approach yields better texture rectification and simultaneously generates material parameters.

\subsection{Diffusion Models and Diffusion Transformers}
Diffusion models~\cite{ho2020denoising, sohl2015deep, song2019generative, song2021scorebased} are state-of-the-art generative models, showing great results across various visual applications such as image synthesis and video generation. The core architecture of diffusion models progressed from simple U-Nets, incorporating self-attention and enhanced upscaling layers~\cite{dhariwal2021diffusion}, prior-based text-to-image model \cite{nichol2021glide, ramesh2022hierarchical}, a VAE~\cite{kingma2022autoencoding} for latent diffusion models (LDM) ~\cite{rombach2022high} and temporal attention layers for video generations~\cite{blattmann2023align, blattmann2023stable}. These image generation methods all rely on a U-Net backbone, a convolutional-based encoder-decoder architecture.

Recently, transformer-based diffusion models, Diffusion Transformers (DiT) were proposed~\cite{peebles2023scalable}, benefiting from the scalability of Transformer models, removing the convolutions inductive bias. PixArt-$\alpha$ presents a DiT-based text-to-image that can synthesize high resolution images with low training cost. Stable Diffusion 3 \cite{esser2024scaling} demonstrates that a multi-modal DiT model trained with Rectified Flow can achieve superior image synthesis quality. Compared to the U-Net architecture, the DiT shows greater flexibility in the representation on the visual data, which is particularly important to video synthesis tasks. Sora~\cite{videoworldsimulators2024}, a DiT-based video diffusion model, encodes video sequences as tokens and uses transformers to denoise these visual tokens, demonstrating the ability to generate minute-long, high-resolution high-quality videos. We adapt a DiT-based video generation model for our purpose and show that it can be flexibly transformed into a multi-channel material generator.

\section{Method} 
\label{sec:method}
\subsection{Diffusion Transformers}
Diffusion models are generative models that iteratively transform an initial noise distribution (e.g. Gaussian noise) into a complex real-world data distribution (e.g., images, or their encodings). The diffusion process relies on a \textit{forward} process that progressively transforms the original data distribution into a noise distribution. For example, this can be achieved by iteratively adding Gaussian noise to the data sample. Given data samples \( x \sim p_{\text{data}} \), corrupted data $p(x_T|x_0)=\prod_{t=1}^{T} p(x_t|x_{t-1}, \epsilon), \epsilon \sim \mathcal{N}(0, I) $ are constructed in $T$ diffusion steps.

To sample the original data distribution $p_{\text{data}}$ from the noise distribution, a \textit{reverse} mapping $p(x_0)=p(x_T) \prod_{t=1}^{T} q(x_{t-1}|x_t, \epsilon_t)$ needs to be modeled where \( \epsilon_t \) is the noise sampled at each step. A neural network $f_{\theta}$ is conditioned on the denoising step $t$ to predict the noise $\epsilon_t$, which is then used to reconstruct $x_{t-1}$ from $x_t$ in each reverse step \cite{ho2020denoising}:
\begin{equation}
   \mathbb{E}_{x \sim p_{\text{data}}, t \sim U(0,T))} \left[ \left\lVert \epsilon_t - f_{\theta}(x_t; c, t) \right\rVert^2 \right],
    \label{eq:ojective}
\end{equation}
where $c$ is conditional inputs (e.g., text prompts or images).

We use a Diffusion Transformer \cite{peebles2023scalable} architecture as a backbone to model $f_{\theta}$. The visual data $x \in \mathbb{R}^{F \times 3 \times H \times W}$ is tokenized patch-wise, resulting in visual tokens $\hat{x} \in \mathbb{R}^{V \times D}$ where $H, W, F$ are the spatial and temporal dimensions of the video, $V$ is the number of tokens and $D$ is the feature dimension. Positional encoding is also added to $\hat{x}$ to specify spatial and temporal order. Any condition $c$ is also embedded as tokens $\hat{c} \in \mathbb{R}^{V' \times D}$ where $V'$ is the number of the tokens for conditional inputs. For example, when $c$ is a text, it is encoded by a pre-trained encoder \cite{radford2021learning} with additional embedding layers to map it into the same feature dimension $D$. The transformer $f_{\theta}(\hat{x}_t; \hat{c}, t)$ is trained to denoise each patch at timestep $t$. The final denoised patches $\hat{x}_0 \in \mathbb{R}^{V \times D}$ are reassembled as visual data $x_0 \in \mathbb{R}^{F \times 3 \times H \times W}$ after decoding through linear layers. Since the number of tokens grows quickly with resolution, we use a variational autoencoder (VAE) model \cite{peebles2023scalable, rombach2022high}  before the tokenizing process, producing a latent representation of $y \in \mathbb{R}^{F' \times D' \times H' \times W'}$ of the original data $x$ for the transformer to process.

\subsection{Datasets}
\label{sec:dataset}
To train our material generative model, we propose two datasets, \textit{Scenes} and \textit{Materials}. Together, these datasets enable joint training for both surface rectification and high quality material generation.

\begin{figure}[t]
    \centering		
    \begin{minipage}{3.4in}
         \begin{minipage}{3.4in}	
            % \centering
            \begin{minipage}{0.118\linewidth}
                \subcaption*{\tiny Render}
                \includegraphics[width=\linewidth]{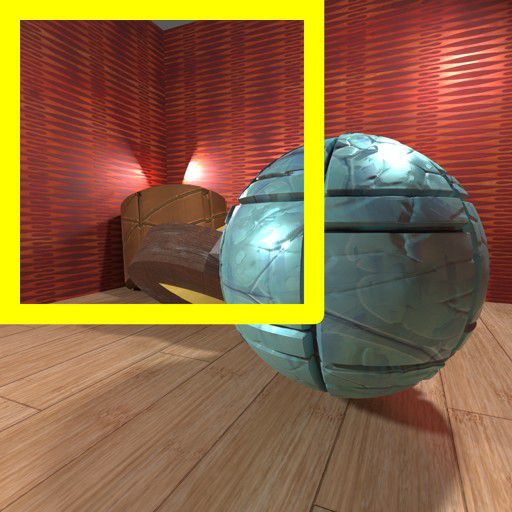}
            \end{minipage}
            \begin{minipage}{0.118\linewidth}
                \subcaption*{\tiny Crop}
                \includegraphics[width=\linewidth]{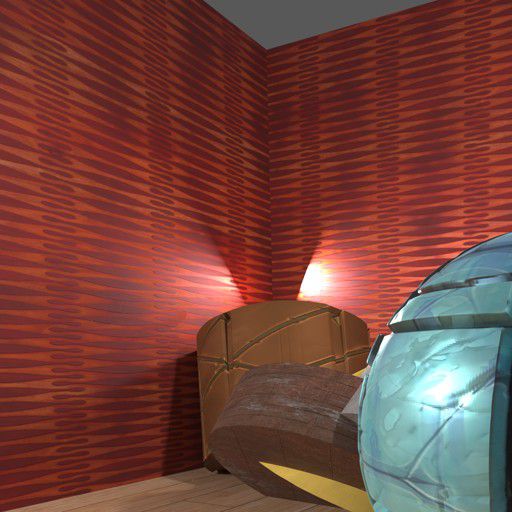}
            \end{minipage}
            \begin{minipage}{0.118\linewidth}
                \subcaption*{\tiny Mask}
                \includegraphics[width=\linewidth]{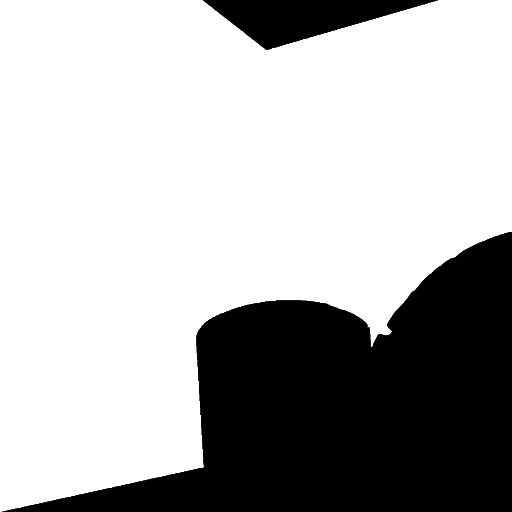}
            \end{minipage}
            \begin{minipage}{0.118\linewidth}
                \subcaption*{\tiny Albedo}
                \includegraphics[width=\linewidth]{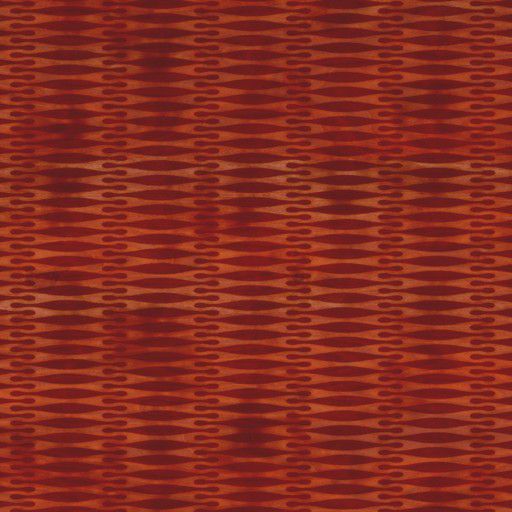}
            \end{minipage}
            \begin{minipage}{0.118\linewidth}
                \subcaption*{\tiny Normal}
                \includegraphics[width=\linewidth]{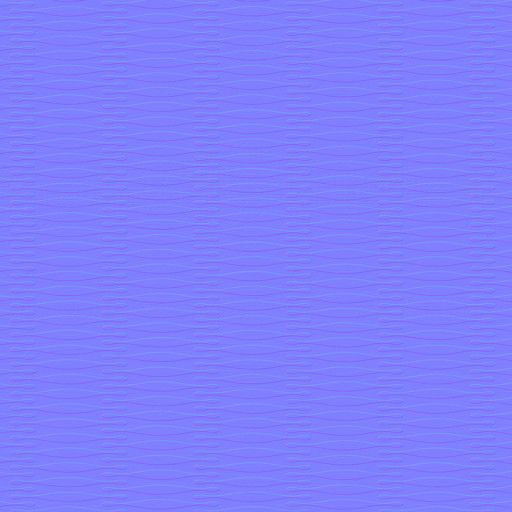}
            \end{minipage}
            \begin{minipage}{0.118\linewidth}
                \subcaption*{\tiny Roughness}
                \includegraphics[width=\linewidth]{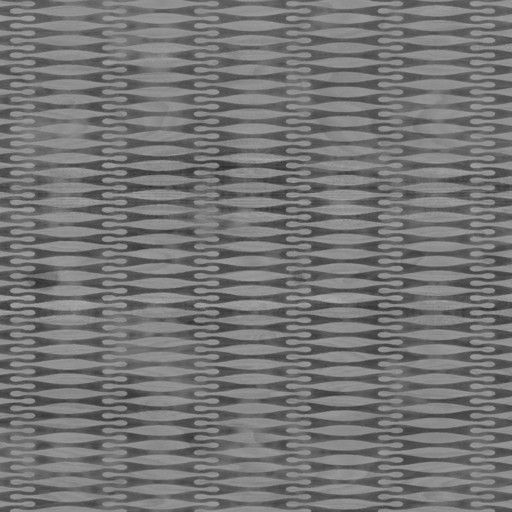}
            \end{minipage}
            \begin{minipage}{0.118\linewidth}
                \subcaption*{\tiny Height}
                \includegraphics[width=\linewidth]{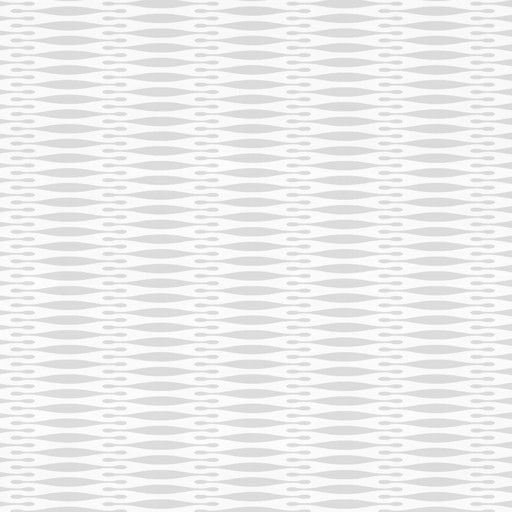}
            \end{minipage}
            \begin{minipage}{0.118\linewidth}
                \subcaption*{\tiny Metallic}
                \includegraphics[width=\linewidth]{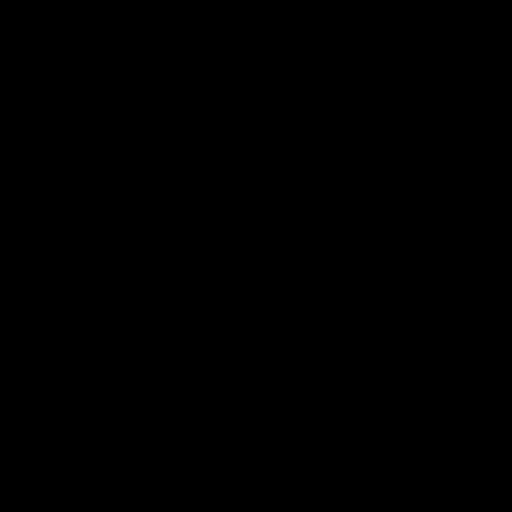}
            \end{minipage}
        \end{minipage}	
        \centering
        \vspace{-0.2cm}
        \subcaption*{\tiny Text: ``Wood, cherry wood laser cut bowling pin"}
    \end{minipage}	

    \begin{minipage}{3.4in}
         \begin{minipage}{3.4in}	
            % \centering
            \begin{minipage}{0.118\linewidth}
            \includegraphics[width=\linewidth]{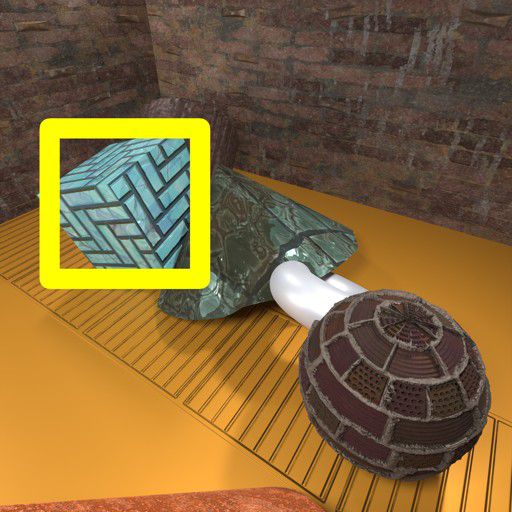}
            \end{minipage}	
            \begin{minipage}{0.118\linewidth}
            \includegraphics[width=\linewidth]{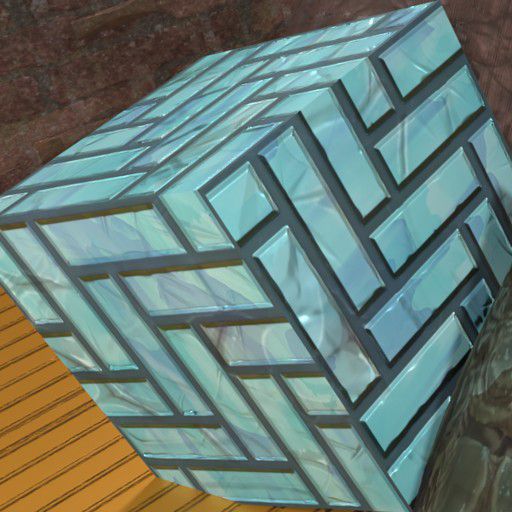}
            \end{minipage}	
            \begin{minipage}{0.118\linewidth}
            \includegraphics[width=\linewidth]{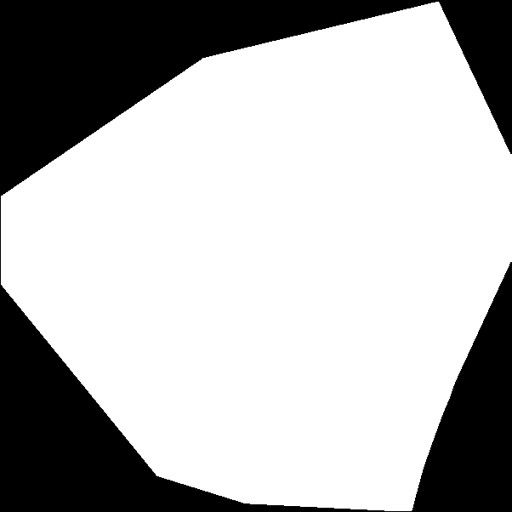}
            \end{minipage}	
            \begin{minipage}{0.118\linewidth}
            \includegraphics[width=\linewidth]{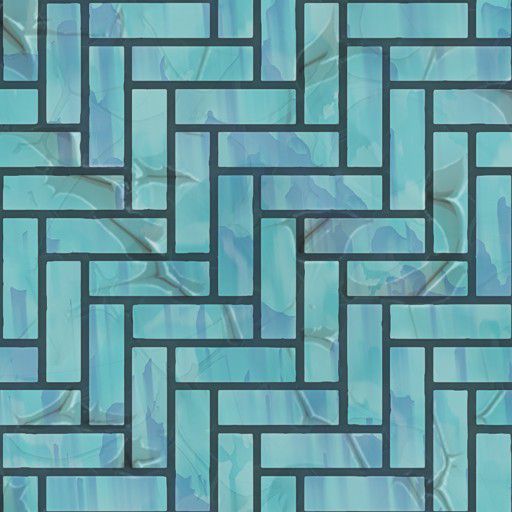}
            \end{minipage}	
            \begin{minipage}{0.118\linewidth}
            \includegraphics[width=\linewidth]{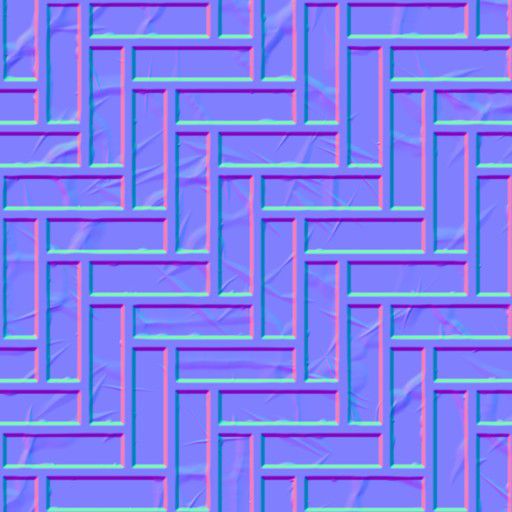}
            \end{minipage}	
            \begin{minipage}{0.118\linewidth}
            \includegraphics[width=\linewidth]{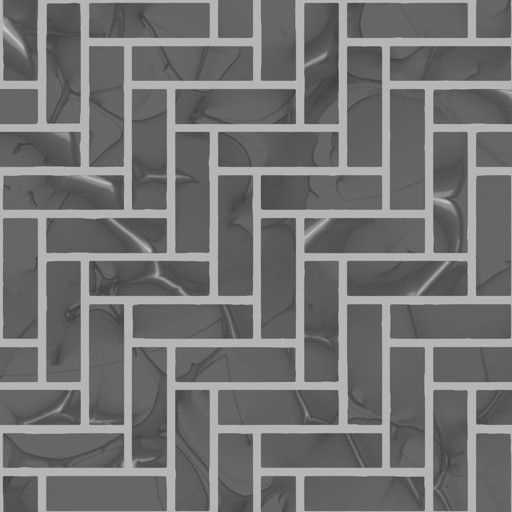}
            \end{minipage}	
            \begin{minipage}{0.118\linewidth}
            \includegraphics[width=\linewidth]{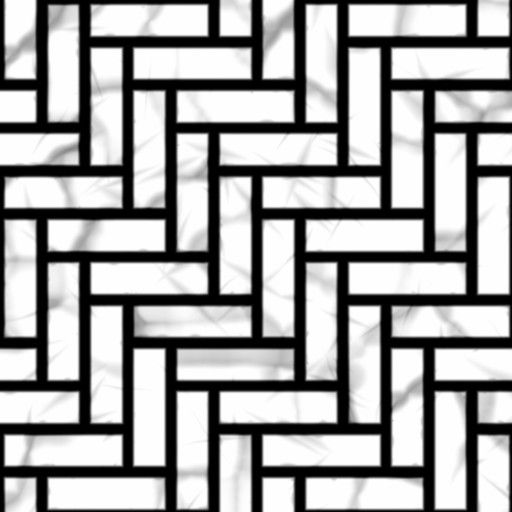}
            \end{minipage}	
            \begin{minipage}{0.118\linewidth}
            \includegraphics[width=\linewidth]{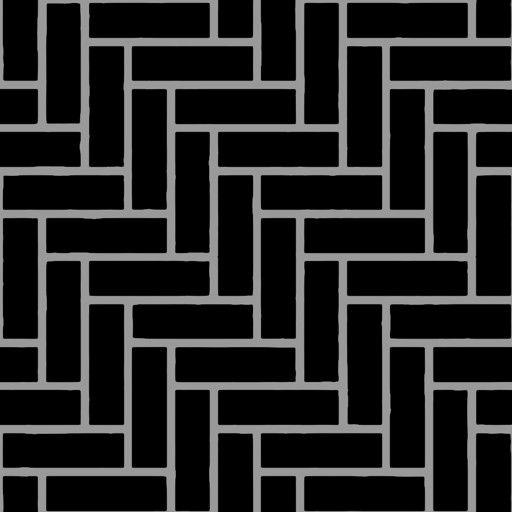}
            \end{minipage}	
        \end{minipage}	
        \centering
        \vspace{-0.2cm}
        \subcaption*{\tiny Text: ``Marble granite, stylized light blue marble herringbone tiles"}
    \end{minipage}	

   \caption{Our \textit{Scenes} dataset. We build random scenes and render paired text/image-to-material dataset with 3K randomly sampled materials. In each row we show a 2K synthetic rendering, a crop with a dominant material, the material mask and corresponding material maps.} 
   \label{fig:dataset}
\end{figure}

For the \textit{Scenes} dataset, we build a set of synthetic indoor scenes with planar floors, walls, and randomly placed 3D objects, such as cubes, spheres, cylinders, cones, and toruses, similar to random Cornell boxes~\cite{cohen1985hemi}. Each object is randomly assigned a unique material from around 3,000 stationary (i.e., approximately shift invariant) materials. We use the Blender implementation of the Disney Principled BSDF model~\cite{disneybrdf} for rendering the dataset and other visualizations in the paper, using base color (albedo), normal, roughness, metallic, and height maps, leaving other parameters as default. Using this approach we create a dataset of 100,000 high-resolution rendered images, with different kinds of light sources, including point lights and area lights, to simulate complex real-world illumination (see Fig.~\ref{fig:dataset}). We randomly place cameras to capture a wide variety of view points and maximize coverage. 

We further crop the rendered images to construct training data, including input images, corresponding material maps, binary material mask, and the material name as an optional text prompt. During cropping, we ensure that the dominant material occupies at least 70\% of the region. Importantly, we rescale the material maps based on UV coordinates to ensure that the rendered crops and target material maps share a matching texture scale. After cropping, this dataset contains 800,000 text-image-mask-material tuples. We will share out dataset creation script, facilitating reproduction using public materials datasets~\cite{vecchio2024matsynth, ma2023opensvbrdf}.

As our \textit{Scenes} dataset only contains stationary materials, it may fail to represent the full diversity of textures in the wild. To enhance the generalization capability, we use an additional \textit{Materials} dataset~\cite{Martin2022}, which we augment to 800,000 cropped material maps. We use the name of the materials as the text prompts for text-to-material generation. These data items can be thought of as text-material pairs. This additional data diversity leads to significant improvement for non-stationary textures in input photographs as discussed in Sec.~\ref{sec:mixed_dataset}.

\subsection{Generative Material Model}
\label{sec:our_model}
We employ a pre-trained DiT-based text-to-video generative model as our base model, with an architecture similar to that of the publicly available HunyuanVideo model~\cite{kong2024hunyuanvideo}. It follows a decoder-only Transformer structure with stacked self-attention blocks. The model takes both text and visual conditions as input: the input frames are first encoded by a 3DVAE encoder, producing a latent representation which is then corrupted by noise during training. Simultaneously, text prompts are processed by a T5-based encoder, producing text embeddings that are appended to the noisy latents. Additionally, timestep embeddings and spatial-temporal positional embeddings are added to the latent sequences to provide temporal and spatial context (i.e., each token's frame number and position within the frame). The DiT backbone denoises the latent sequence, which is decoded into video frames using the 3DVAE decoder. We retarget this architecture into a multi-channel material generator.

To retarget the model while preserving its learned prior knowledge, we stack the material maps $M$ (albedo map, normal map, height map, roughness map and metallicity map) into a ``video'' of 5 frames, and compute the temporal positional embedding assuming their time stamp interval is 1 e.g., fps=1. Since DiT flexibly generates tokenized data, as opposed to a U-Net architecture~\cite{blattmann2023stable}, the number of frames it is able to produce is not fixed, allowing us to adapt the original video generator to generate the right number of ``frames'' to meet our requirement. Note that we treat all material maps as ``keyframes'' (images) in the 3DVAE, with no motion prediction between the frames.

For image-conditioned material generation, we consider the input image $I$ as the first frame, with the model generating the stacked material maps $M$ as the subsequent frames, similar to a video extension model. This setup allows the transformer’s self-attention mechanism to jointly reason over both the input image and the predicted material maps, while tolerating pixel misalignment due to perspective distortion or varying camera poses.

Our decision to use a video-generation backbone is further motivated by its inherent capacity to enforce temporal consistency, which, in this context, translates to spatial alignment across material maps. The DiT model implicitly learns that all frames beyond the first should remain temporally coherent, a property that aligns well with the goal of generating consistent texture channels. Convolution-based architectures such as SDXL~\cite{podell2023sdxl} led to worse results in our initial experiments; we hypothesize this is because they prefer (approximate) pixel alignment between input and output.

This design also avoids architectural modifications that disrupt pretrained knowledge. Image diffusion models are typically trained to generate 3 channels (RGB) and need to be non-trivially adjusted to generate more channels~\cite{liu2023hyperhuman}, or generate a single material map at a time and repurpose the input text prompt as a "switch" ~\cite{zeng2024rgb}. Recent work~\cite{vecchio2024controlmat} modified the architecture to generate multiple maps, but had to train the model from random weights, missing the rich prior provided by large scale image datasets about material appearance. Our use of a video model enables the generation of multiple maps with minimally invasive architecture modifications, inheriting strong priors from pretrained video diffusion models.

Our solution preserves compatibility with inference-time techniques such as noise rolling~\cite{vecchio2024controlmat}, TexSliders~\cite{guerrero2024texsliders}, SDEdit~\cite{meng2021sdedit}, etc., which could broaden its range of applications. Furthermore, since the DiT backbone operates on tokenized representations rather than fixed-size tensors (as in U-Net-based architectures), it is extensible, as additional maps (e.g., opacity maps) can be incorporated by appending new frames at the end of the token sequence.
It also remains computationally efficient, since only a small number of frames is generated.

We additionally train our material generator to produce a segmentation mask for the dominant material in the crop. Typically, the user-provided crop is not entirely covered by a single material (see Fig.~\ref{fig:material_genertaion_real}). Performing conservative cropping on an image may reduce the number of usable pixels, while using an additional segmentation mask requires additional user input or a separate segmentation model~\cite{Sharma2023materialistic}. Instead, our model automatically identifies the dominant material~\cite{Lu2009Dominant} in the image. We add a mask $S$ to be inferred from the input image as the second frame. Our training data $x$ can thus be represented as $x = \text{stack}(I, S, M)$, where $x \in \mathbb{R}^{7 \times 3 \times H \times W}$; we have 7 RGB frames: input, mask, and five material maps. Since mask, height, roughness, and metallic maps are single-channel, we convert them into RGB images before concatenating them with other frames. Noise $\epsilon_t$ is applied only to the last six frames occupied by $S$ and $M$, resulting in $x_t = \text{stack}(I, S_t, M_t)$, with the first frame (input image) remaining free of noise. Our objective from Eq.~\ref{eq:ojective} is
\begin{equation}
   \mathbb{E}_{x \sim p_{\text{data}}, t \sim U(0,T)} \left[ \left\lVert \epsilon_t - f_{\theta}(x_t; c, t)[-6:]\right\rVert^2 \right],
\end{equation}
where $c$ denotes the text material description. This process can be seen as frames completion (mask and material channels) given the input image and text condition. The notation $[-6:]$ refers to the last 6 frames generated by the Transformer.
When the input consists solely of $c$ without $I$, $x = \text{stack}(S, M)$ where $S$ is a uniformly white RGB image. The computation of the loss remains unchanged.

\subsection{Training and Inference}
\label{sec:implementation}
We finetune the pre-trained DiT model using the AdamW optimizer on 8 Nvidia A100 GPUs. The learning rate is set at $0.99\times10^{-4}$ with an effective batch size of 64. The model is finetuned on $256 \times 256$ resolution for about 70K steps, which takes 90 hours. During training, we feed data from our two training datasets \textit{Scenes} and \textit{Materials} in a 5:3 ratio, prioritizing the task of image-conditioned material generation. For text-only or unconditional generation, the mask is replaced by a completely white image placeholder. 

Our model completes a generation in 12 seconds using DDIM \cite{song2020denoising} with 50 diffusion steps on an Nvidia A100. The model natively outputs a resolution of 256 due to limited computational resources. We apply an upsampler \cite{waifu2x-ncnn-vulkan} to increase the resolution of each material map to 512 $\times$ 512.

\section{Results}
\label{sec:results}
We evaluate the performance of our \method across multiple dimensions. First, we perform qualitative and quantitative comparisons with Material Palette~\cite{lopes2024material} on material extraction using both synthetic and real-world images (Sec.~\ref{sec:image_conditioned_generation}). Next, we compare with a material acquisition method~\cite{vecchio2024controlmat} and a texture rectification method on real-world images (Sec.~\ref{sec:image_conditioned_generation}), and with MatGen~\cite{vecchio2024controlmat} and MatFuse~\cite{vecchio2024matfuse} on text-to-material generation (Sec.~\ref{sec:text_conditioned_generation}). Finally, we conduct ablation studies on multi-modality, dataset design, the usage of a mask, and evaluate the impact of the input image scale and the robustness to distortion and lighting/shadowing. We also demonstrate our generalizability to complex patterns and the ability to generate seamless results (Sec.~\ref{sec:ablation}). 

\subsection{Evaluation dataset and metrics}

\subsubsection{Synthetic evaluation dataset.}
\label{sec:evaluation_dataset}
For systematic evaluation, we build a synthetic evaluation dataset by gathering a diverse set of 531 materials from PolyHaven\footnote{\url{https://polyhaven.com/}}, applied to three interior scenes from the Archinteriors collection~\cite{evermotion2021archinterior} (completely independent from our training set). For each scene, we sequentially apply the 531 collected materials to a designated object inside the scene, and render 2D images using Blender Cycles~\cite{blender2018} with the scene's default illumination setup. We generate 1,593 synthetic renderings, and crop a square around the location of the object with replaced materials.

\subsubsection{Real photographs evaluation dataset.}
To validate the generalization of our models, we curate an evaluation dataset containing real photographs captured by smartphones. This dataset covers a comprehensive set of real-world materials observed under both natural outdoor lighting and complex indoor illumination. We crop the photographs with a primary focus on our target material, without strictly limiting the cropping boundaries. 

\subsubsection{Evaluation metrics.}
Since we do not target pixel-aligned material capture, per-pixel metrics cannot be used for our results. Instead, we focus on the \textit{appearance similarity} of the materials extracted from the photo inputs. Following related work on high-fidelity image synthesis such as DreamBooth~\cite{ruiz2023dreambooth}, we leverage CLIP-I, which is the average pairwise cosine similarity between ViT-L-14 CLIP~\cite{radford2021learning} embeddings of two sets of images. We also use the DINO metric~\cite{ruiz2023dreambooth} to measure the average pairwise cosine similarity between ViT-L-16 DINO embeddings.

\subsection{Image Conditioned Generation}
\label{sec:image_conditioned_generation}

\begin{figure}[t]
    \centering		
    \begin{minipage}{3.4in}
         \begin{minipage}{3.4in}	
            \centering
            \begin{minipage}{0.32\linewidth}
            \subcaption*{\small Ground-Truth}
            \includegraphics[width=\linewidth]{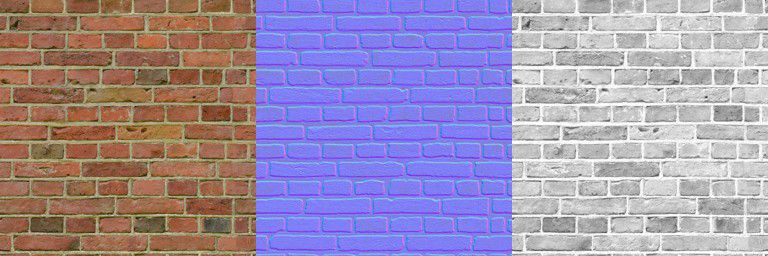}
            \end{minipage}	
            \begin{minipage}{0.32\linewidth}
            \subcaption*{\small Ours}
            \includegraphics[width=\linewidth]{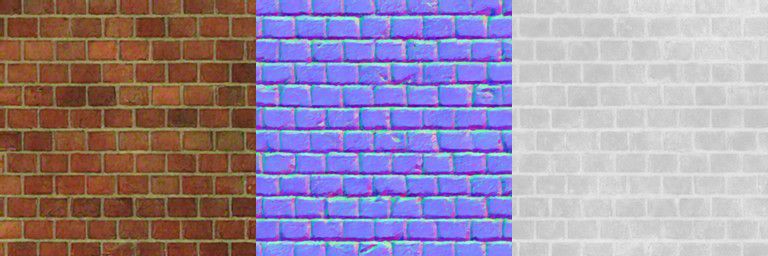}
            \end{minipage}	
            \begin{minipage}{0.32\linewidth}
            \subcaption*{\small Material Palette}
            \includegraphics[width=\linewidth]{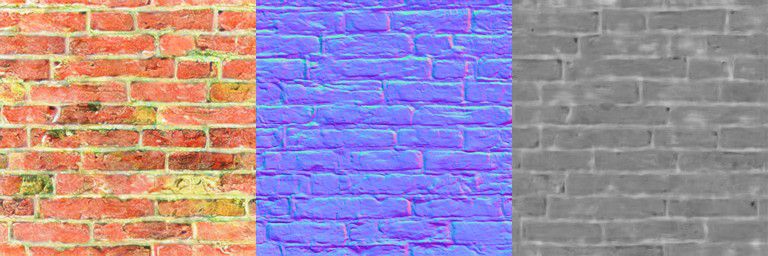}
            \end{minipage}	
        \end{minipage}	
    \end{minipage}	
    
    \begin{minipage}{3.4in}
         \begin{minipage}{3.4in}	
            \centering
            \begin{minipage}{0.32\linewidth}
                \includegraphics[width=\linewidth]{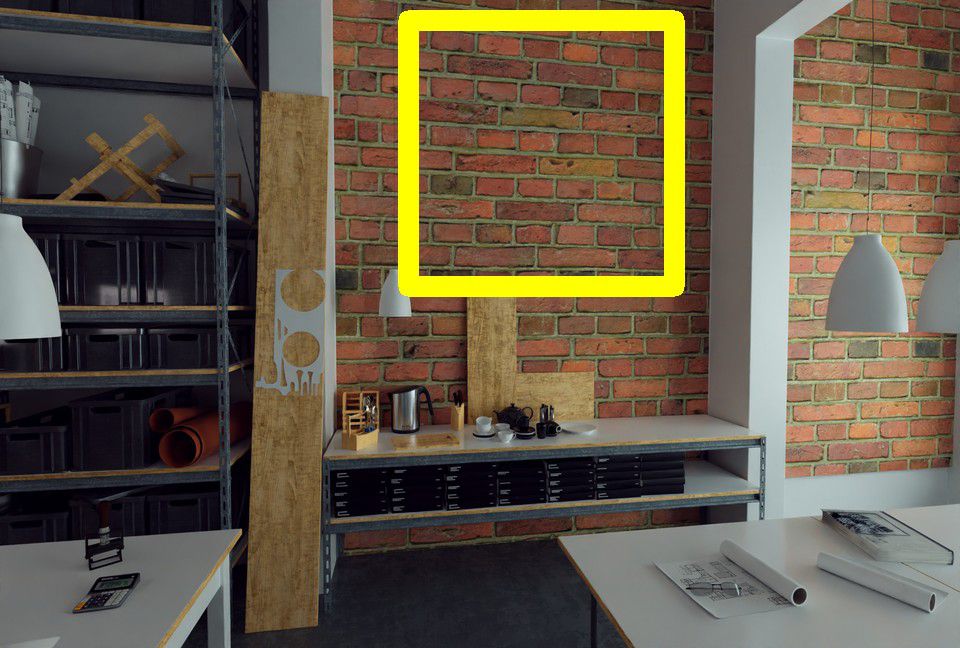}
            \end{minipage}
            \begin{minipage}{0.32\linewidth}
                \includegraphics[width=\linewidth]{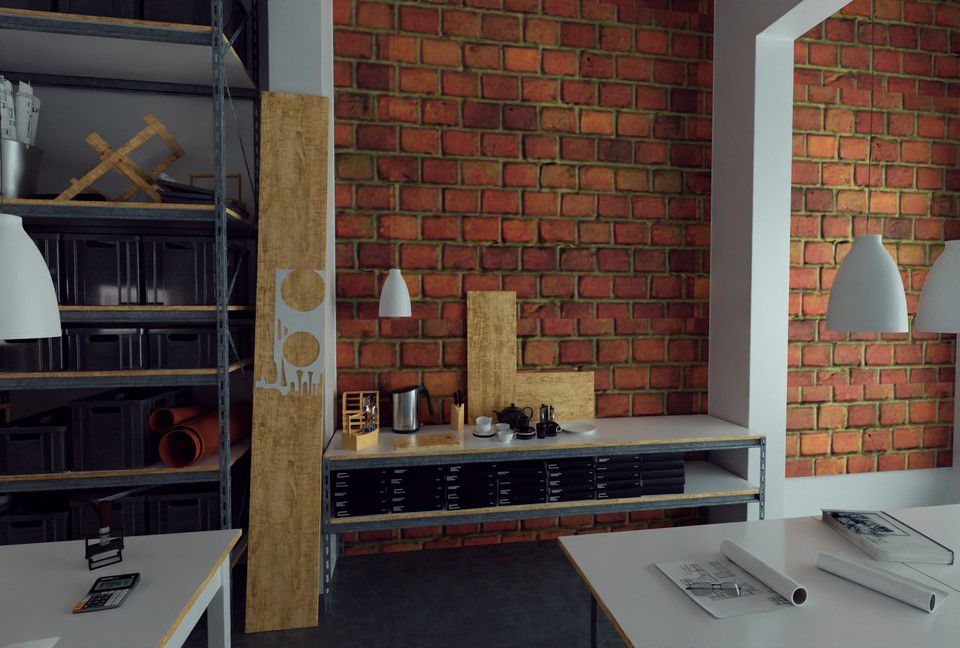}
            \end{minipage}
            \begin{minipage}{0.32\linewidth}
                \includegraphics[width=\linewidth]{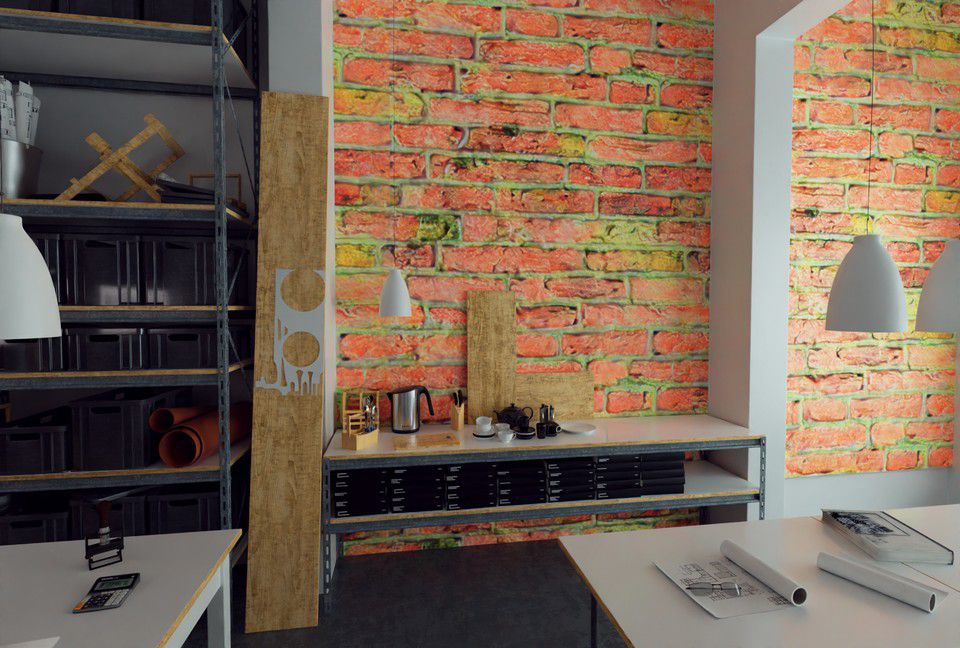}
            \end{minipage}
        \end{minipage}	
    \end{minipage}	
    
    \vspace{0.05in}
    
    \begin{minipage}{3.4in}
         \begin{minipage}{3.4in}	
            \centering
            \begin{minipage}{0.32\linewidth}
            \includegraphics[width=\linewidth]{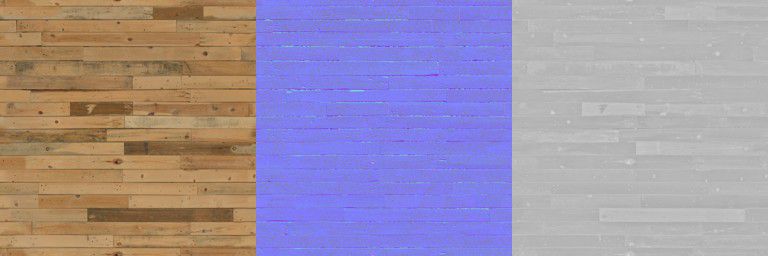}
            \end{minipage}	
            \begin{minipage}{0.32\linewidth}
            \includegraphics[width=\linewidth]{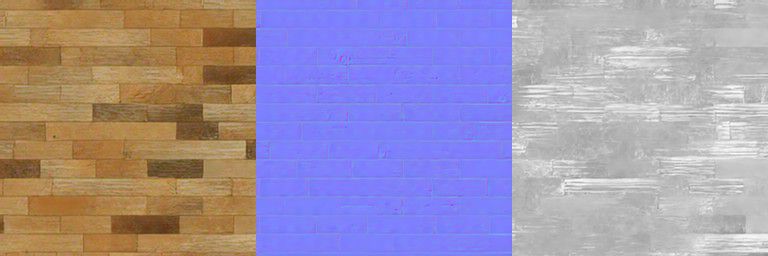}
            \end{minipage}	
            \begin{minipage}{0.32\linewidth}
            \includegraphics[width=\linewidth]{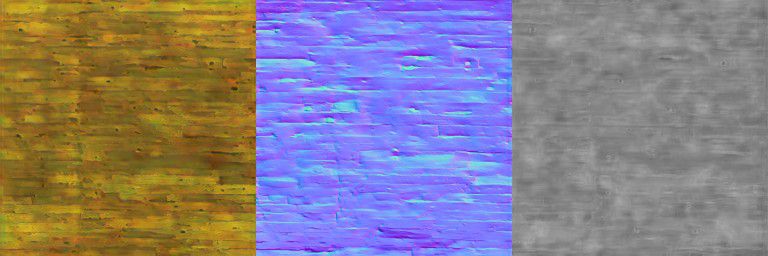}
            \end{minipage}	
        \end{minipage}	
    \end{minipage}	
    
    \begin{minipage}{3.4in}
         \begin{minipage}{3.4in}	
            \centering
            \begin{minipage}{0.32\linewidth}
            \includegraphics[width=\linewidth]{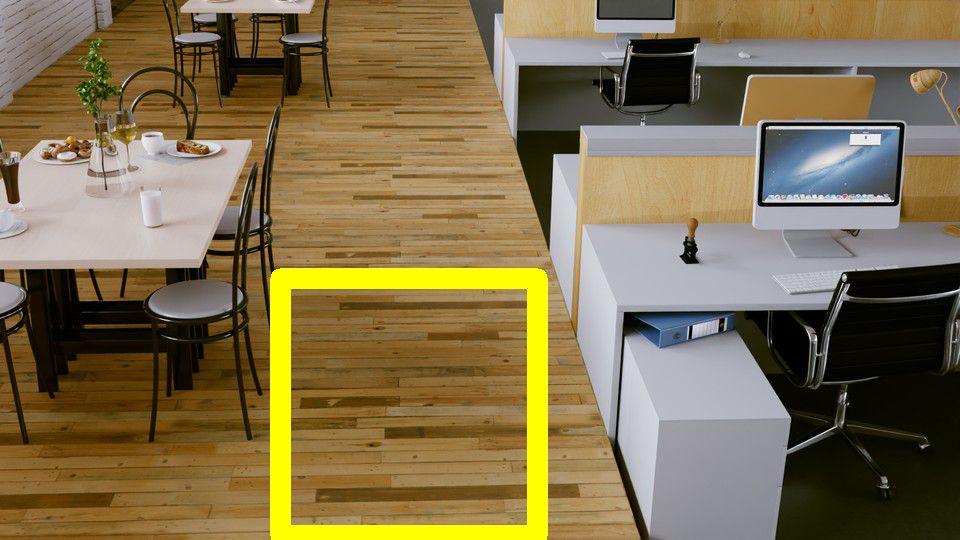}
            \end{minipage}	
            \begin{minipage}{0.32\linewidth}
            \includegraphics[width=\linewidth]{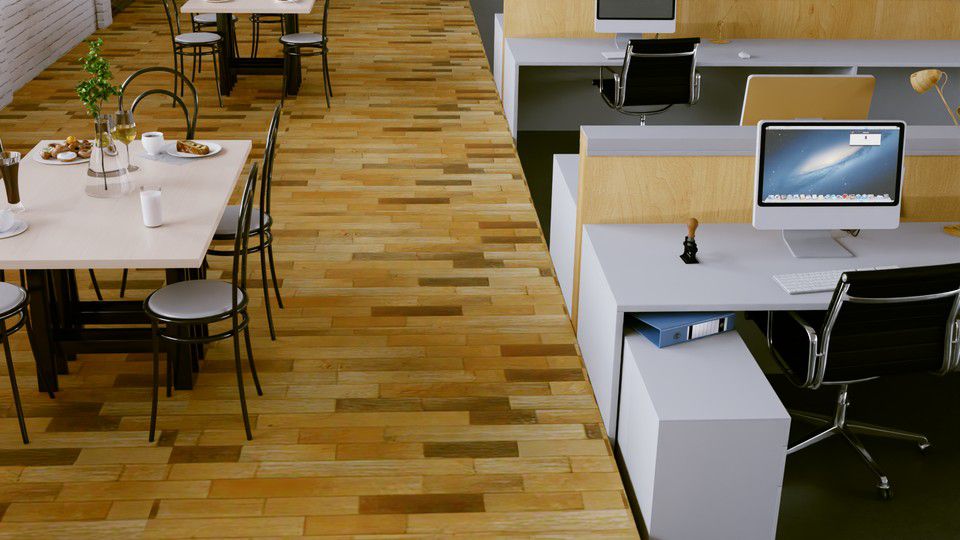}
            \end{minipage}	
            \begin{minipage}{0.32\linewidth}
            \includegraphics[width=\linewidth]{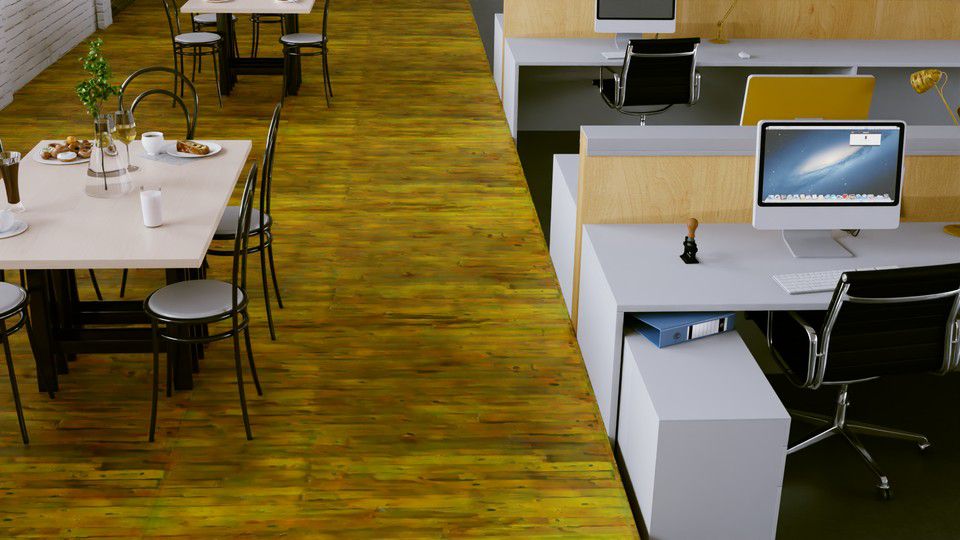}
            \end{minipage}	
        \end{minipage}	
    \end{minipage}	
    
    \vspace{0.05in}
    
    \begin{minipage}{3.4in}
         \begin{minipage}{3.4in}	
            \centering
            \begin{minipage}{0.32\linewidth}
            \includegraphics[width=\linewidth]{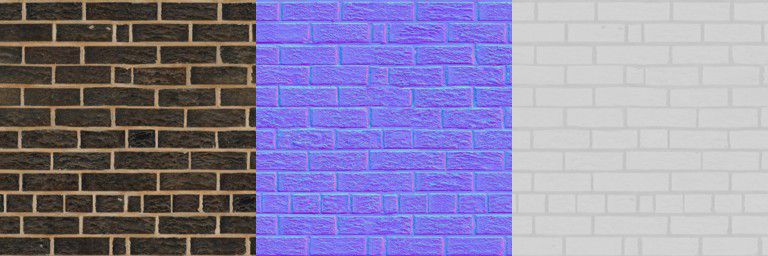}
            \end{minipage}	
            \begin{minipage}{0.32\linewidth}
            \includegraphics[width=\linewidth]{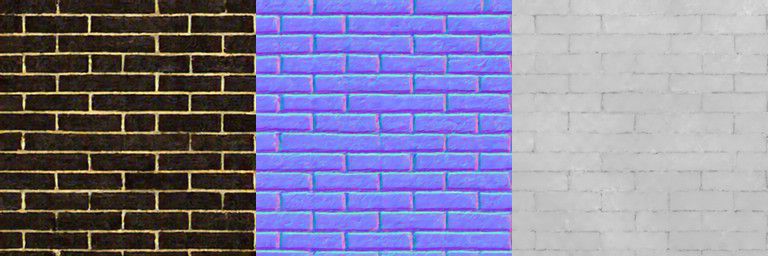}
            \end{minipage}	
            \begin{minipage}{0.32\linewidth}
            \includegraphics[width=\linewidth]{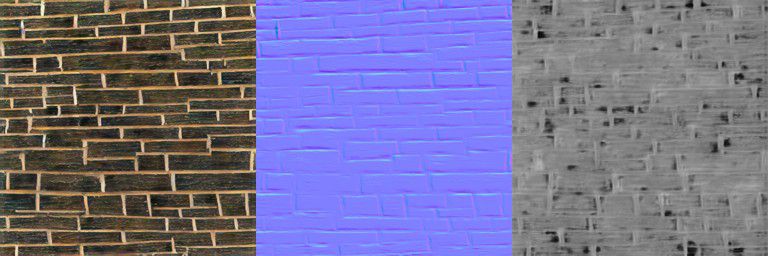}
            \end{minipage}	
        \end{minipage}	
    \end{minipage}	
    
    \begin{minipage}{3.4in}
         \begin{minipage}{3.4in}	
            \centering
            \begin{minipage}{0.32\linewidth}
            \includegraphics[width=\linewidth]{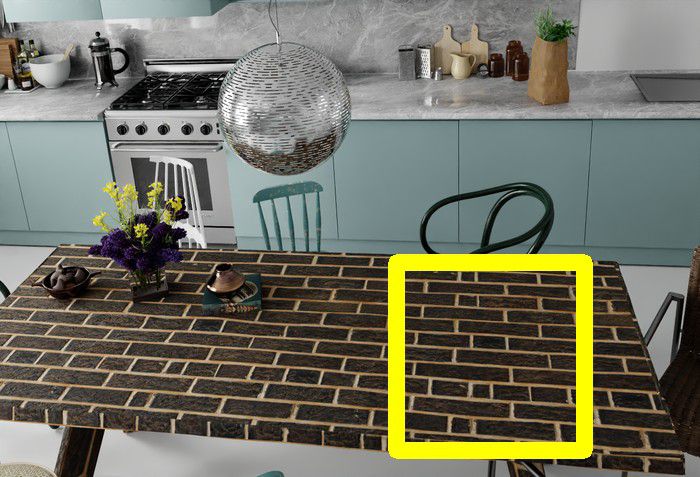}
            \end{minipage}	
            \begin{minipage}{0.32\linewidth}
            \includegraphics[width=\linewidth]{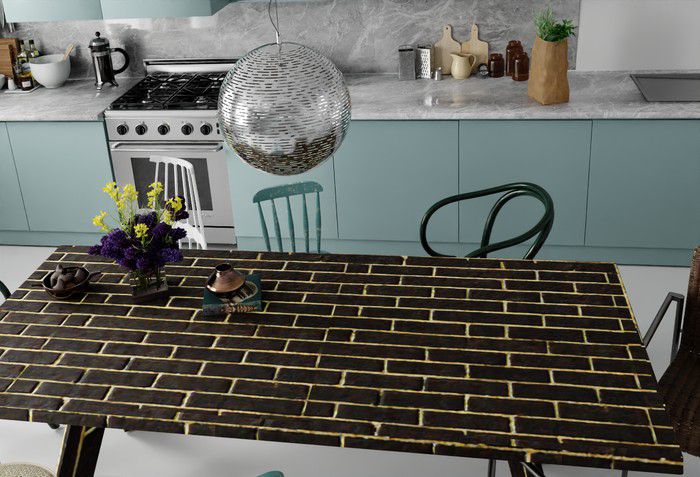}
            \end{minipage}	
            \begin{minipage}{0.32\linewidth}
            \includegraphics[width=\linewidth]{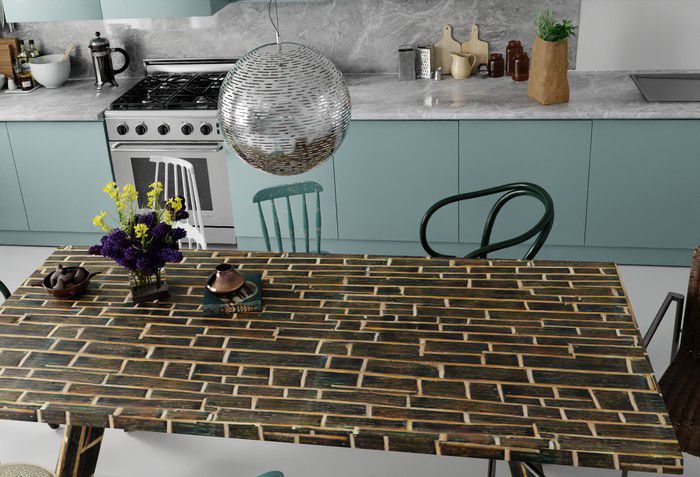}
            \end{minipage}	
        \end{minipage}	
    \end{minipage}	
    
   \caption{Comparisons with Material Palette~\cite{lopes2024material} on synthetic dataset for material extraction. The first column shows the ground truth material maps from PolyHaven, with the rendered scene below. The yellow square area indicates the crop used as the input for both models. The second and third columns show the material maps extracted by our model and Material Palette, along with the re-rendered images. We can see that our approach better matches the ground truth appearance.} 
   \label{fig:synthetic_results}
\end{figure}

We evaluate the performance of our model on both synthetic images and real photographs. We first show a visual comparison with the state-of-the-art method Material Palette~\cite{lopes2024material} on our synthetic evaluation dataset (Sec.~\ref{sec:evaluation_dataset}). Since Material Palette generates only three material maps (albedo, normal, and roughness), we present both qualitative and quantitative results for these channels, along with the re-rendered images using these generated material maps. Our method takes 12 seconds to generate a material while Material Palette takes 3 minutes, on the same Nvidia A100 GPU, a 15 times speedup. Furthermore, our model can generate materials in batches. In Fig.~\ref{fig:synthetic_results} we show that our model produces material maps with a closer texture appearance and better matching the ground-truth material maps. In contrast, Material Palette struggles to reconstruct structured textures often resulting in distorted lines. We also observe that in the rendered images, our generated materials better matches the original input images.  

We include a quantitative comparison and 95\% confidence interval with Material Palette on the entire synthetic dataset in Tab.~\ref{tab:quantitative_score}. We find that our model performs better on two metrics for the vast majority of generated materials channels, with the exception of the Albedo for which the intervals overlap. Our re-rendered images also show consistently higher alignment with the ground truth.

\begin{table}[t]
\centering
\caption{Quantitative results of material extraction. We compare with Material Palette \cite{lopes2024material} and report the average CLIP-I metric $\uparrow$ and DINO metric $\uparrow$ between the output material maps and ground truth alongside the 95\% confidence interval.}
\label{tab:quantitative_score}
\begin{tabular}{@{} c @{\hspace{8pt}} c @{\hspace{8pt}} c @{\hspace{8pt}} c @{\hspace{8pt}} c @{}}
\toprule
\textbf{CLIP$\uparrow$} & \textbf{Albedo} & \textbf{Normal} & \textbf{Roughness} & \textbf{Render} \\
\midrule
\textbf{Mat-Palette} & 0.816$\pm 0.03$ & 0.867$\pm 0.03$ & 0.791$\pm 0.03$ & 0.955$\pm 0.01$ \\
\textbf{Ours}  & \textbf{0.857}$\pm 0.02$ & \textbf{0.874}$\pm 0.02$ & \textbf{0.866}$\pm 0.03$ & \textbf{0.967}$\pm 0.01$ \\
\bottomrule
\textbf{DINO$\uparrow$} & \textbf{Albedo} & \textbf{Normal} & \textbf{Roughness} & \textbf{Render} \\
\midrule
\textbf{Mat-Palette} & \textbf{0.503}$\pm 0.1$ & 0.631$\pm 0.09$ & 0.502$\pm 0.09$ & 0.797$\pm 0.05$ \\
\textbf{Ours}  & 0.494$\pm 0.1$ & \textbf{0.672}$\pm 0.08$ & \textbf{0.566}$\pm 0.1$ & \textbf{0.863}$\pm 0.04$ \\
\bottomrule
\end{tabular}
\end{table}

\begin{figure}[htbp!]
    \centering		
    \begin{minipage}{3.4in}
        \begin{minipage}{0.02in}	
            \centering
                \vspace{0.1in}
                \rotatebox{90}{\parbox{1cm}{\centering\tiny \vspace{0.05cm} Ours}}
        \end{minipage}	
        \hspace{0.02in}
             \begin{minipage}{3.3in}	
                \centering
                \begin{minipage}{0.13\linewidth}
                    \subcaption*{\tiny Input}
                    \includegraphics[width=\linewidth]{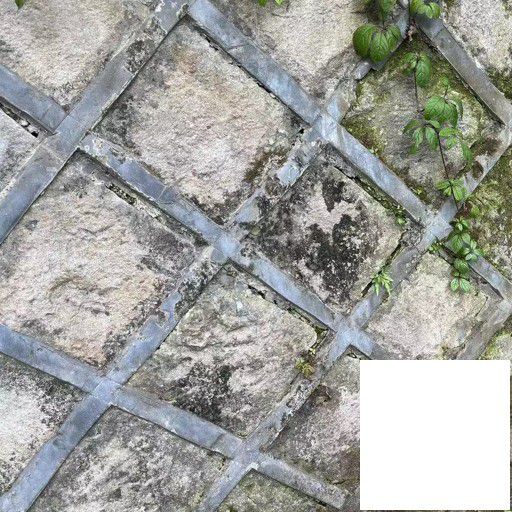}
                \end{minipage}
                \begin{minipage}{0.13\linewidth}
                    \subcaption*{\tiny Albedo}
                    \includegraphics[width=\linewidth]{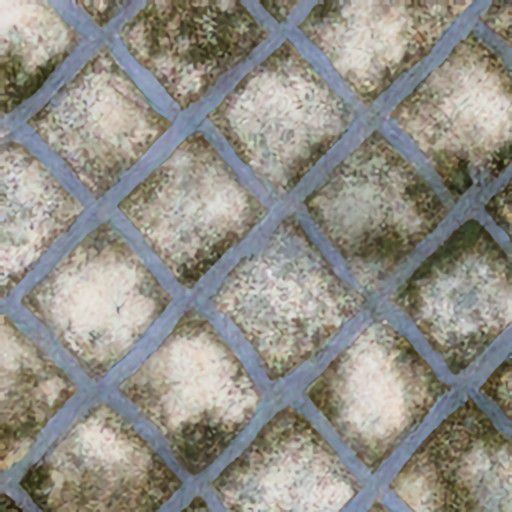}
                \end{minipage}
                \begin{minipage}{0.13\linewidth}
                    \subcaption*{\tiny Normal}
                    \includegraphics[width=\linewidth]{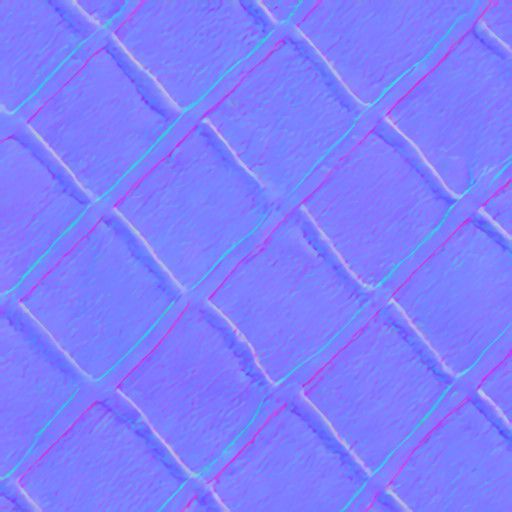}
                \end{minipage}
                \begin{minipage}{0.13\linewidth}
                    \subcaption*{\tiny Roughness}
                    \includegraphics[width=\linewidth]{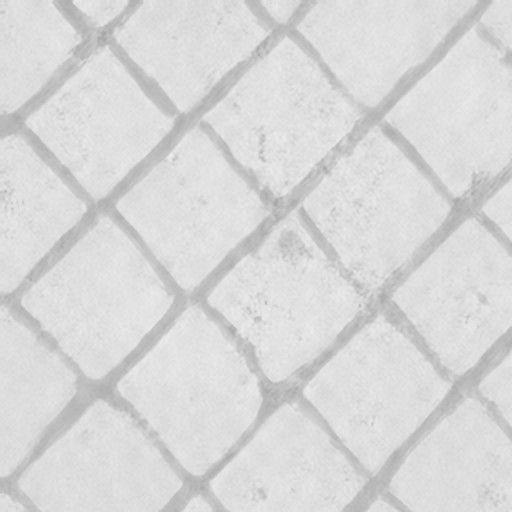}
                \end{minipage}
                \begin{minipage}{0.13\linewidth}
                    \subcaption*{\tiny Height}
                    \includegraphics[width=\linewidth]{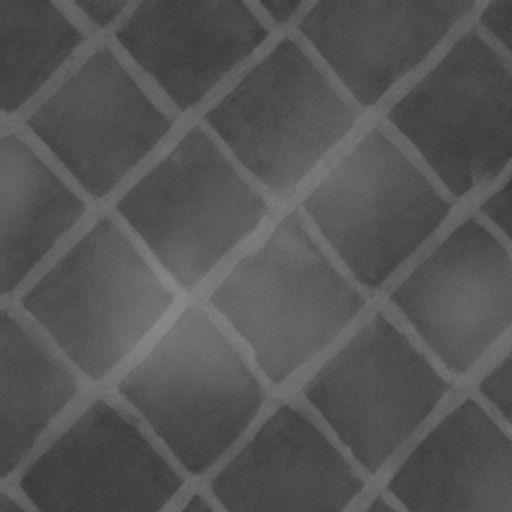}
                \end{minipage}
                \begin{minipage}{0.13\linewidth}
                    \subcaption*{\tiny Metallic}
                    \includegraphics[width=\linewidth]{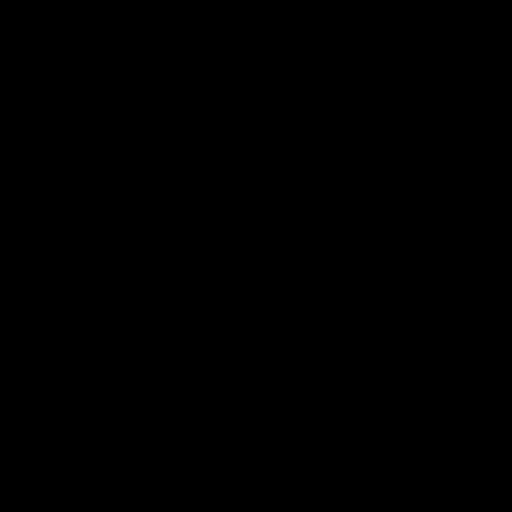}
                \end{minipage}
                \begin{minipage}{0.13\linewidth}
                    \subcaption*{\tiny Render}
                    \includegraphics[width=\linewidth]{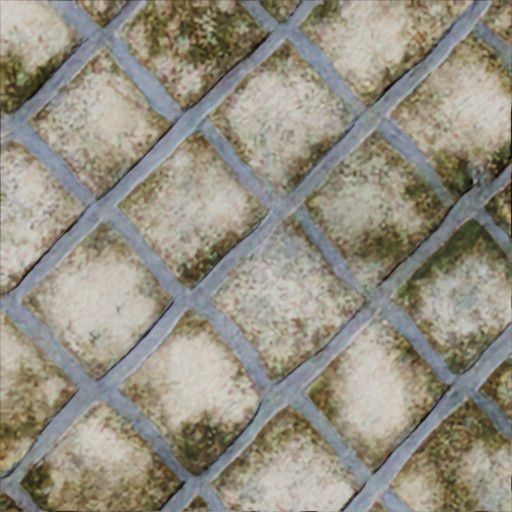}
                \end{minipage}
            \end{minipage}	
        \end{minipage}	

    \begin{minipage}{3.4in}
        \begin{minipage}{0.02in}	
            \centering
                \rotatebox{90}{\parbox{1cm}{\centering\tiny Material\vspace{-0.05cm}\\Palette}}
        \end{minipage}	
        \hspace{0.02in}
         \begin{minipage}{3.3in}	
            \centering
            \includegraphics[width=0.13\linewidth]{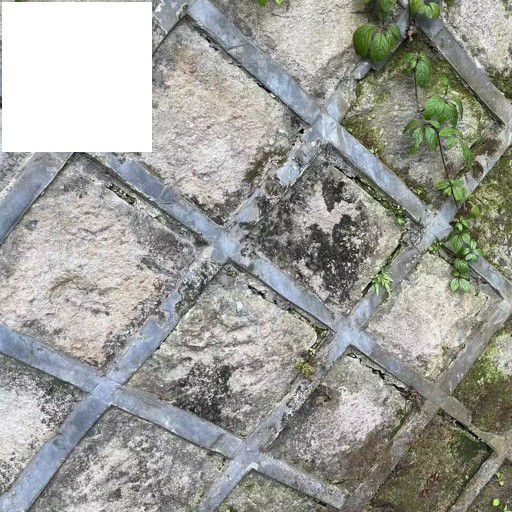}
            \includegraphics[width=0.13\linewidth]{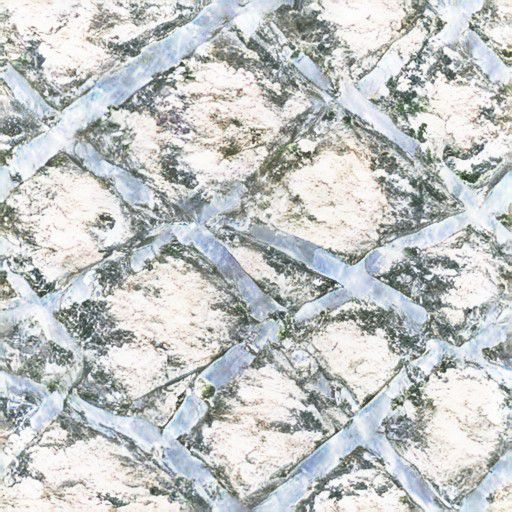}
            \includegraphics[width=0.13\linewidth]{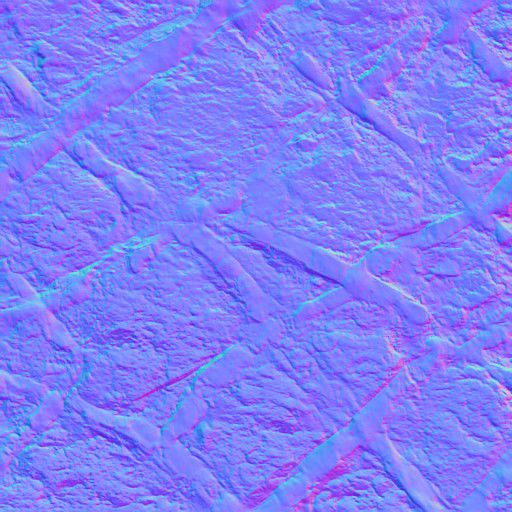}
            \includegraphics[width=0.13\linewidth]{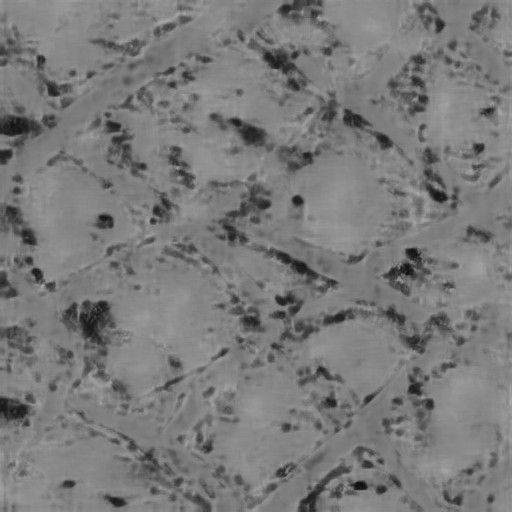}
            \includegraphics[width=0.13\linewidth]{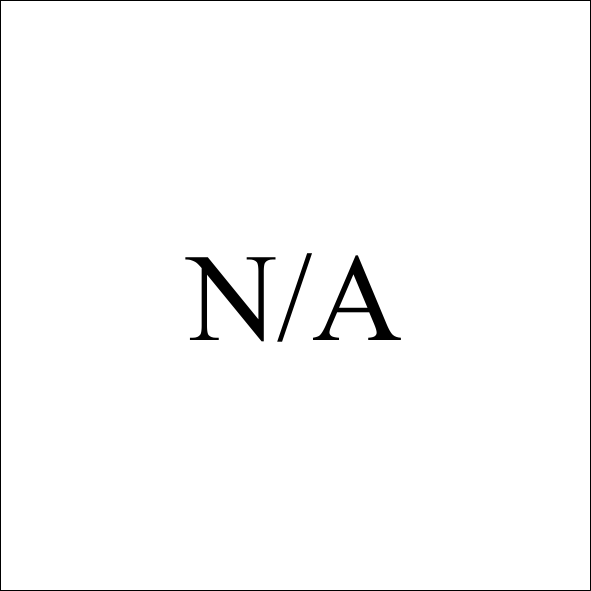}
            \includegraphics[width=0.13\linewidth]{fig/NA.pdf}
            \includegraphics[width=0.13\linewidth]{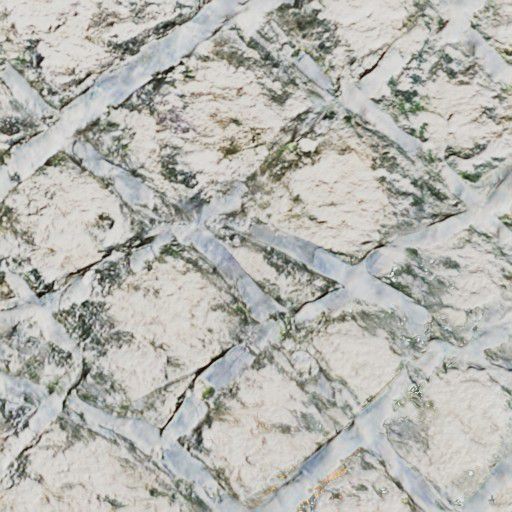}
        \end{minipage}	
    \end{minipage}	

    \begin{minipage}{3.4in}
        \begin{minipage}{0.02in}	
            \centering
            \rotatebox{90}{\parbox{1cm}{\centering\tiny \vspace{0.05cm} Ours}}
        \end{minipage}	
        \hspace{0.02in}
         \begin{minipage}{3.3in}	
            \centering
            \includegraphics[width=0.13\linewidth]{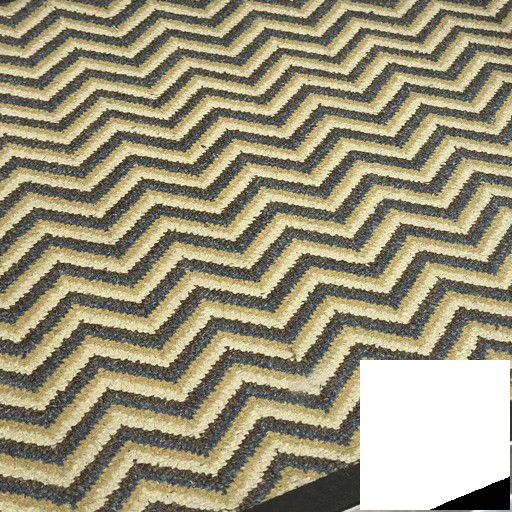}
            \includegraphics[width=0.13\linewidth]{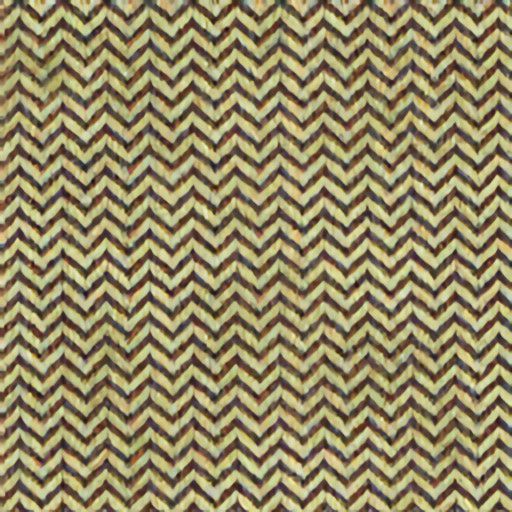}
            \includegraphics[width=0.13\linewidth]{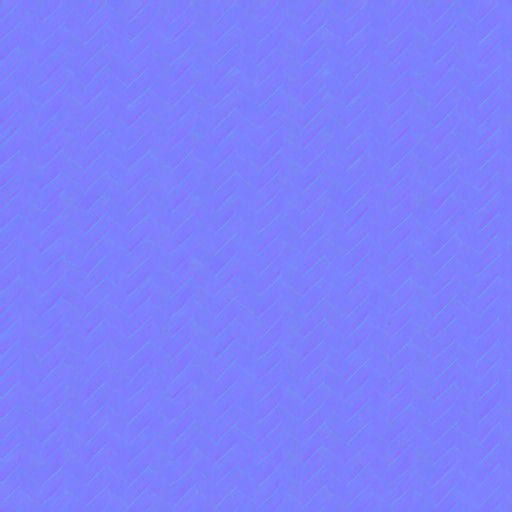}
            \includegraphics[width=0.13\linewidth]{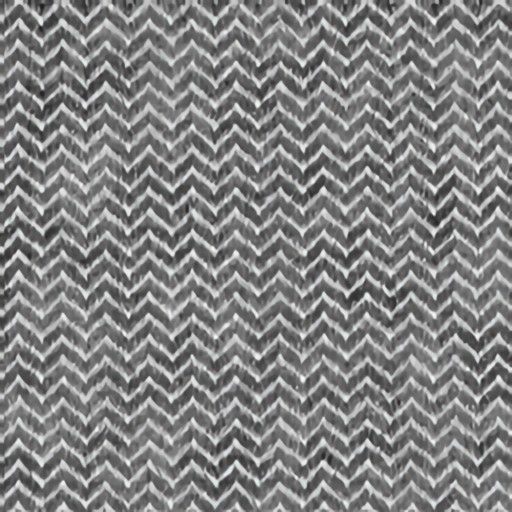}
            \includegraphics[width=0.13\linewidth]{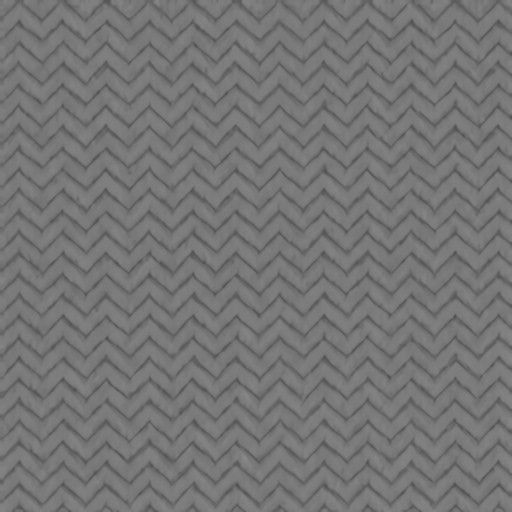}
            \includegraphics[width=0.13\linewidth]{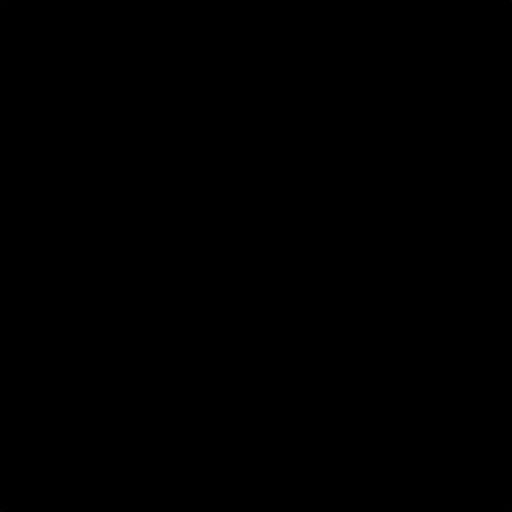}
            \includegraphics[width=0.13\linewidth]{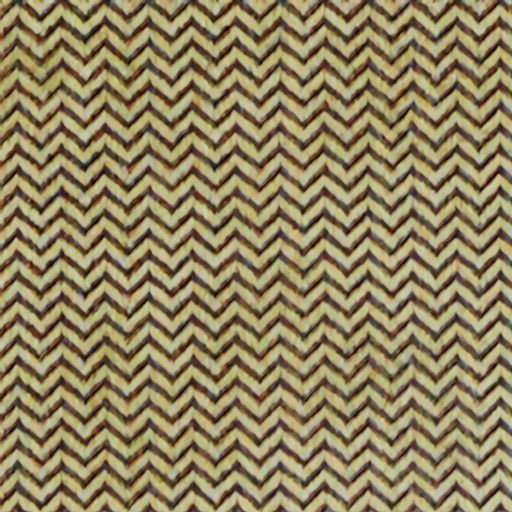}
        \end{minipage}
    \end{minipage}

    \begin{minipage}{3.4in}
        \begin{minipage}{0.02in}	
            \centering
                \rotatebox{90}{\parbox{1cm}{\centering\tiny Material\vspace{-0.05cm}\\Palette}}
        \end{minipage}	
        \hspace{0.02in}
         \begin{minipage}{3.3in}	
            \centering
            \includegraphics[width=0.13\linewidth]{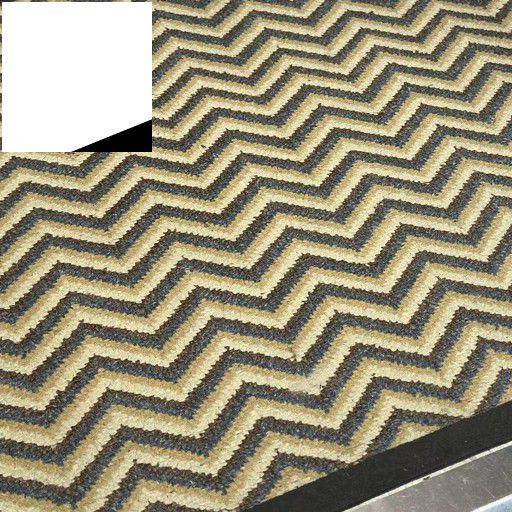}
            \includegraphics[width=0.13\linewidth]{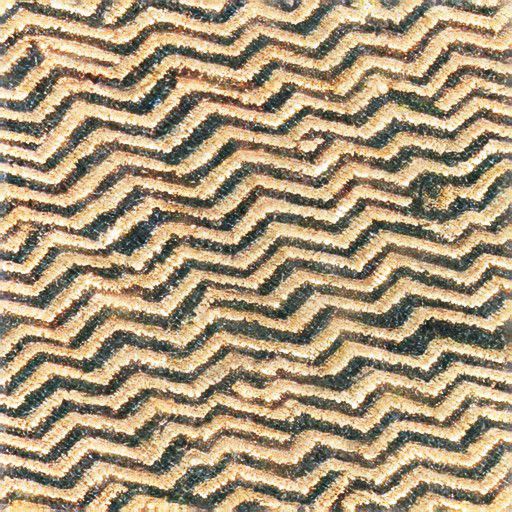}
            \includegraphics[width=0.13\linewidth]{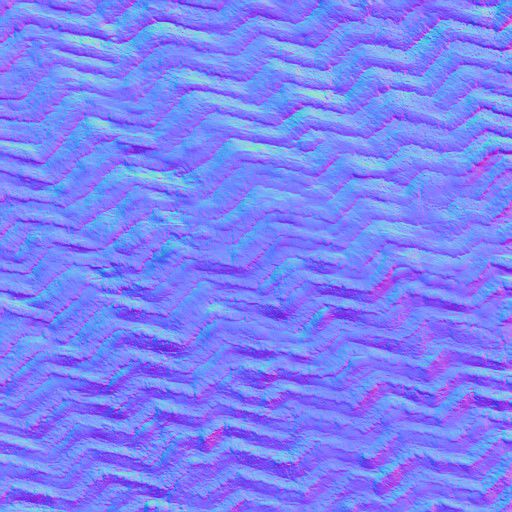}
            \includegraphics[width=0.13\linewidth]{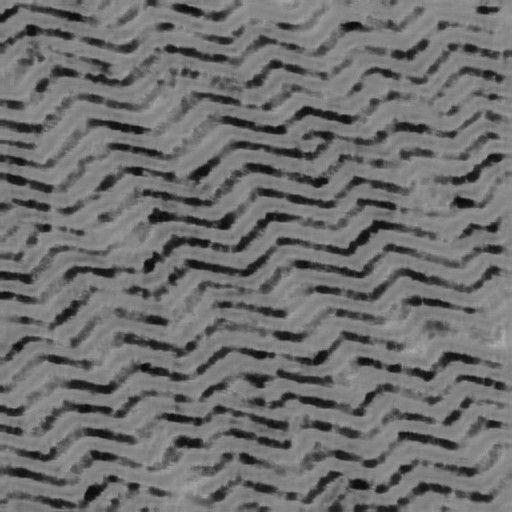}
            \includegraphics[width=0.13\linewidth]{fig/NA.pdf}
            \includegraphics[width=0.13\linewidth]{fig/NA.pdf}
            \includegraphics[width=0.13\linewidth]{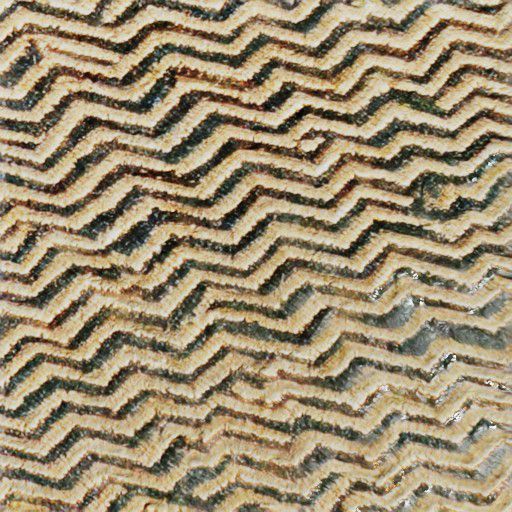}
        \end{minipage}
    \end{minipage}

    \begin{minipage}{3.4in}
        \begin{minipage}{0.02in}	
            \centering
                \rotatebox{90}{\parbox{1cm}{\centering\tiny \vspace{0.05cm} Ours}}
        \end{minipage}	
        \hspace{0.02in}
         \begin{minipage}{3.3in}	
            \centering
            \includegraphics[width=0.13\linewidth]{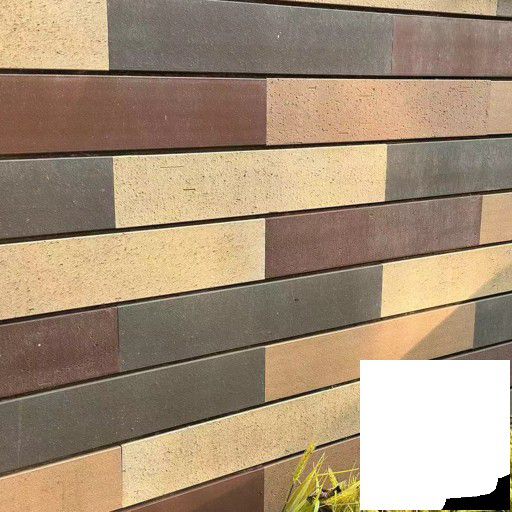}
            \includegraphics[width=0.13\linewidth]{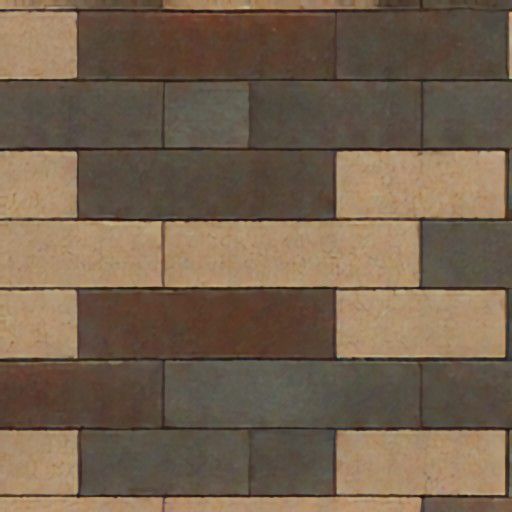}
            \includegraphics[width=0.13\linewidth]{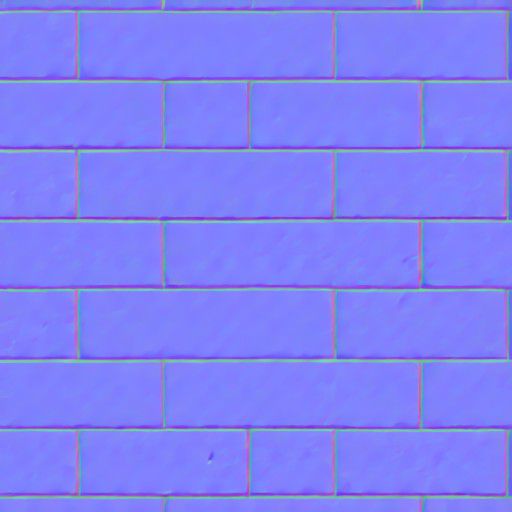}
            \includegraphics[width=0.13\linewidth]{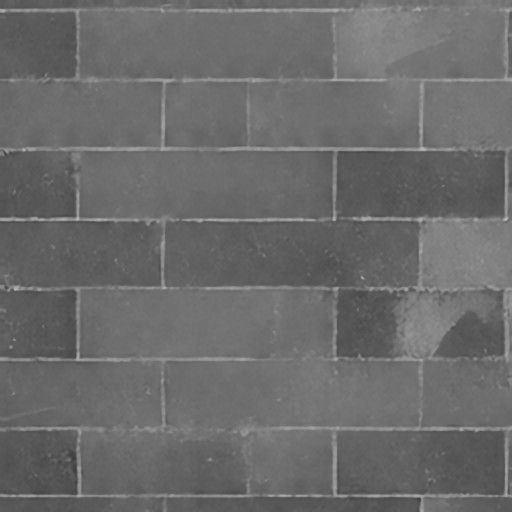}
            \includegraphics[width=0.13\linewidth]{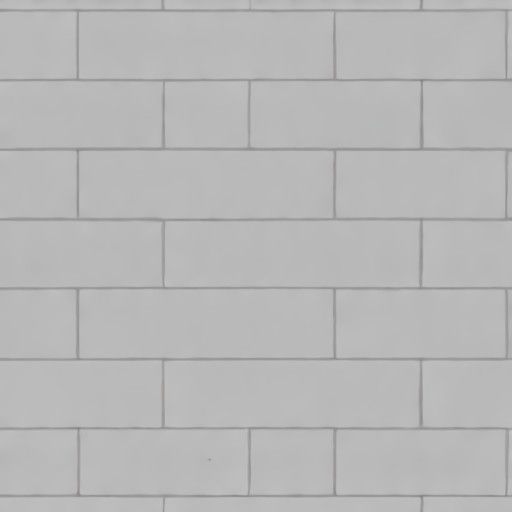}
            \includegraphics[width=0.13\linewidth]{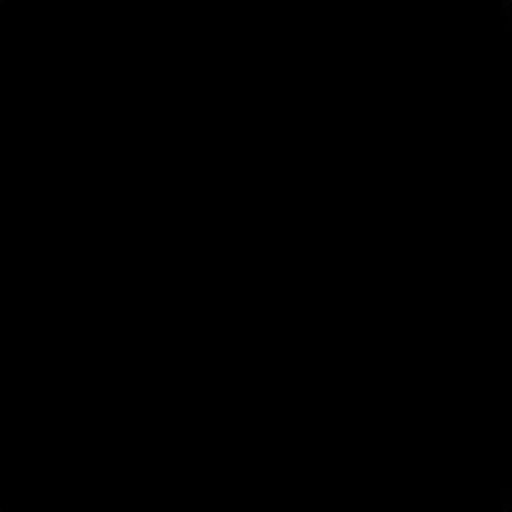}
            \includegraphics[width=0.13\linewidth]{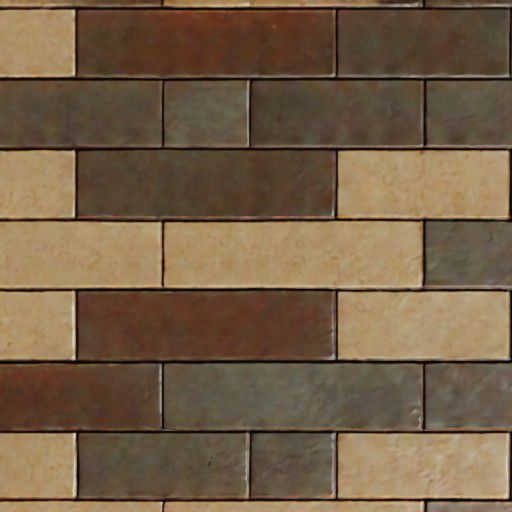}
        \end{minipage}
    \end{minipage}

    \begin{minipage}{3.4in}
        \begin{minipage}{0.02in}	
            \centering
                \rotatebox{90}{\parbox{1cm}{\centering\tiny Material\vspace{-0.05cm}\\Palette}}
        \end{minipage}	
        \hspace{0.02in}
         \begin{minipage}{3.3in}	
            \centering
            \includegraphics[width=0.13\linewidth]{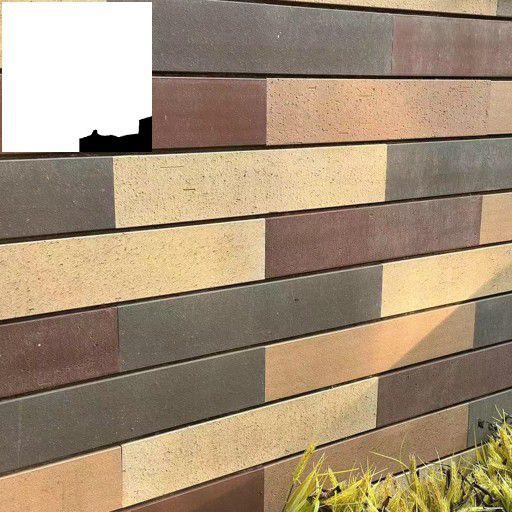}
            \includegraphics[width=0.13\linewidth]{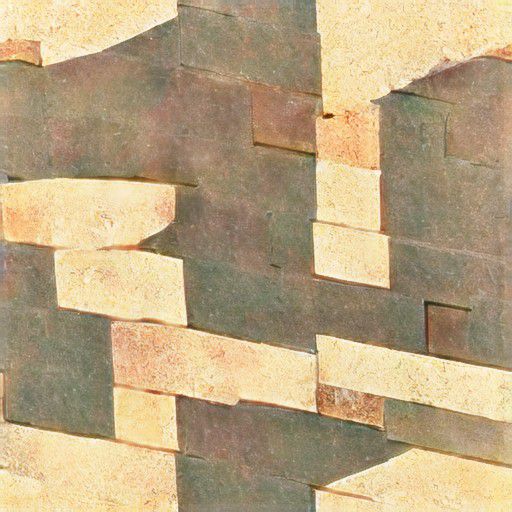}
            \includegraphics[width=0.13\linewidth]{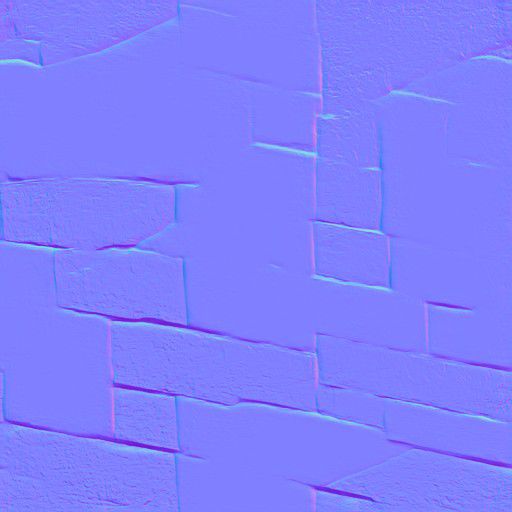}
            \includegraphics[width=0.13\linewidth]{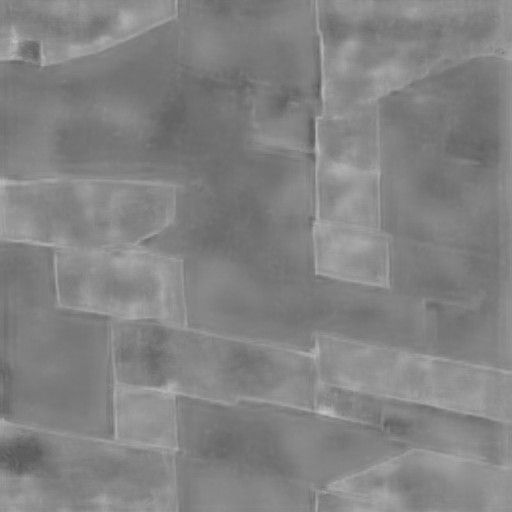}
            \includegraphics[width=0.13\linewidth]{fig/NA.pdf}
            \includegraphics[width=0.13\linewidth]{fig/NA.pdf}
            \includegraphics[width=0.13\linewidth]{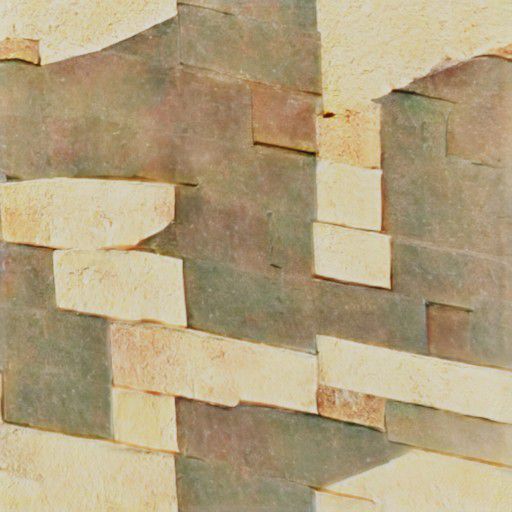}
        \end{minipage}
    \end{minipage}

    \begin{minipage}{3.4in}
        \begin{minipage}{0.02in}	
            \centering
                \rotatebox{90}{\parbox{1cm}{\centering\tiny \vspace{0.05cm} Ours}}
        \end{minipage}	
        \hspace{0.02in}
         \begin{minipage}{3.3in}	
            \centering
            \includegraphics[width=0.13\linewidth]{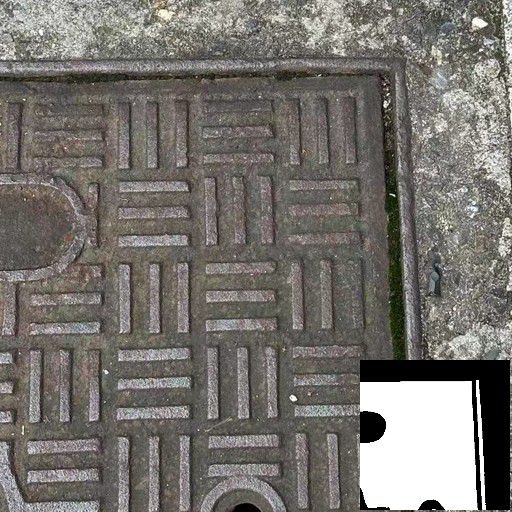}
            \includegraphics[width=0.13\linewidth]{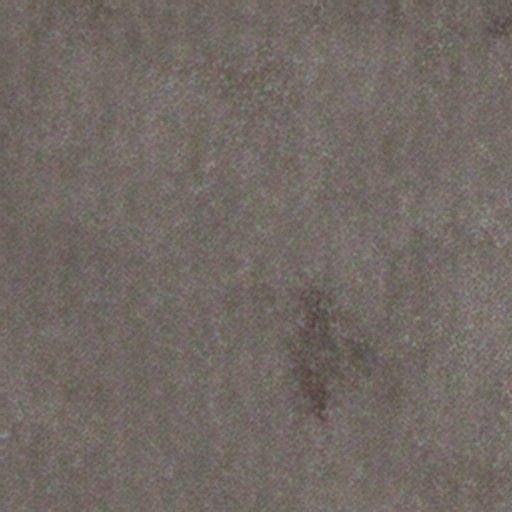}
            \includegraphics[width=0.13\linewidth]{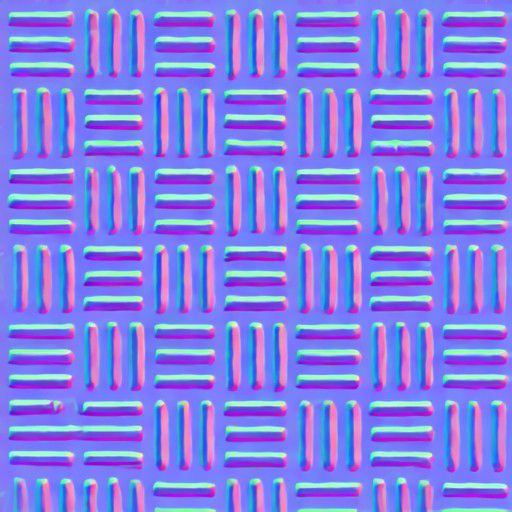}
            \includegraphics[width=0.13\linewidth]{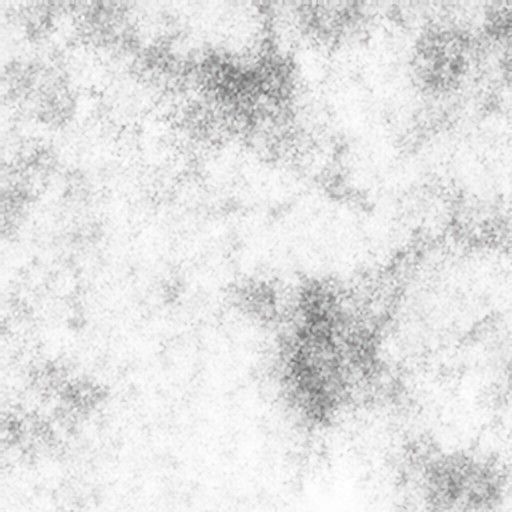}
            \includegraphics[width=0.13\linewidth]{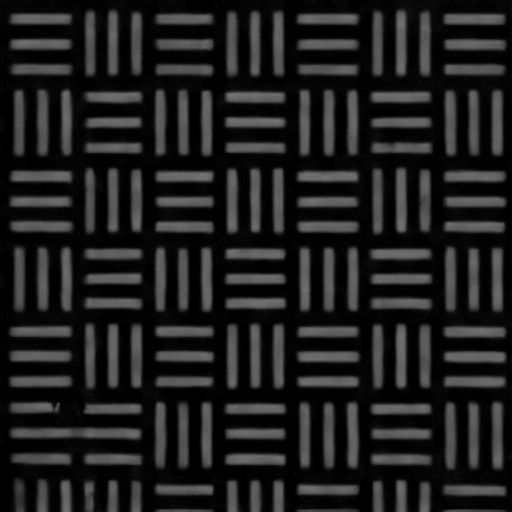}
            \includegraphics[width=0.13\linewidth]{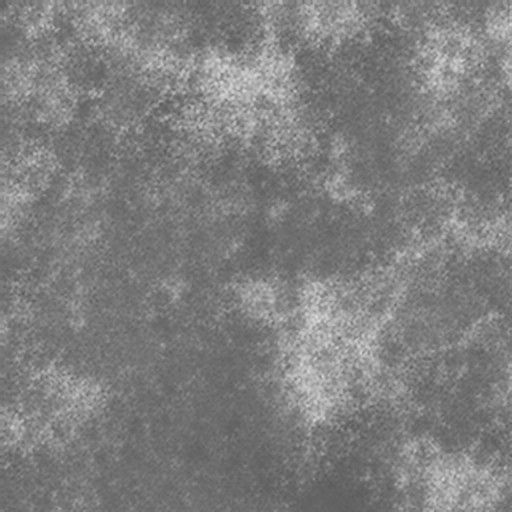}
            \includegraphics[width=0.13\linewidth]{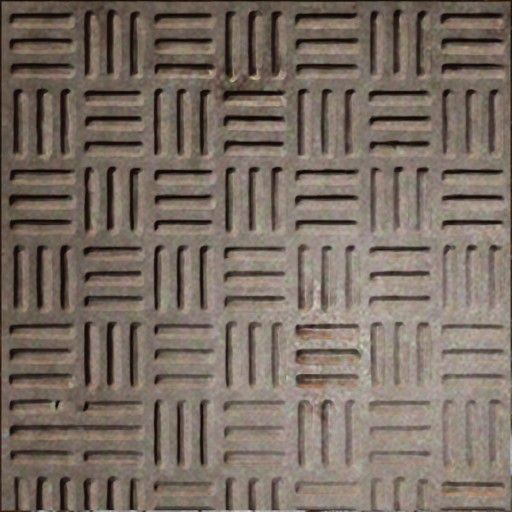}
        \end{minipage}
    \end{minipage}

    \begin{minipage}{3.4in}
        \begin{minipage}{0.02in}	
            \centering
                \rotatebox{90}{\parbox{1cm}{\centering\tiny Material\vspace{-0.05cm}\\Palette}}
        \end{minipage}	
        \hspace{0.02in}
         \begin{minipage}{3.3in}	
            \centering
            \includegraphics[width=0.13\linewidth]{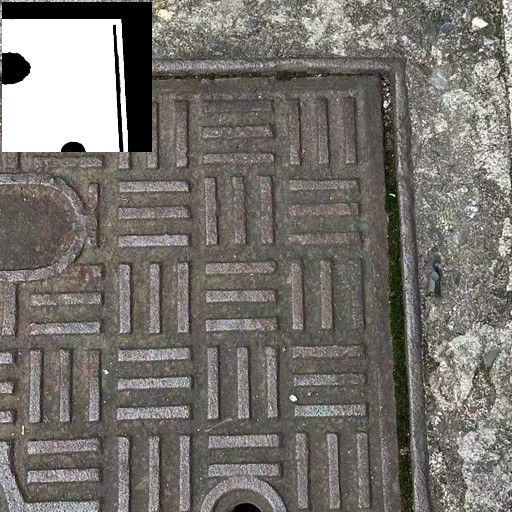}
            \includegraphics[width=0.13\linewidth]{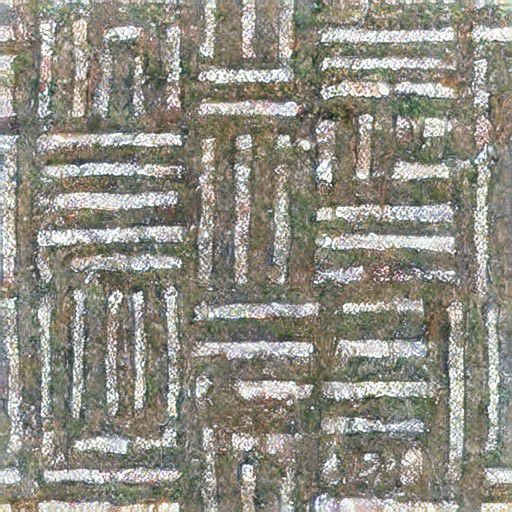}
            \includegraphics[width=0.13\linewidth]{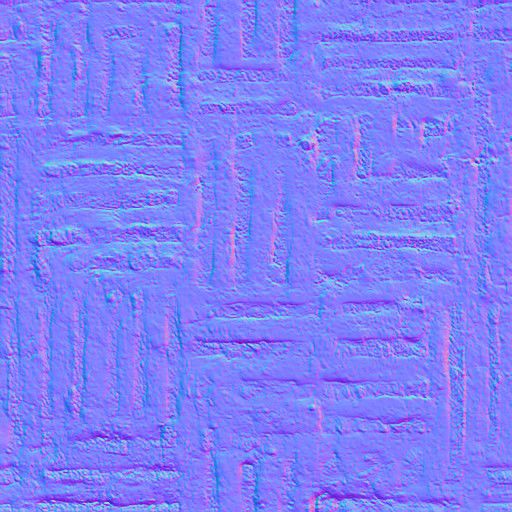}
            \includegraphics[width=0.13\linewidth]{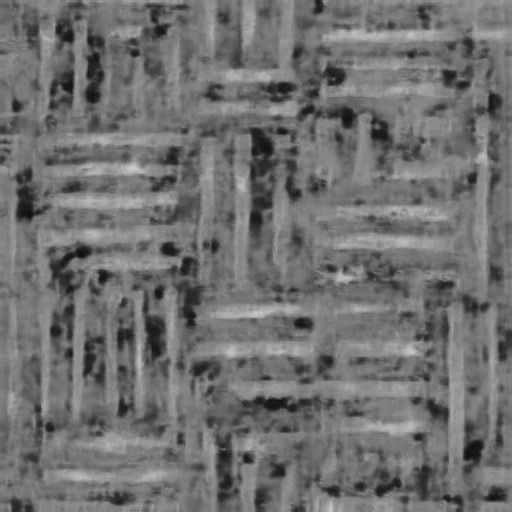}
            \includegraphics[width=0.13\linewidth]{fig/NA.pdf}
            \includegraphics[width=0.13\linewidth]{fig/NA.pdf}
            \includegraphics[width=0.13\linewidth]{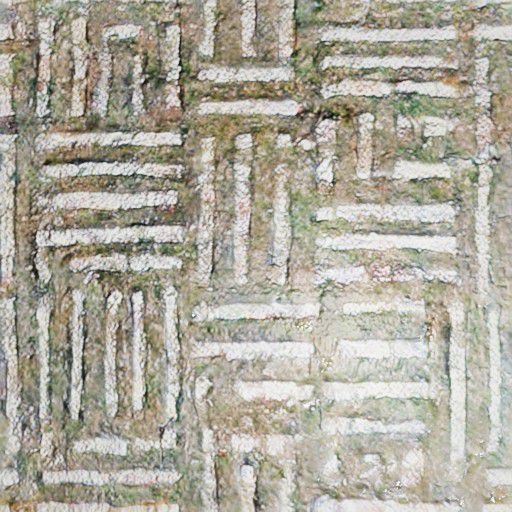}
        \end{minipage}
    \end{minipage}

    \begin{minipage}{3.4in}
        \begin{minipage}{0.02in}	
            \centering
            \rotatebox{90}{\parbox{1cm}{\centering\tiny \vspace{0.05cm} Ours}}
        \end{minipage}	
        \hspace{0.02in}
         \begin{minipage}{3.3in}	
            \centering
            \includegraphics[width=0.13\linewidth]{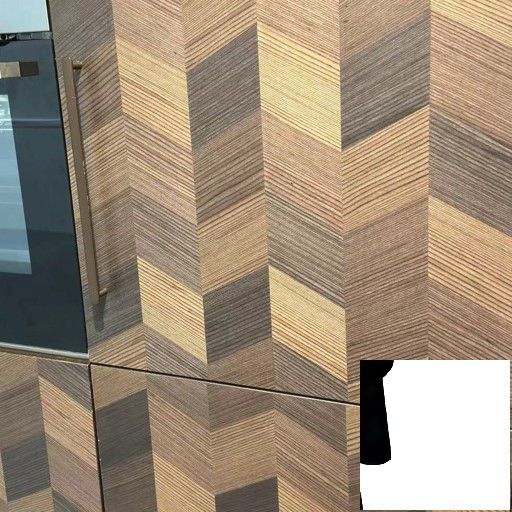}
            \includegraphics[width=0.13\linewidth]{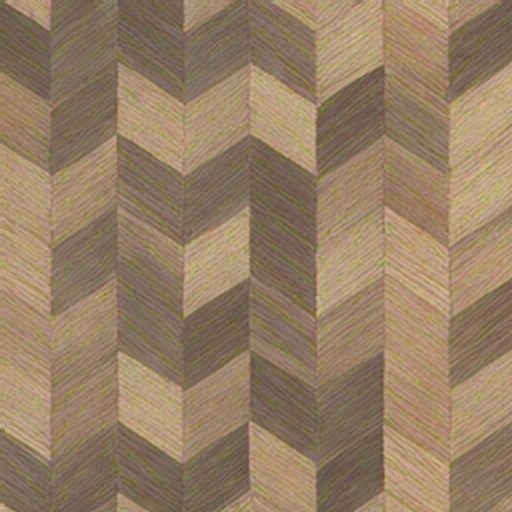}
            \includegraphics[width=0.13\linewidth]{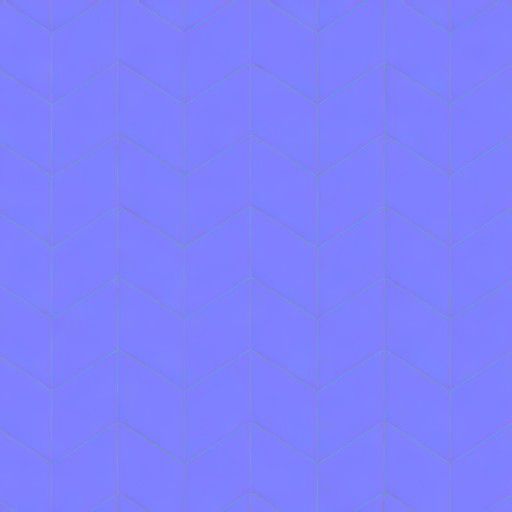}
            \includegraphics[width=0.13\linewidth]{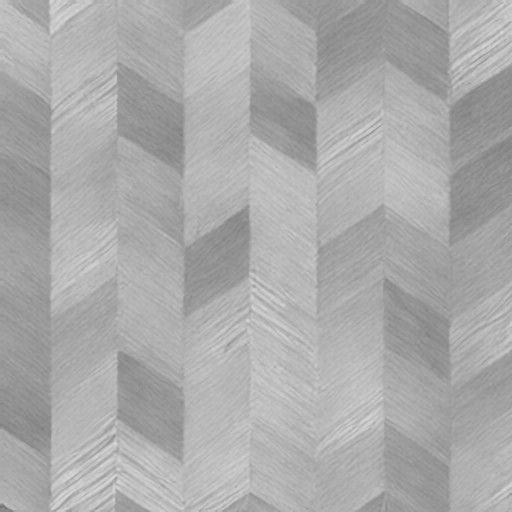}
            \includegraphics[width=0.13\linewidth]{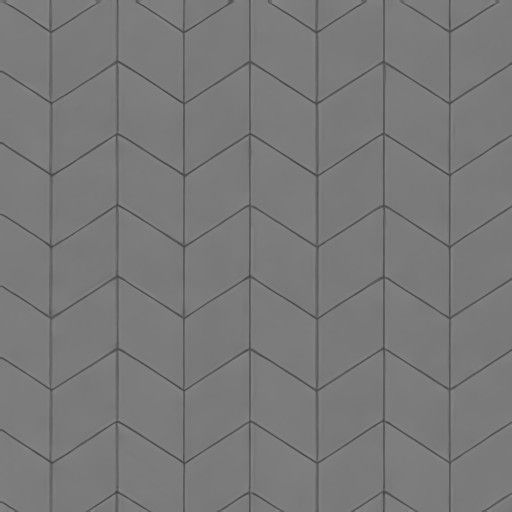}
            \includegraphics[width=0.13\linewidth]{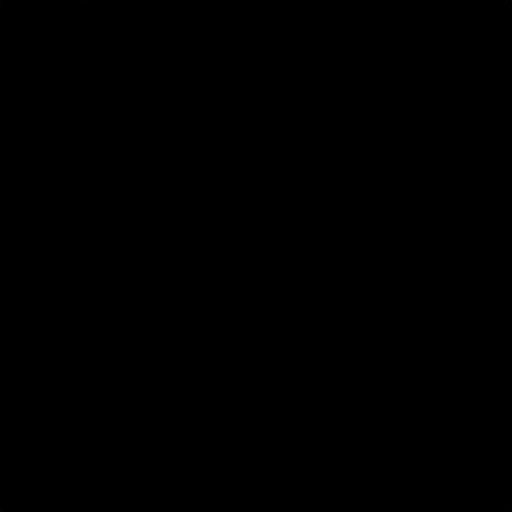}
            \includegraphics[width=0.13\linewidth]{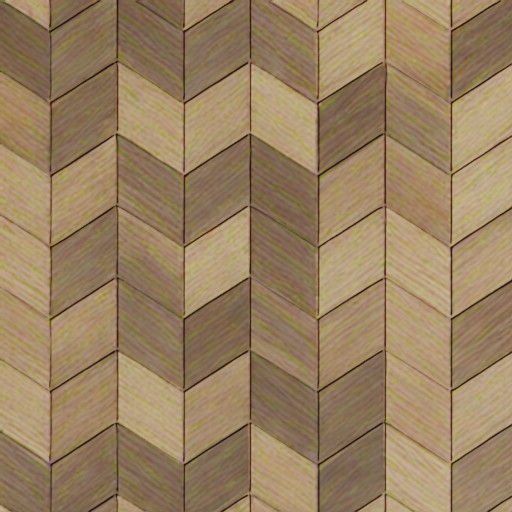}
        \end{minipage}
    \end{minipage}

    \begin{minipage}{3.4in}
        \begin{minipage}{0.02in}	
            \centering
                \rotatebox{90}{\parbox{1cm}{\centering\tiny Material\vspace{-0.05cm}\\Palette}}
        \end{minipage}	
        \hspace{0.02in}
         \begin{minipage}{3.3in}	
            \centering
            \includegraphics[width=0.13\linewidth]{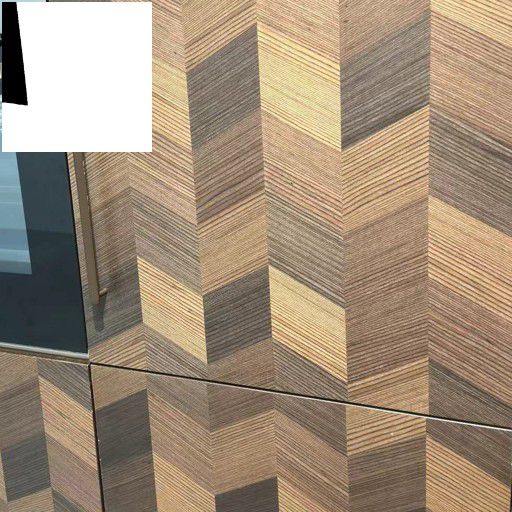}
            \includegraphics[width=0.13\linewidth]{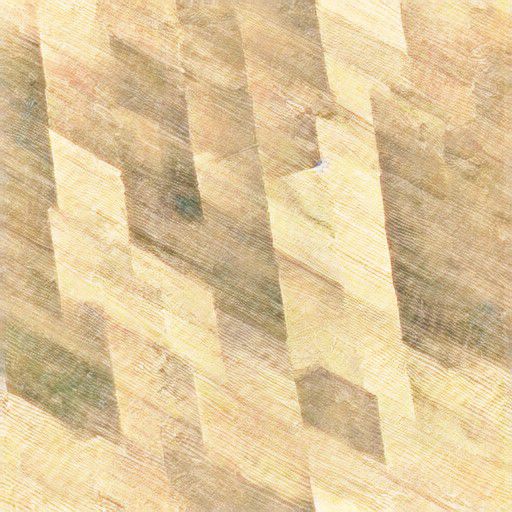}
            \includegraphics[width=0.13\linewidth]{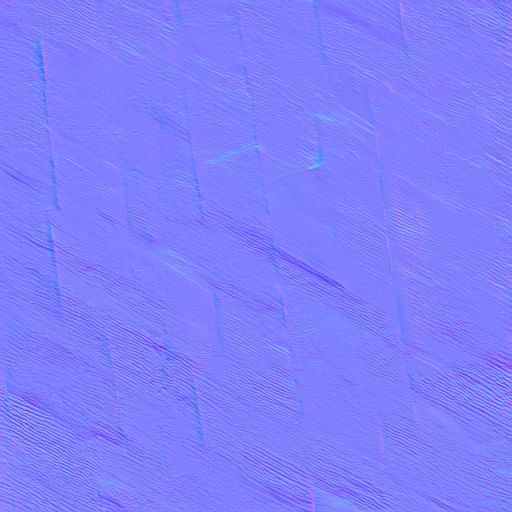}
            \includegraphics[width=0.13\linewidth]{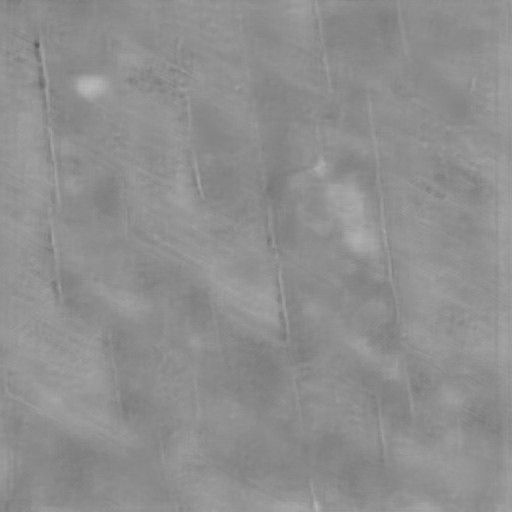}
            \includegraphics[width=0.13\linewidth]{fig/NA.pdf}
            \includegraphics[width=0.13\linewidth]{fig/NA.pdf}
            \includegraphics[width=0.13\linewidth]{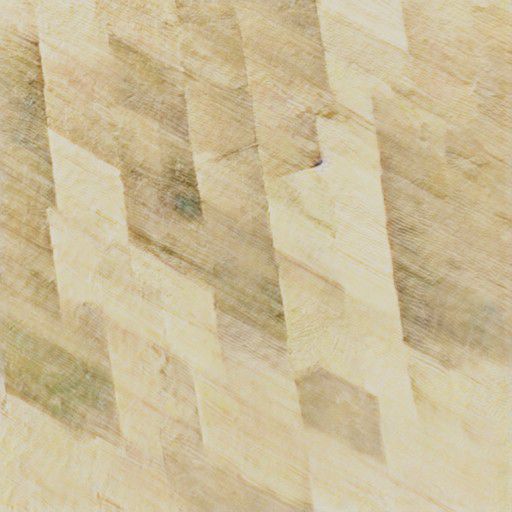}
        \end{minipage}
    \end{minipage}

    \begin{minipage}{3.4in}
        \begin{minipage}{0.02in}	
            \centering
            \rotatebox{90}{\parbox{1cm}{\centering\tiny \vspace{0.05cm} Ours}}
        \end{minipage}	
        \hspace{0.02in}
         \begin{minipage}{3.3in}	
            \centering
            \includegraphics[width=0.13\linewidth]{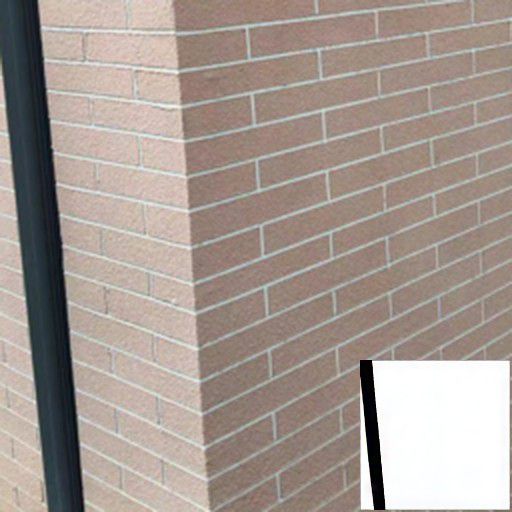}
            \includegraphics[width=0.13\linewidth]{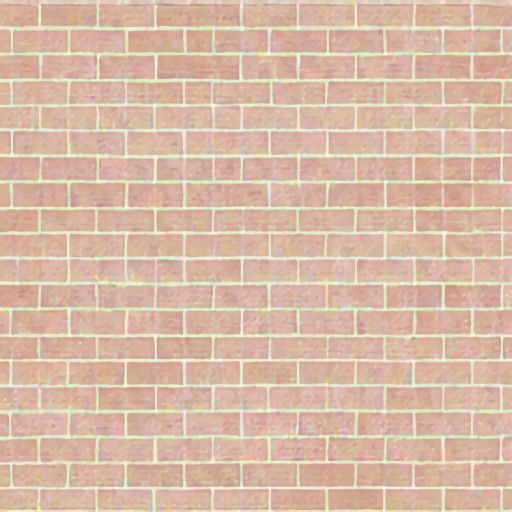}
            \includegraphics[width=0.13\linewidth]{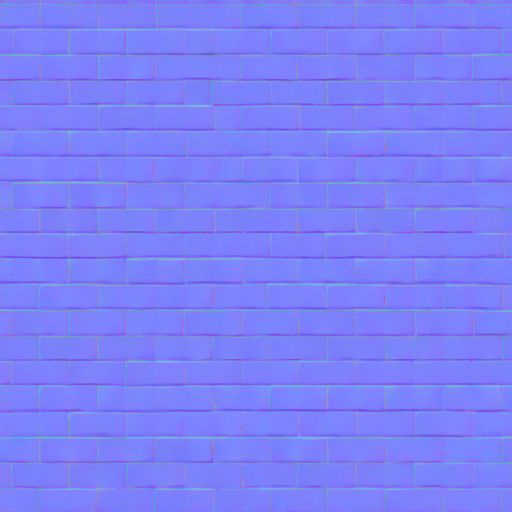}
            \includegraphics[width=0.13\linewidth]{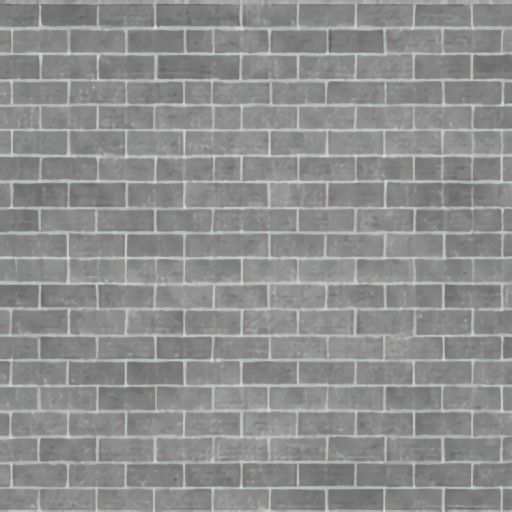}
            \includegraphics[width=0.13\linewidth]{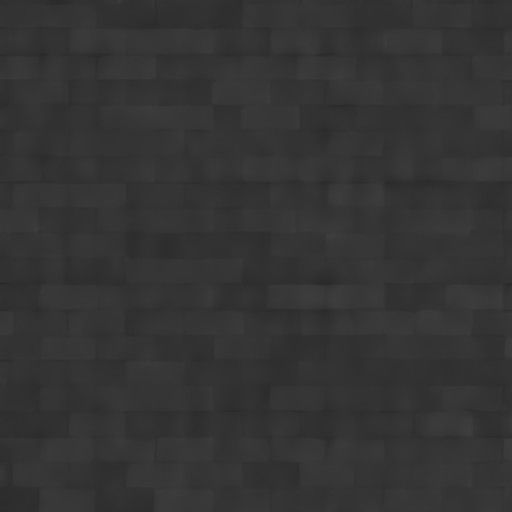}
            \includegraphics[width=0.13\linewidth]{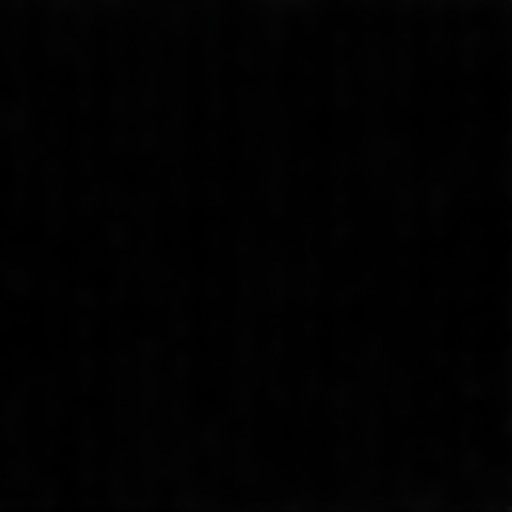}
            \includegraphics[width=0.13\linewidth]{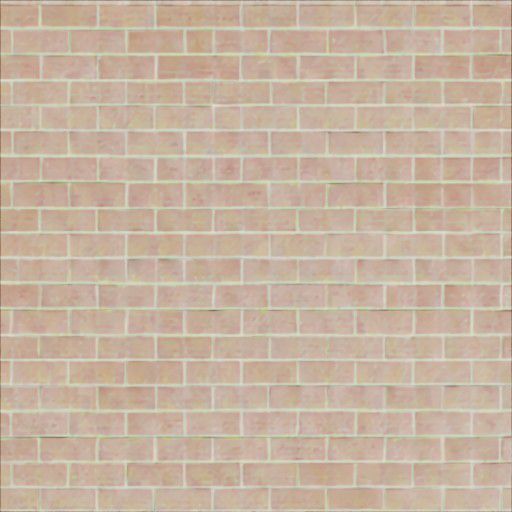}
        \end{minipage}
    \end{minipage}

    \begin{minipage}{3.4in}
        \begin{minipage}{0.02in}	
            \centering
                \rotatebox{90}{\parbox{1cm}{\centering\tiny Material\vspace{-0.05cm}\\Palette}}
        \end{minipage}	
        \hspace{0.02in}
         \begin{minipage}{3.3in}	
            \centering
            \includegraphics[width=0.13\linewidth]{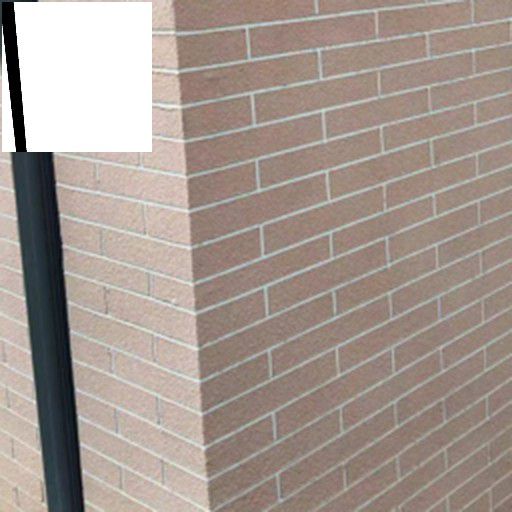}
            \includegraphics[width=0.13\linewidth]{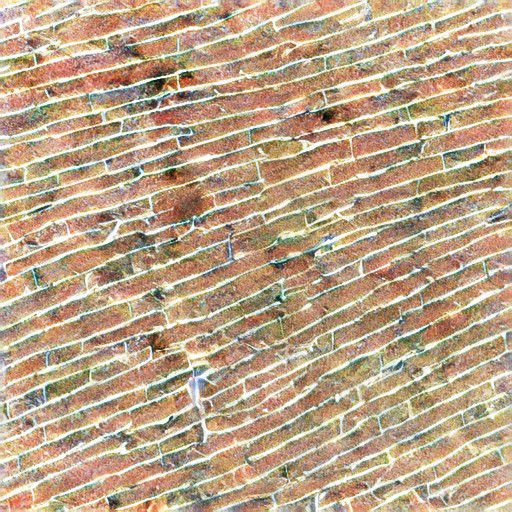}
            \includegraphics[width=0.13\linewidth]{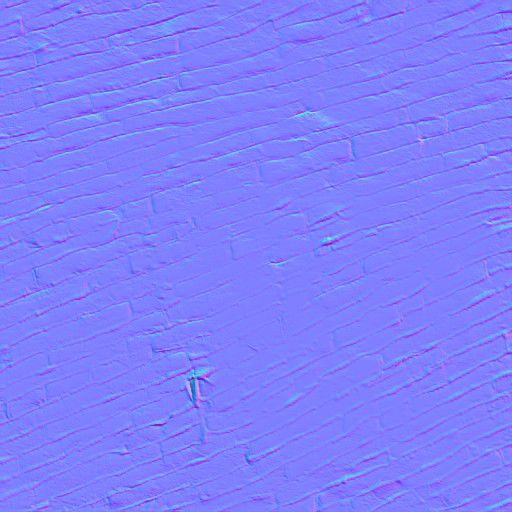}
            \includegraphics[width=0.13\linewidth]{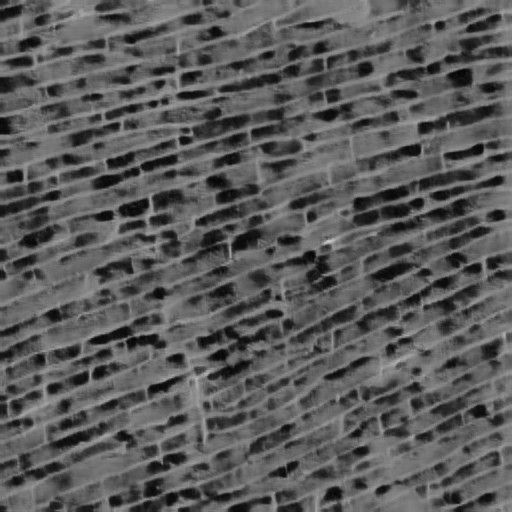}
            \includegraphics[width=0.13\linewidth]{fig/NA.pdf}
            \includegraphics[width=0.13\linewidth]{fig/NA.pdf}
            \includegraphics[width=0.13\linewidth]{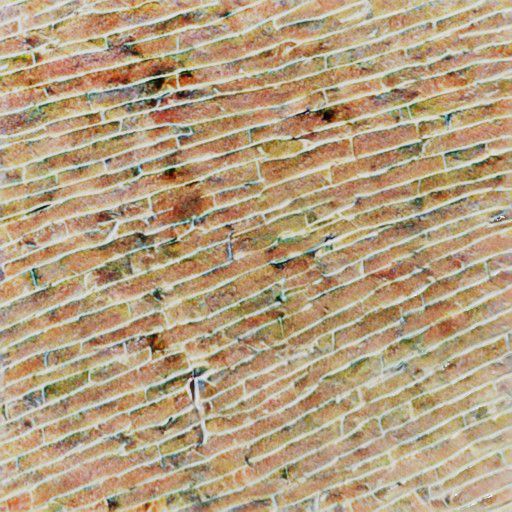}
        \end{minipage}
    \end{minipage}
    
   \caption{Comparison between our method and Material Palette~\cite{lopes2024material} on material extraction for real photographs. The first column shows the input images and the generated (ours bottom-right)/provided (Material Palette top-left) masks. The second to sixth columns show the generated material maps and rendering under an environment map. We see that our approach better corrects for distortion and match the original appearance.}
   \label{fig:material_genertaion_real}
\end{figure}

We show qualitative evaluation on real photographs in Fig.~\ref{fig:material_genertaion_real} where we see that our model generalizes well to photographs of materials from various angles. We render the generated materials on a planar surface under environment lighting, showing strong visual similarity to the original input images. Unlike Material Palette, which requires input masks from a separate segmentation step~\cite{Sharma2023materialistic}, our model operates out-of-box with an input image only, showcasing its potential as a lightweight \textit{\method}. 

\begin{figure}[htbp]
    \centering		
    \begin{minipage}{3.4in}
        \begin{minipage}{0.02in}	
            \centering
                \vspace{0.1in}
                \rotatebox{90}{\tiny Ours}
        \end{minipage}	
        \hspace{0.02in}
         \begin{minipage}{3.3in}	
            \centering
            \begin{minipage}{0.13\linewidth}
                \subcaption*{\tiny Input}
                \includegraphics[width=\linewidth]{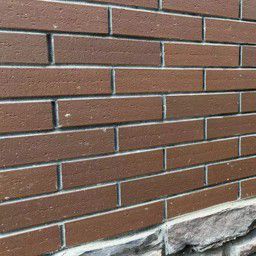}
            \end{minipage}
            \begin{minipage}{0.13\linewidth}
                \subcaption*{\tiny Albedo}
                \includegraphics[width=\linewidth]{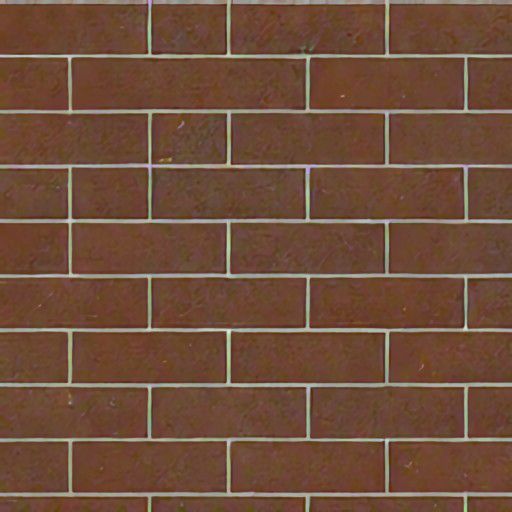}
            \end{minipage}
            \begin{minipage}{0.13\linewidth}
                \subcaption*{\tiny Normal}
                \includegraphics[width=\linewidth]{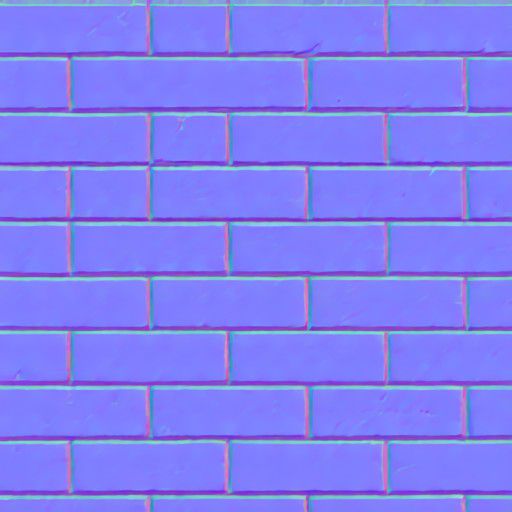}
            \end{minipage}
            \begin{minipage}{0.13\linewidth}
                \subcaption*{\tiny Roughness}
                \includegraphics[width=\linewidth]{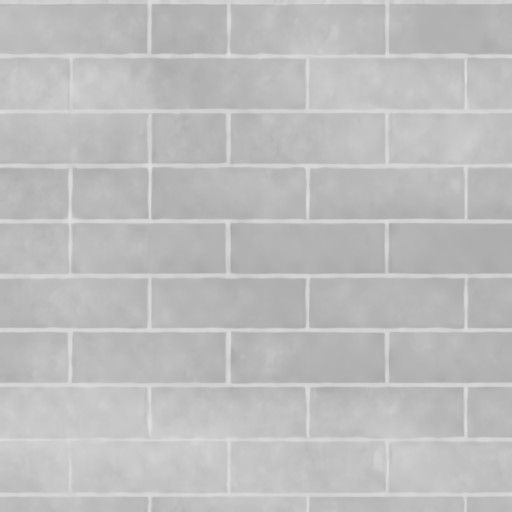}
            \end{minipage}
            \begin{minipage}{0.13\linewidth}
                \subcaption*{\tiny Height}
                \includegraphics[width=\linewidth]{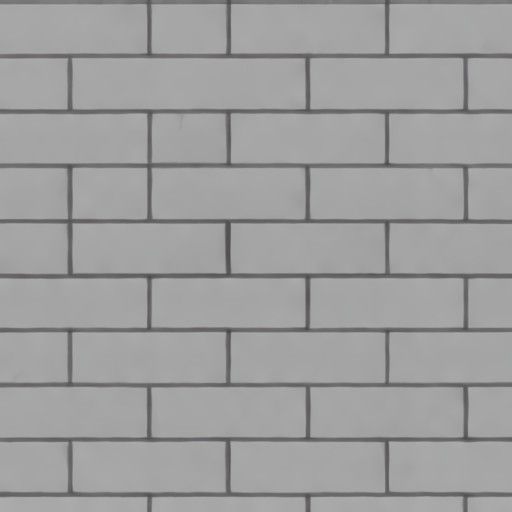}
            \end{minipage}
            \begin{minipage}{0.13\linewidth}
                \subcaption*{\tiny Metallic}
                \includegraphics[width=\linewidth]{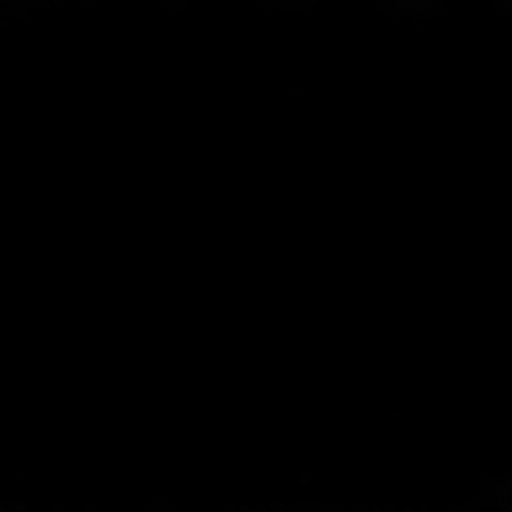}
            \end{minipage}
            \begin{minipage}{0.13\linewidth}
                \subcaption*{\tiny Render}
                \includegraphics[width=\linewidth]{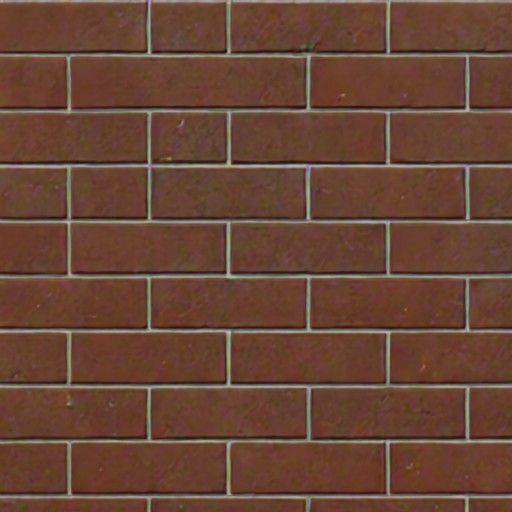}
            \end{minipage}
        \end{minipage}	
        \end{minipage}	

    \begin{minipage}{3.4in}
        \begin{minipage}{0.02in}	
            \centering
                \rotatebox{90}{\tiny ControlMat}
        \end{minipage}	
        \hspace{0.02in}
         \begin{minipage}{3.3in}	
            \centering
            \begin{minipage}{0.13\linewidth}
            \includegraphics[width=\linewidth]{fig/compare_controlmat/0_input.jpg}
            \end{minipage}	
            \begin{minipage}{0.13\linewidth}
            \includegraphics[width=\linewidth]{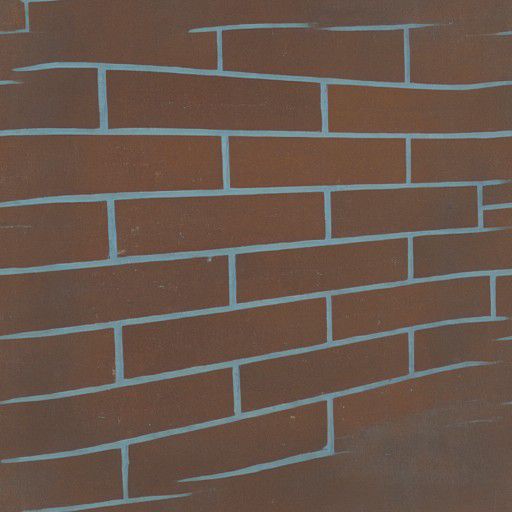}
            \end{minipage}	
            \begin{minipage}{0.13\linewidth}
            \includegraphics[width=\linewidth]{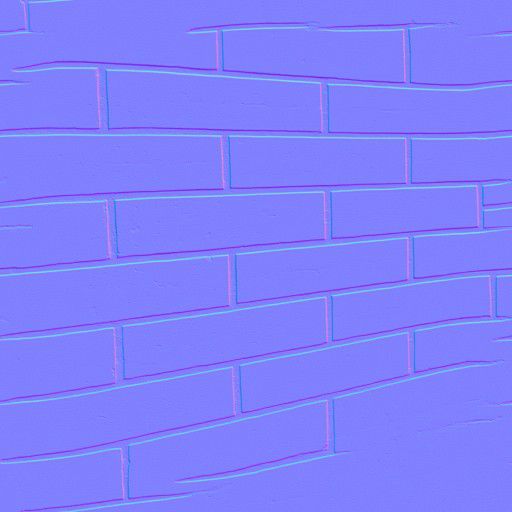}
            \end{minipage}	
            \begin{minipage}{0.13\linewidth}
            \includegraphics[width=\linewidth]{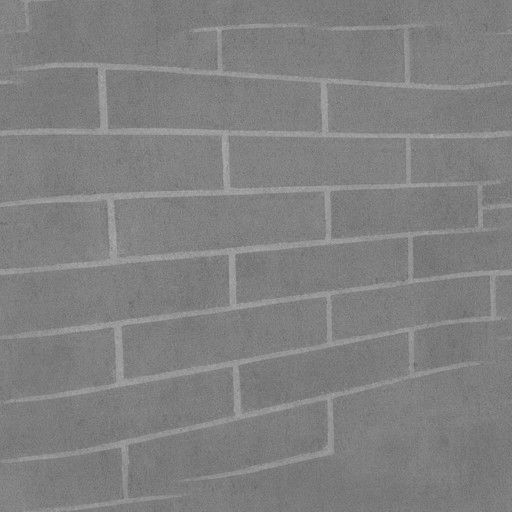}
            \end{minipage}	
            \begin{minipage}{0.13\linewidth}
            \includegraphics[width=\linewidth]{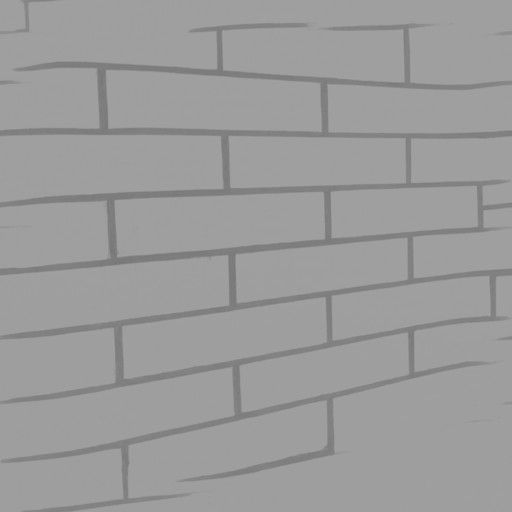}
            \end{minipage}	
            \begin{minipage}{0.13\linewidth}
            \includegraphics[width=\linewidth]{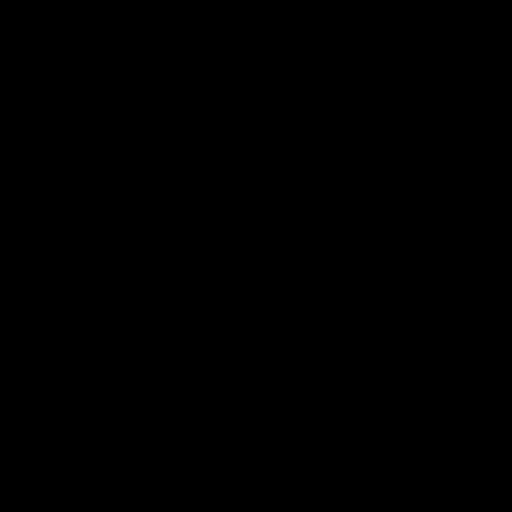}
            \end{minipage}	
            \begin{minipage}{0.13\linewidth}
            \includegraphics[width=\linewidth]{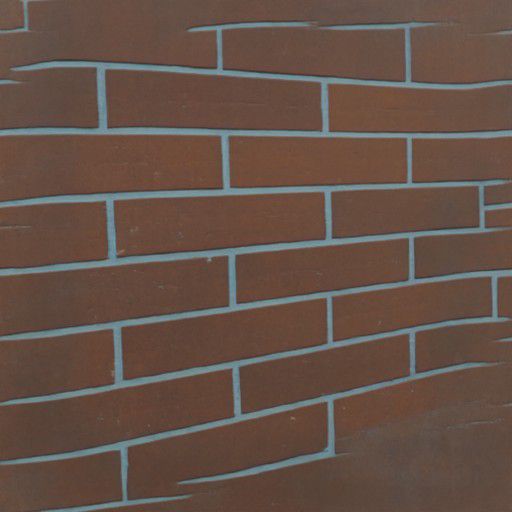}
            \end{minipage}	
        \end{minipage}	
    \end{minipage}	

    \begin{minipage}{3.4in}
        \begin{minipage}{0.02in}	
            \centering
                \rotatebox{90}{\tiny Ours}
        \end{minipage}	
        \hspace{0.02in}
         \begin{minipage}{3.3in}	
            \centering
            \begin{minipage}{0.13\linewidth}
            \includegraphics[width=\linewidth]{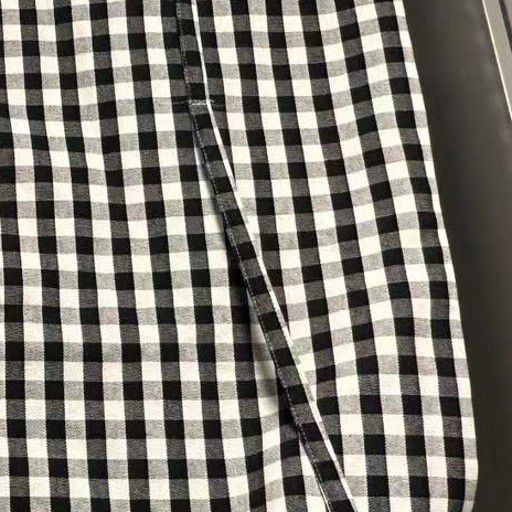}
            \end{minipage}	
            \begin{minipage}{0.13\linewidth}
            \includegraphics[width=\linewidth]{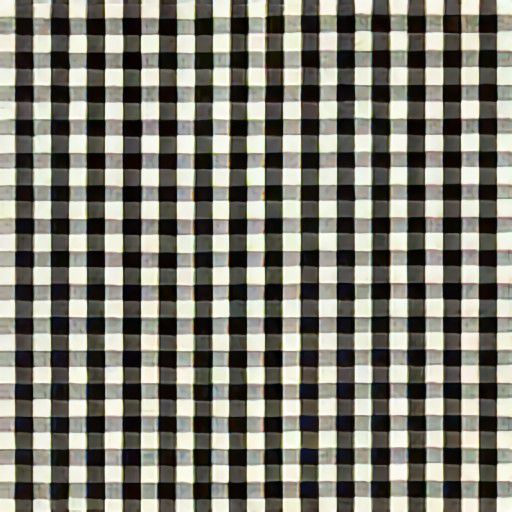}
            \end{minipage}	
            \begin{minipage}{0.13\linewidth}
            \includegraphics[width=\linewidth]{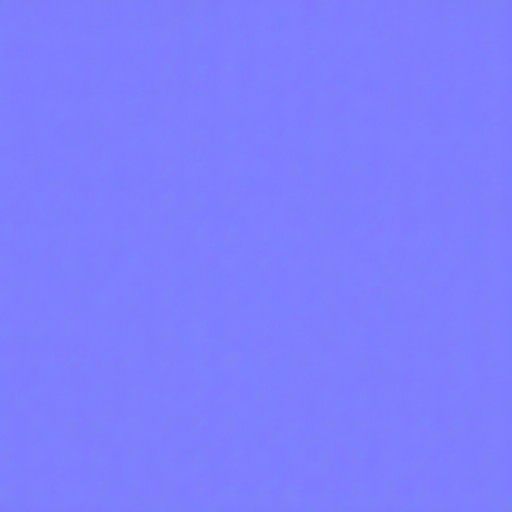}
            \end{minipage}	
            \begin{minipage}{0.13\linewidth}
            \includegraphics[width=\linewidth]{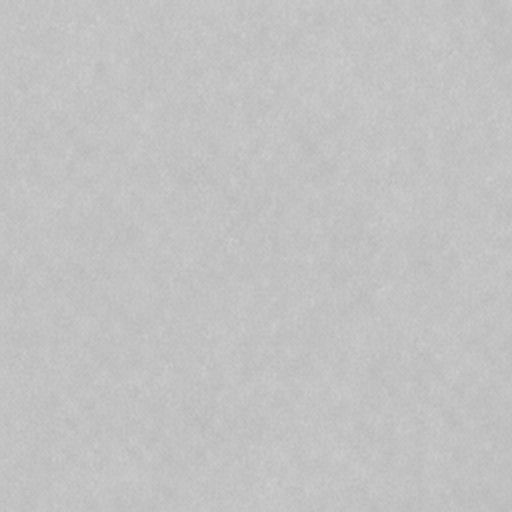}
            \end{minipage}	
            \begin{minipage}{0.13\linewidth}
            \includegraphics[width=\linewidth]{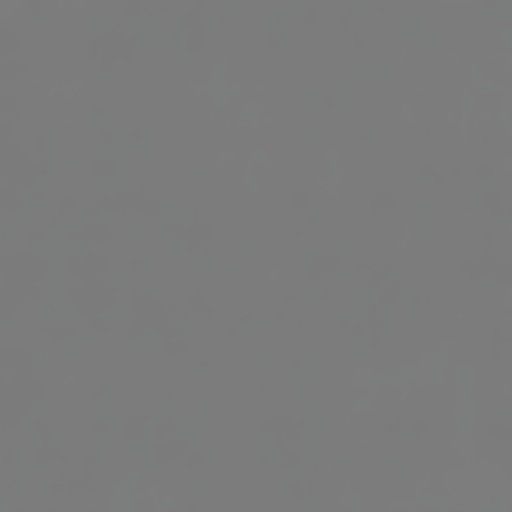}
            \end{minipage}	
            \begin{minipage}{0.13\linewidth}
            \includegraphics[width=\linewidth]{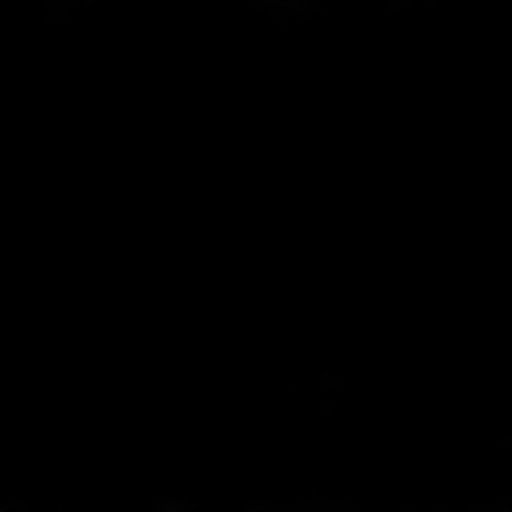}
            \end{minipage}	
            \begin{minipage}{0.13\linewidth}
            \includegraphics[width=\linewidth]{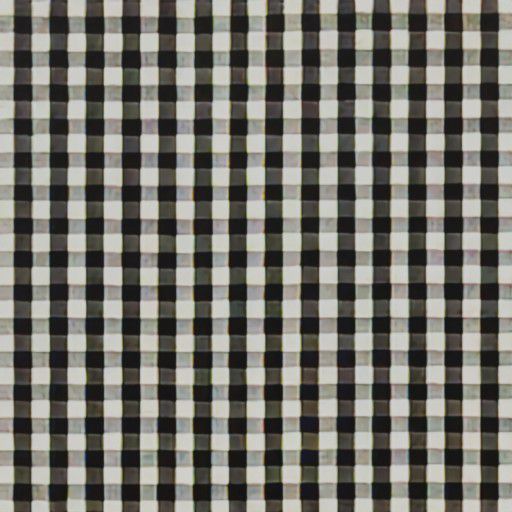}
            \end{minipage}	
        \end{minipage}	
    \end{minipage}	

    \begin{minipage}{3.4in}
        \begin{minipage}{0.02in}	
            \centering
                \rotatebox{90}{\tiny ControlMat}
        \end{minipage}	
        \hspace{0.02in}
         \begin{minipage}{3.3in}	
            \centering
            \begin{minipage}{0.13\linewidth}
            \includegraphics[width=\linewidth]{fig/compare_controlmat/1_input.jpg}
            \end{minipage}	
            \begin{minipage}{0.13\linewidth}
            \includegraphics[width=\linewidth]{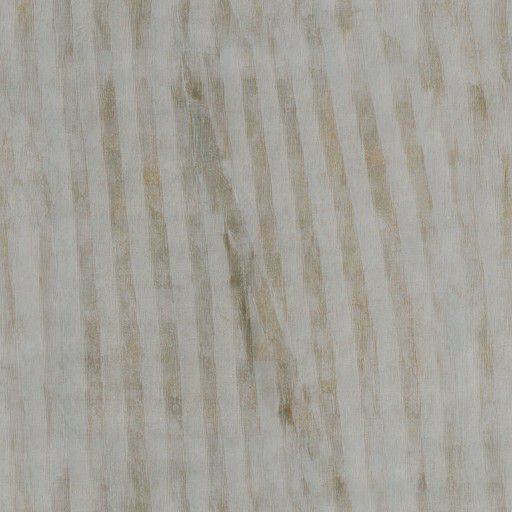}
            \end{minipage}	
            \begin{minipage}{0.13\linewidth}
            \includegraphics[width=\linewidth]{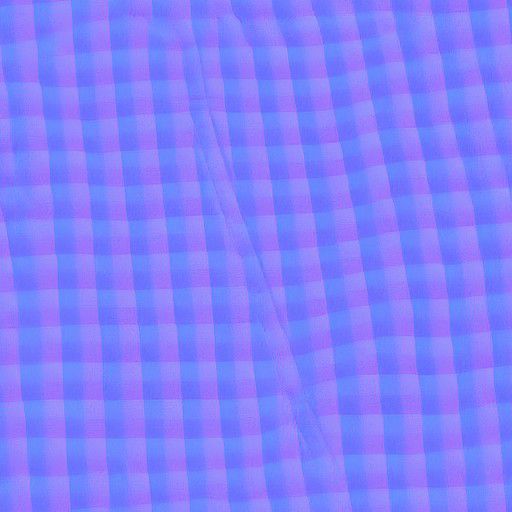}
            \end{minipage}	
            \begin{minipage}{0.13\linewidth}
            \includegraphics[width=\linewidth]{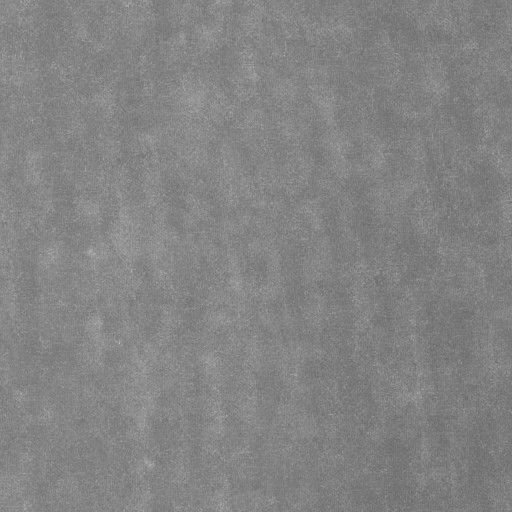}
            \end{minipage}	
            \begin{minipage}{0.13\linewidth}
            \includegraphics[width=\linewidth]{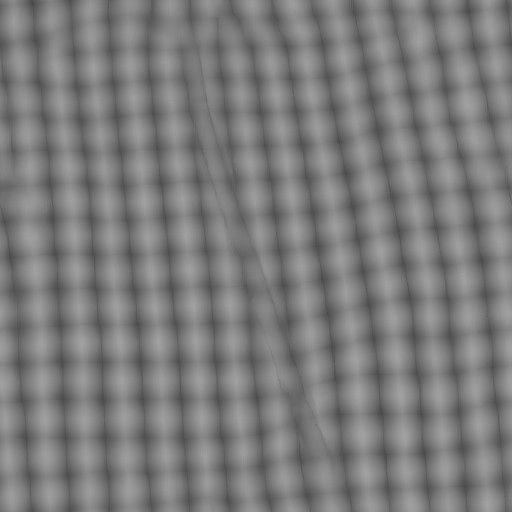}
            \end{minipage}	
            \begin{minipage}{0.13\linewidth}
            \includegraphics[width=\linewidth]{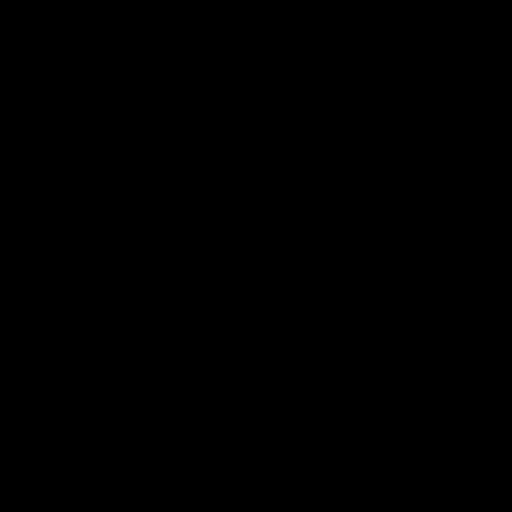}
            \end{minipage}	
            \begin{minipage}{0.13\linewidth}
            \includegraphics[width=\linewidth]{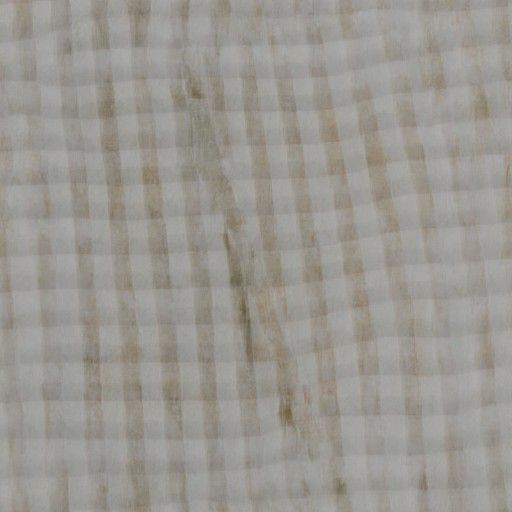}
            \end{minipage}	
        \end{minipage}	
    \end{minipage}

    \begin{minipage}{3.4in}
        \begin{minipage}{0.02in}	
            \centering
                \rotatebox{90}{\tiny Ours}
        \end{minipage}	
        \hspace{0.02in}
         \begin{minipage}{3.3in}	
            \centering
            \begin{minipage}{0.13\linewidth}
            \includegraphics[width=\linewidth]{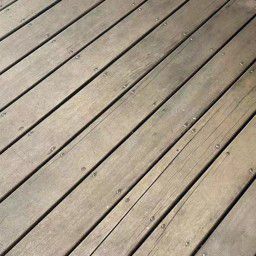}
            \end{minipage}	
            \begin{minipage}{0.13\linewidth}
            \includegraphics[width=\linewidth]{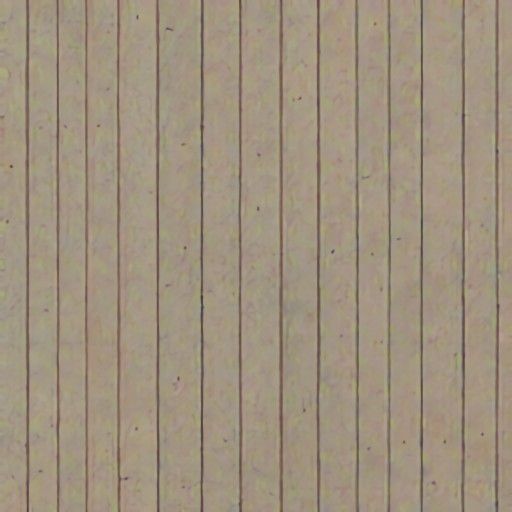}
            \end{minipage}	
            \begin{minipage}{0.13\linewidth}
            \includegraphics[width=\linewidth]{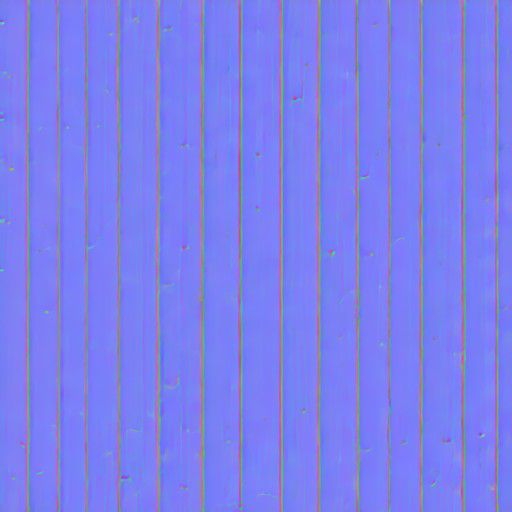}
            \end{minipage}	
            \begin{minipage}{0.13\linewidth}
            \includegraphics[width=\linewidth]{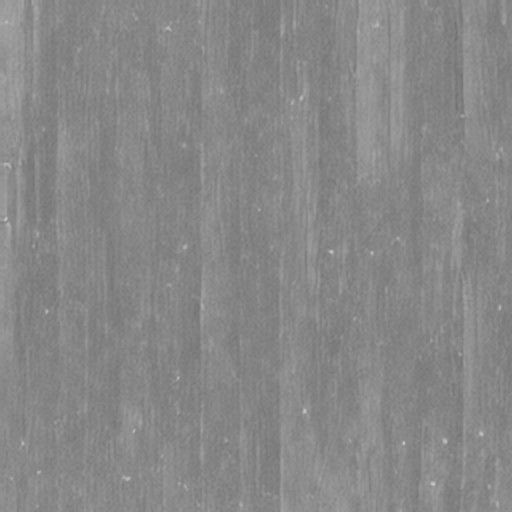}
            \end{minipage}	
            \begin{minipage}{0.13\linewidth}
            \includegraphics[width=\linewidth]{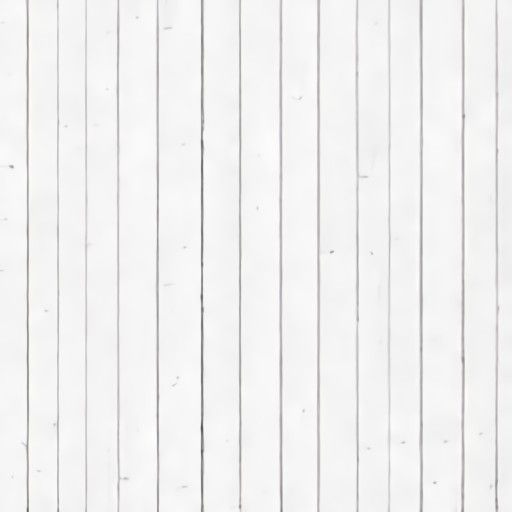}
            \end{minipage}	
            \begin{minipage}{0.13\linewidth}
            \includegraphics[width=\linewidth]{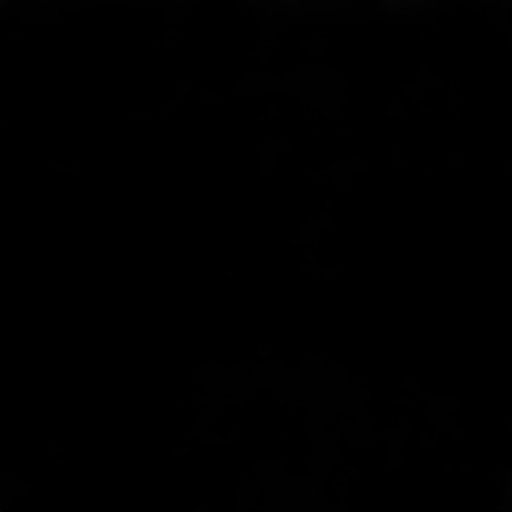}
            \end{minipage}	
            \begin{minipage}{0.13\linewidth}
            \includegraphics[width=\linewidth]{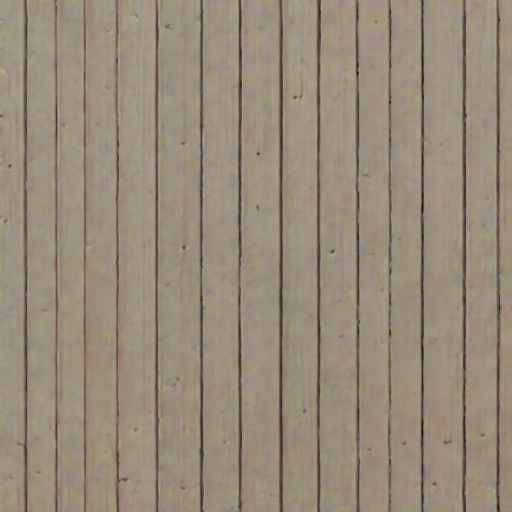}
            \end{minipage}	
        \end{minipage}	
    \end{minipage}	

    \begin{minipage}{3.4in}
        \begin{minipage}{0.02in}	
            \centering
                \rotatebox{90}{\tiny ControlMat}
        \end{minipage}	
        \hspace{0.02in}
         \begin{minipage}{3.3in}	
            \centering
            \begin{minipage}{0.13\linewidth}
            \includegraphics[width=\linewidth]{fig/compare_controlmat/62_input.jpg}
            \end{minipage}	
            \begin{minipage}{0.13\linewidth}
            \includegraphics[width=\linewidth]{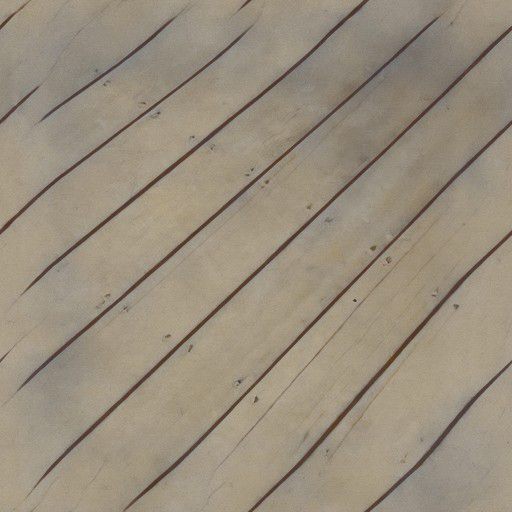}
            \end{minipage}	
            \begin{minipage}{0.13\linewidth}
            \includegraphics[width=\linewidth]{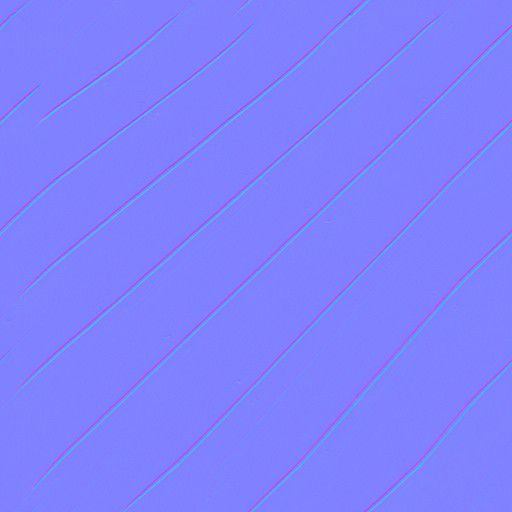}
            \end{minipage}	
            \begin{minipage}{0.13\linewidth}
            \includegraphics[width=\linewidth]{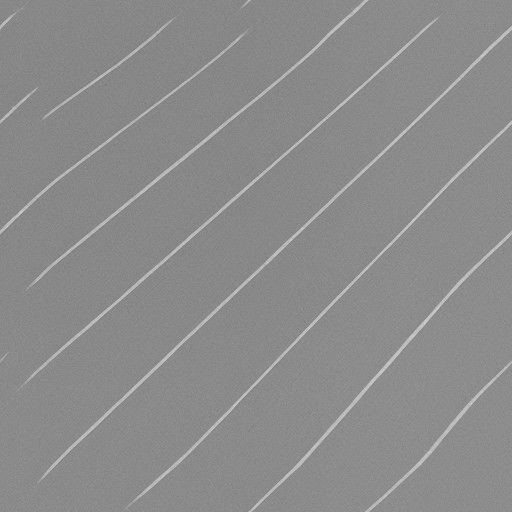}
            \end{minipage}	
            \begin{minipage}{0.13\linewidth}
            \includegraphics[width=\linewidth]{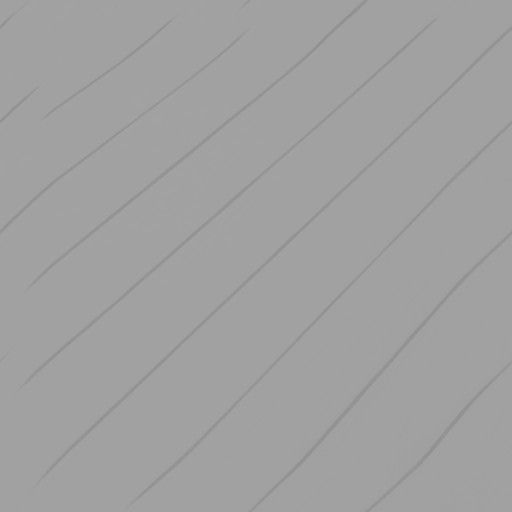}
            \end{minipage}	
            \begin{minipage}{0.13\linewidth}
            \includegraphics[width=\linewidth]{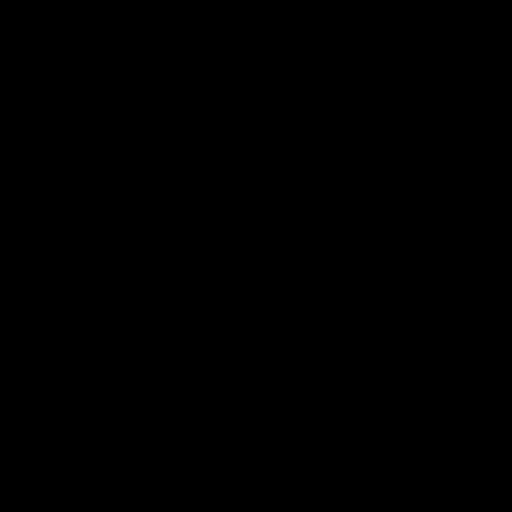}
            \end{minipage}	
            \begin{minipage}{0.13\linewidth}
            \includegraphics[width=\linewidth]{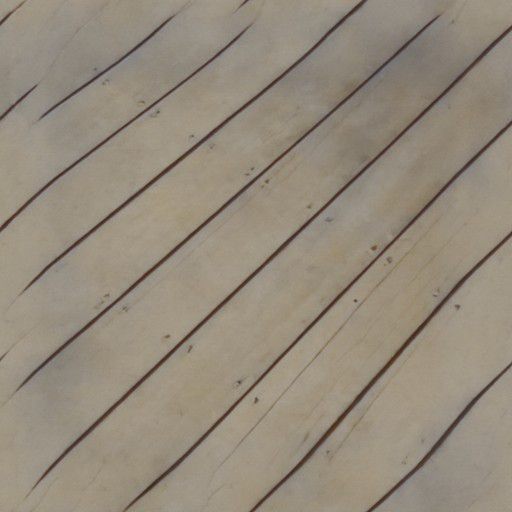}
            \end{minipage}	
        \end{minipage}	
    \end{minipage}

    \begin{minipage}{3.4in}
        \begin{minipage}{0.02in}	
            \centering
                \rotatebox{90}{\tiny Ours}
        \end{minipage}	
        \hspace{0.02in}
         \begin{minipage}{3.3in}	
            \centering
            \begin{minipage}{0.13\linewidth}
            \includegraphics[width=\linewidth]{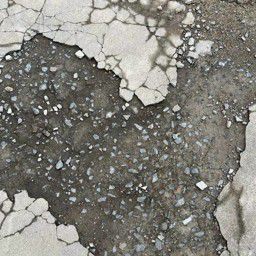}
            \end{minipage}	
            \begin{minipage}{0.13\linewidth}
            \includegraphics[width=\linewidth]{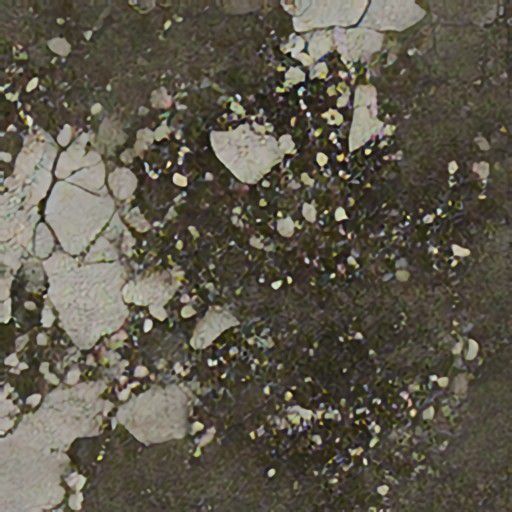}
            \end{minipage}	
            \begin{minipage}{0.13\linewidth}
            \includegraphics[width=\linewidth]{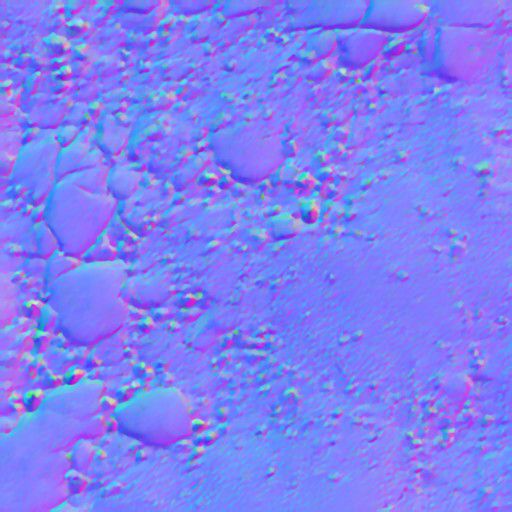}
            \end{minipage}	
            \begin{minipage}{0.13\linewidth}
            \includegraphics[width=\linewidth]{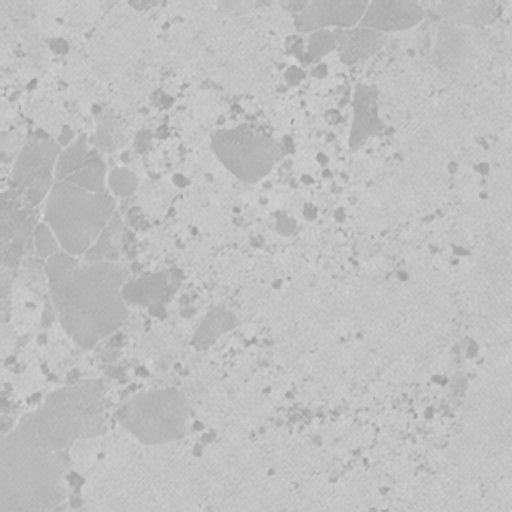}
            \end{minipage}	
            \begin{minipage}{0.13\linewidth}
            \includegraphics[width=\linewidth]{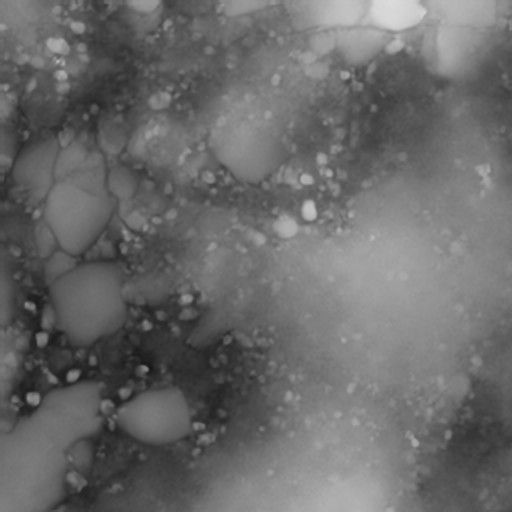}
            \end{minipage}	
            \begin{minipage}{0.13\linewidth}
            \includegraphics[width=\linewidth]{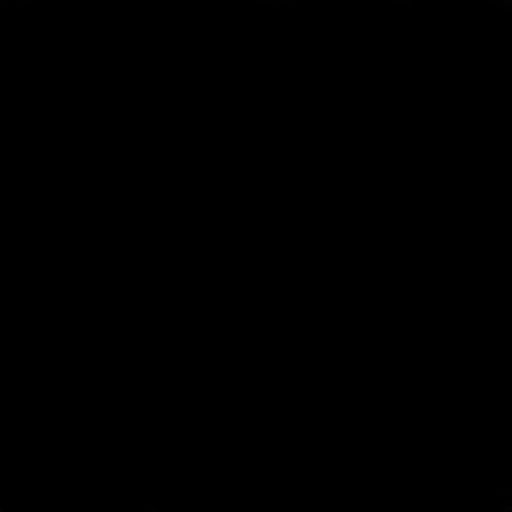}
            \end{minipage}	
            \begin{minipage}{0.13\linewidth}
            \includegraphics[width=\linewidth]{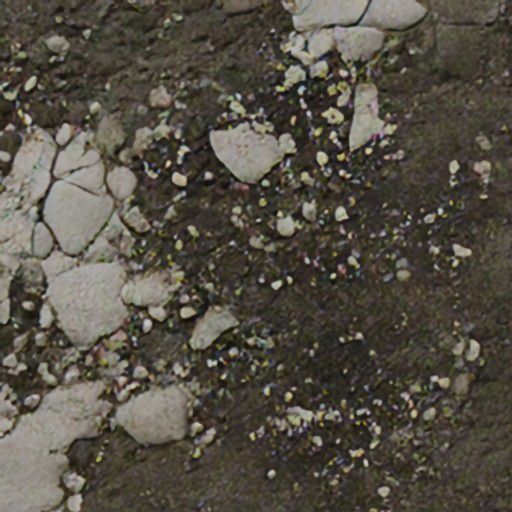}
            \end{minipage}	
        \end{minipage}	
    \end{minipage}	

    \begin{minipage}{3.4in}
        \begin{minipage}{0.02in}	
            \centering
                \rotatebox{90}{\tiny ControlMat}
        \end{minipage}	
        \hspace{0.02in}
         \begin{minipage}{3.3in}	
            \centering
            \begin{minipage}{0.13\linewidth}
            \includegraphics[width=\linewidth]{fig/compare_controlmat/71_input.jpg}
            \end{minipage}	
            \begin{minipage}{0.13\linewidth}
            \includegraphics[width=\linewidth]{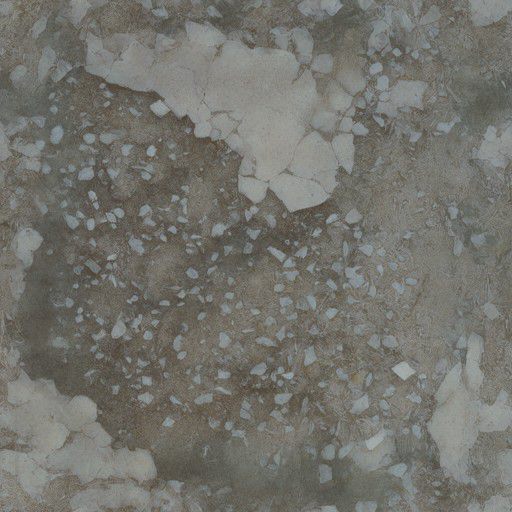}
            \end{minipage}	
            \begin{minipage}{0.13\linewidth}
            \includegraphics[width=\linewidth]{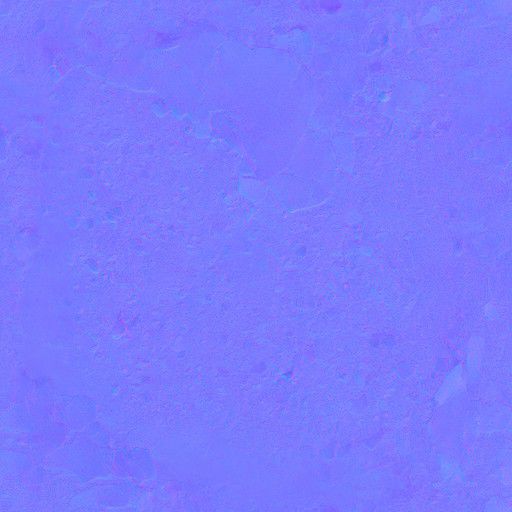}
            \end{minipage}	
            \begin{minipage}{0.13\linewidth}
            \includegraphics[width=\linewidth]{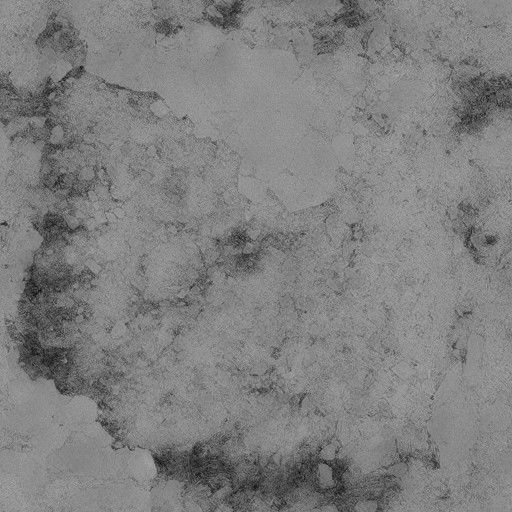}
            \end{minipage}	
            \begin{minipage}{0.13\linewidth}
            \includegraphics[width=\linewidth]{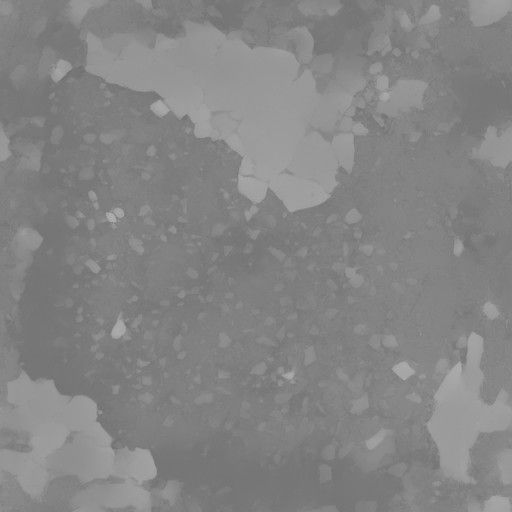}
            \end{minipage}	
            \begin{minipage}{0.13\linewidth}
            \includegraphics[width=\linewidth]{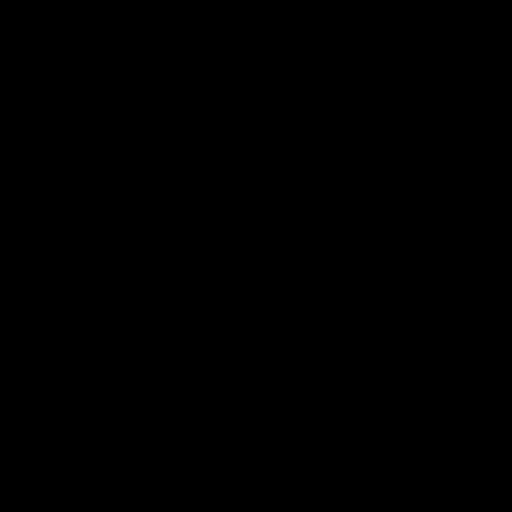}
            \end{minipage}	
            \begin{minipage}{0.13\linewidth}
            \includegraphics[width=\linewidth]{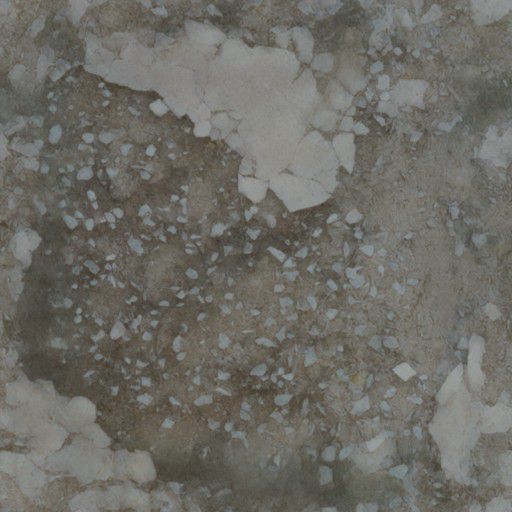}
            \end{minipage}	
        \end{minipage}	
    \end{minipage}

   \caption{Comparison between our method and ControlMat~\cite{vecchio2024controlmat} on material extraction for real photographs with distortions. The first column shows the input images. The second to the sixth columns show the generated material maps and rendering under an environment map. As ControlMat relies on well aligned conditions through ControlNet, our approach demonstrates superior performance in correcting perspectives and distortions. Further, as ControlMat is trained only on material data, it does not benefit from the learned priors from a pre-trained model like our fine-tuning, making their method less effective at generalizing to unseen patterns and lighting conditions (e.g. in the 4th row).}
   \label{fig:compare_controlmat}
\end{figure}

We show a comparison with ControlMat~\cite{vecchio2024controlmat} 
in Fig.~\ref{fig:compare_controlmat}. Since ControlMat uses local features of the images extracted by ControlNet~\cite{zhang2023adding} to guide the diffusion process, it is trained to generate results that align with the input, therefore fail to handle imperfect perspectives and distortions. Further, since the generation of multichannel materials in ControlMat depends on modifications to the VAE, it must be trained from scratch without leveraging any image or video priors from which our method benefits. This limits ControlMat generalization to complex textures captured under unseen lighting conditions (e.g. 4th row).

\begin{figure*}
    \begin{minipage}{\linewidth}
         \centering
            \begin{minipage}{0.085\linewidth}
                \subcaption*{\tiny Input}
                \includegraphics[width=\linewidth]{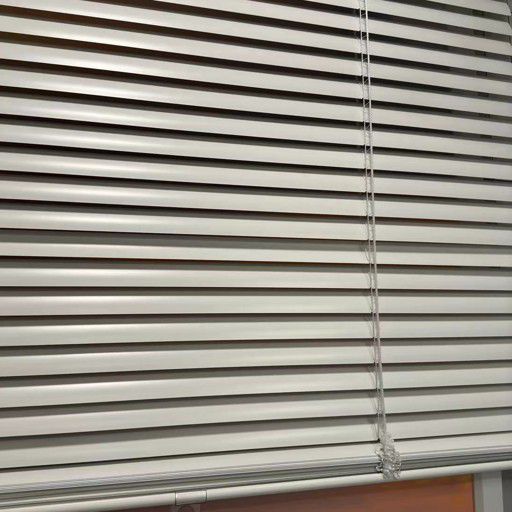}
            \end{minipage}
            \begin{minipage}{0.085\linewidth}
                \subcaption*{\tiny Mask (Input)}
                \includegraphics[width=\linewidth]{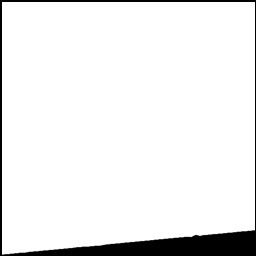}
            \end{minipage}
            \begin{minipage}{0.085\linewidth}
                \subcaption*{\tiny [Hao et al. 2023]}
                \includegraphics[width=\linewidth]{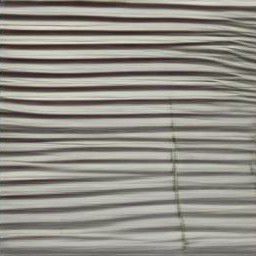}
            \end{minipage}
            \begin{minipage}{0.085\linewidth}
            % \vspace{-0.08in}
                % \subcaption*{\fontsize{5}{3}\selectfont Our Mask(Output)}
                \parbox{1.1\linewidth}{\subcaption*{\hspace{-0.05in}\tiny Our Mask (Output)}}
                \includegraphics[width=\linewidth]{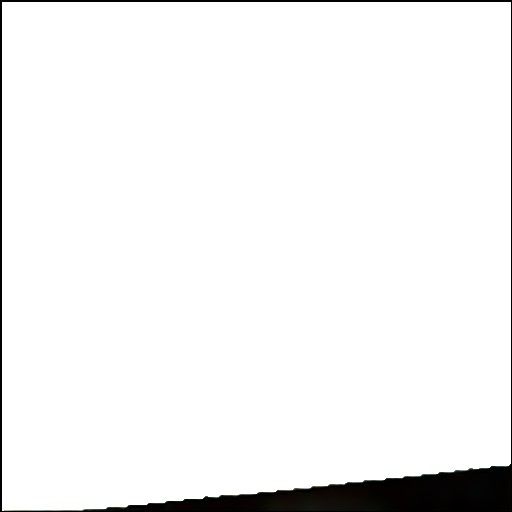}
            \end{minipage}
            \begin{minipage}{0.085\linewidth}
                \subcaption*{\tiny Ours (Render 1)}
                \includegraphics[width=\linewidth]{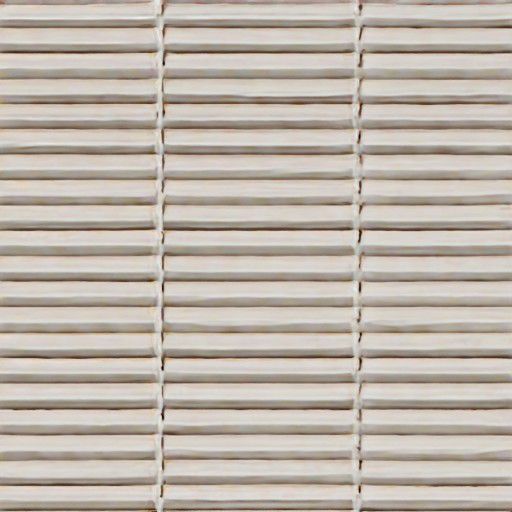}
            \end{minipage}
            \begin{minipage}{0.085\linewidth}
                \subcaption*{\tiny Ours (Render 2)}
                \includegraphics[width=\linewidth]{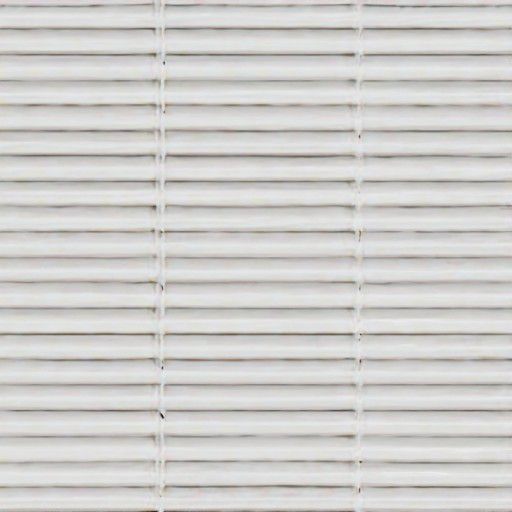}
            \end{minipage}
            \begin{minipage}{0.085\linewidth}
                \subcaption*{\tiny Albedo}
                \includegraphics[width=\linewidth]{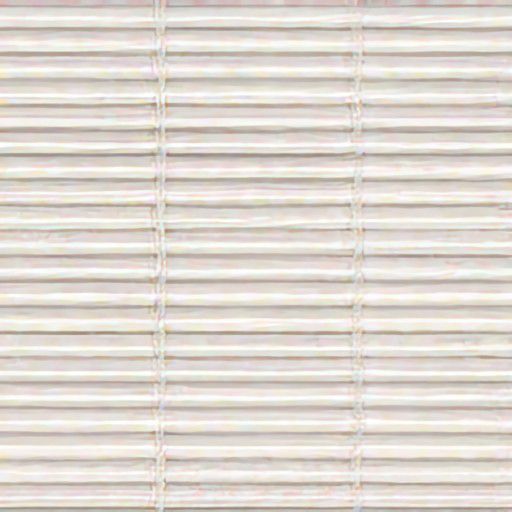}
            \end{minipage}
            \begin{minipage}{0.085\linewidth}
                \subcaption*{\tiny Normal}
                \includegraphics[width=\linewidth]{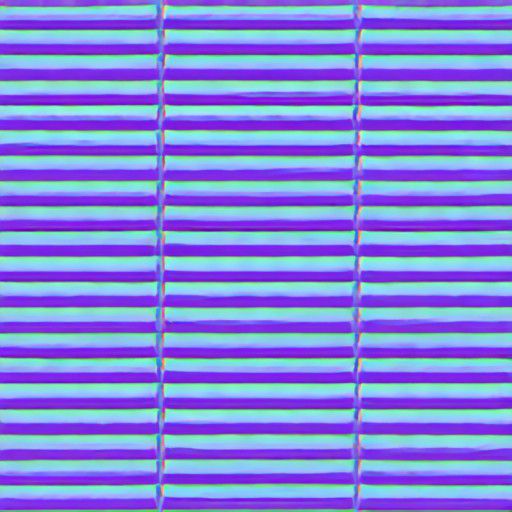}
            \end{minipage}
            \begin{minipage}{0.085\linewidth}
                \subcaption*{\tiny Roughness}
                \includegraphics[width=\linewidth]{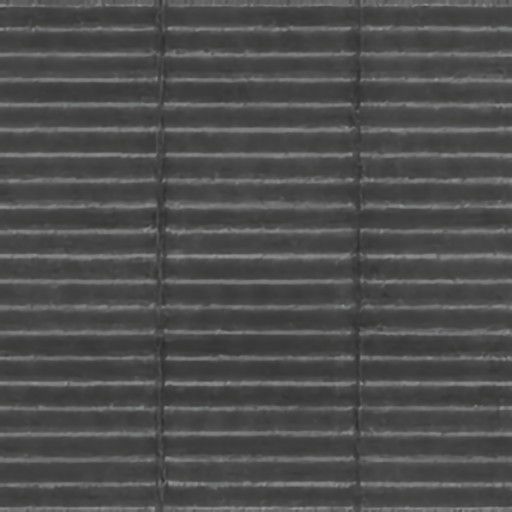}
            \end{minipage}
            \begin{minipage}{0.085\linewidth}
                \subcaption*{\tiny Height}
                \includegraphics[width=\linewidth]{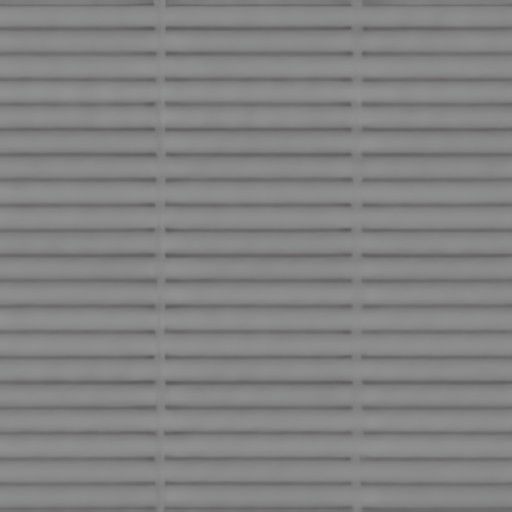}
            \end{minipage}
            \begin{minipage}{0.085\linewidth}
                \subcaption*{\tiny Metallic}
                \includegraphics[width=\linewidth]{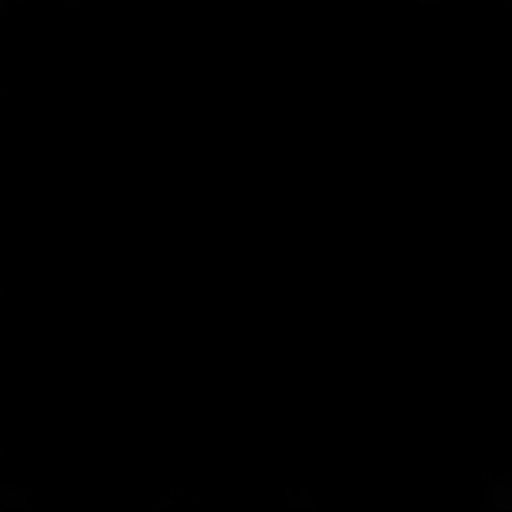}
            \end{minipage}
        
    \end{minipage}	

    \begin{minipage}{\linewidth}
         \centering
            \begin{minipage}{0.085\linewidth}
            \includegraphics[width=\linewidth]{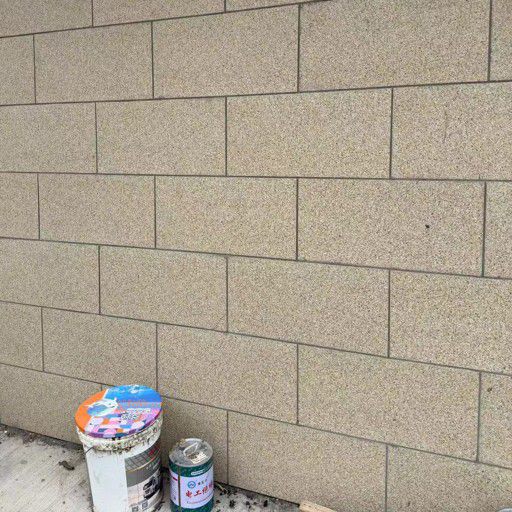}
            \end{minipage}	
            \begin{minipage}{0.085\linewidth}
            \includegraphics[width=\linewidth]{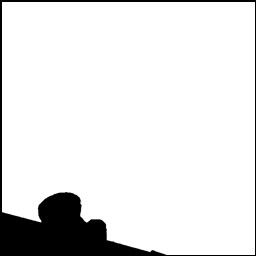}
            \end{minipage}	
            \begin{minipage}{0.085\linewidth}
            \includegraphics[width=\linewidth]{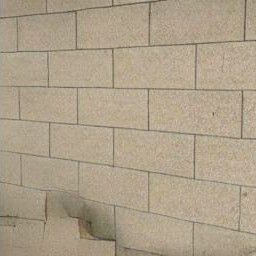}
            \end{minipage}	
            \begin{minipage}{0.085\linewidth}
            \includegraphics[width=\linewidth]{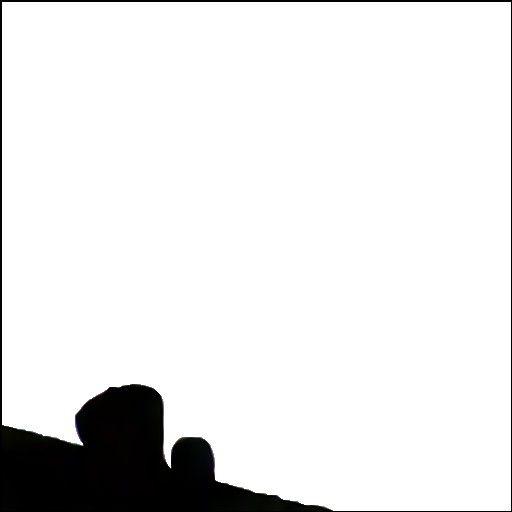}
            \end{minipage}	
            \begin{minipage}{0.085\linewidth}
            \includegraphics[width=\linewidth]{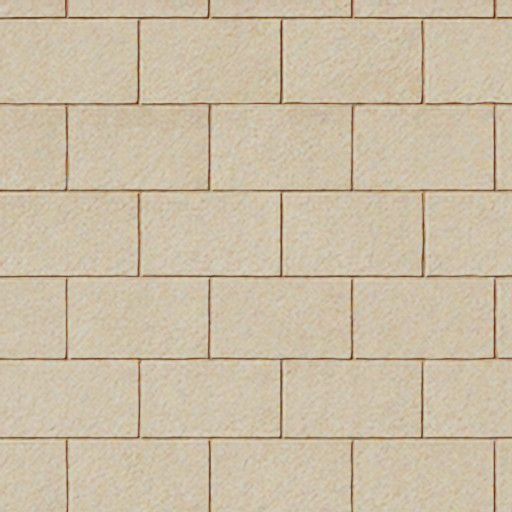}
            \end{minipage}	
            \begin{minipage}{0.085\linewidth}
            \includegraphics[width=\linewidth]{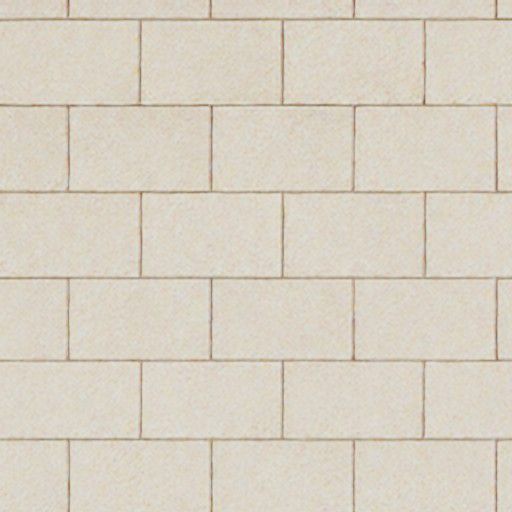}
            \end{minipage}	
            \begin{minipage}{0.085\linewidth}
            \includegraphics[width=\linewidth]{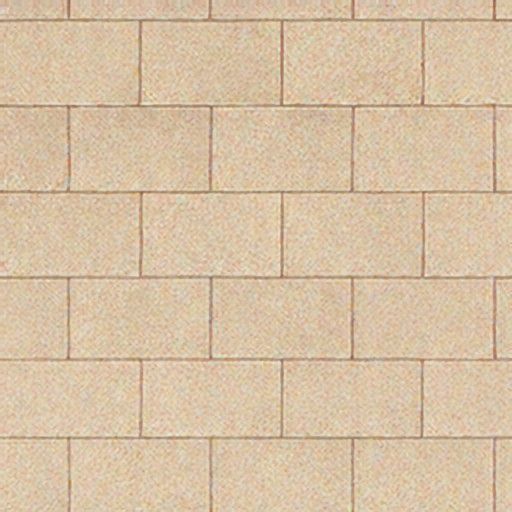}
            \end{minipage}	
            \begin{minipage}{0.085\linewidth}
            \includegraphics[width=\linewidth]{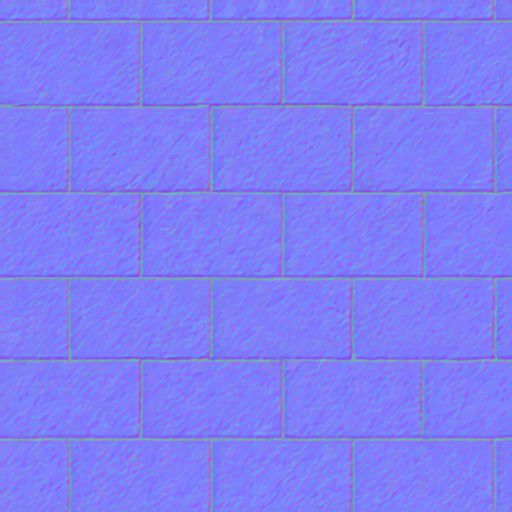}
            \end{minipage}	
            \begin{minipage}{0.085\linewidth}
            \includegraphics[width=\linewidth]{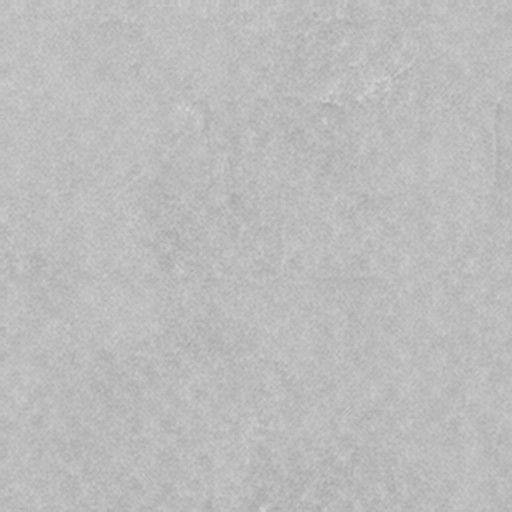}
            \end{minipage}	
            \begin{minipage}{0.085\linewidth}
            \includegraphics[width=\linewidth]{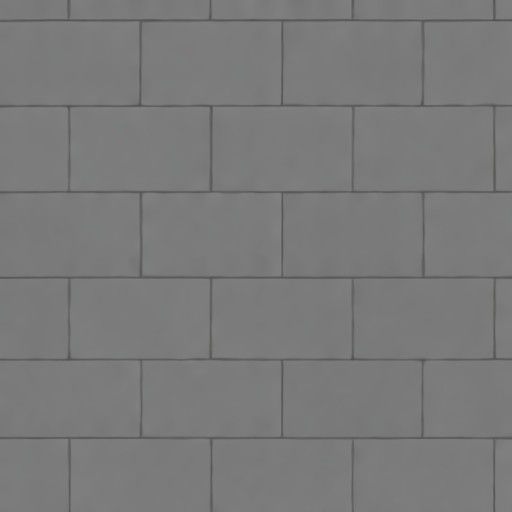}
            \end{minipage}	
            \begin{minipage}{0.085\linewidth}
            \includegraphics[width=\linewidth]{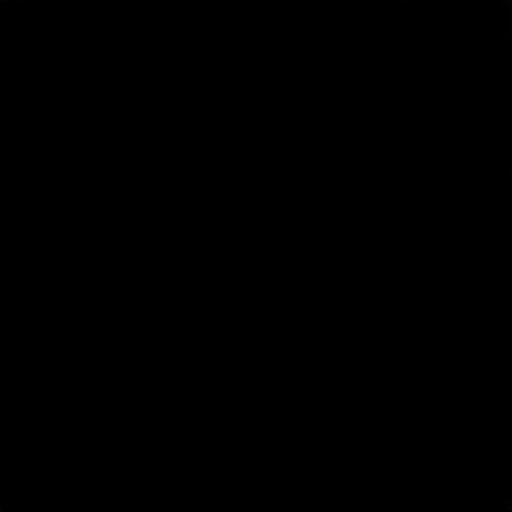}
            \end{minipage}	
    \end{minipage}	

     \begin{minipage}{\linewidth}
         \centering
            \begin{minipage}{0.085\linewidth}
            \includegraphics[width=\linewidth]{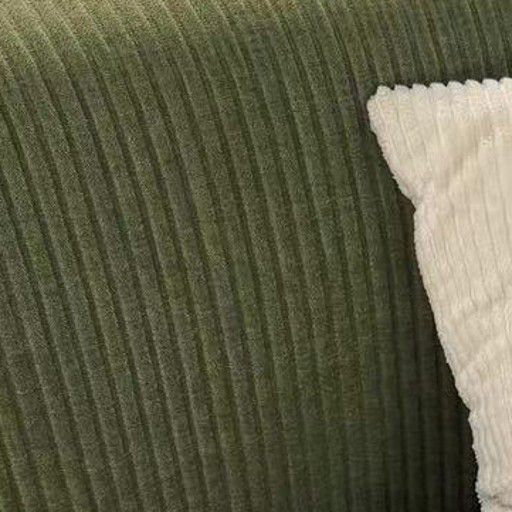}
            \end{minipage}	
            \begin{minipage}{0.085\linewidth}
            \includegraphics[width=\linewidth]{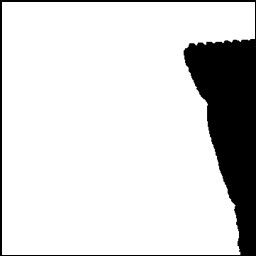}
            \end{minipage}	
            \begin{minipage}{0.085\linewidth}
            \includegraphics[width=\linewidth]{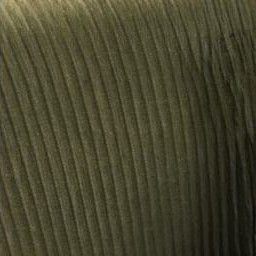}
            \end{minipage}	
            \begin{minipage}{0.085\linewidth}
            \includegraphics[width=\linewidth]{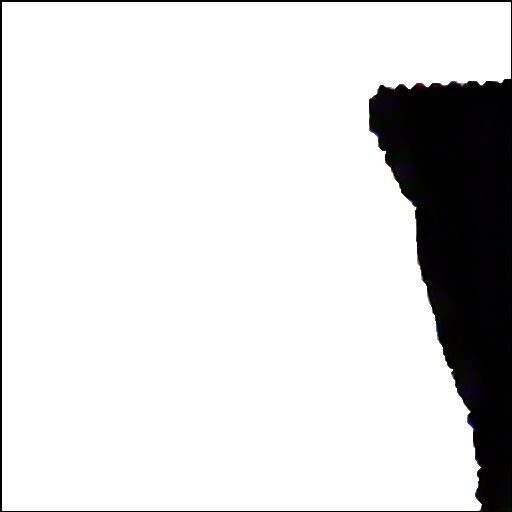}
            \end{minipage}	
            \begin{minipage}{0.085\linewidth}
            \includegraphics[width=\linewidth]{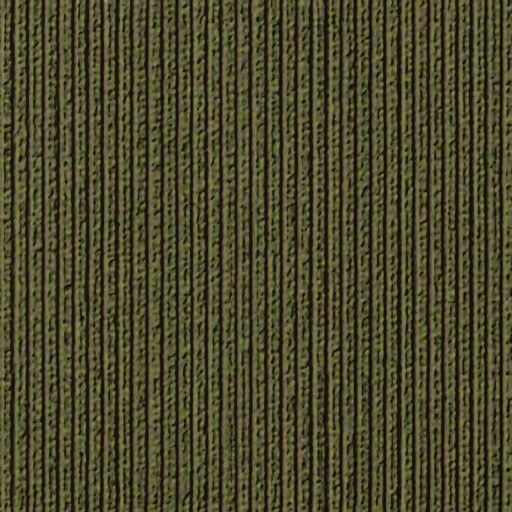}
            \end{minipage}
            \begin{minipage}{0.085\linewidth}
            \includegraphics[width=\linewidth]{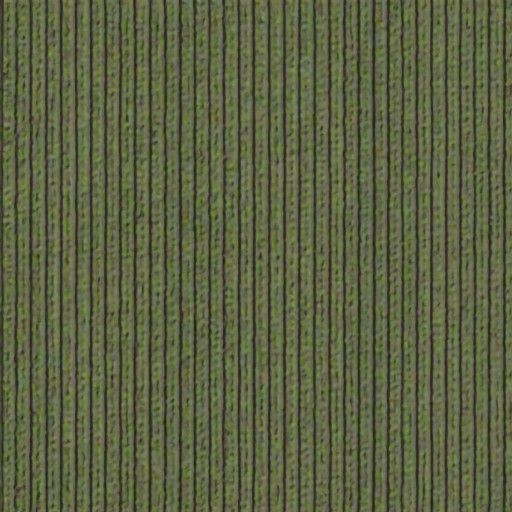}
            \end{minipage}	
            \begin{minipage}{0.085\linewidth}
            \includegraphics[width=\linewidth]{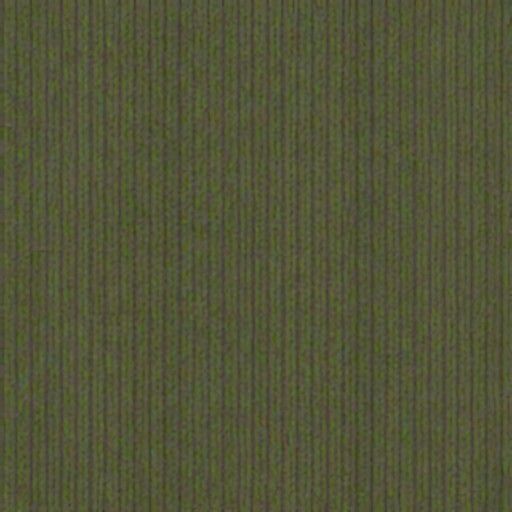}
            \end{minipage}	
            \begin{minipage}{0.085\linewidth}
            \includegraphics[width=\linewidth]{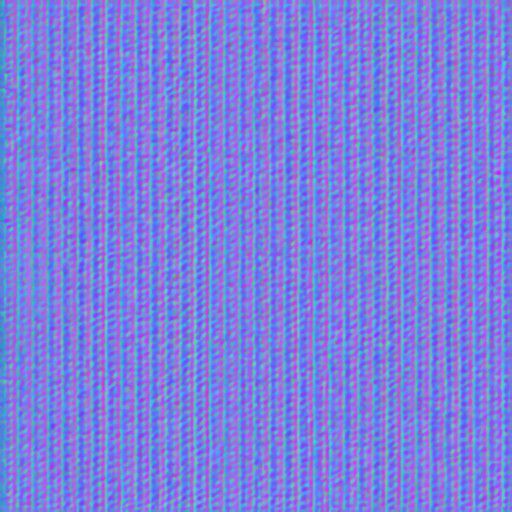}
            \end{minipage}	
            \begin{minipage}{0.085\linewidth}
            \includegraphics[width=\linewidth]{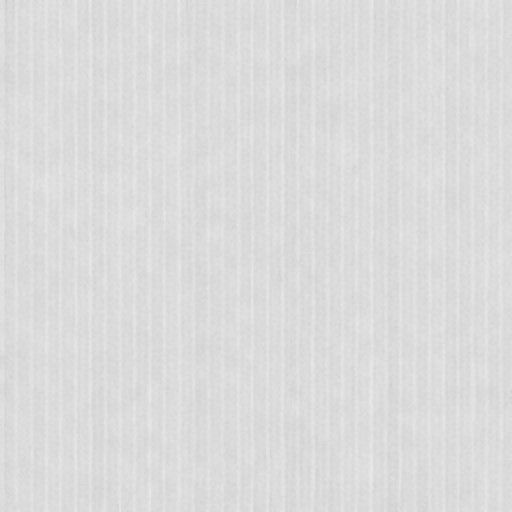}
            \end{minipage}	
            \begin{minipage}{0.085\linewidth}
            \includegraphics[width=\linewidth]{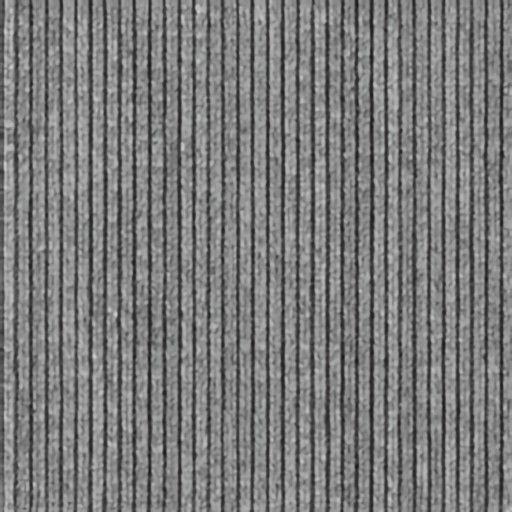}
            \end{minipage}	
            \begin{minipage}{0.085\linewidth}
            \includegraphics[width=\linewidth]{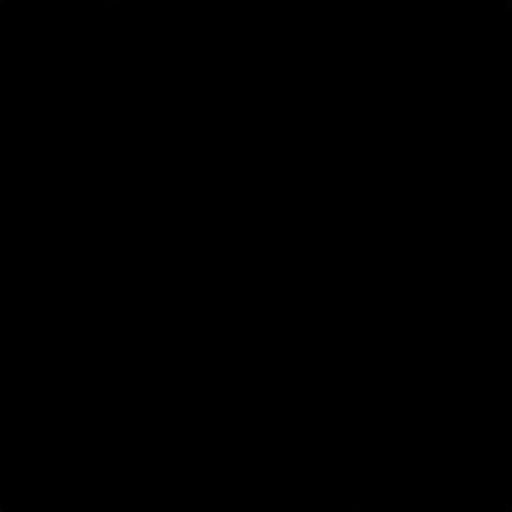}
            \end{minipage}	
    \end{minipage}	
    
    \begin{minipage}{\linewidth}
         \centering
            \begin{minipage}{0.085\linewidth}
            \includegraphics[width=\linewidth]{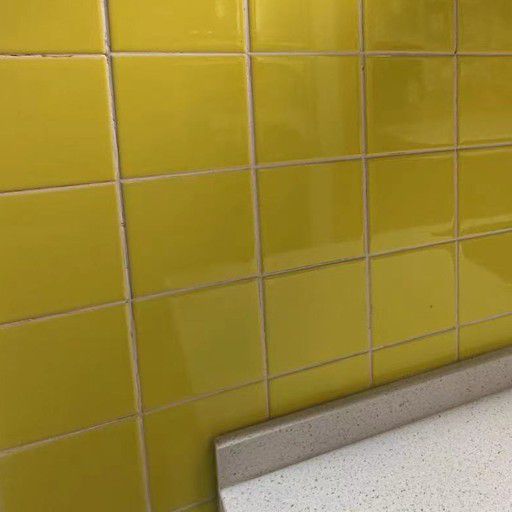}
            \end{minipage}	
            \begin{minipage}{0.085\linewidth}
            \includegraphics[width=\linewidth]{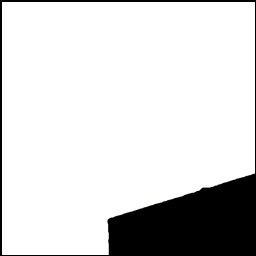}
            \end{minipage}	
            \begin{minipage}{0.085\linewidth}
            \includegraphics[width=\linewidth]{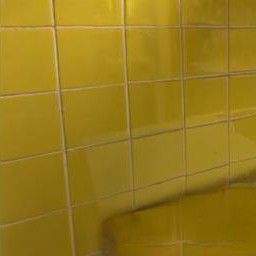}
            \end{minipage}	
            \begin{minipage}{0.085\linewidth}
            \includegraphics[width=\linewidth]{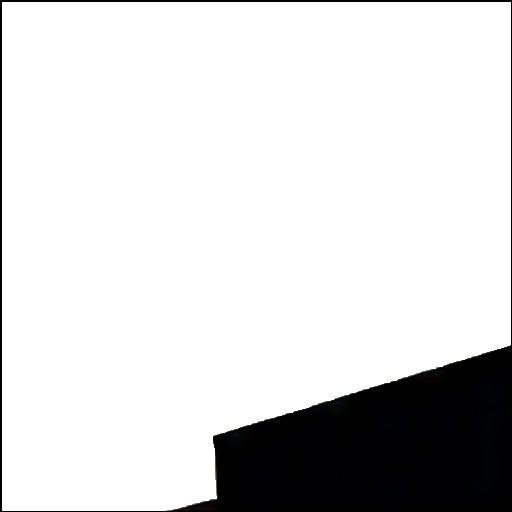}
            \end{minipage}	
            \begin{minipage}{0.085\linewidth}
            \includegraphics[width=\linewidth]{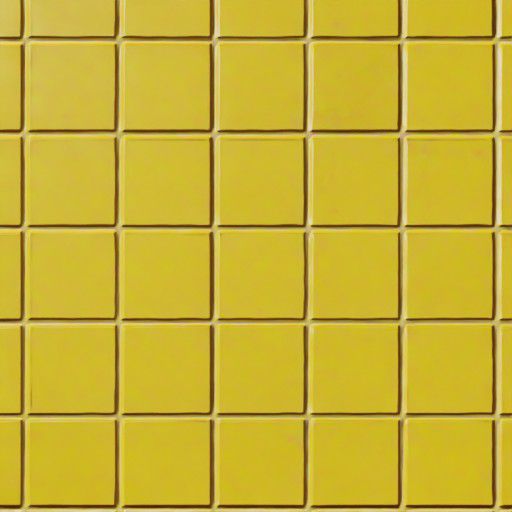}
            \end{minipage}	
            \begin{minipage}{0.085\linewidth}
            \includegraphics[width=\linewidth]{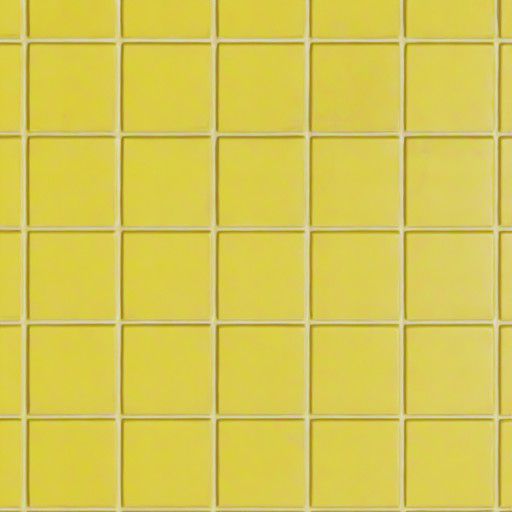}
            \end{minipage}	
            \begin{minipage}{0.085\linewidth}
            \includegraphics[width=\linewidth]{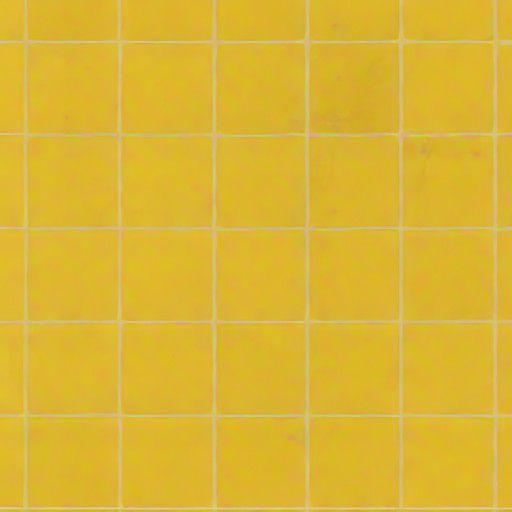}
            \end{minipage}	
            \begin{minipage}{0.085\linewidth}
            \includegraphics[width=\linewidth]{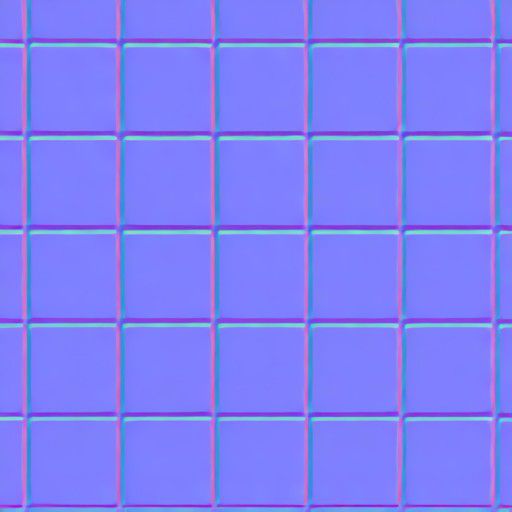}
            \end{minipage}	
            \begin{minipage}{0.085\linewidth}
            \includegraphics[width=\linewidth]{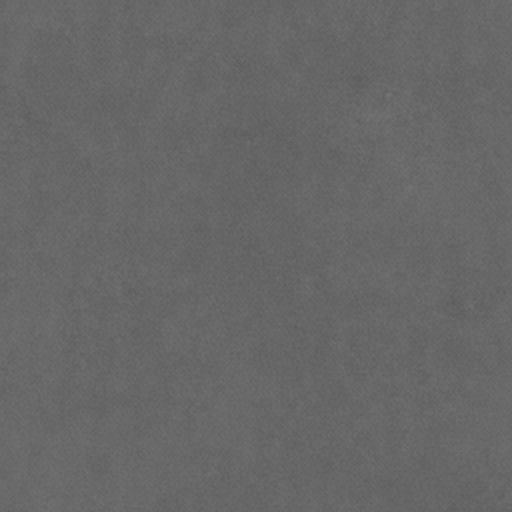}
            \end{minipage}	
            \begin{minipage}{0.085\linewidth}
            \includegraphics[width=\linewidth]{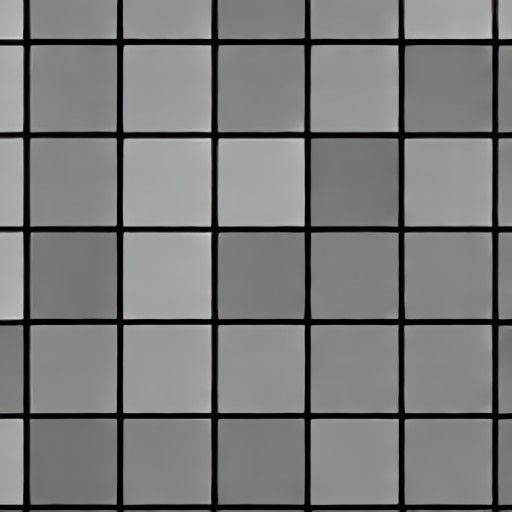}
            \end{minipage}	
            \begin{minipage}{0.085\linewidth}
            \includegraphics[width=\linewidth]{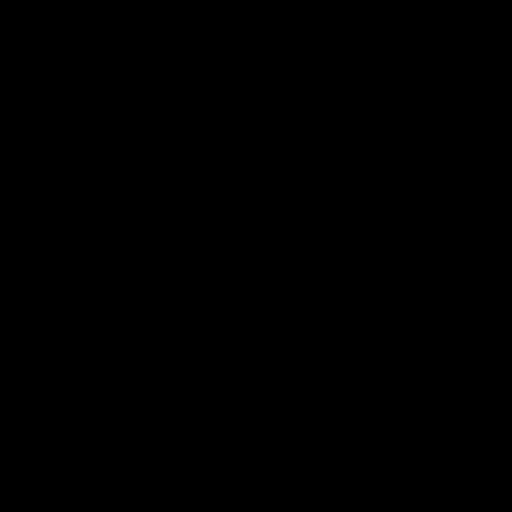}
            \end{minipage}	
        	
    \end{minipage}	

    \begin{minipage}{\linewidth}
         \centering
            \begin{minipage}{0.085\linewidth}
            \includegraphics[width=\linewidth]{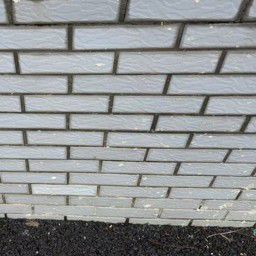}
            \end{minipage}	
            \begin{minipage}{0.085\linewidth}
            \includegraphics[width=\linewidth]{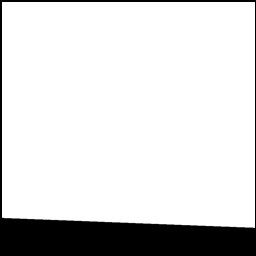}
            \end{minipage}	
            \begin{minipage}{0.085\linewidth}
            \includegraphics[width=\linewidth]{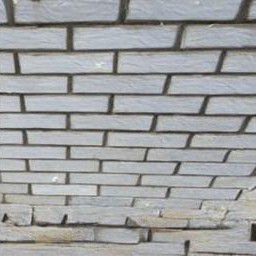}
            \end{minipage}	
            \begin{minipage}{0.085\linewidth}
            \includegraphics[width=\linewidth]{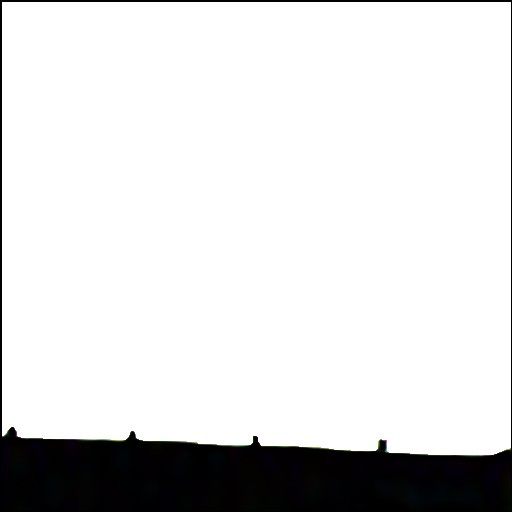}
            \end{minipage}	
            \begin{minipage}{0.085\linewidth}
            \includegraphics[width=\linewidth]{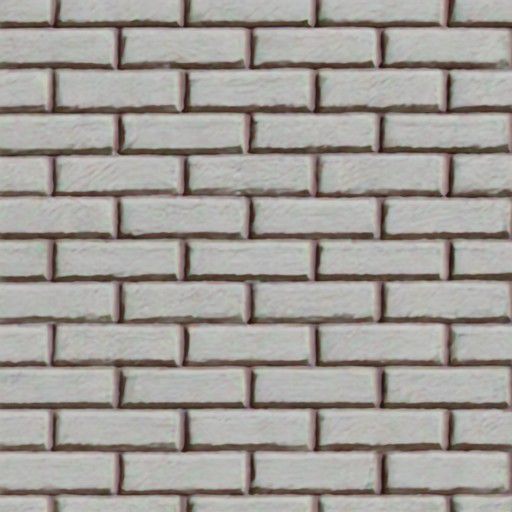}
            \end{minipage}	
            \begin{minipage}{0.085\linewidth}
            \includegraphics[width=\linewidth]{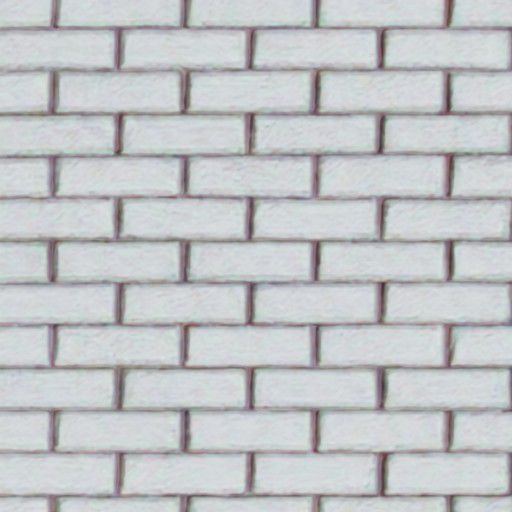}
            \end{minipage}	
            \begin{minipage}{0.085\linewidth}
            \includegraphics[width=\linewidth]{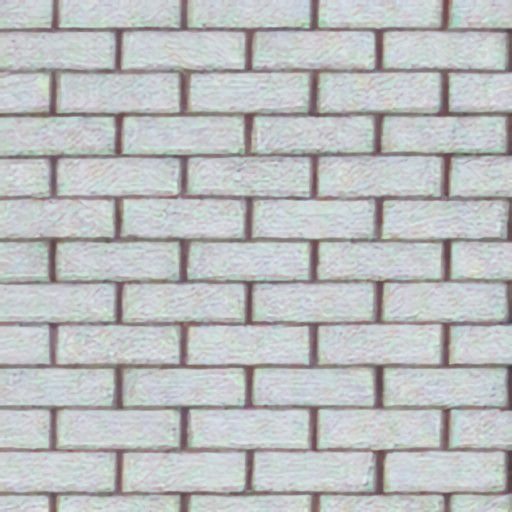}
            \end{minipage}	
            \begin{minipage}{0.085\linewidth}
            \includegraphics[width=\linewidth]{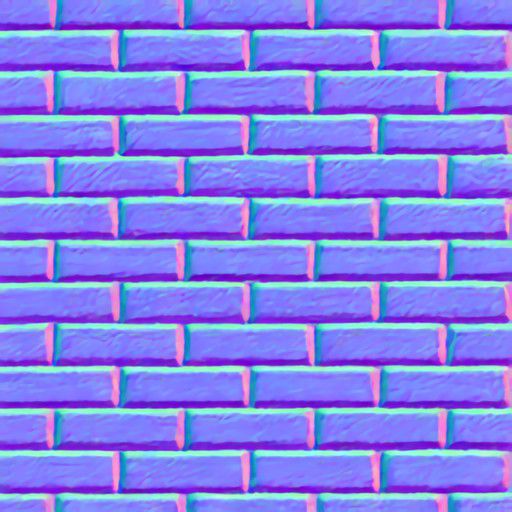}
            \end{minipage}	
            \begin{minipage}{0.085\linewidth}
            \includegraphics[width=\linewidth]{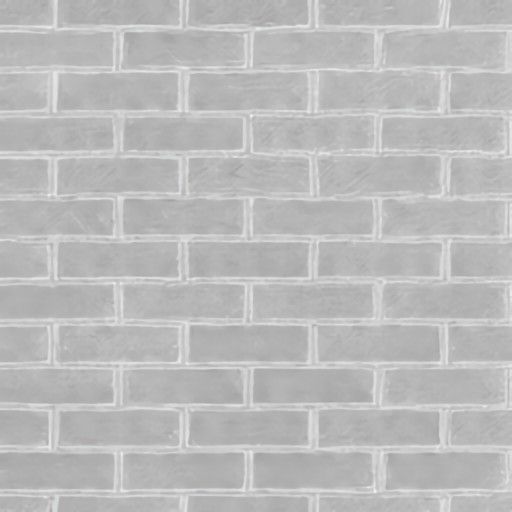}
            \end{minipage}	
            \begin{minipage}{0.085\linewidth}
            \includegraphics[width=\linewidth]{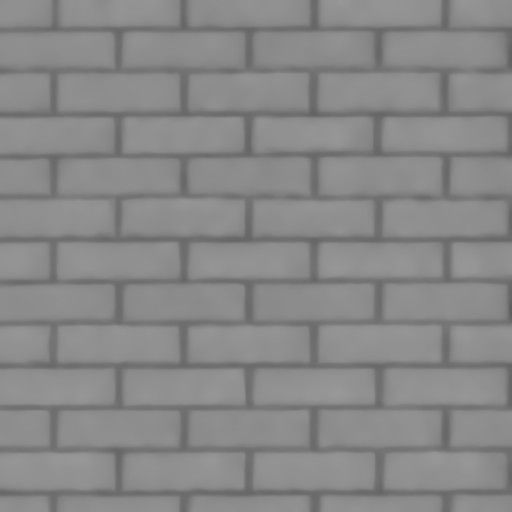}
            \end{minipage}	
            \begin{minipage}{0.085\linewidth}
            \includegraphics[width=\linewidth]{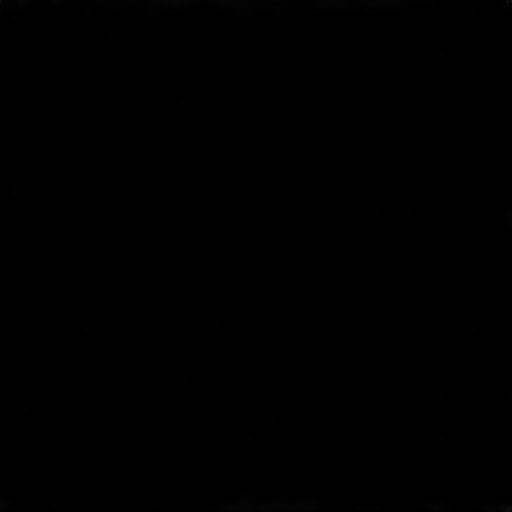}
            \end{minipage}	
    \end{minipage}	
    \begin{tikzpicture}
        \node (txt) at (0,0) {};
        \begin{scope}[shift={(txt.east)}, overlay, remember picture]
            \draw[dashed, color=darkred,line width=0.1pt] (-3.97cm,0.1cm) -- ++(0,8.0cm);
        \end{scope}
    \end{tikzpicture}
    \begin{tikzpicture}
        \node (txt) at (0,0) {};
        \begin{scope}[shift={(txt.east)}, overlay, remember picture]
            \draw[dashed, color=darkred,line width=0.1pt] (-7.47cm,0.1cm) -- ++(0,8.cm);
        \end{scope}
    \end{tikzpicture}
   \caption{Comparison with \citet{hao2023diffusion} on texture rectification for real photos. The first column shows the input photos. The second and third columns are the required input masks and output textures of \citeauthor{hao2023diffusion} The fourth column shows masks generated by our model, followed by two renderings (Render 1 \& Render 2) of our generated material maps (in the last five columns) under different environment maps. Despite not requiring an input mask, our method better rectifies perspective and distortions. Further, as we support material extraction, our result does not contain shading from the input image.} 
   \label{fig:texture_synthesis}
\end{figure*}

Since our model automatically performs perspective rectification on the generated materials, we further compare against another state-of-the-art texture rectification and synthesis method~\cite{hao2023diffusion}. In Fig.~\ref{fig:texture_synthesis}, we evaluate both methods using real photographs. Since our model directly outputs material maps, instead of textures, we present our results by rendering them under different environment maps. We find that the compared method does not generalize well to real-world photographs, taken from non-frontal and/or non-parallel setups and fails to correct distortion in these cases. In contrast, our approach synthesizes a fronto-parallel view and remains robust across various real-world lighting conditions and viewing angles. Finally, as previously, our model does not require detailed masks as input,  directly rectifying the dominant texture in the input image. 

\subsection{Text Conditioned Generation}
\label{sec:text_conditioned_generation}
\begin{figure}[t]
    \centering		
    \begin{minipage}{3.4in}
        \begin{minipage}{0.02in}	
            \centering
                \vspace{0.1in}
                \rotatebox{90}{\tiny Ours}
        \end{minipage}	
        \hspace{0.02in}
         \begin{minipage}{3.3in}	
            \centering
            \begin{minipage}{0.13\linewidth}
                \subcaption*{\tiny Text}
                \includegraphics[width=\linewidth]{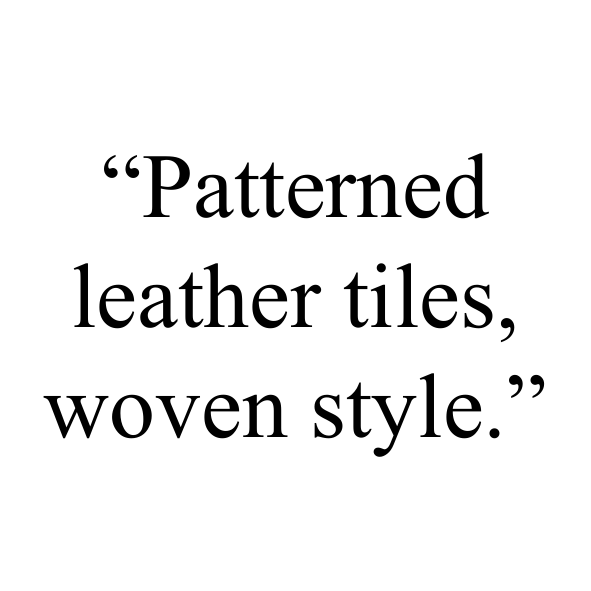}
            \end{minipage}
            \begin{minipage}{0.13\linewidth}
                \subcaption*{\tiny Albedo}
                \includegraphics[width=\linewidth]{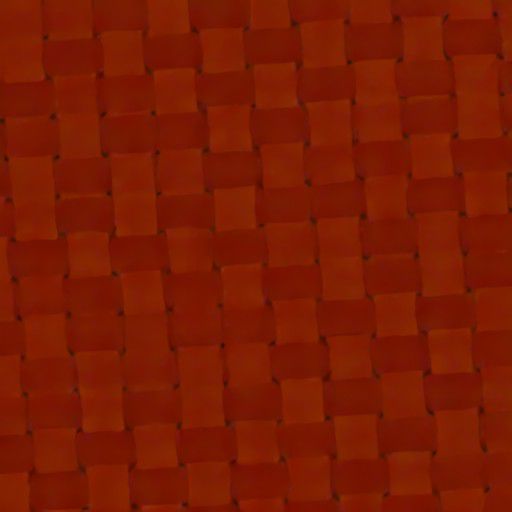}
            \end{minipage}
            \begin{minipage}{0.13\linewidth}
                \subcaption*{\tiny Normal}
                \includegraphics[width=\linewidth]{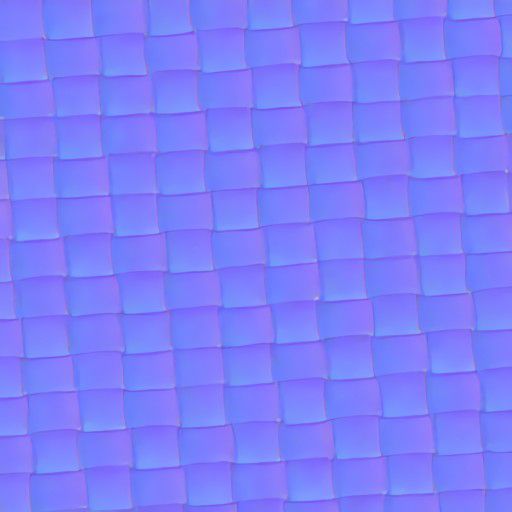}
            \end{minipage}
            \begin{minipage}{0.13\linewidth}
                \subcaption*{\tiny Roughness}
                \includegraphics[width=\linewidth]{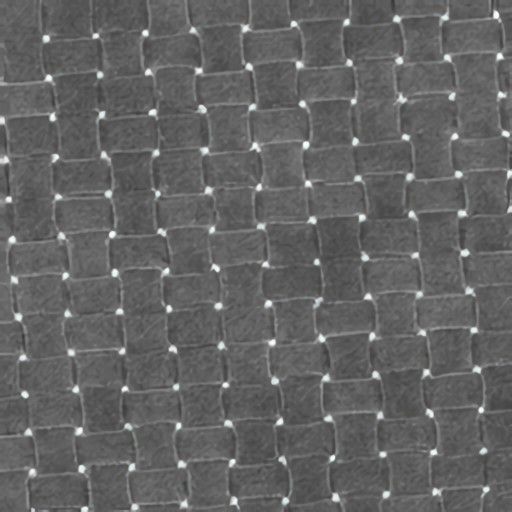}
            \end{minipage}
            \begin{minipage}{0.13\linewidth}
                \subcaption*{\tiny Height}
                \includegraphics[width=\linewidth]{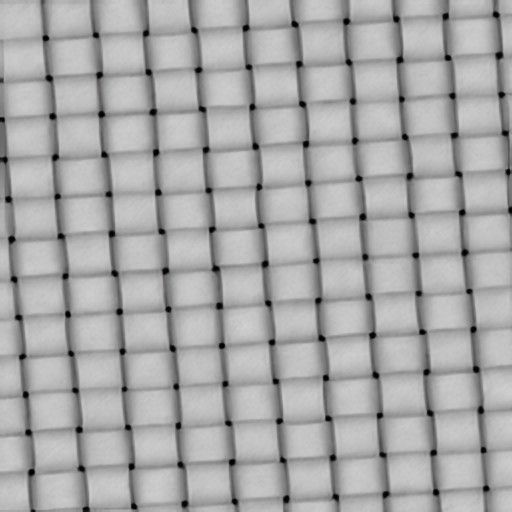}
            \end{minipage}
            \begin{minipage}{0.13\linewidth}
                \subcaption*{\tiny Metal. / Spec.}
                \includegraphics[width=\linewidth]{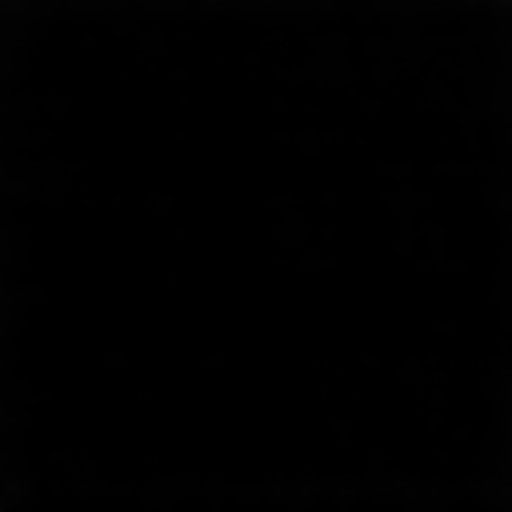}
            \end{minipage}
            \begin{minipage}{0.13\linewidth}
                \subcaption*{\tiny Render}
                \includegraphics[width=\linewidth]{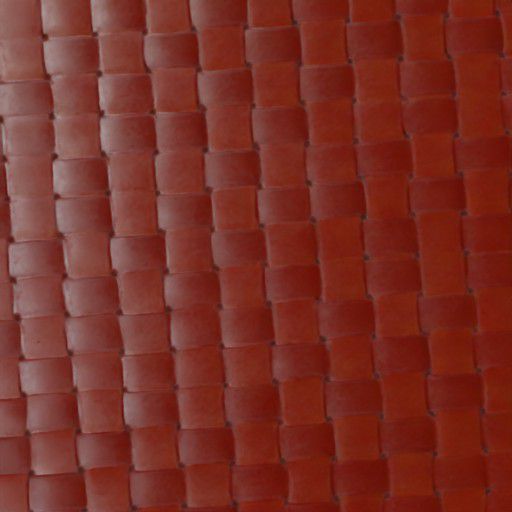}
                \put(-21,2){\tiny \textbf{\color{white}0.291}}
            \end{minipage}
        \end{minipage}	
        \end{minipage}	

    \begin{minipage}{3.4in}
        \begin{minipage}{0.02in}	
            \centering
                \rotatebox{90}{\tiny MatGen}
        \end{minipage}	
        \hspace{0.02in}
         \begin{minipage}{3.3in}	
            \centering
            \begin{minipage}{0.13\linewidth}
            \includegraphics[width=\linewidth]{fig/text_gen_comparison/leather.pdf}
            \end{minipage}	
            \begin{minipage}{0.13\linewidth}
            \includegraphics[width=\linewidth]{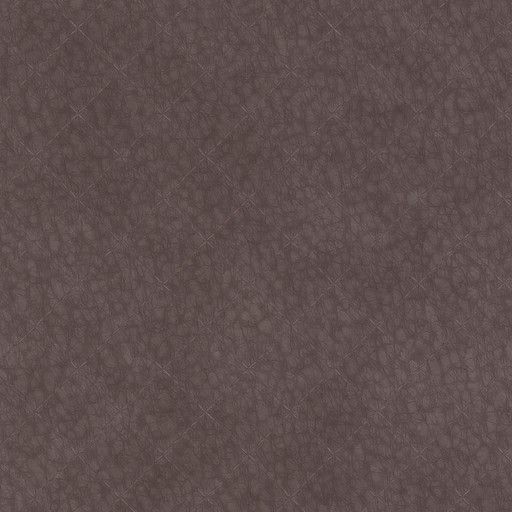}
            \end{minipage}	
            \begin{minipage}{0.13\linewidth}
            \includegraphics[width=\linewidth]{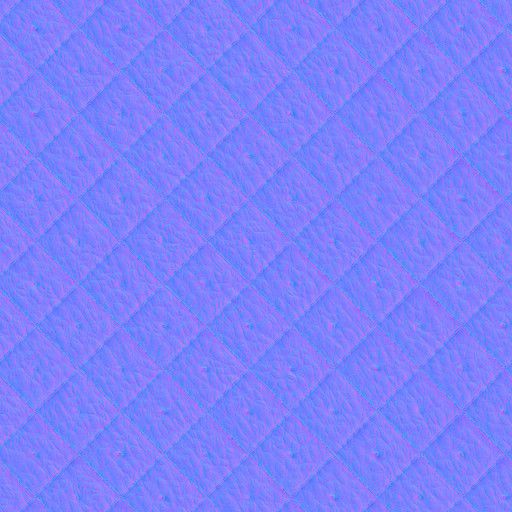}
            \end{minipage}	
            \begin{minipage}{0.13\linewidth}
            \includegraphics[width=\linewidth]{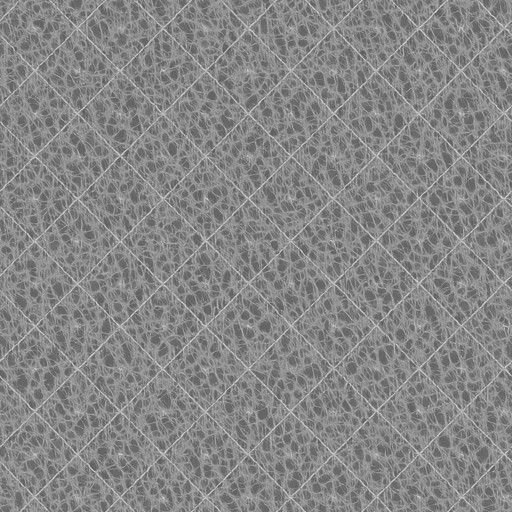}
            \end{minipage}	
            \begin{minipage}{0.13\linewidth}
            \includegraphics[width=\linewidth]{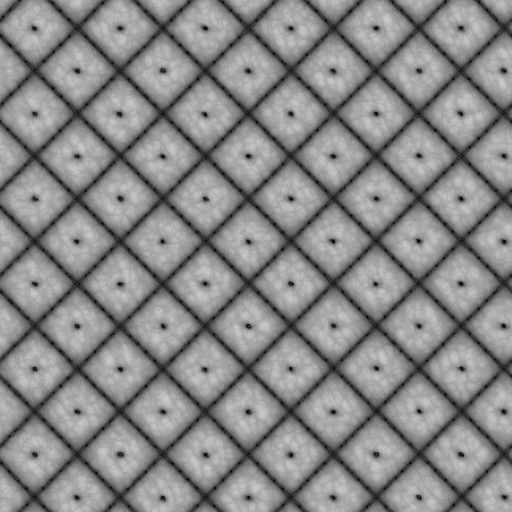}
            \end{minipage}	
            \begin{minipage}{0.13\linewidth}
            \includegraphics[width=\linewidth]{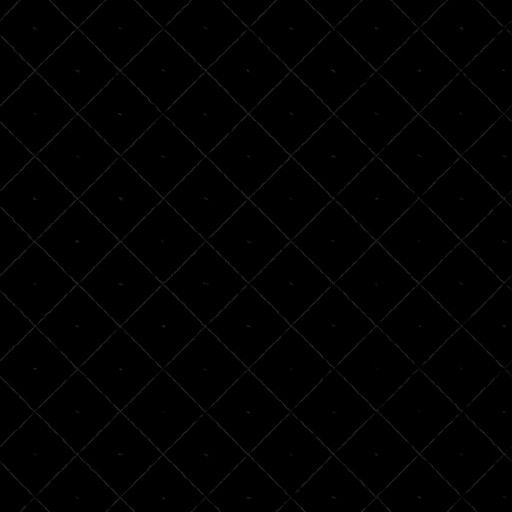}
            \end{minipage}	
            \begin{minipage}{0.13\linewidth}
            \includegraphics[width=\linewidth]{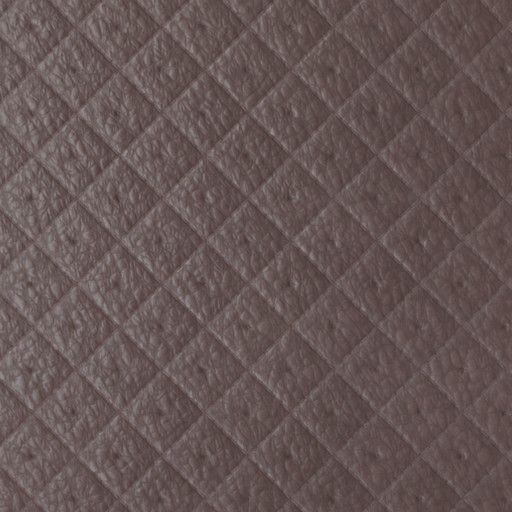}
            \put(-21,2){\tiny \color{white}0.260}
            \end{minipage}	
        \end{minipage}	
    \end{minipage}	

    \begin{minipage}{3.4in}
        \begin{minipage}{0.02in}	
            \centering
                \rotatebox{90}{\tiny MatFuse}
        \end{minipage}	
        \hspace{0.02in}
         \begin{minipage}{3.3in}	
            \centering
            \begin{minipage}{0.13\linewidth}
            \includegraphics[width=\linewidth]{fig/text_gen_comparison/leather.pdf}
            \end{minipage}	
            \begin{minipage}{0.13\linewidth}
            \includegraphics[width=\linewidth]{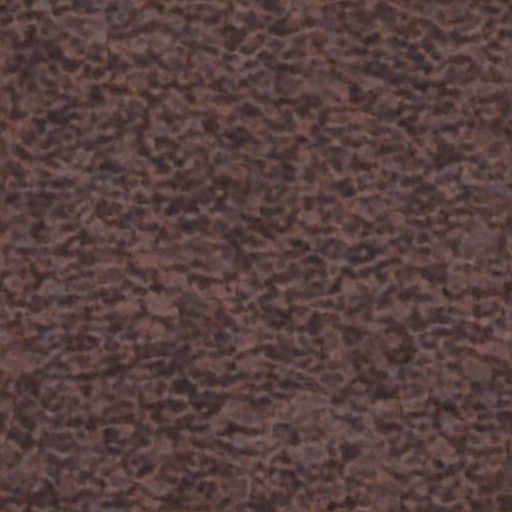}
            \end{minipage}	
            \begin{minipage}{0.13\linewidth}
            \includegraphics[width=\linewidth]{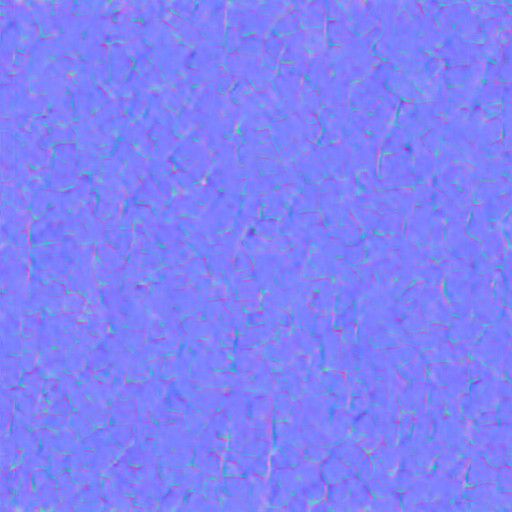}
            \end{minipage}	
            \begin{minipage}{0.13\linewidth}
            \includegraphics[width=\linewidth]{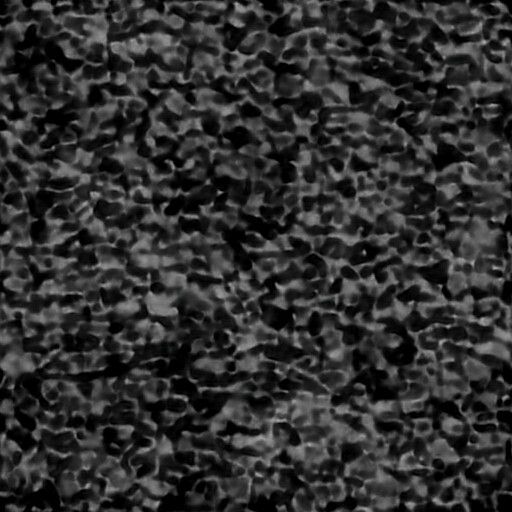}
            \end{minipage}	
            \begin{minipage}{0.13\linewidth}
            \includegraphics[width=\linewidth]{fig/NA.pdf}
            \end{minipage}	
            \begin{minipage}{0.13\linewidth}
            \includegraphics[width=\linewidth]{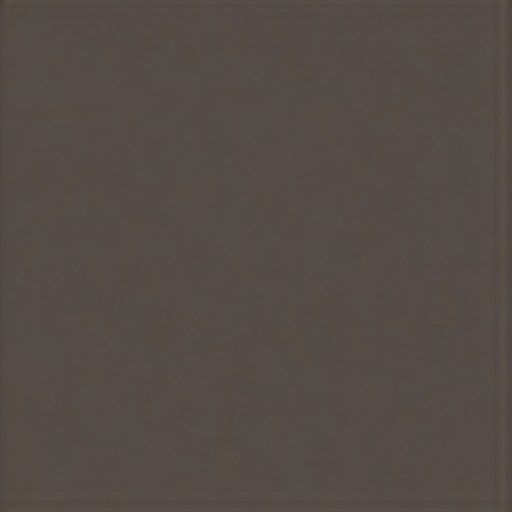}
            \end{minipage}	
            \begin{minipage}{0.13\linewidth}
            \includegraphics[width=\linewidth]{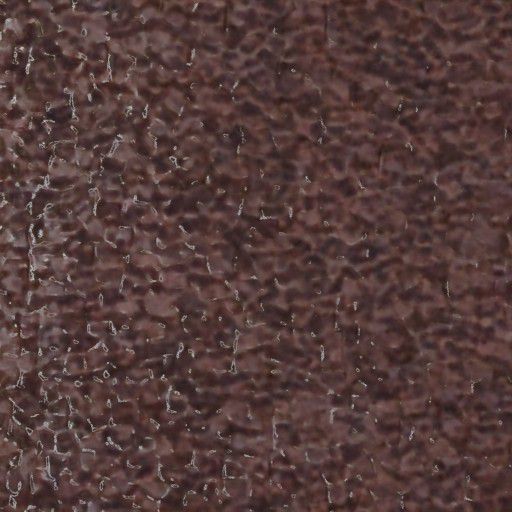}
            \put(-21,2){\tiny \color{white}0.243}
            \end{minipage}	
        \end{minipage}	
    \end{minipage}	
    \begin{tikzpicture}
    \draw[dashed, color=darkred,line width=0.1pt] (-5,0) -- (3.5,0); 
    \end{tikzpicture}
    \begin{minipage}{3.4in}
        \begin{minipage}{0.02in}	
            \centering
                \rotatebox{90}{\tiny Ours}
        \end{minipage}	
        \hspace{0.02in}
         \begin{minipage}{3.3in}	
            \centering
            \begin{minipage}{0.13\linewidth}
            \includegraphics[width=\linewidth]{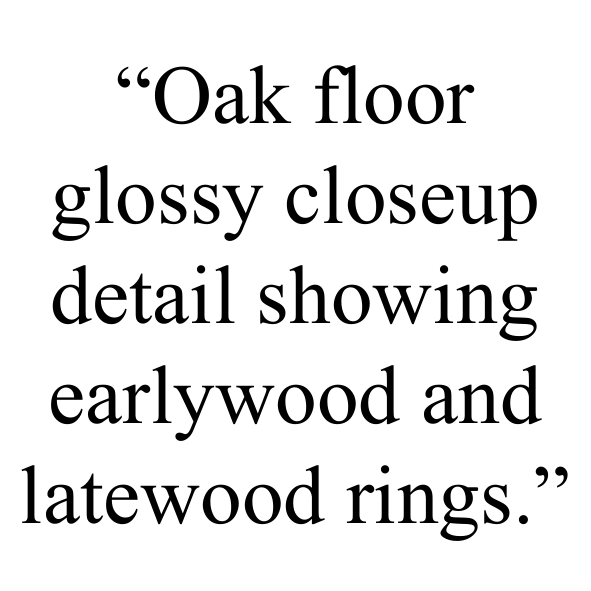}
            \end{minipage}	
            \begin{minipage}{0.13\linewidth}
            \includegraphics[width=\linewidth]{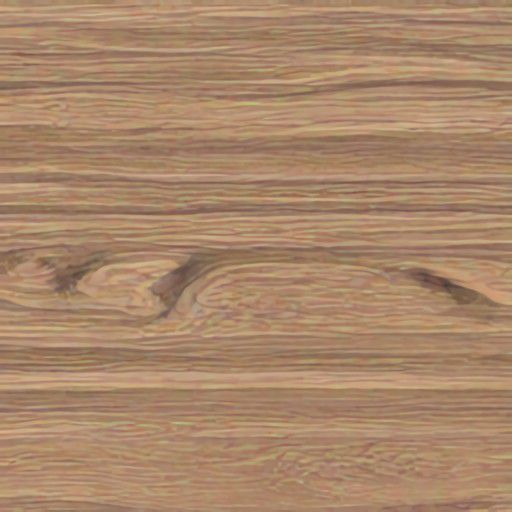}
            \end{minipage}	
            \begin{minipage}{0.13\linewidth}
            \includegraphics[width=\linewidth]{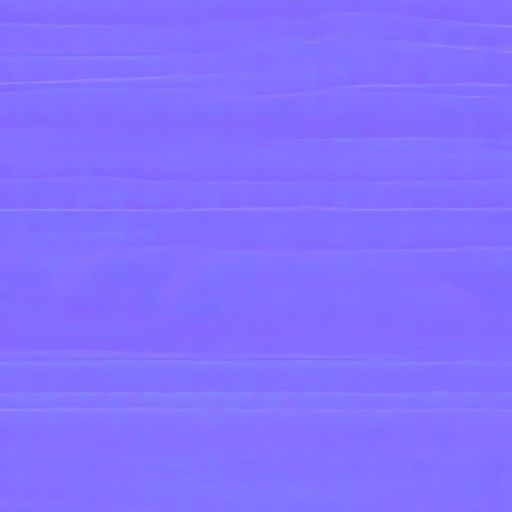}
            \end{minipage}	
            \begin{minipage}{0.13\linewidth}
            \includegraphics[width=\linewidth]{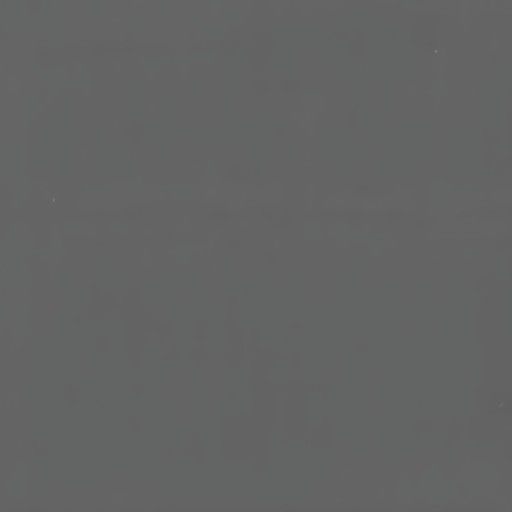}
            \end{minipage}	
            \begin{minipage}{0.13\linewidth}
            \includegraphics[width=\linewidth]{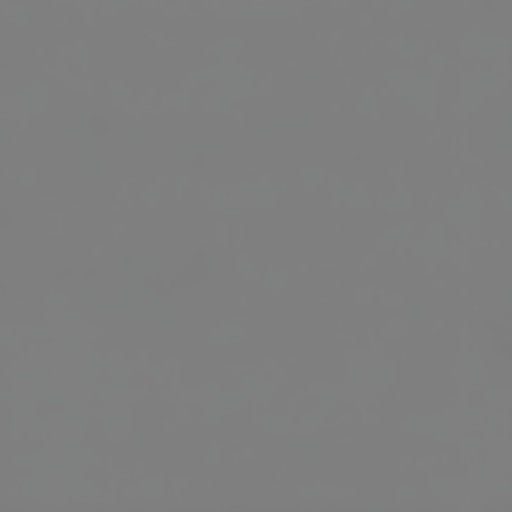}
            \end{minipage}	
            \begin{minipage}{0.13\linewidth}
            \includegraphics[width=\linewidth]{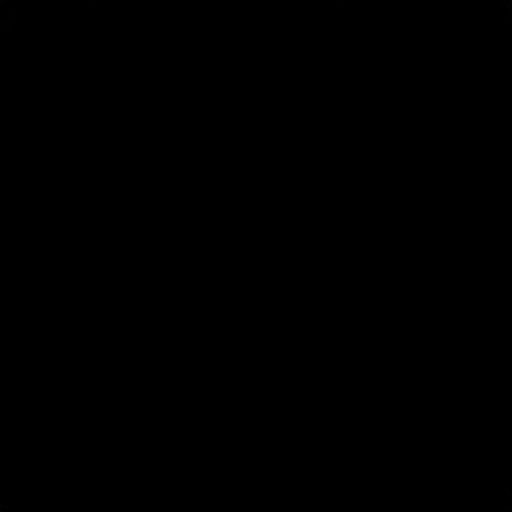}
            \end{minipage}	
            \begin{minipage}{0.13\linewidth}
            \includegraphics[width=\linewidth]{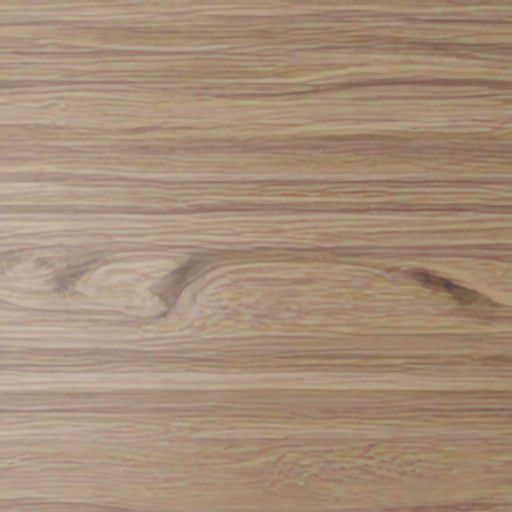}
            \put(-21,2){\tiny \textbf {\color{white}0.228}}
            \end{minipage}	
        \end{minipage}	
    \end{minipage}	

    \begin{minipage}{3.4in}
        \begin{minipage}{0.02in}	
            \centering
                \rotatebox{90}{\tiny MatGen}
        \end{minipage}	
        \hspace{0.02in}
         \begin{minipage}{3.3in}	
            \centering
            \begin{minipage}{0.13\linewidth}
            \includegraphics[width=\linewidth]{fig/text_gen_comparison/wood.pdf}
            \end{minipage}	
            \begin{minipage}{0.13\linewidth}
            \includegraphics[width=\linewidth]{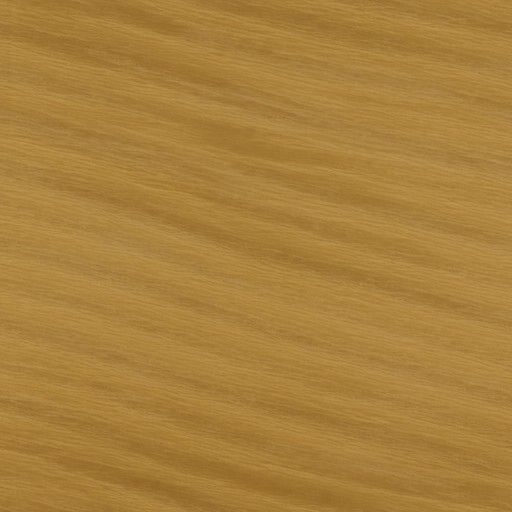}
            \end{minipage}	
            \begin{minipage}{0.13\linewidth}
            \includegraphics[width=\linewidth]{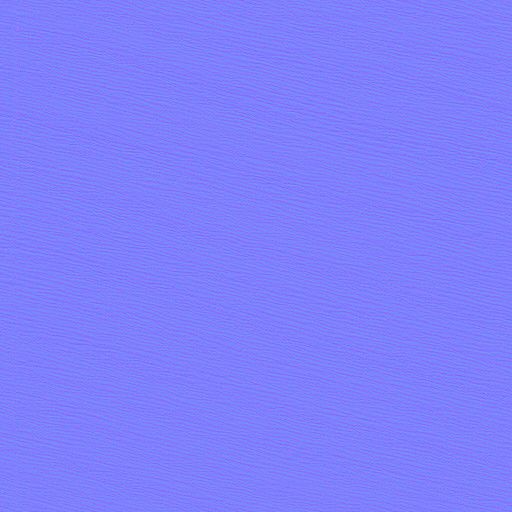}
            \end{minipage}	
            \begin{minipage}{0.13\linewidth}
            \includegraphics[width=\linewidth]{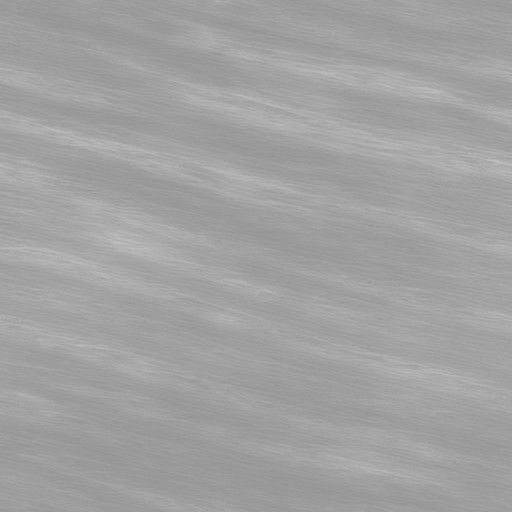}
            \end{minipage}	
            \begin{minipage}{0.13\linewidth}
            \includegraphics[width=\linewidth]{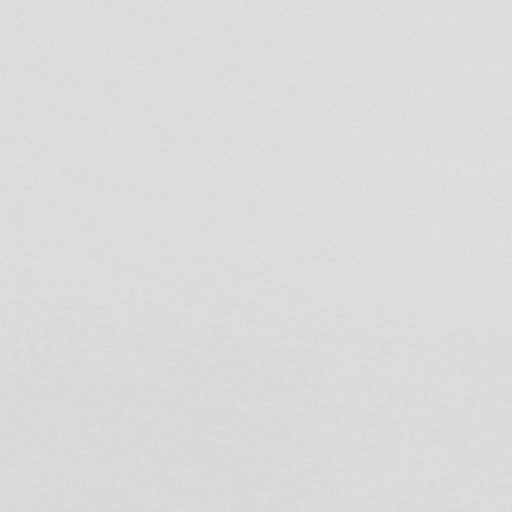}
            \end{minipage}	
            \begin{minipage}{0.13\linewidth}
            \includegraphics[width=\linewidth]{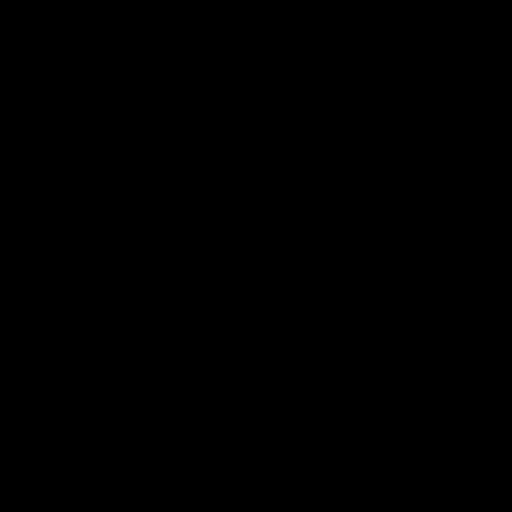}
            \end{minipage}	
            \begin{minipage}{0.13\linewidth}
            \includegraphics[width=\linewidth]{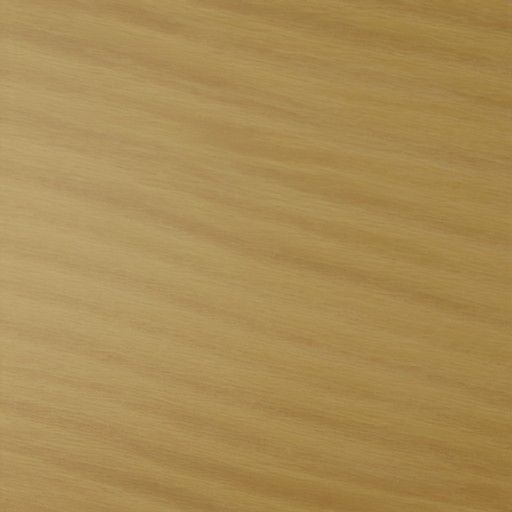}
            \put(-21,2){\tiny \color{white}0.218}
            \end{minipage}	
        \end{minipage}	
    \end{minipage}	

    \begin{minipage}{3.4in}
        \begin{minipage}{0.02in}	
            \centering
                \rotatebox{90}{\tiny MatFuse}
        \end{minipage}	
        \hspace{0.02in}
         \begin{minipage}{3.3in}	
            \centering
            \begin{minipage}{0.13\linewidth}
            \includegraphics[width=\linewidth]{fig/text_gen_comparison/wood.pdf}
            \end{minipage}	
            \begin{minipage}{0.13\linewidth}
            \includegraphics[width=\linewidth]{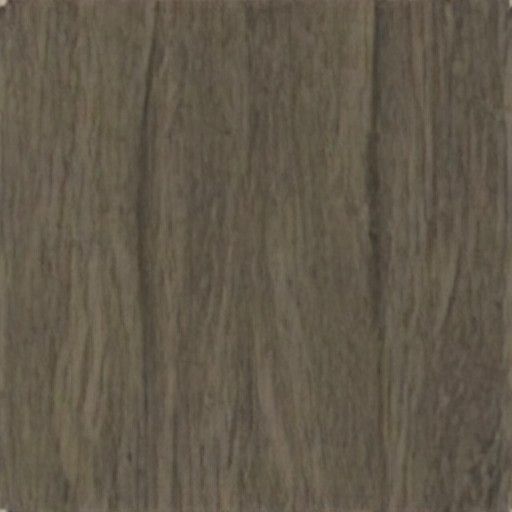}
            \end{minipage}	
            \begin{minipage}{0.13\linewidth}
            \includegraphics[width=\linewidth]{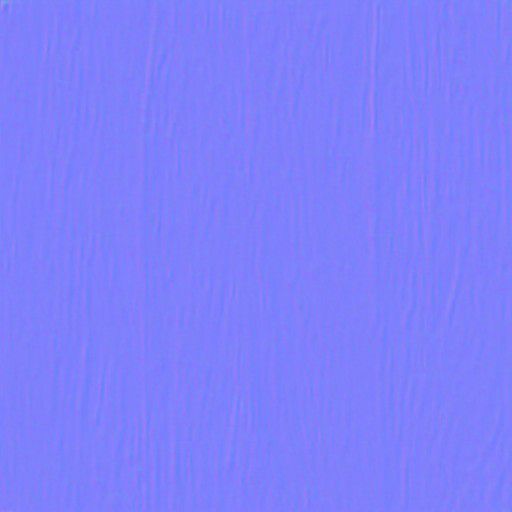}
            \end{minipage}	
            \begin{minipage}{0.13\linewidth}
            \includegraphics[width=\linewidth]{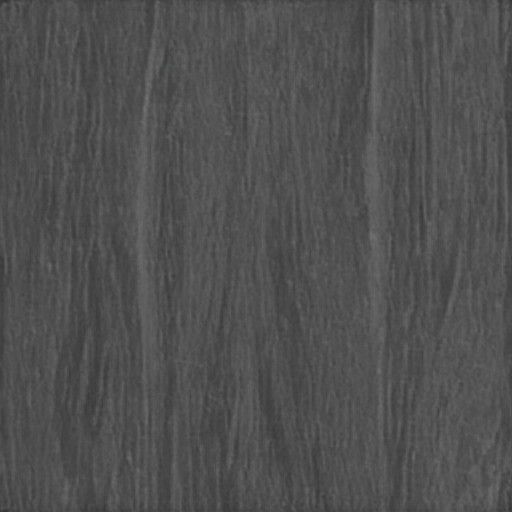}
            \end{minipage}	
            \begin{minipage}{0.13\linewidth}
            \includegraphics[width=\linewidth]{fig/NA.pdf}
            \end{minipage}	
            \begin{minipage}{0.13\linewidth}
            \includegraphics[width=\linewidth]{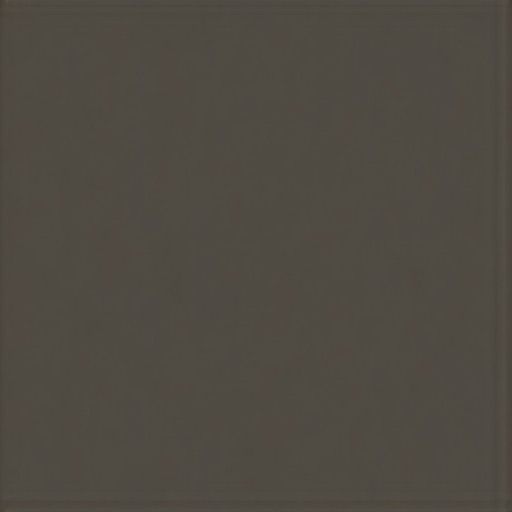}
            \end{minipage}	
            \begin{minipage}{0.13\linewidth}
            \includegraphics[width=\linewidth]{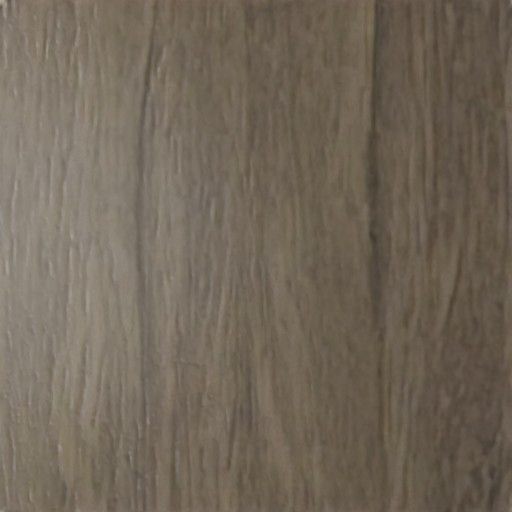}
            \put(-21,2){\tiny \color{white}0.220}
            \end{minipage}	
        \end{minipage}	
    \end{minipage}	
    \begin{tikzpicture}
    \draw[dashed, color=darkred,line width=0.1pt] (-5,0) -- (3.5,0); 
    \end{tikzpicture}
    \begin{minipage}{3.4in}
        \begin{minipage}{0.02in}	
            \centering
                \rotatebox{90}{\tiny Ours}
        \end{minipage}	
        \hspace{0.02in}
         \begin{minipage}{3.3in}	
            \centering
            \begin{minipage}{0.13\linewidth}
            \includegraphics[width=\linewidth]{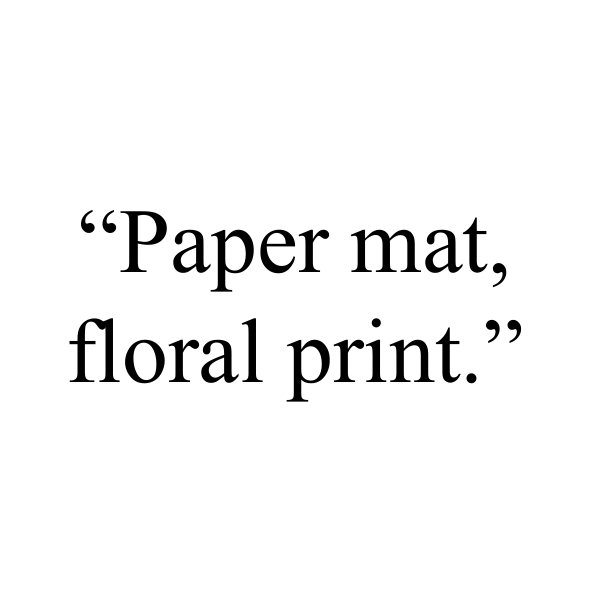}
            \end{minipage}	
            \begin{minipage}{0.13\linewidth}
            \includegraphics[width=\linewidth]{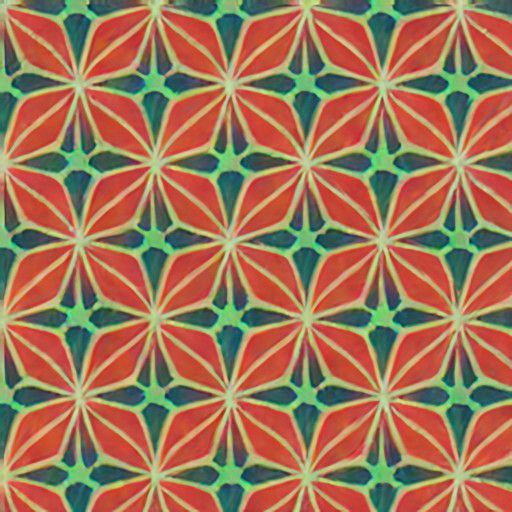}
            \end{minipage}	
            \begin{minipage}{0.13\linewidth}
            \includegraphics[width=\linewidth]{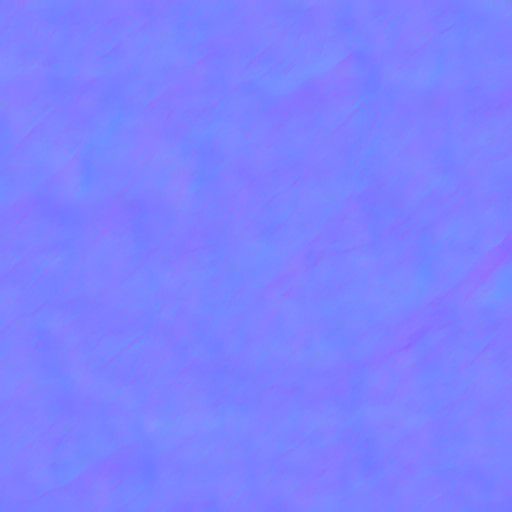}
            \end{minipage}	
            \begin{minipage}{0.13\linewidth}
            \includegraphics[width=\linewidth]{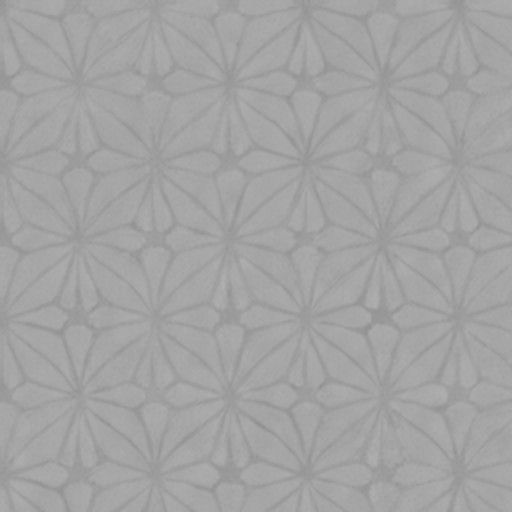}
            \end{minipage}	
            \begin{minipage}{0.13\linewidth}
            \includegraphics[width=\linewidth]{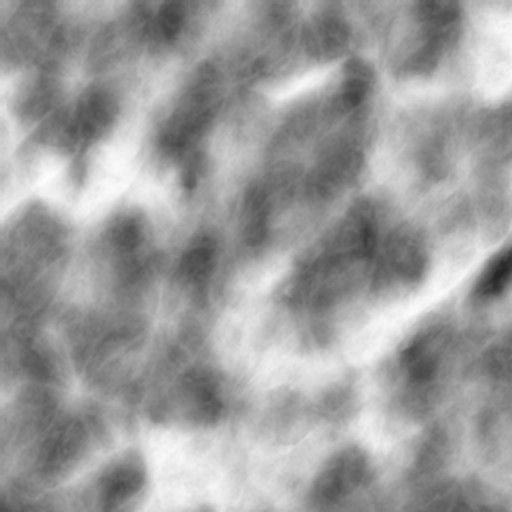}
            \end{minipage}	
            \begin{minipage}{0.13\linewidth}
            \includegraphics[width=\linewidth]{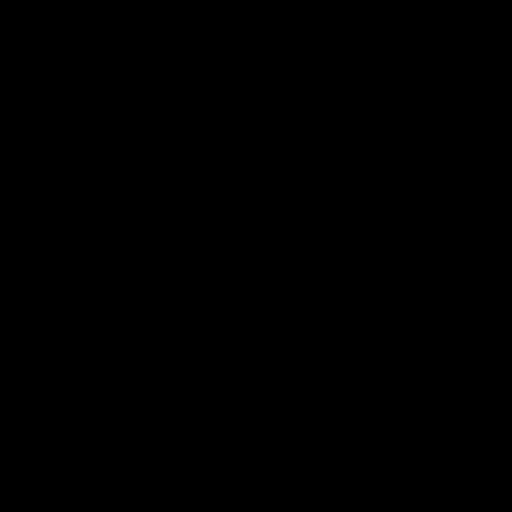}
            \end{minipage}	
            \begin{minipage}{0.13\linewidth}
            \includegraphics[width=\linewidth]{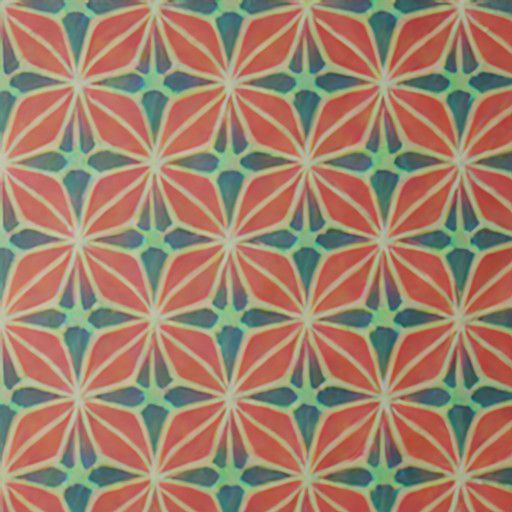}
            \put(-21,2){\tiny \textbf {\color{white}0.237}}
            \end{minipage}	
        \end{minipage}	
    \end{minipage}	
    \begin{minipage}{3.4in}
        \begin{minipage}{0.02in}	
            \centering
                \rotatebox{90}{\tiny MatGen}
        \end{minipage}	
        \hspace{0.02in}
         \begin{minipage}{3.3in}	
            \centering
            \begin{minipage}{0.13\linewidth}
            \includegraphics[width=\linewidth]{fig/text_gen_comparison/paper.pdf}
            \end{minipage}	
            \begin{minipage}{0.13\linewidth}
            \includegraphics[width=\linewidth]{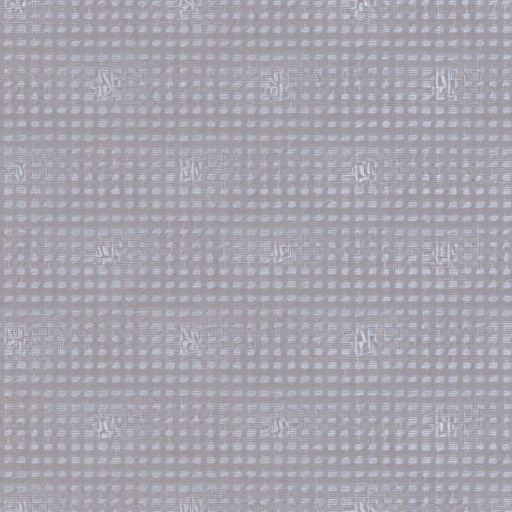}
            \end{minipage}	
            \begin{minipage}{0.13\linewidth}
            \includegraphics[width=\linewidth]{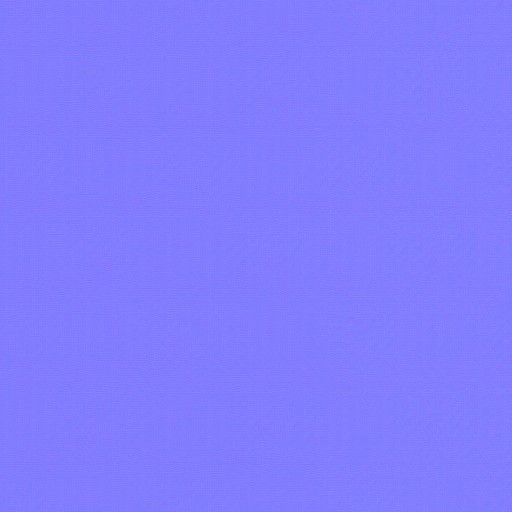}
            \end{minipage}	
            \begin{minipage}{0.13\linewidth}
            \includegraphics[width=\linewidth]{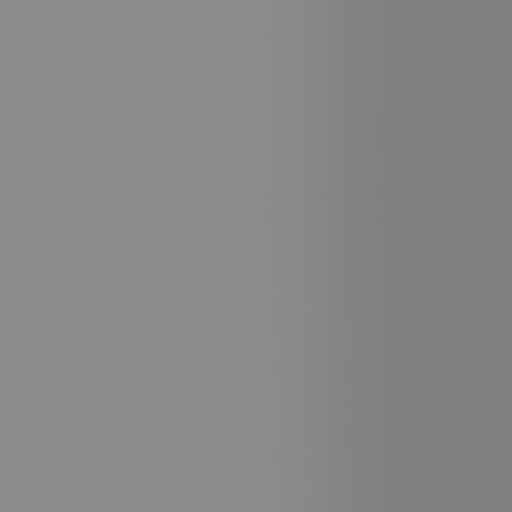}
            \end{minipage}	
            \begin{minipage}{0.13\linewidth}
            \includegraphics[width=\linewidth]{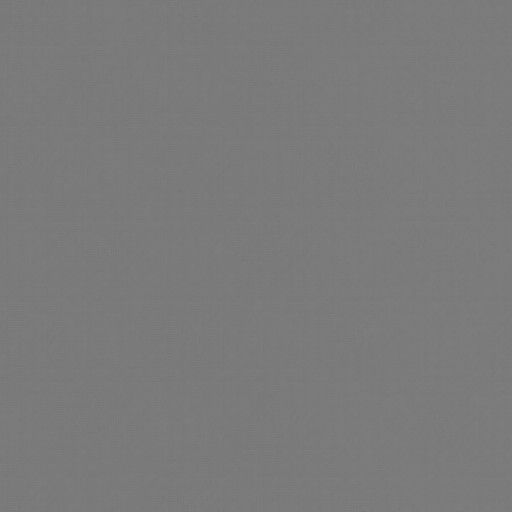}
            \end{minipage}	
            \begin{minipage}{0.13\linewidth}
            \includegraphics[width=\linewidth]{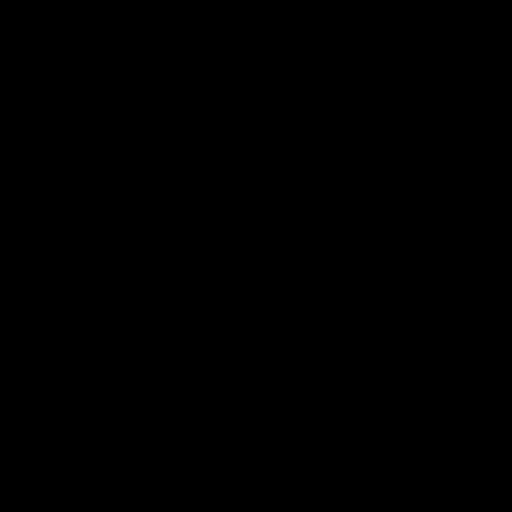}
            \end{minipage}	
            \begin{minipage}{0.13\linewidth}
            \includegraphics[width=\linewidth]{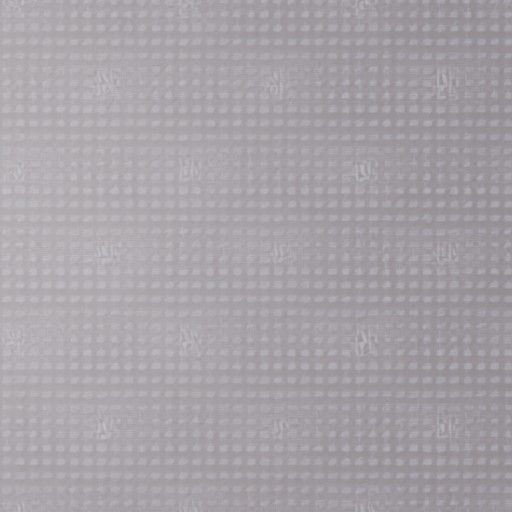}
            \put(-21,2){\tiny \color{white}0.220}
            \end{minipage}	
        \end{minipage}	
    \end{minipage}	

    \begin{minipage}{3.4in}
        \begin{minipage}{0.02in}	
            \centering
                \rotatebox{90}{\tiny MatFuse}
        \end{minipage}	
        \hspace{0.02in}
         \begin{minipage}{3.3in}	
            \centering
            \begin{minipage}{0.13\linewidth}
            \includegraphics[width=\linewidth]{fig/text_gen_comparison/paper.pdf}
            \end{minipage}	
            \begin{minipage}{0.13\linewidth}
            \includegraphics[width=\linewidth]{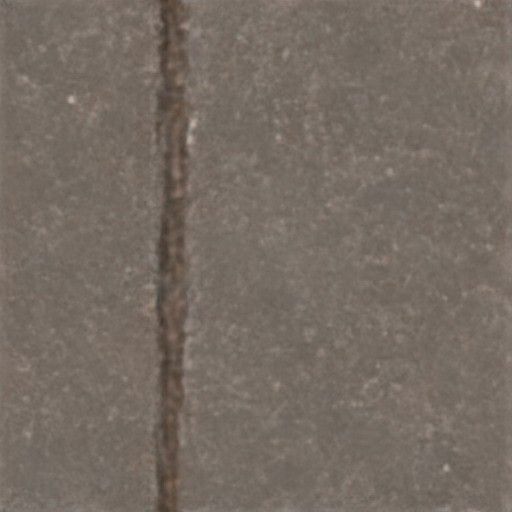}
            \end{minipage}	
            \begin{minipage}{0.13\linewidth}
            \includegraphics[width=\linewidth]{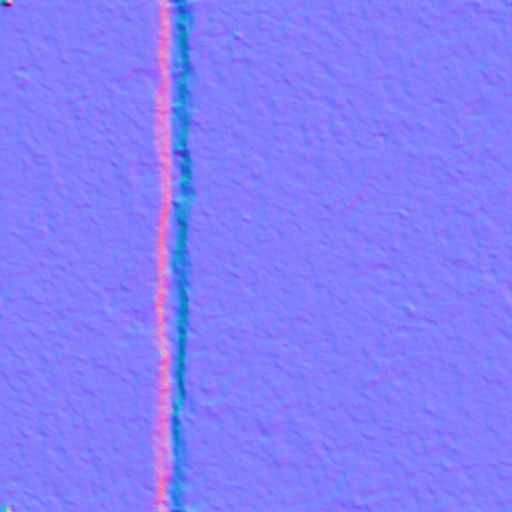}
            \end{minipage}	
            \begin{minipage}{0.13\linewidth}
            \includegraphics[width=\linewidth]{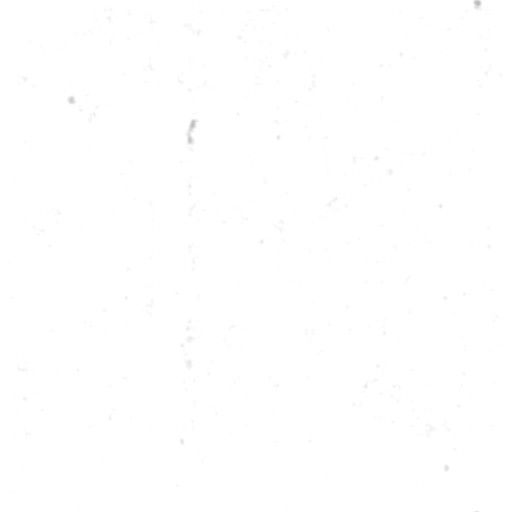}
            \end{minipage}	
            \begin{minipage}{0.13\linewidth}
            \includegraphics[width=\linewidth]{fig/NA.pdf}
            \end{minipage}	
            \begin{minipage}{0.13\linewidth}
            \includegraphics[width=\linewidth]{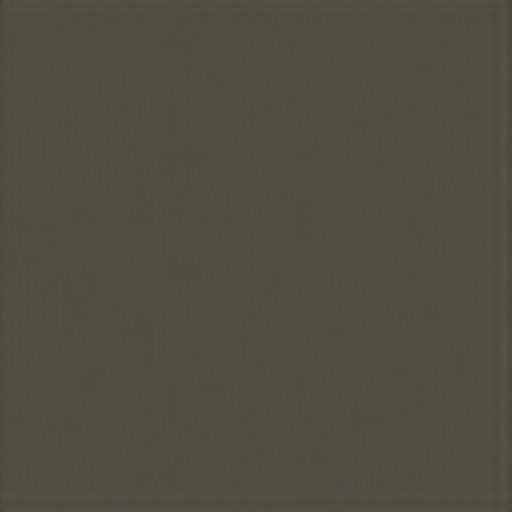}
            \end{minipage}	
            \begin{minipage}{0.13\linewidth}
            \includegraphics[width=\linewidth]{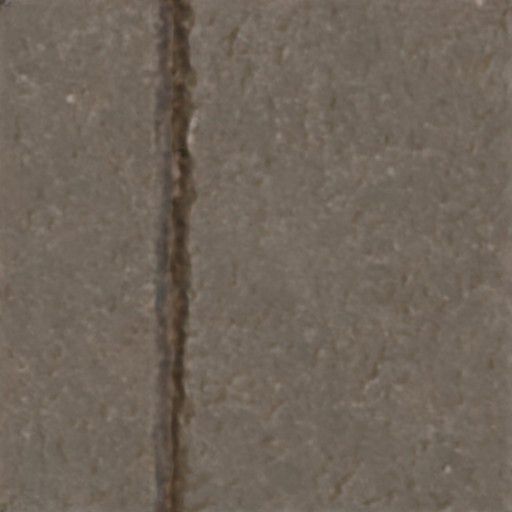}
            \put(-21,2){\tiny \color{white}0.132}
            \end{minipage}	
        \end{minipage}	
    \end{minipage}	
   \caption{Comparison of text-to-material generation between our model, MatGen~\cite{vecchio2024controlmat}, and MatFuse~\cite{vecchio2024matfuse}. The "Text" column contains the input text conditions. The second to last columns show the generated material maps, along with a rendering under environment lighting. The CLIP score between the rendering and the text is reported at the bottom of each rendered image (higher is better). We note that MatFuse generates a specular map rather than a metallic map. } 
   \label{fig:text_generation}
\end{figure}

Although the primary focus of our method is the generation of materials from photos, our multi-modal model also supports text-conditioned generation without image inputs. We evaluate its performance on the text-to-material task, comparing it with two state-of-the-art diffusion-based generative models for material synthesis: MatFuse~\cite{vecchio2024matfuse} and MatGen~\cite{vecchio2024controlmat}. As shown in Fig.~\ref{fig:text_generation}, our model demonstrates strong text-to-material synthesis capability, producing high-quality material samples, comparable to other state-of-the-art approaches. We report the cosine similarity between each rendered image and the text condition, calculated using ViT-L-14 CLIP~\cite{radford2021learning} embeddings. Leveraging a pretrained text-to-video model as a prior, our model can interpret complex semantics beyond the material-only training set, such as "wood rings" and "floral" patterns. 

\subsection{Ablation Study}
\label{sec:ablation}
\subsubsection{Multi-modality}
\begin{figure}[htbp!]
    \centering		
    \begin{minipage}{3.5in}
        
         \begin{minipage}{3.5in}	
            % \centering
            \begin{minipage}{0.13\linewidth}
                \subcaption*{\tiny Input}
                \includegraphics[width=\linewidth]{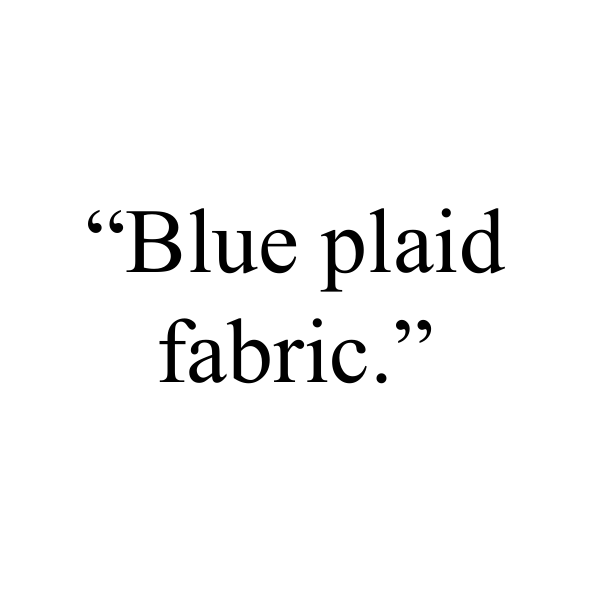}
            \end{minipage}
            \begin{minipage}{0.13\linewidth}
                \subcaption*{\tiny Albedo}
                \includegraphics[width=\linewidth]{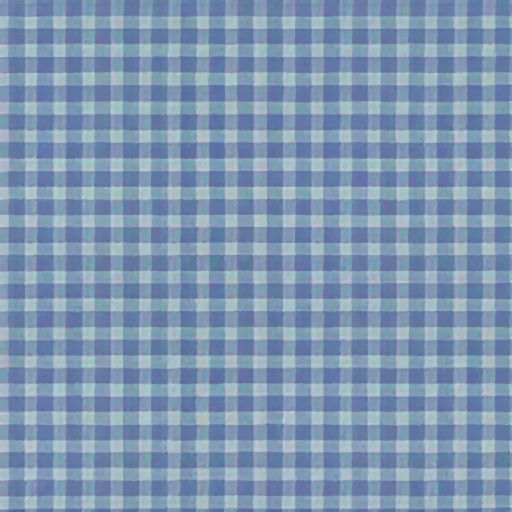}
            \end{minipage}
            \begin{minipage}{0.13\linewidth}
                \subcaption*{\tiny Normal}
                \includegraphics[width=\linewidth]{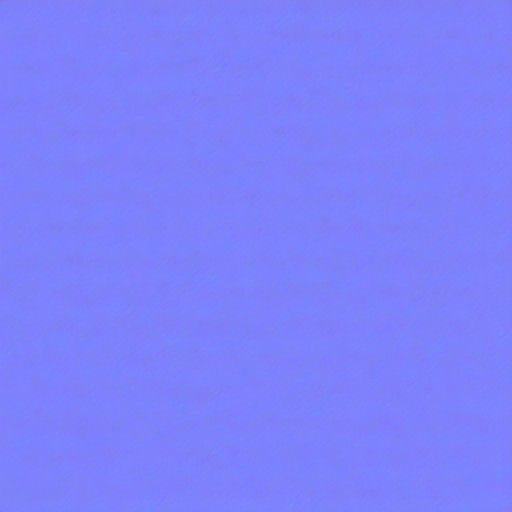}
            \end{minipage}
            \begin{minipage}{0.13\linewidth}
                \subcaption*{\tiny Roughness}
                \includegraphics[width=\linewidth]{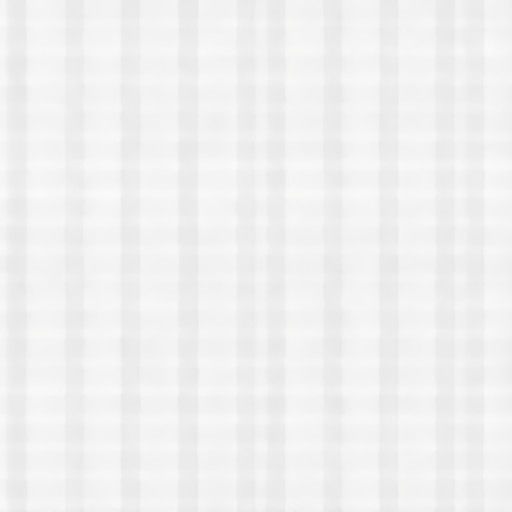}
            \end{minipage}
            \begin{minipage}{0.13\linewidth}
                \subcaption*{\tiny Height}
                \includegraphics[width=\linewidth]{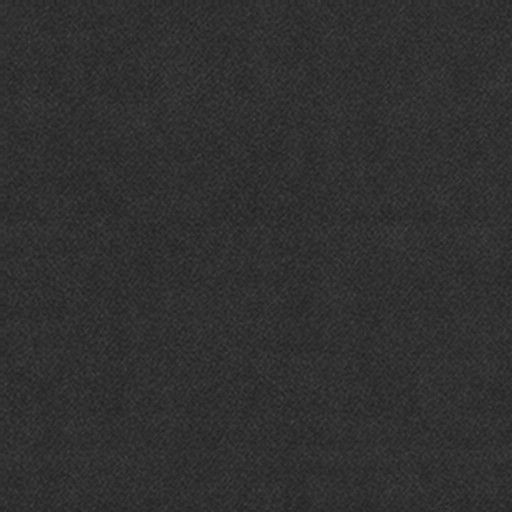}
            \end{minipage}
            \begin{minipage}{0.13\linewidth}
                \subcaption*{\tiny Metallic}
                \includegraphics[width=\linewidth]{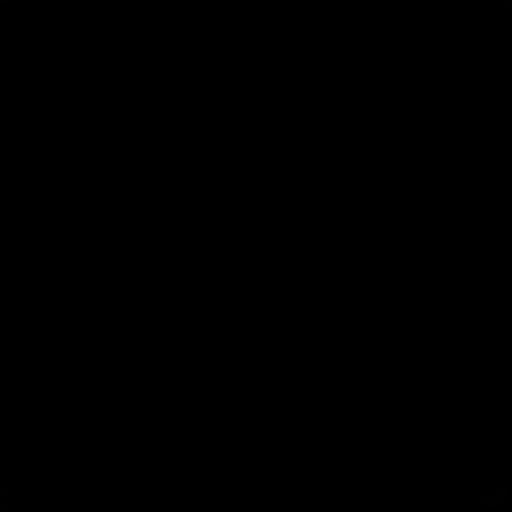}
            \end{minipage}
            \begin{minipage}{0.13\linewidth}
                \subcaption*{\tiny Render}
                \includegraphics[width=\linewidth]{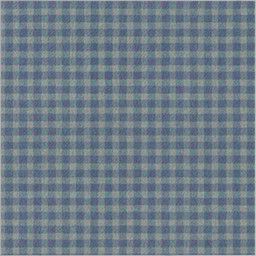}
            \end{minipage}
        \end{minipage}	
    \end{minipage}	

    \begin{minipage}{3.5in}
         \begin{minipage}{3.5in}	
            % \centering
            \begin{minipage}{0.13\linewidth}
            \includegraphics[width=\linewidth]{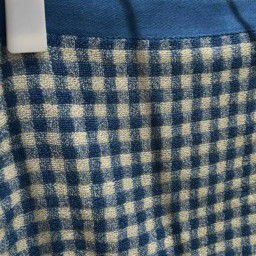}
            \end{minipage}	
            \begin{minipage}{0.13\linewidth}
            \includegraphics[width=\linewidth]{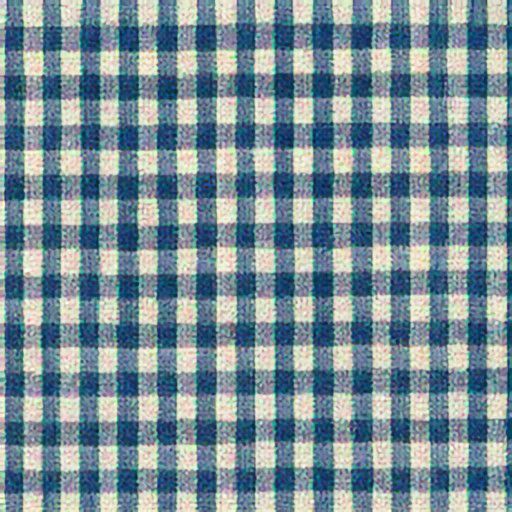}
            \end{minipage}	
            \begin{minipage}{0.13\linewidth}          
            \includegraphics[width=\linewidth]{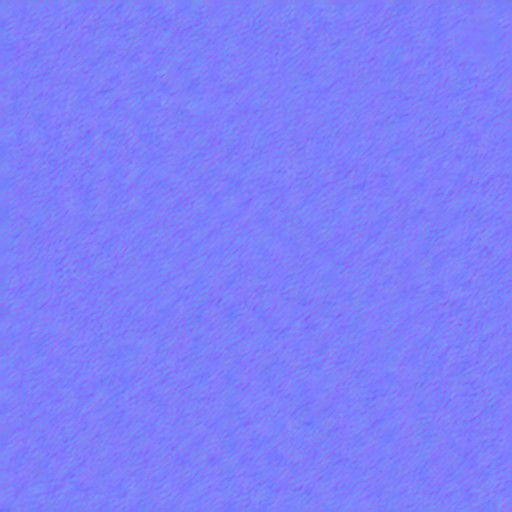}
            \end{minipage}	
            \begin{minipage}{0.13\linewidth}          
            \includegraphics[width=\linewidth]{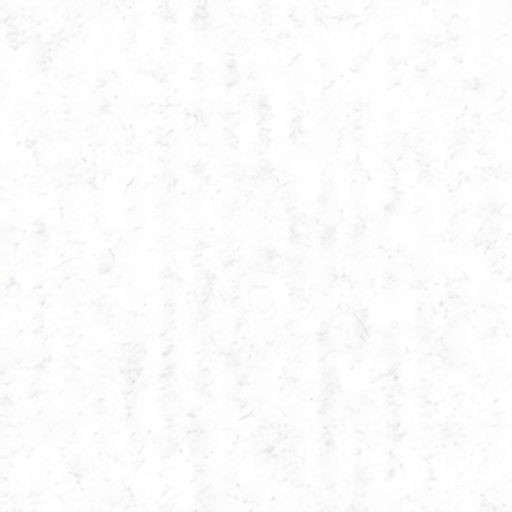}
            \end{minipage}	
            \begin{minipage}{0.13\linewidth}          
            \includegraphics[width=\linewidth]{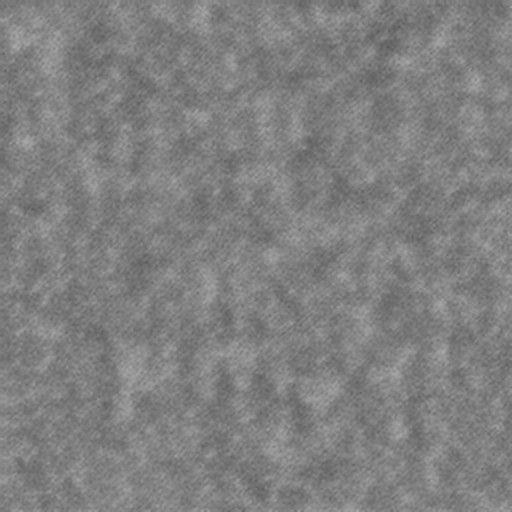}
            \end{minipage}	
            \begin{minipage}{0.13\linewidth}           
            \includegraphics[width=\linewidth]{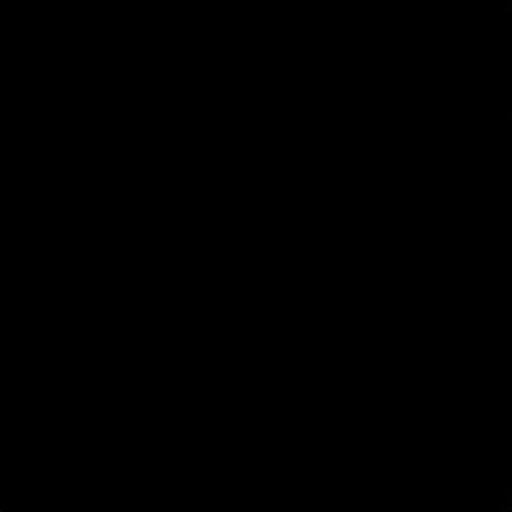}
            \end{minipage}	
            \begin{minipage}{0.13\linewidth}           
            \includegraphics[width=\linewidth]{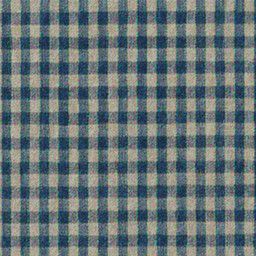}
            \end{minipage}	
        \end{minipage}	
    \end{minipage}	

    \begin{minipage}{3.5in}
         \begin{minipage}{3.5in}	
            % \centering
            \begin{minipage}{0.13\linewidth}
            \includegraphics[width=\linewidth]{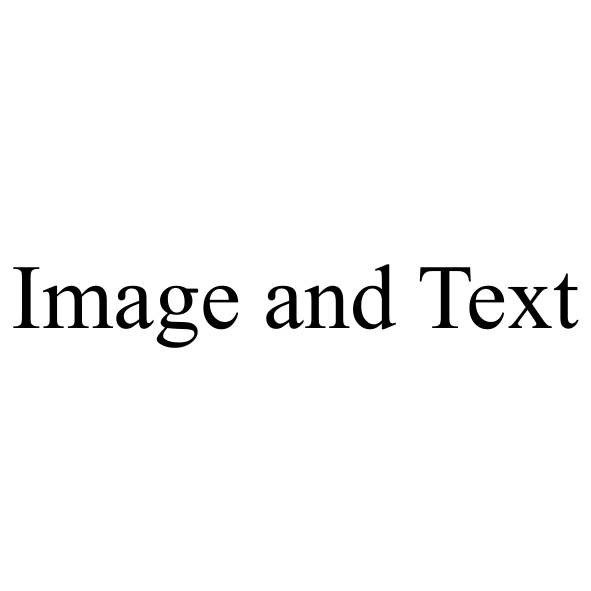}
            \end{minipage}	
            \begin{minipage}{0.13\linewidth}
            \includegraphics[width=\linewidth]{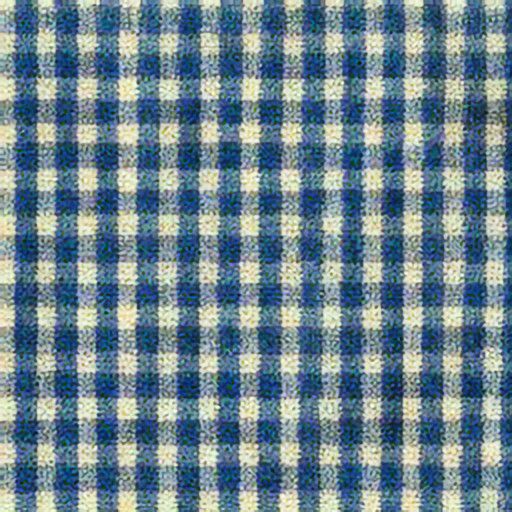}
            \end{minipage}	
            \begin{minipage}{0.13\linewidth}            
            \includegraphics[width=\linewidth]{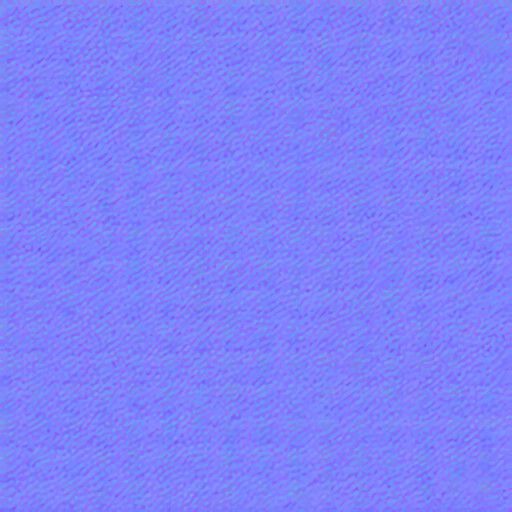}
            \end{minipage}	
            \begin{minipage}{0.13\linewidth}           
            \includegraphics[width=\linewidth]{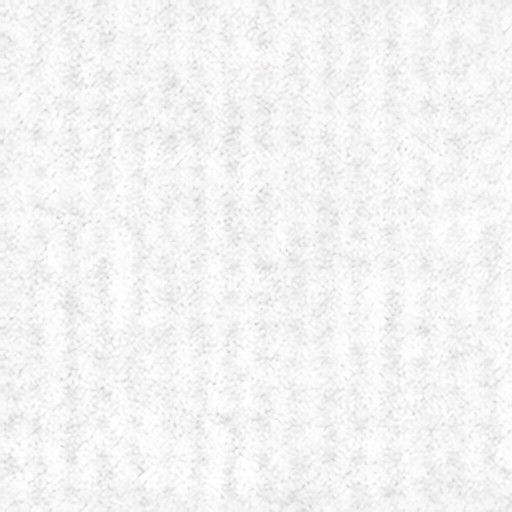}
            \end{minipage}	
            \begin{minipage}{0.13\linewidth}           
            \includegraphics[width=\linewidth]{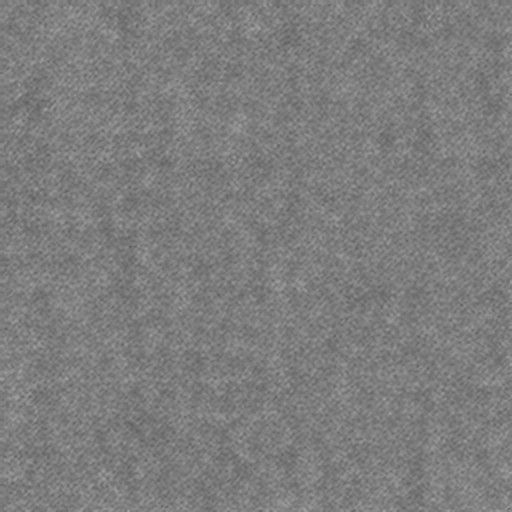}
            \end{minipage}	
            \begin{minipage}{0.13\linewidth}           
            \includegraphics[width=\linewidth]{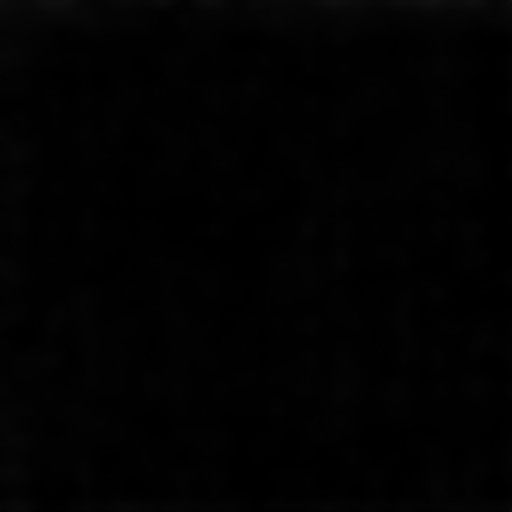}
            \end{minipage}	
            \begin{minipage}{0.13\linewidth}           
            \includegraphics[width=\linewidth]{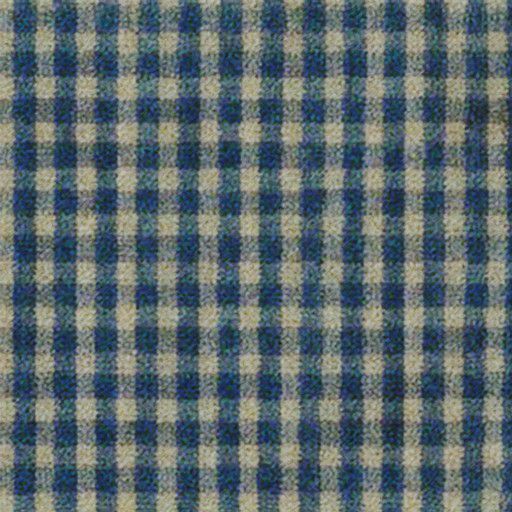}
            \end{minipage}	
        \end{minipage}	
    \end{minipage}

    \begin{minipage}{3.5in}
         \begin{minipage}{3.5in}	
            % \centering
            \begin{minipage}{0.13\linewidth}
            \includegraphics[width=\linewidth]{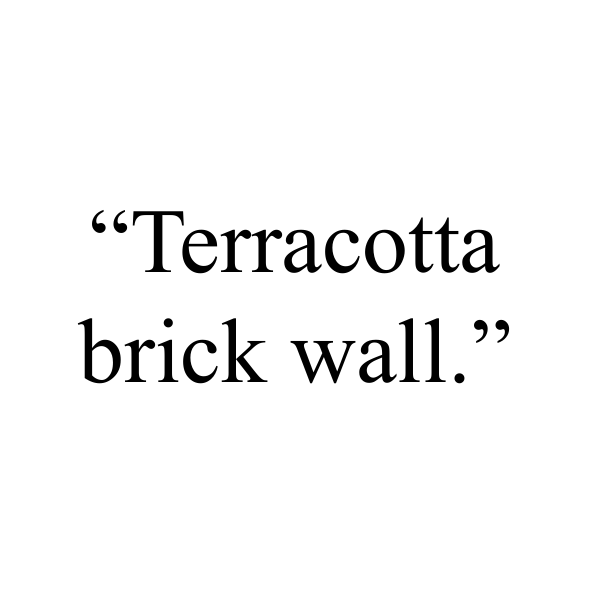}
            \end{minipage}	
            \begin{minipage}{0.13\linewidth}
            \includegraphics[width=\linewidth]{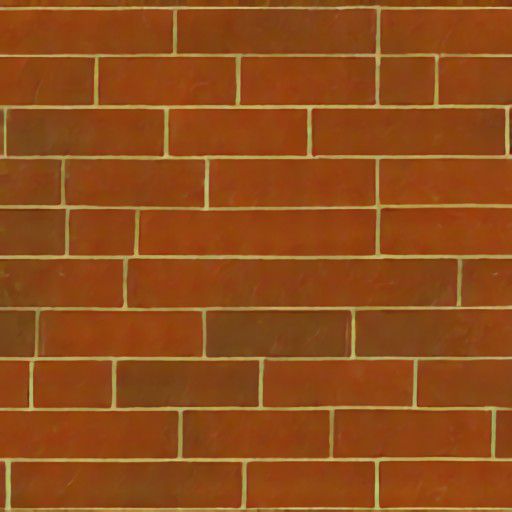}
            \end{minipage}	
            \begin{minipage}{0.13\linewidth}            
            \includegraphics[width=\linewidth]{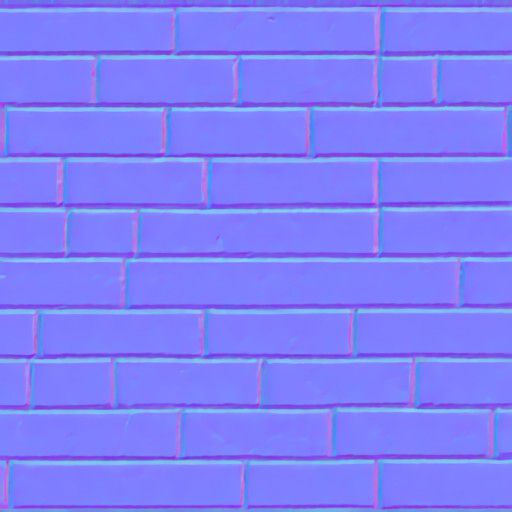}
            \end{minipage}	
            \begin{minipage}{0.13\linewidth}           
            \includegraphics[width=\linewidth]{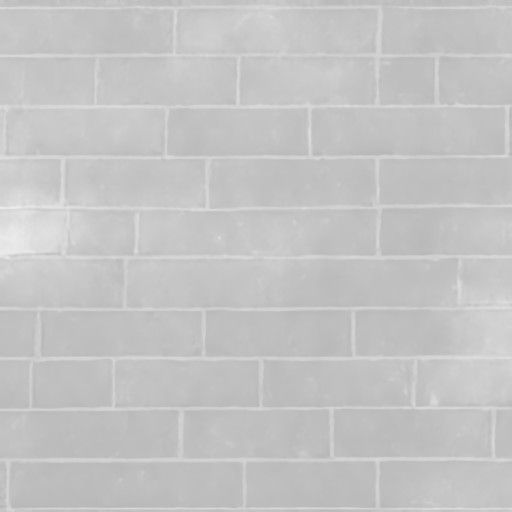}
            \end{minipage}	
            \begin{minipage}{0.13\linewidth}           
            \includegraphics[width=\linewidth]{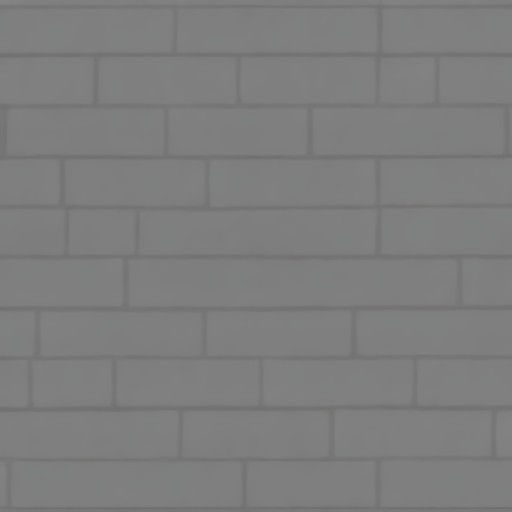}
            \end{minipage}	
            \begin{minipage}{0.13\linewidth}           
            \includegraphics[width=\linewidth]{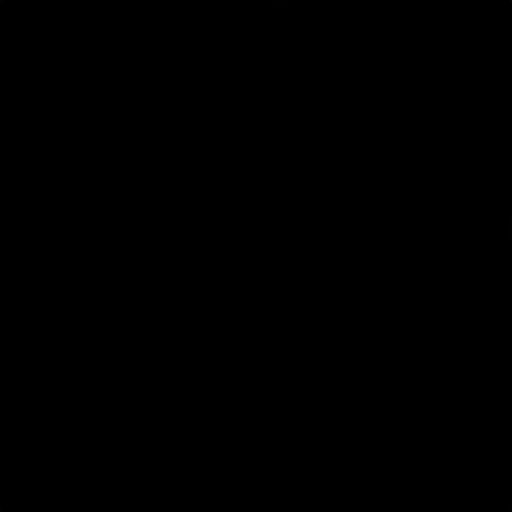}
            \end{minipage}	
            \begin{minipage}{0.13\linewidth}           
            \includegraphics[width=\linewidth]{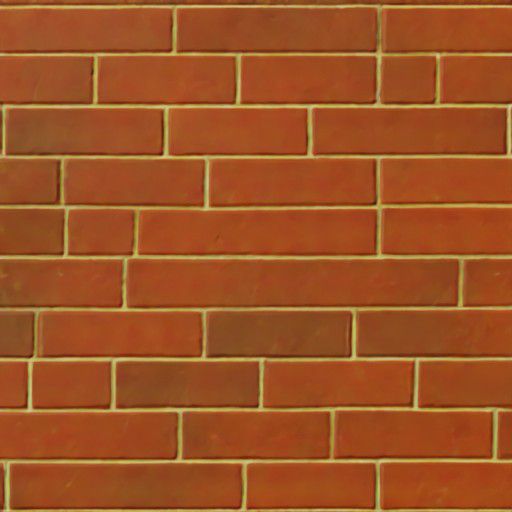}
            \end{minipage}	
        \end{minipage}	
    \end{minipage}

    \begin{minipage}{3.5in}
         \begin{minipage}{3.5in}	
            % \centering
            \begin{minipage}{0.13\linewidth}
            \includegraphics[width=\linewidth]{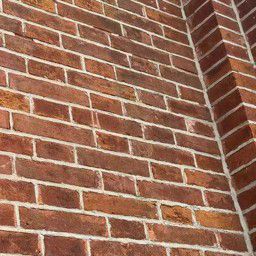}
            \end{minipage}	
            \begin{minipage}{0.13\linewidth}
            \includegraphics[width=\linewidth]{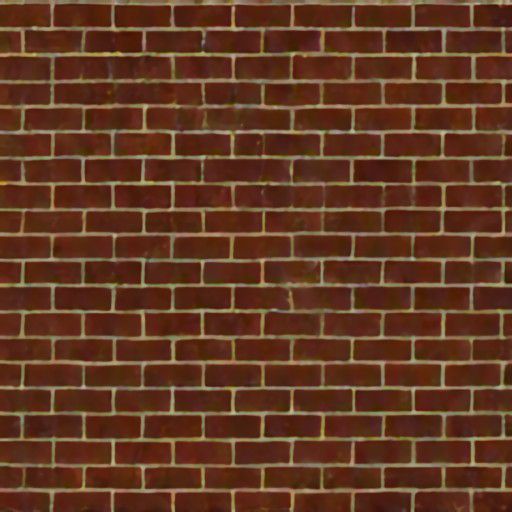}
            \end{minipage}	
            \begin{minipage}{0.13\linewidth}            
            \includegraphics[width=\linewidth]{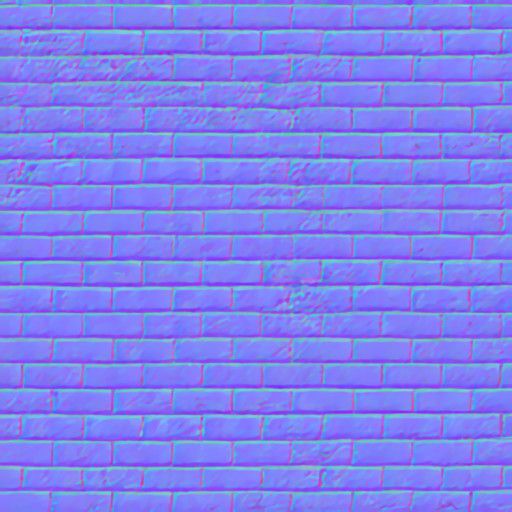}
            \end{minipage}	
            \begin{minipage}{0.13\linewidth}           
            \includegraphics[width=\linewidth]{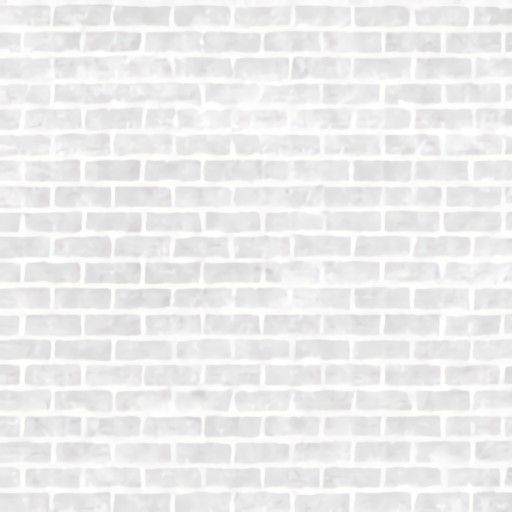}
            \end{minipage}	
            \begin{minipage}{0.13\linewidth}           
            \includegraphics[width=\linewidth]{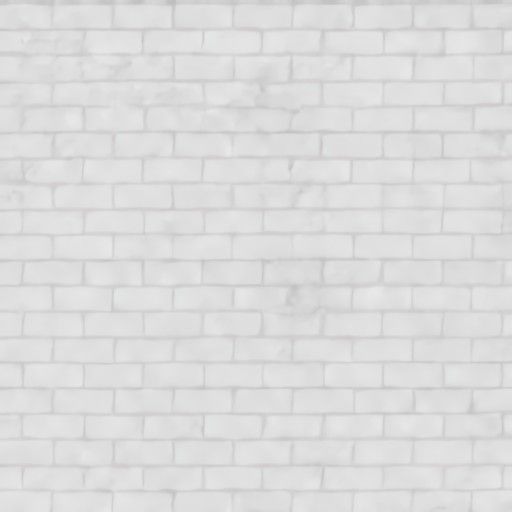}
            \end{minipage}	
            \begin{minipage}{0.13\linewidth}           
            \includegraphics[width=\linewidth]{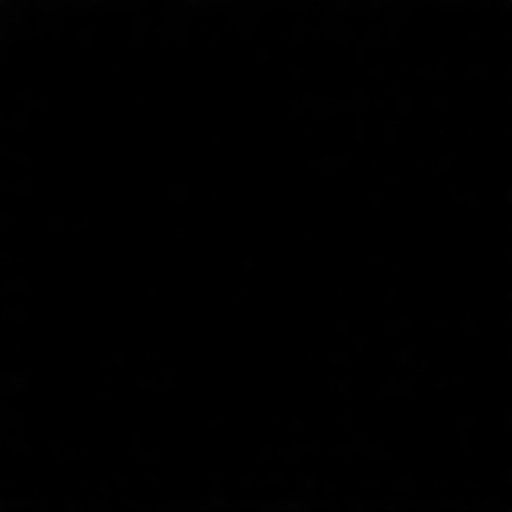}
            \end{minipage}	
            \begin{minipage}{0.13\linewidth}           
            \includegraphics[width=\linewidth]{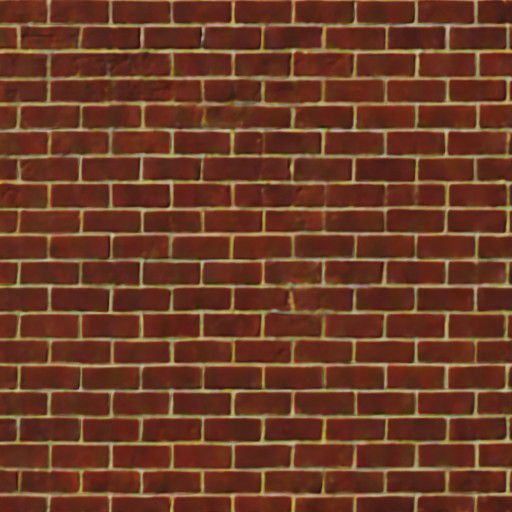}
            \end{minipage}	
        \end{minipage}	
    \end{minipage}

    \begin{minipage}{3.5in}
         \begin{minipage}{3.5in}	
            % \centering
            \begin{minipage}{0.13\linewidth}
            \includegraphics[width=\linewidth]{fig/multi_modal/cond.pdf}
            % \put(-28,13){\parbox{80mm}{\centering \fontsize{5}{5}\selectfont Text + \\ Image}}
            \end{minipage}	
            \begin{minipage}{0.13\linewidth}
            \includegraphics[width=\linewidth]{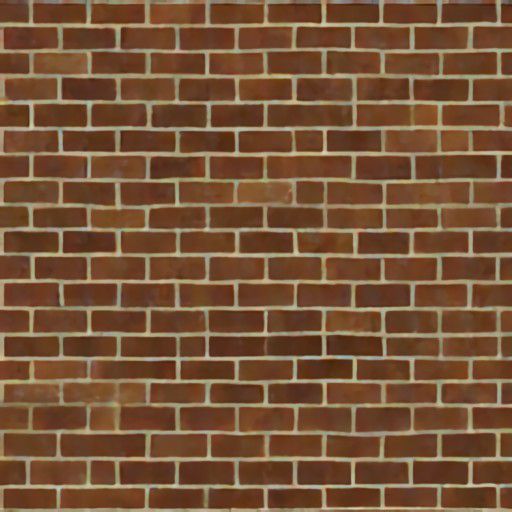}
            \end{minipage}	
            \begin{minipage}{0.13\linewidth}            
            \includegraphics[width=\linewidth]{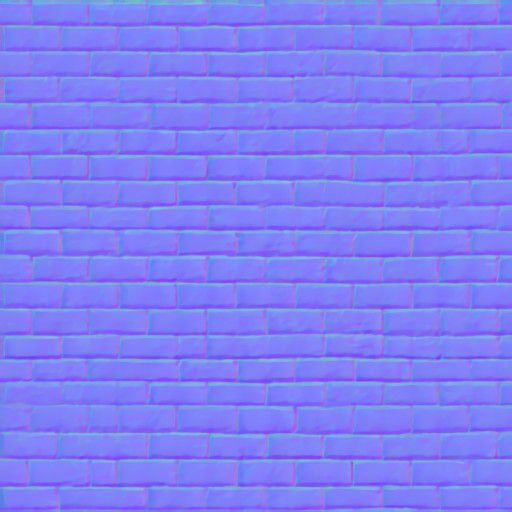}
            \end{minipage}	
            \begin{minipage}{0.13\linewidth}           
            \includegraphics[width=\linewidth]{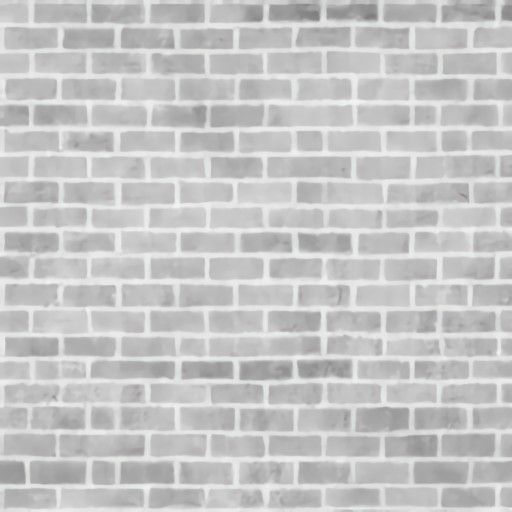}
            \end{minipage}	
            \begin{minipage}{0.13\linewidth}           
            \includegraphics[width=\linewidth]{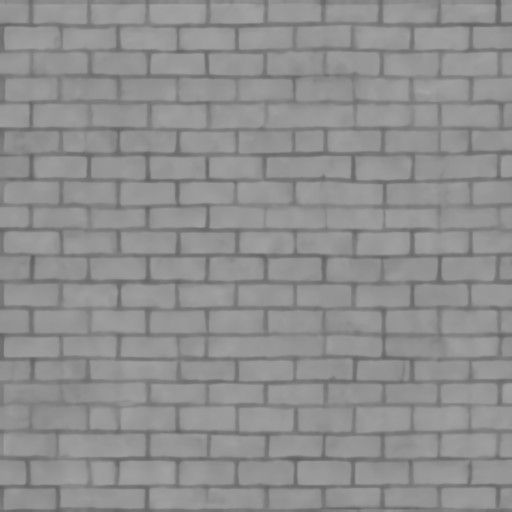}
            \end{minipage}	
            \begin{minipage}{0.13\linewidth}           
            \includegraphics[width=\linewidth]{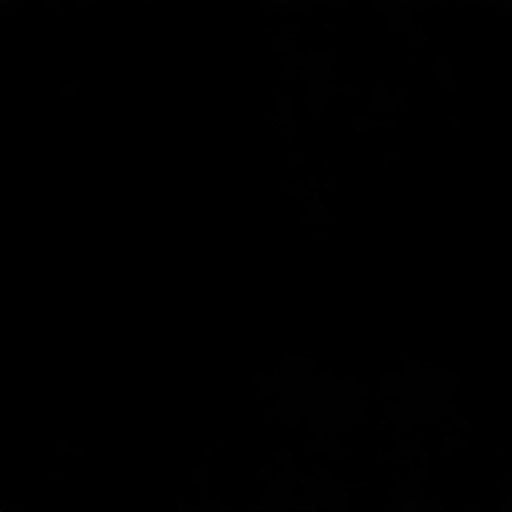}
            \end{minipage}	
            \begin{minipage}{0.13\linewidth}           
            \includegraphics[width=\linewidth]{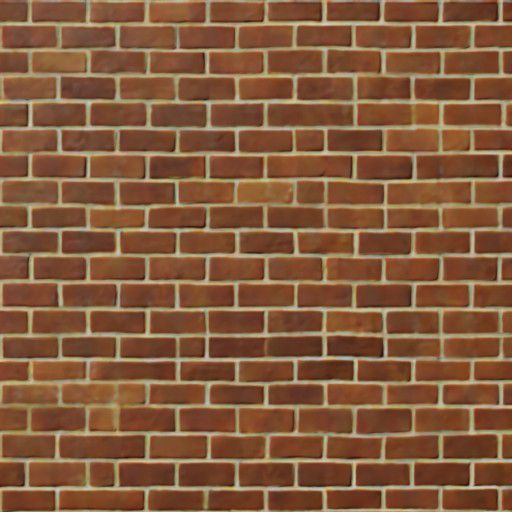}
            \end{minipage}	
        \end{minipage}	
    \end{minipage}

    \begin{minipage}{3.5in}
         \begin{minipage}{3.5in}	
            % \centering
            \begin{minipage}{0.13\linewidth}
            \includegraphics[width=\linewidth]{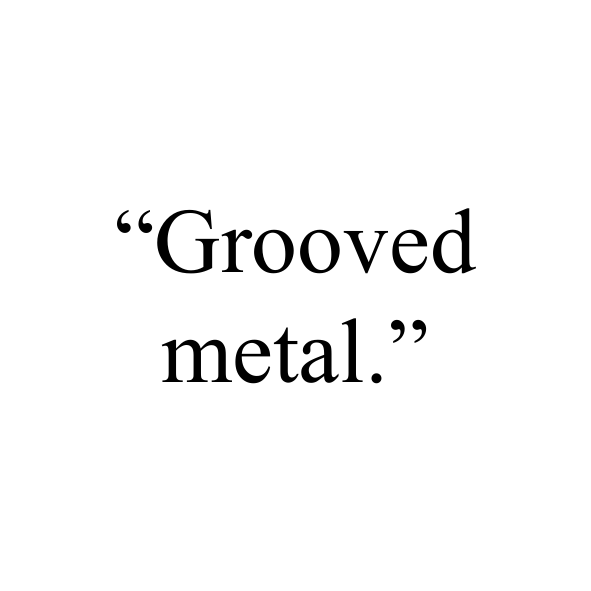}
            \end{minipage}	
            \begin{minipage}{0.13\linewidth}
            \includegraphics[width=\linewidth]{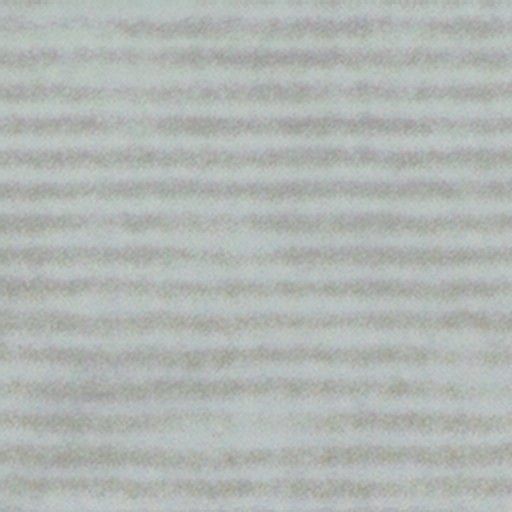}
            \end{minipage}	
            \begin{minipage}{0.13\linewidth}            
            \includegraphics[width=\linewidth]{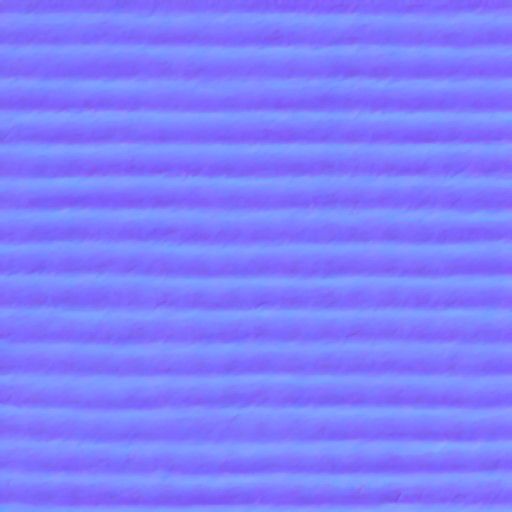}
            \end{minipage}	
            \begin{minipage}{0.13\linewidth}           
            \includegraphics[width=\linewidth]{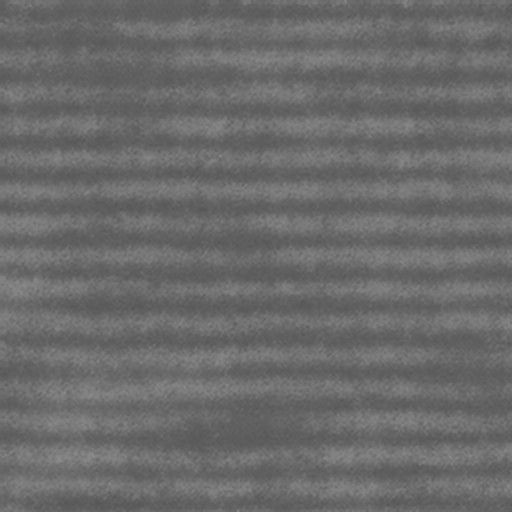}
            \end{minipage}	
            \begin{minipage}{0.13\linewidth}           
            \includegraphics[width=\linewidth]{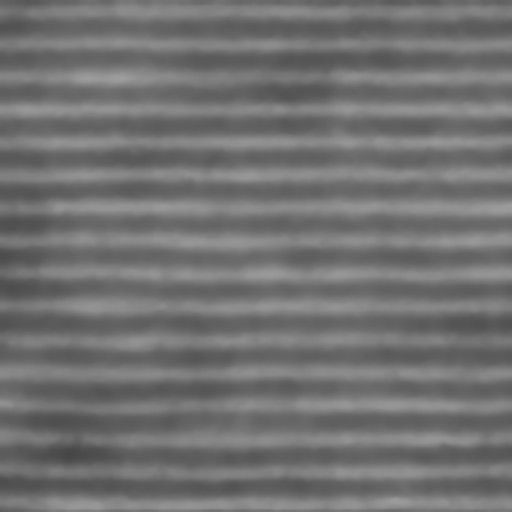}
            \end{minipage}	
            \begin{minipage}{0.13\linewidth}           
            \includegraphics[width=\linewidth]{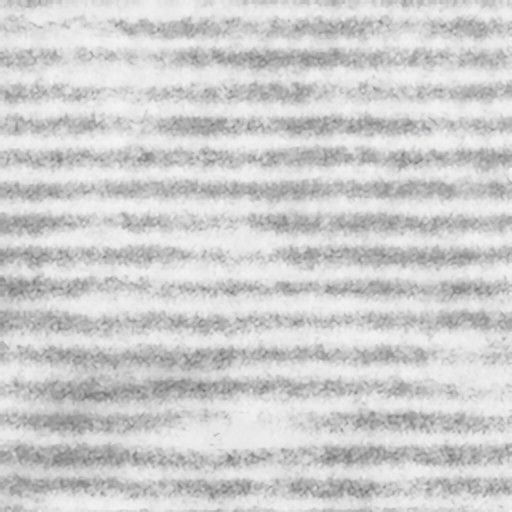}
            \end{minipage}	
            \begin{minipage}{0.13\linewidth}           
            \includegraphics[width=\linewidth]{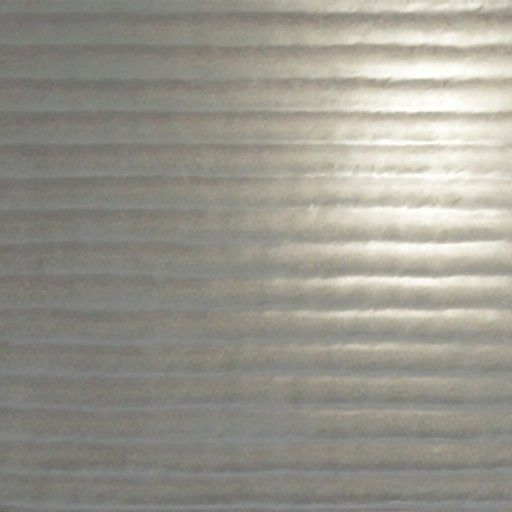}
            \end{minipage}	
        \end{minipage}	
    \end{minipage}

    \begin{minipage}{3.5in}
         \begin{minipage}{3.5in}	
            % \centering
            \begin{minipage}{0.13\linewidth}
            \includegraphics[width=\linewidth]{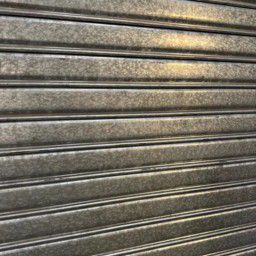}
            \end{minipage}	
            \begin{minipage}{0.13\linewidth}
            \includegraphics[width=\linewidth]{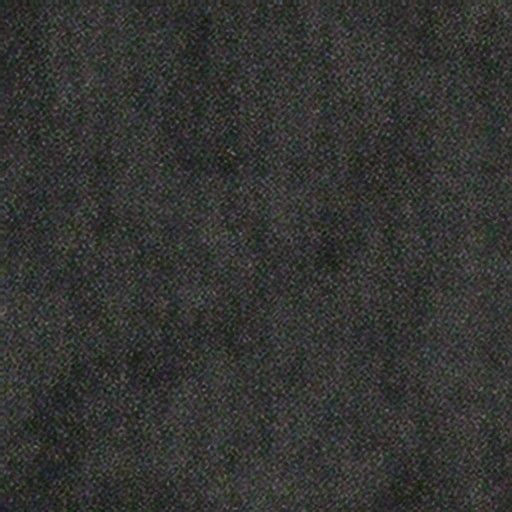}
            \end{minipage}	
            \begin{minipage}{0.13\linewidth}            
            \includegraphics[width=\linewidth]{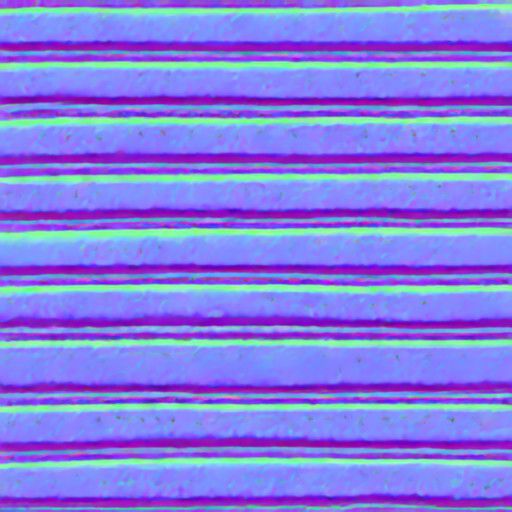}
            \end{minipage}	
            \begin{minipage}{0.13\linewidth}           
            \includegraphics[width=\linewidth]{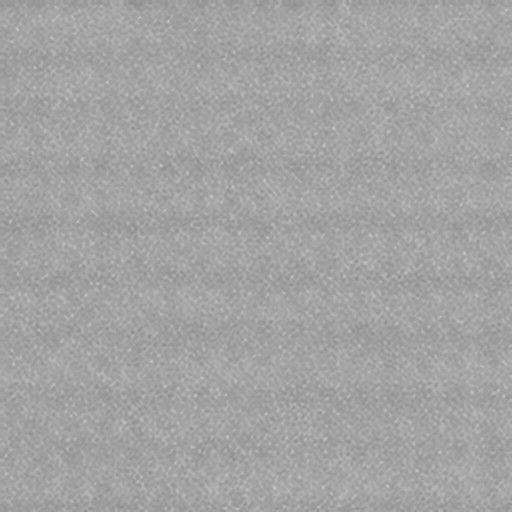}
            \end{minipage}	
            \begin{minipage}{0.13\linewidth}           
            \includegraphics[width=\linewidth]{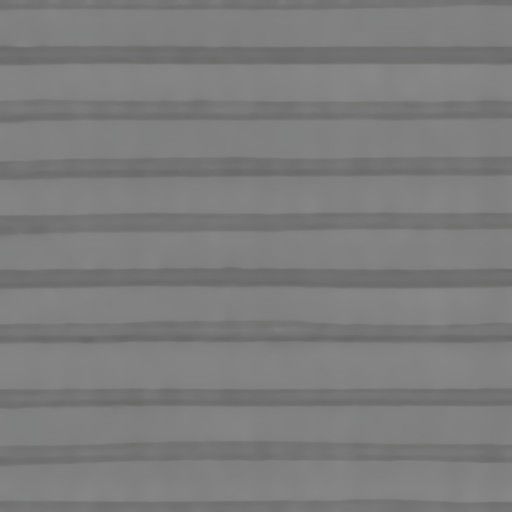}
            \end{minipage}	
            \begin{minipage}{0.13\linewidth}           
            \includegraphics[width=\linewidth]{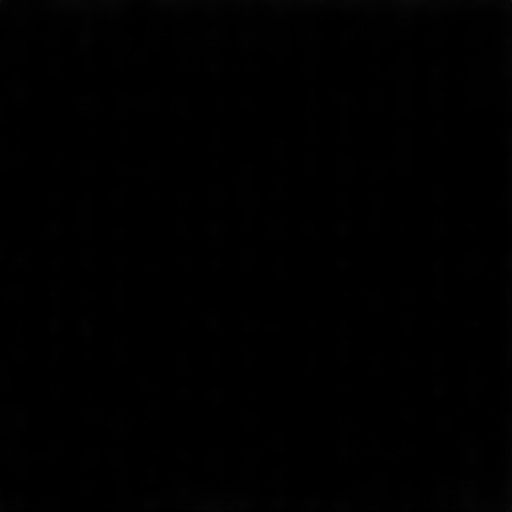}
            \end{minipage}	
            \begin{minipage}{0.13\linewidth}           
            \includegraphics[width=\linewidth]{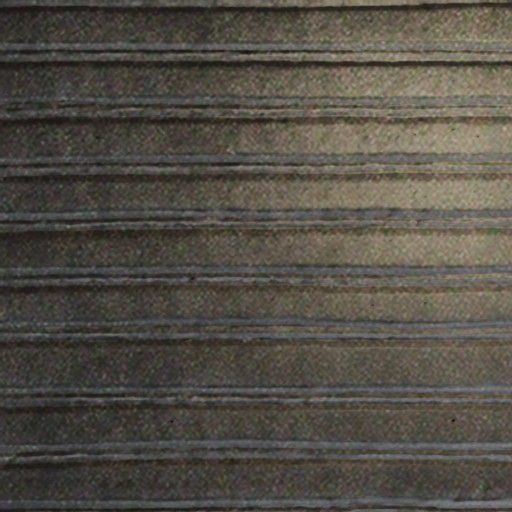}
            \end{minipage}	
        \end{minipage}	
    \end{minipage}

    \begin{minipage}{3.5in}
         \begin{minipage}{3.5in}	
            % \centering
            \begin{minipage}{0.13\linewidth}
            \includegraphics[width=\linewidth]{fig/multi_modal/cond.pdf}
            \end{minipage}	
            \begin{minipage}{0.13\linewidth}
            \includegraphics[width=\linewidth]{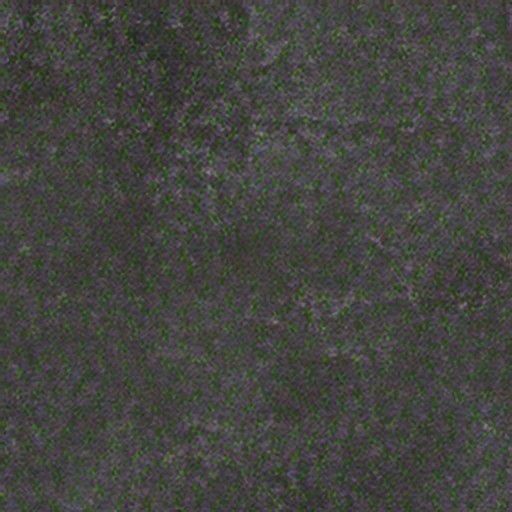}
            \end{minipage}	
            \begin{minipage}{0.13\linewidth}            
            \includegraphics[width=\linewidth]{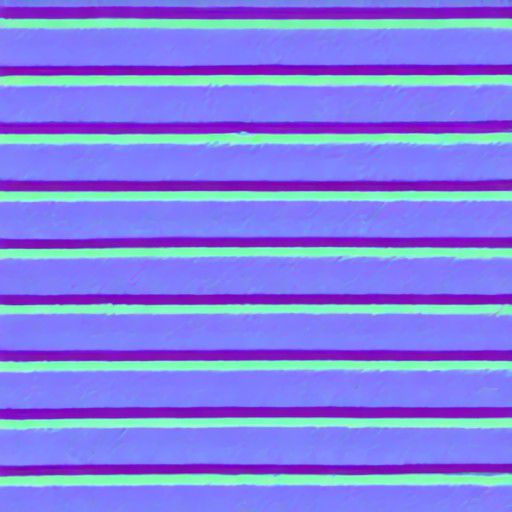}
            \end{minipage}	
            \begin{minipage}{0.13\linewidth}           
            \includegraphics[width=\linewidth]{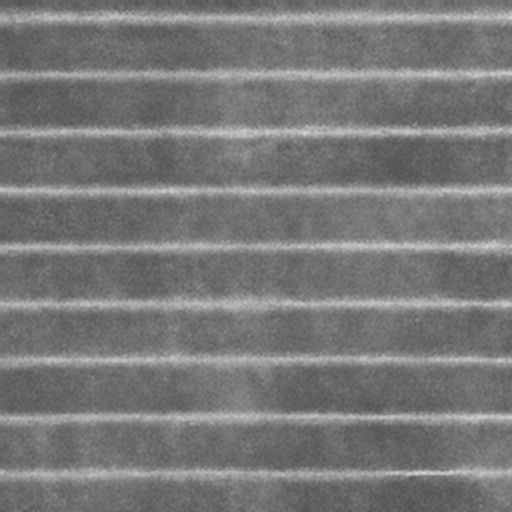}
            \end{minipage}	
            \begin{minipage}{0.13\linewidth}           
            \includegraphics[width=\linewidth]{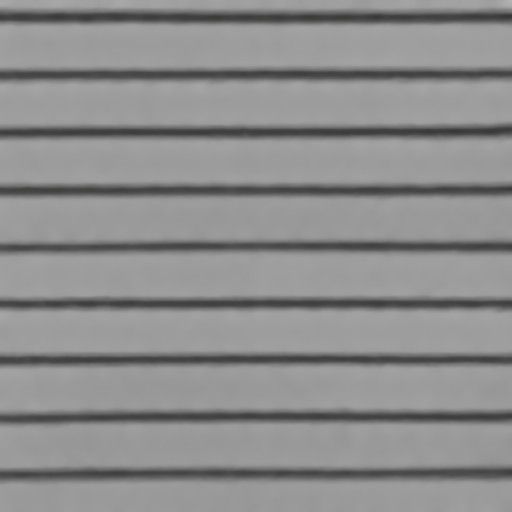}
            \end{minipage}	
            \begin{minipage}{0.13\linewidth}           
            \includegraphics[width=\linewidth]{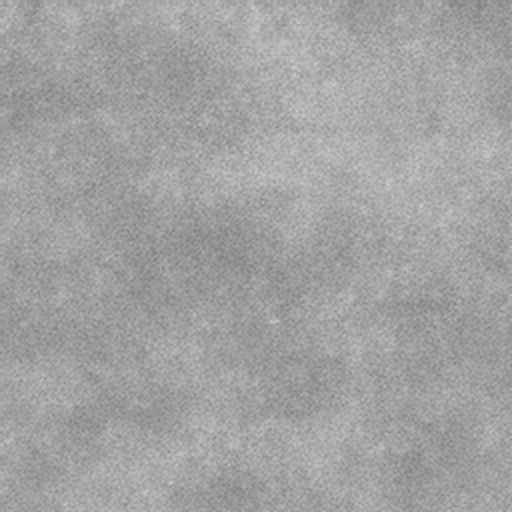}
            \end{minipage}	
            \begin{minipage}{0.13\linewidth}           
            \includegraphics[width=\linewidth]{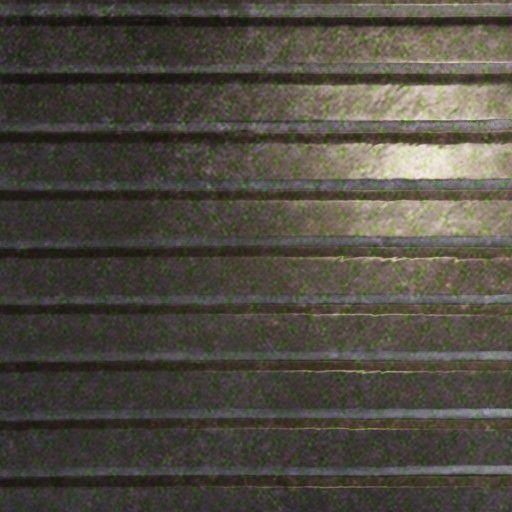}
            \end{minipage}	
        \end{minipage}	
    \end{minipage}

   \caption{Comparison of different input combinations. The first column shows the input condition. Text conditioning provides only coarse guidance for material generation, while image inputs offer explicit cues on material appearance. However, image inputs remain ambiguous with respect to material properties, as seen in the third example. Using both text and image conditioning simultaneously reduces this ambiguity, enhancing control and quality.} 
   \label{fig:conditioning_options}
\end{figure}

Our generative material model takes advantages of its multi-modality. Though it is designed to create material maps from input photographs, it can benefit from additional signal to reduce the ambiguity of a single in-the-wild photograph. We present different combinations of input conditions in Fig.~\ref{fig:conditioning_options} including 1) text condition only; 2) image condition only and 3) text+image dual conditions. We found that text conditioning provides high level guidance for material generation. On the other hand, image conditioning contains ambiguities, as lighting and camera poses are uncontrolled. Combining both options enables text prompts to guide the model in identifying the reflective properties of a material. For instance, by prompting the model with appropriate text, it can better differentiate between metallic and non-metallic materials, as shown in the third example in Fig.~\ref{fig:conditioning_options}.

\subsubsection{Mixed Dataset} \label{sec:mixed_dataset}
\begin{figure}[htbp!]
    \centering		
    \begin{minipage}{3.4in}
        \begin{minipage}{0.02in}	
            \centering
                \vspace{0.15in}
                \rotatebox{90}{\parbox{1cm}{\centering\tiny \vspace{-0.15cm} ``Woven\vspace{-0.05cm}\\ rattan"\vspace{-0.05cm}\\Mixed dataset}}
        \end{minipage}	
        \hspace{0.02in}
             \begin{minipage}{3.3in}	
                \centering
                \begin{minipage}{0.13\linewidth}
                    \subcaption*{\tiny Input}
                    \includegraphics[width=\linewidth]{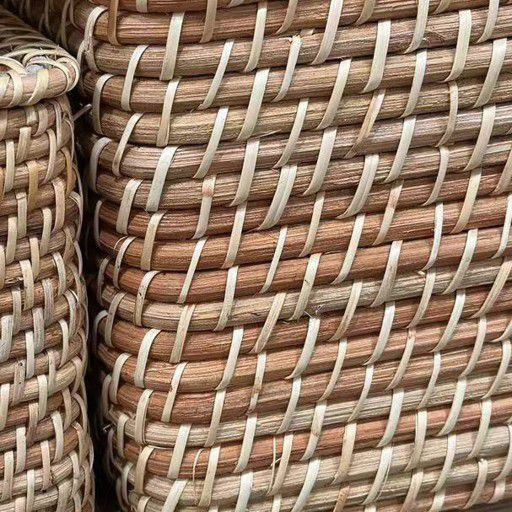}
                \end{minipage}
                \begin{minipage}{0.13\linewidth}
                    \subcaption*{\tiny Albedo}
                    \includegraphics[width=\linewidth]{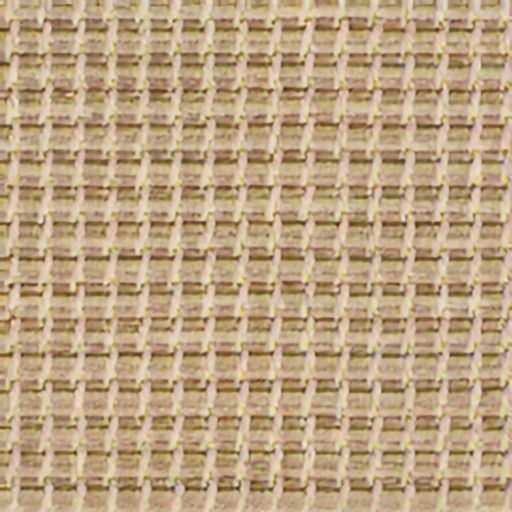}
                \end{minipage}
                \begin{minipage}{0.13\linewidth}
                    \subcaption*{\tiny Normal}
                    \includegraphics[width=\linewidth]{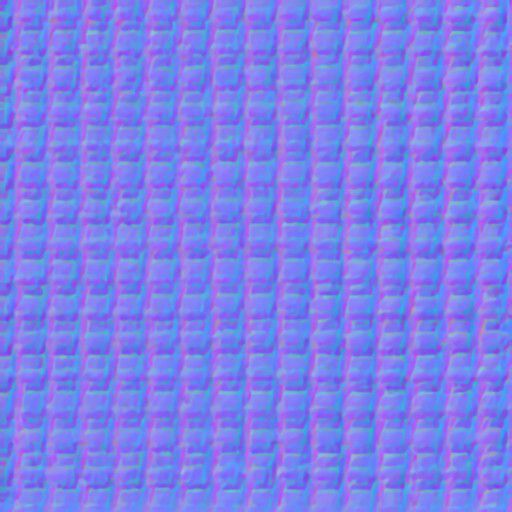}
                \end{minipage}
                \begin{minipage}{0.13\linewidth}
                    \subcaption*{\tiny Roughness}
                    \includegraphics[width=\linewidth]{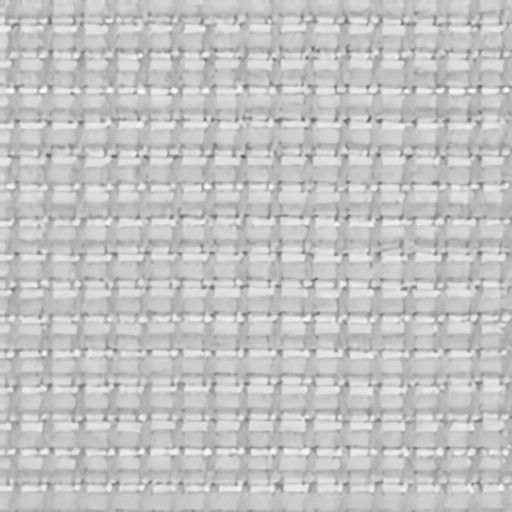}
                \end{minipage}
                \begin{minipage}{0.13\linewidth}
                    \subcaption*{\tiny Height}
                    \includegraphics[width=\linewidth]{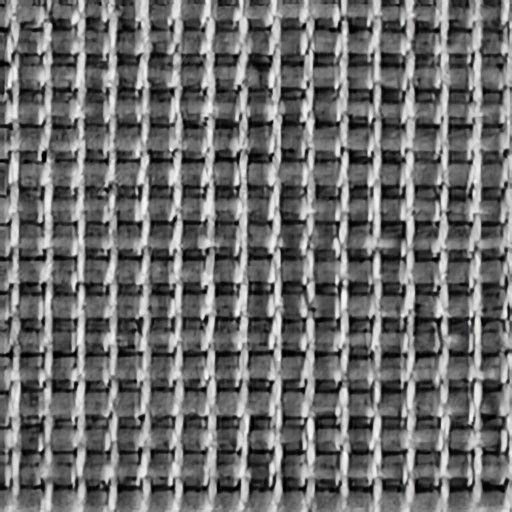}
                \end{minipage}
                \begin{minipage}{0.13\linewidth}
                    \subcaption*{\tiny Metallic}
                    \includegraphics[width=\linewidth]{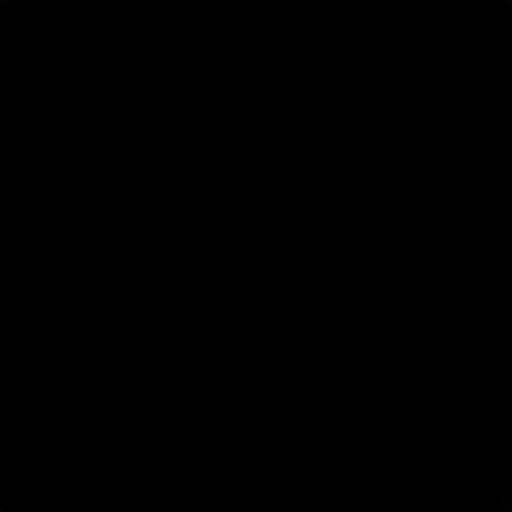}
                \end{minipage}
                \begin{minipage}{0.13\linewidth}
                    \subcaption*{\tiny Render}
                    \includegraphics[width=\linewidth]{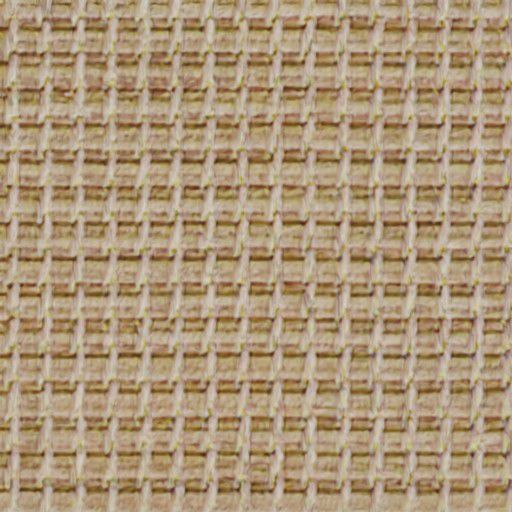}
                \end{minipage}
            \end{minipage}	
        \end{minipage}	

    \begin{minipage}{3.4in}
        \begin{minipage}{0.02in}	
            \centering
            \rotatebox{90}{\parbox{1cm}{\centering\tiny \vspace{-0.15cm} ``Woven\vspace{-0.05cm}\\ ratten"\vspace{-0.05cm}\\Single dataset}}
        \end{minipage}	
        \hspace{0.02in}
         \begin{minipage}{3.3in}	
            \centering
            \begin{minipage}{0.13\linewidth}
            \includegraphics[width=\linewidth]{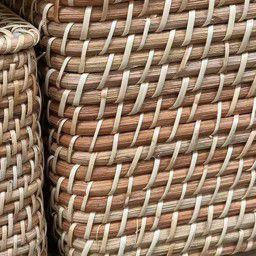}
            \end{minipage}	
            \begin{minipage}{0.13\linewidth}
            \includegraphics[width=\linewidth]{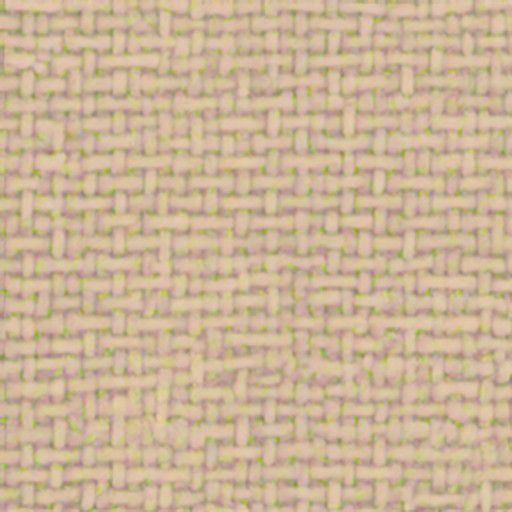}
            \end{minipage}	
            \begin{minipage}{0.13\linewidth}
            \includegraphics[width=\linewidth]{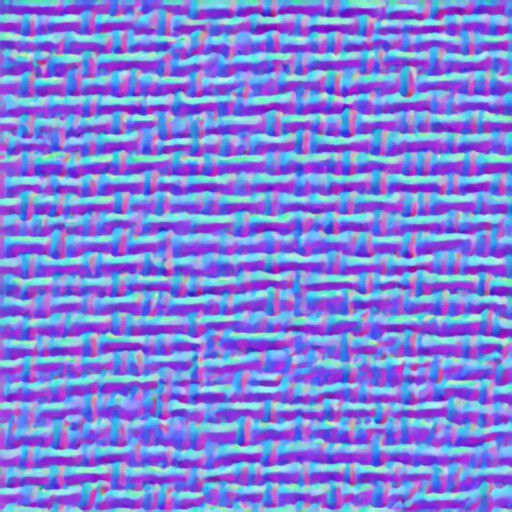}
            \end{minipage}	
            \begin{minipage}{0.13\linewidth}
            \includegraphics[width=\linewidth]{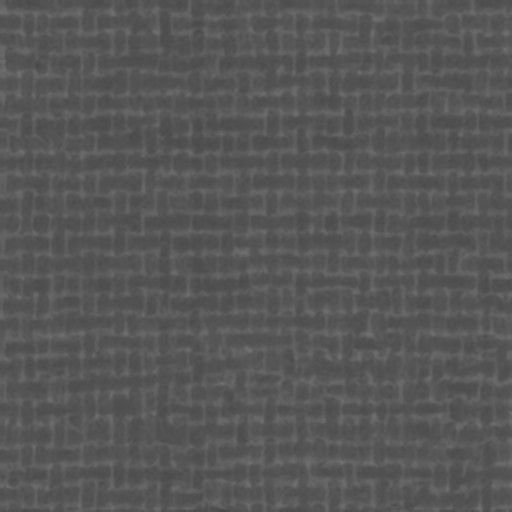}
            \end{minipage}	
            \begin{minipage}{0.13\linewidth}
            \includegraphics[width=\linewidth]{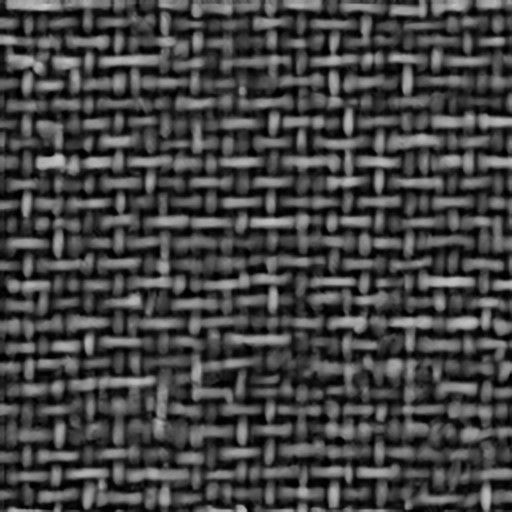}
            \end{minipage}	
            \begin{minipage}{0.13\linewidth}
            \includegraphics[width=\linewidth]{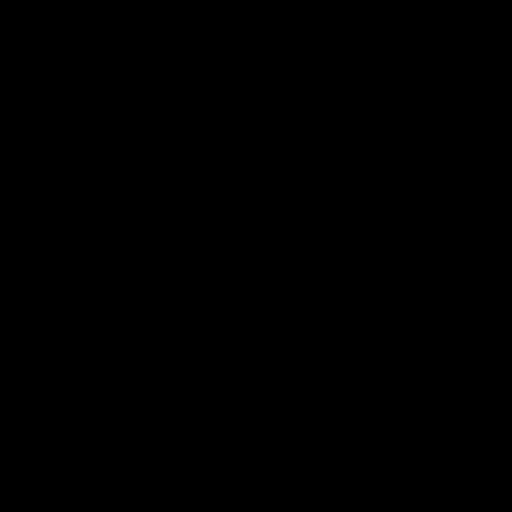}
            \end{minipage}	
            \begin{minipage}{0.13\linewidth}
            \includegraphics[width=\linewidth]{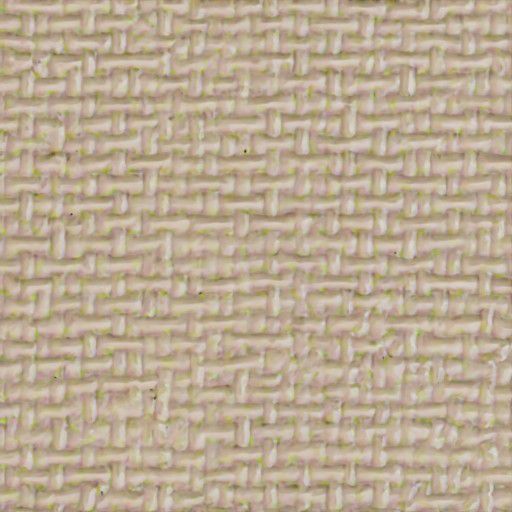}
            \end{minipage}	
        \end{minipage}	
    \end{minipage}	

    \begin{minipage}{3.4in}
        \begin{minipage}{0.02in}	
            \centering
            \rotatebox{90}{\parbox{1cm}{\centering\tiny \vspace{-0.15cm} ``Manhole\vspace{-0.05cm}\\ cover"\vspace{-0.05cm}\\Mixed dataset}}
        \end{minipage}	
        \hspace{0.02in}
         \begin{minipage}{3.3in}	
            \centering
            \begin{minipage}{0.13\linewidth}
            \includegraphics[width=\linewidth]{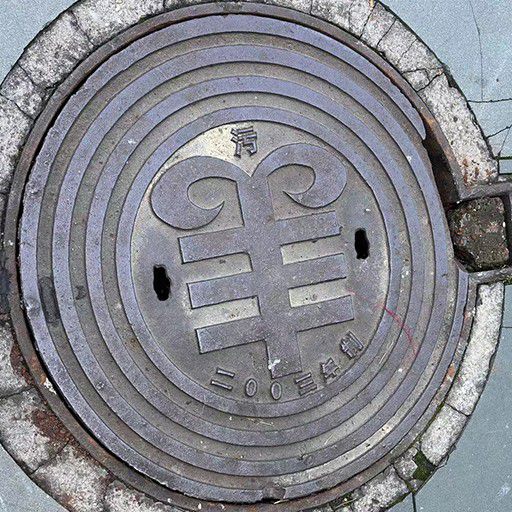}
            \end{minipage}	
            \begin{minipage}{0.13\linewidth}
            \includegraphics[width=\linewidth]{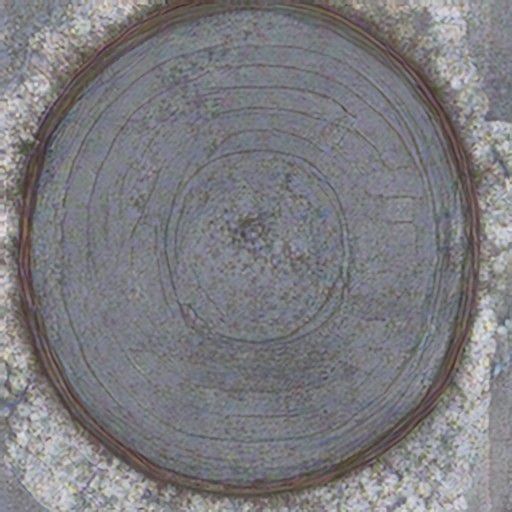}
            \end{minipage}	
            \begin{minipage}{0.13\linewidth}
            \includegraphics[width=\linewidth]{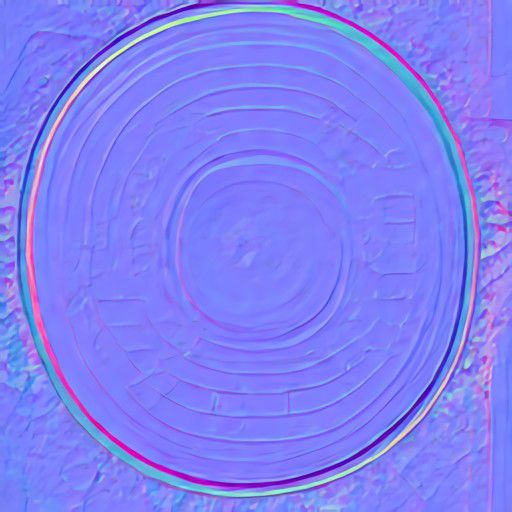}
            \end{minipage}	
            \begin{minipage}{0.13\linewidth}
            \includegraphics[width=\linewidth]{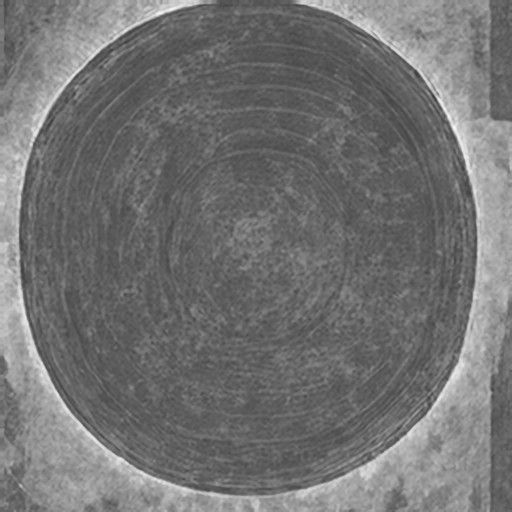}
            \end{minipage}	
            \begin{minipage}{0.13\linewidth}
            \includegraphics[width=\linewidth]{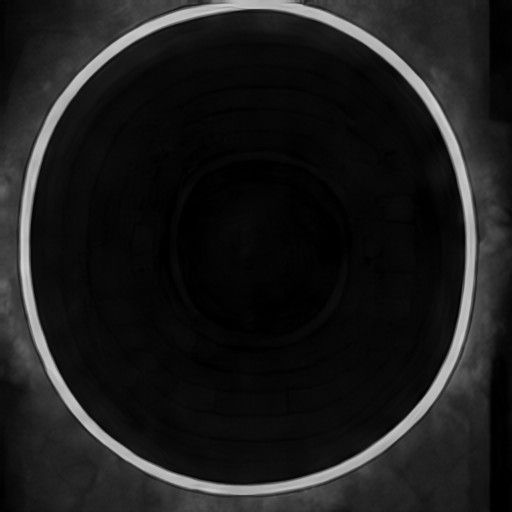}
            \end{minipage}	
            \begin{minipage}{0.13\linewidth}
            \includegraphics[width=\linewidth]{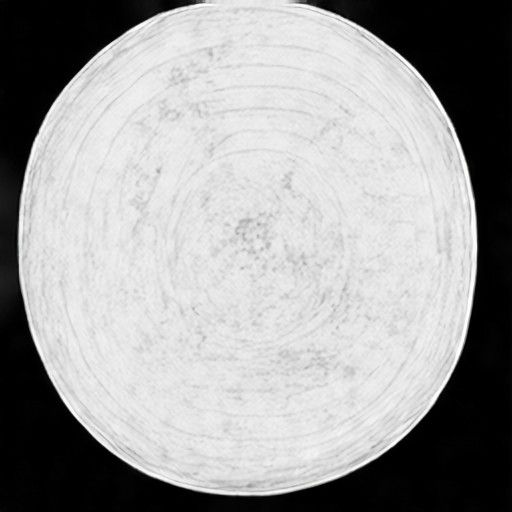}
            \end{minipage}	
            \begin{minipage}{0.13\linewidth}
            \includegraphics[width=\linewidth]{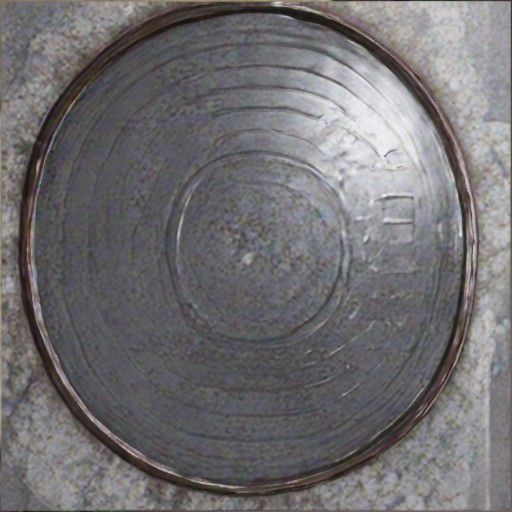}
            \end{minipage}	
        \end{minipage}	
    \end{minipage}

    \begin{minipage}{3.4in}
        \begin{minipage}{0.02in}	
            \centering
            \rotatebox{90}{\parbox{1cm}{\centering\tiny \vspace{-0.15cm} ``Manhole\vspace{-0.05cm}\\ cover"\vspace{-0.05cm}\\Single dataset}}
        \end{minipage}	
        \hspace{0.02in}
         \begin{minipage}{3.3in}	
            \centering
            \begin{minipage}{0.13\linewidth}
            \includegraphics[width=\linewidth]{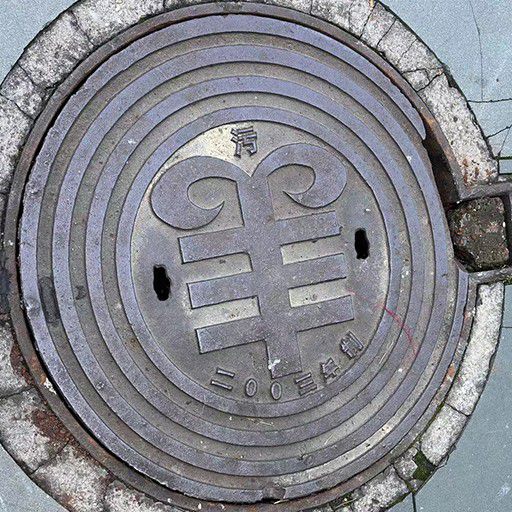}
            \end{minipage}	
            \begin{minipage}{0.13\linewidth}
            \includegraphics[width=\linewidth]{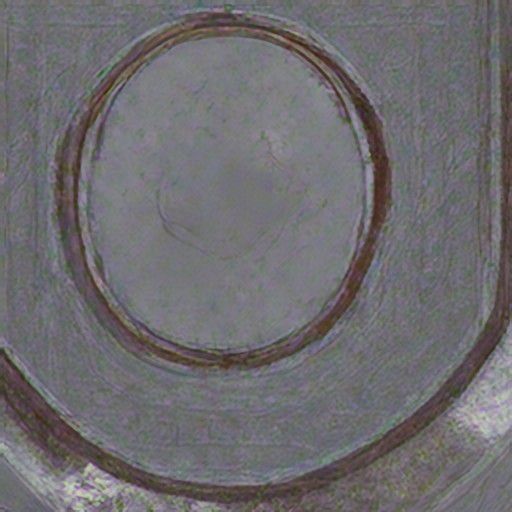}
            \end{minipage}	
            \begin{minipage}{0.13\linewidth}
            \includegraphics[width=\linewidth]{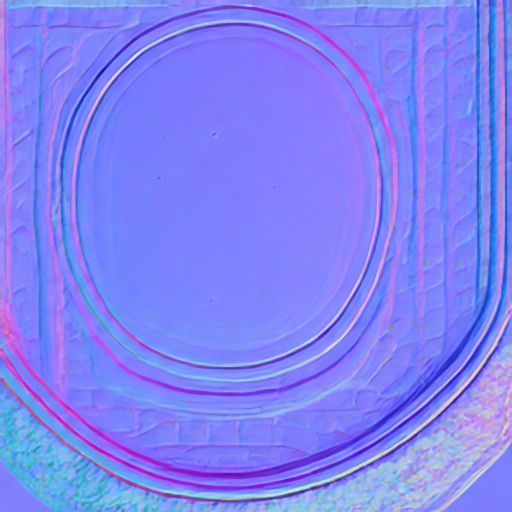}
            \end{minipage}	
            \begin{minipage}{0.13\linewidth}
            \includegraphics[width=\linewidth]{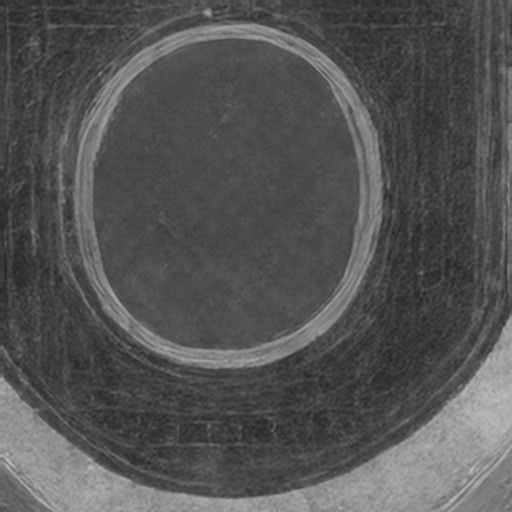}
            \end{minipage}	
            \begin{minipage}{0.13\linewidth}
            \includegraphics[width=\linewidth]{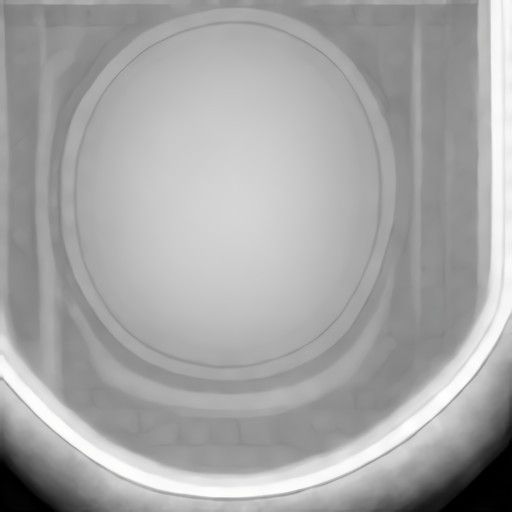}
            \end{minipage}	
            \begin{minipage}{0.13\linewidth}
            \includegraphics[width=\linewidth]{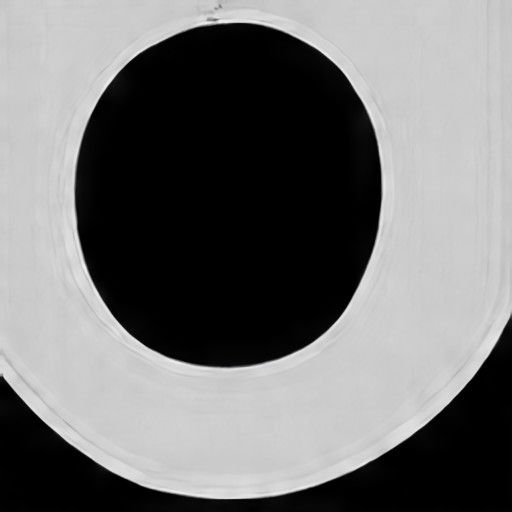}
            \end{minipage}	
            \begin{minipage}{0.13\linewidth}
            \includegraphics[width=\linewidth]{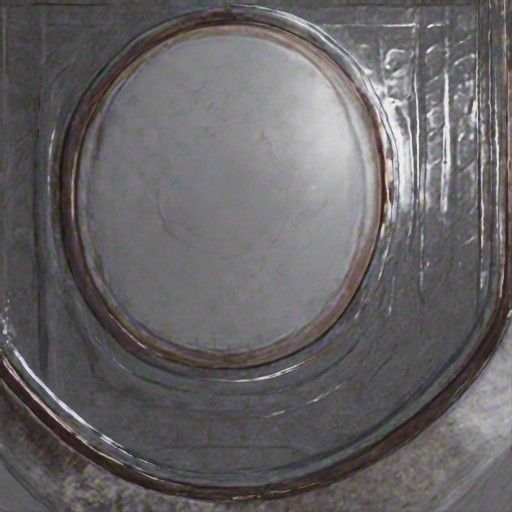}
            \end{minipage}	
        \end{minipage}	
    \end{minipage}

   \caption{Impact of our text-to-material synthetic dataset on generation and generalization. For each sample, the first row shows the generation results from our baseline model, trained on mixed datasets (Sec.~\ref{sec:dataset}), and the second row shows results from a model trained only on the \textit{Synthetic scenes}. The model trained with mixed dataset is able to synthesize better non-stationary, realistic textures. The leftmost side of each row is labeled with the text conditioning input used.} 
   \label{fig:mixed_dataset}
\end{figure}

In Sec.~\ref{sec:dataset}, we introduce two datasets used to train our model. To confirm that using both datasets help, we train a variant using only the \textit{Scene} dataset. Since this dataset primarily contains stationary materials, training exclusively on it reduces our model's generalization for complex texture patterns commonly found in real-world scenarios as shown in Fig. \ref{fig:mixed_dataset}. By mixing additional training data, our model synthesizes more diverse texture patterns and features such as woven pattern or the texture of a manhole cover.

\subsubsection{Mask as Input or Output}\label{sec:ablation_mask}
\begin{figure}[htbp!]
    \centering		
    \begin{minipage}{3.4in}
        \begin{minipage}{0.02in}	
            \centering
                \vspace{0.1in}
                \rotatebox{90}{\parbox{1cm}{\centering\tiny "Ceramic\vspace{-0.05cm}\\tiles"}}
        \end{minipage}	
        \hspace{0.02in}
             \begin{minipage}{3.3in}	
                \centering
                \begin{minipage}{0.13\linewidth}
                    \subcaption*{\tiny Input}
                    \includegraphics[width=\linewidth]{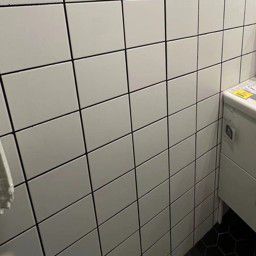}
                \end{minipage}
                \begin{minipage}{0.13\linewidth}
                    \subcaption*{\tiny Mask}
                    \includegraphics[width=\linewidth]{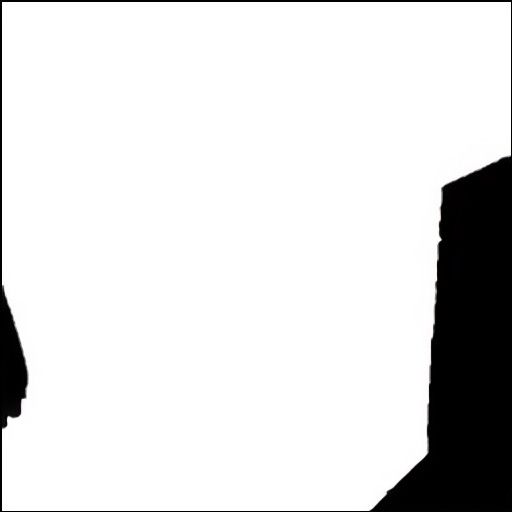}
                    \put(-21.5,25){\tiny\color{black}Input}
                \end{minipage}
                \begin{minipage}{0.13\linewidth}
                    \subcaption*{\tiny Albedo}
                    \includegraphics[width=\linewidth]{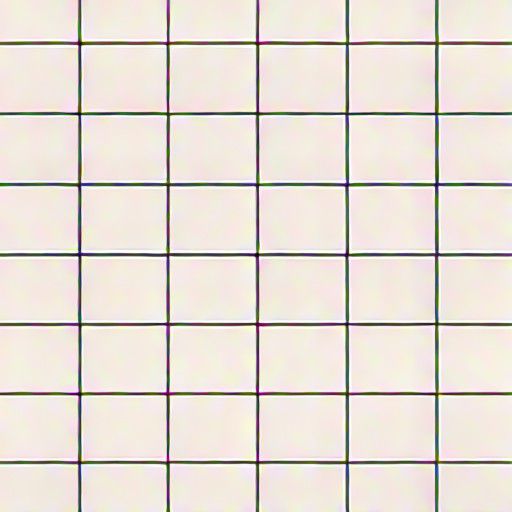}
                \end{minipage}
                \begin{minipage}{0.13\linewidth}
                    \subcaption*{\tiny Normal}
                    \includegraphics[width=\linewidth]{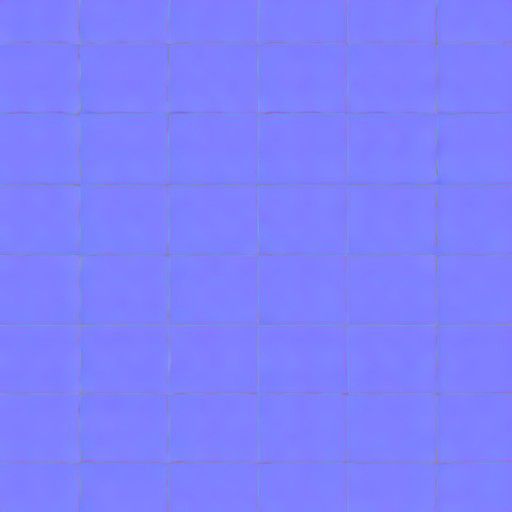}
                \end{minipage}
                \begin{minipage}{0.13\linewidth}
                    \subcaption*{\tiny Roughness}
                    \includegraphics[width=\linewidth]{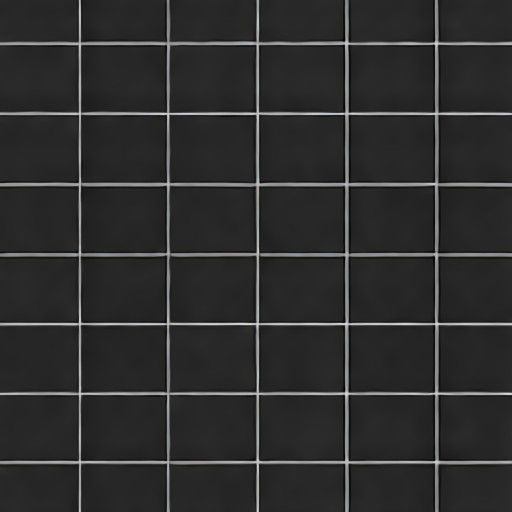}
                \end{minipage}
                \begin{minipage}{0.13\linewidth}
                    \subcaption*{\tiny Height}
                    \includegraphics[width=\linewidth]{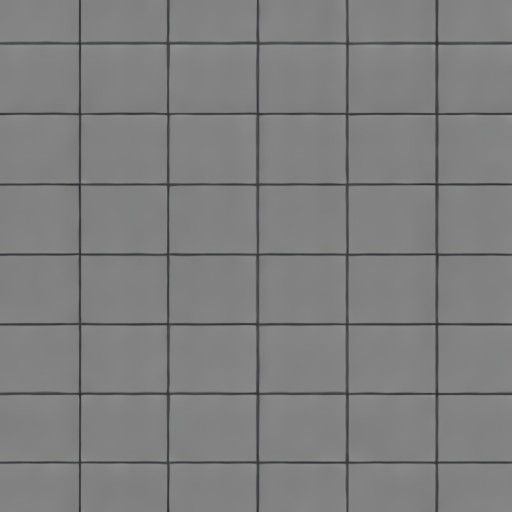}
                \end{minipage}
                \begin{minipage}{0.13\linewidth}
                    \subcaption*{\tiny Metallic}
                    \includegraphics[width=\linewidth]{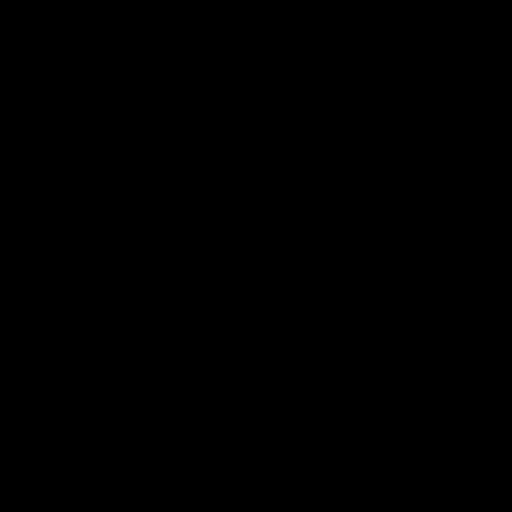}
                \end{minipage}
            \end{minipage}	
        \end{minipage}	

    \begin{minipage}{3.4in}
    \begin{minipage}{0.02in}	
            \centering
            \rotatebox{90}{\parbox{1cm}{\centering\tiny "Ceramic\vspace{-0.05cm}\\tiles"}}
        \end{minipage}	
        \hspace{0.02in}
         \begin{minipage}{3.3in}	
            \centering
            \begin{minipage}{0.13\linewidth}
            \includegraphics[width=\linewidth]{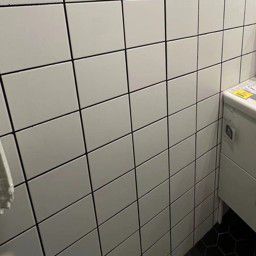}
            \end{minipage}	
            \begin{minipage}{0.13\linewidth}
            \includegraphics[width=\linewidth]{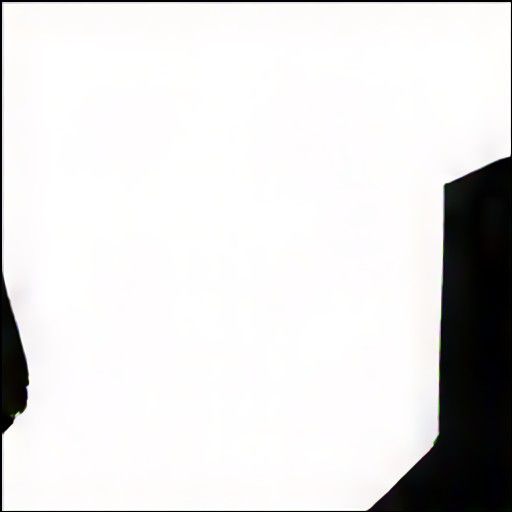}
            \put(-22,25){\tiny\color{black}Output}
            \end{minipage}	
            \begin{minipage}{0.13\linewidth}
            \includegraphics[width=\linewidth]{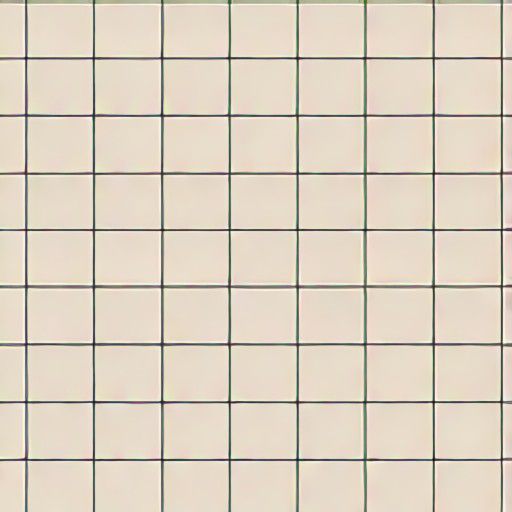}
            \end{minipage}	
            \begin{minipage}{0.13\linewidth}
            \includegraphics[width=\linewidth]{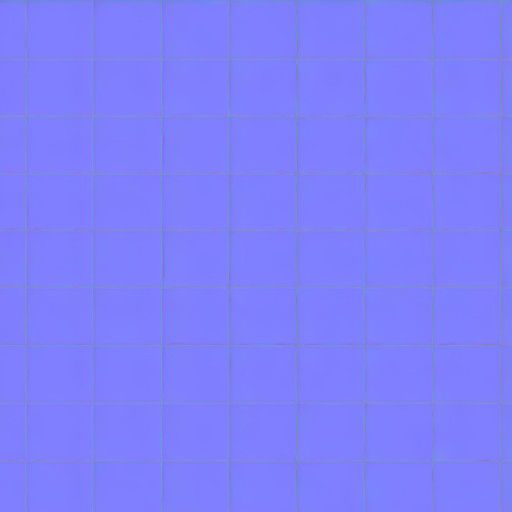}
            \end{minipage}	
            \begin{minipage}{0.13\linewidth}
            \includegraphics[width=\linewidth]{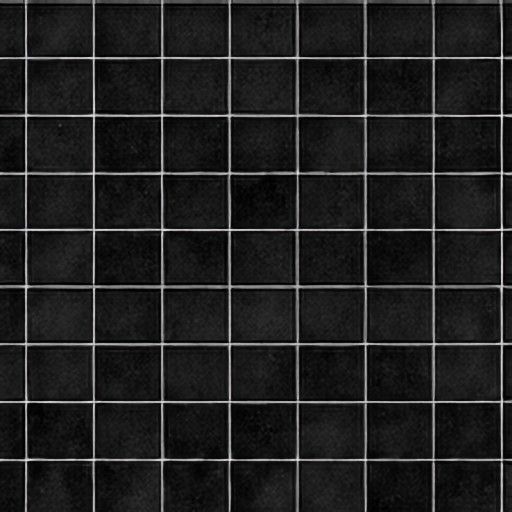}
            \end{minipage}	
            \begin{minipage}{0.13\linewidth}
            \includegraphics[width=\linewidth]{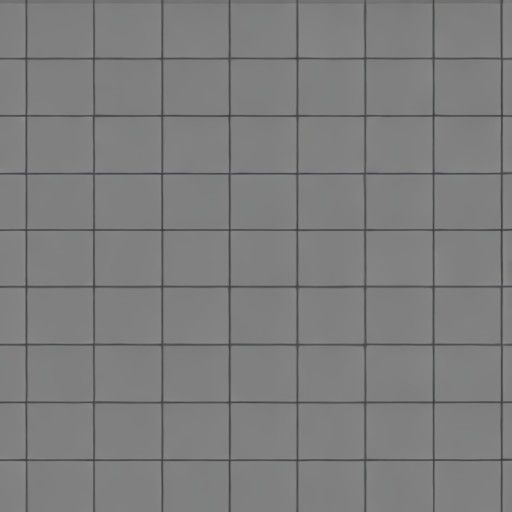}
            \end{minipage}	
            \begin{minipage}{0.13\linewidth}
            \includegraphics[width=\linewidth]{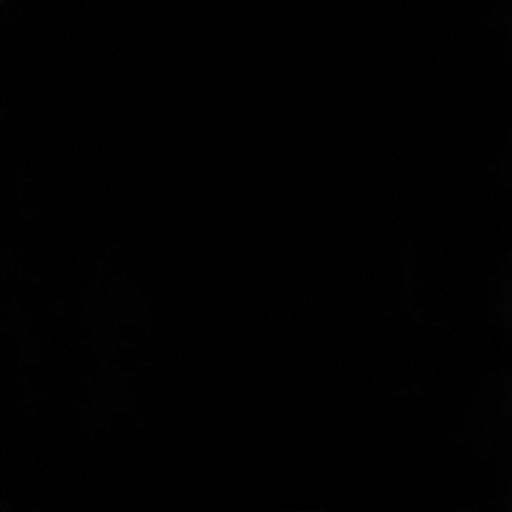}
            \end{minipage}	
        \end{minipage}	
    \end{minipage}	

    \begin{minipage}{3.4in}
        \begin{minipage}{0.02in}	
            \centering
            \rotatebox{90}{\parbox{1cm}{\centering\tiny \vspace{0.05cm} "Stone wall"}}
        \end{minipage}	
        \hspace{0.02in}
         \begin{minipage}{3.3in}	
            \centering
            \begin{minipage}{0.13\linewidth}
            \includegraphics[width=\linewidth]{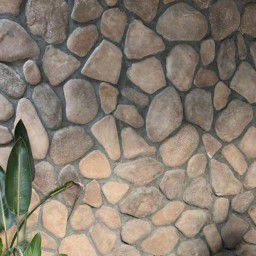}
            \end{minipage}	
            \begin{minipage}{0.13\linewidth}
            \includegraphics[width=\linewidth]{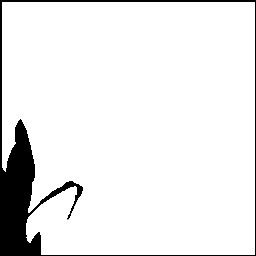}
            \put(-21.5,25){\tiny\color{black}Input}
            \end{minipage}	
            \begin{minipage}{0.13\linewidth}
            \includegraphics[width=\linewidth]{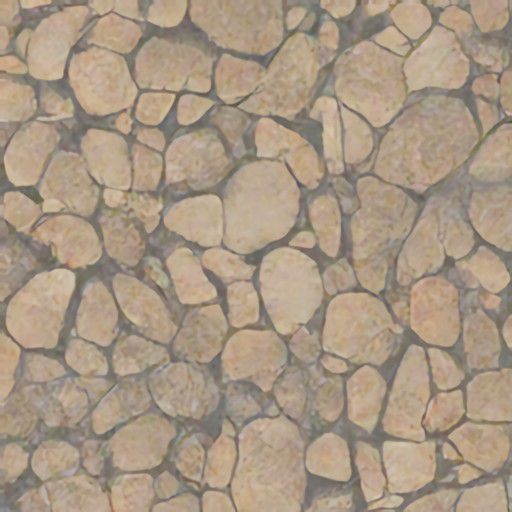}
            \end{minipage}	
            \begin{minipage}{0.13\linewidth}
            \includegraphics[width=\linewidth]{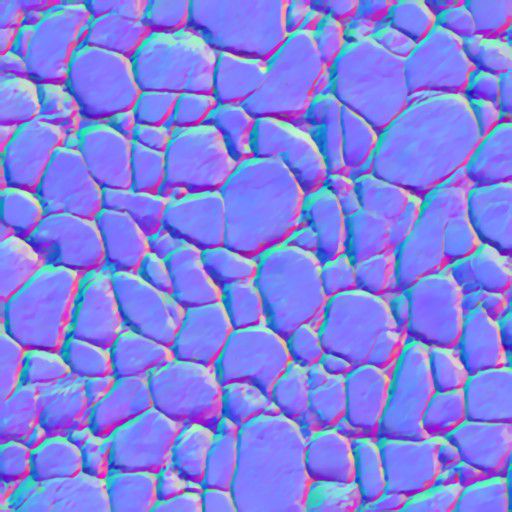}
            \end{minipage}	
            \begin{minipage}{0.13\linewidth}
            \includegraphics[width=\linewidth]{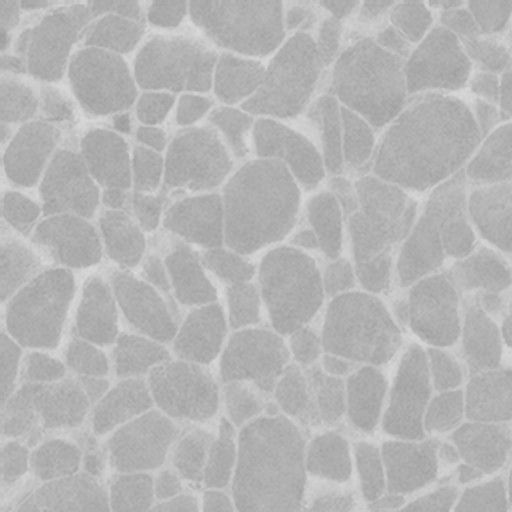}
            \end{minipage}	
            \begin{minipage}{0.13\linewidth}
            \includegraphics[width=\linewidth]{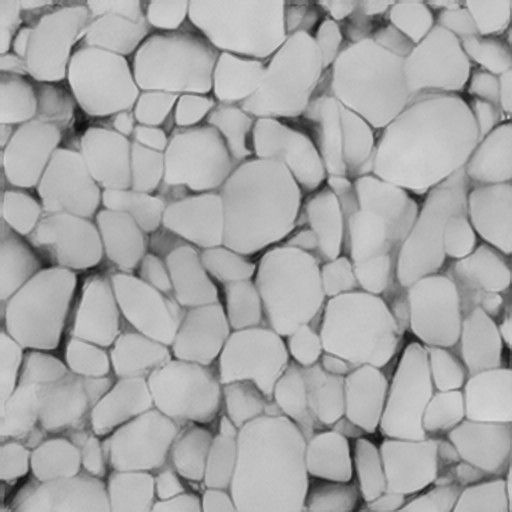}
            \end{minipage}	
            \begin{minipage}{0.13\linewidth}
            \includegraphics[width=\linewidth]{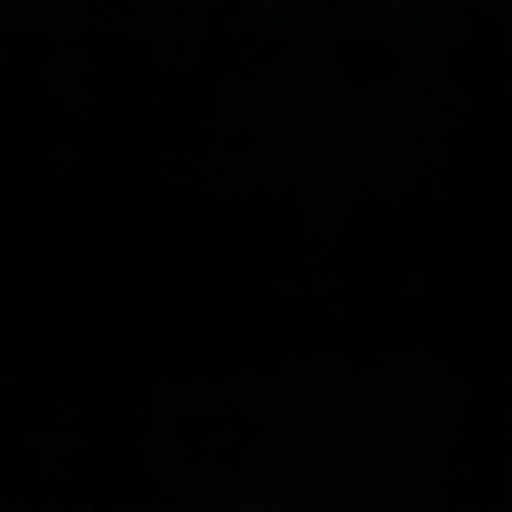}
            \end{minipage}	
        \end{minipage}	
    \end{minipage}	

    \begin{minipage}{3.4in}
        \begin{minipage}{0.02in}	
            \centering
            \rotatebox{90}{\parbox{1cm}{\centering\tiny \vspace{0.05cm} "Stone wall"}}
        \end{minipage}	
        \hspace{0.02in}
         \begin{minipage}{3.3in}	
            \centering
            \begin{minipage}{0.13\linewidth}
            \includegraphics[width=\linewidth]{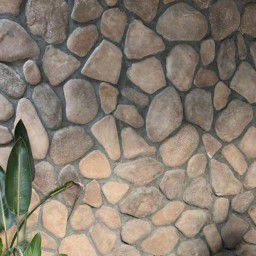}
            \end{minipage}	
            \begin{minipage}{0.13\linewidth}
            \includegraphics[width=\linewidth]{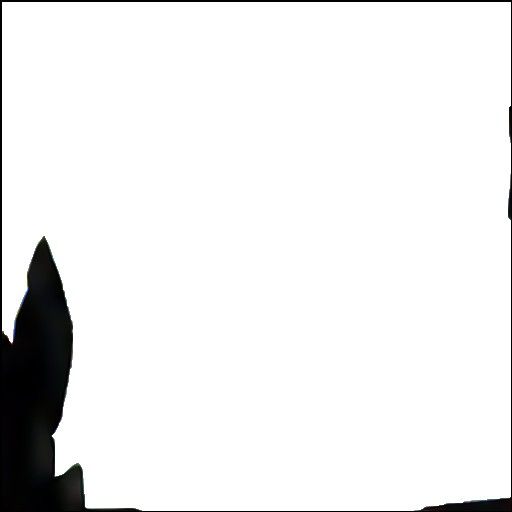}
            \put(-22,25){\tiny\color{black}Output}
            \end{minipage}	
            \begin{minipage}{0.13\linewidth}
            \includegraphics[width=\linewidth]{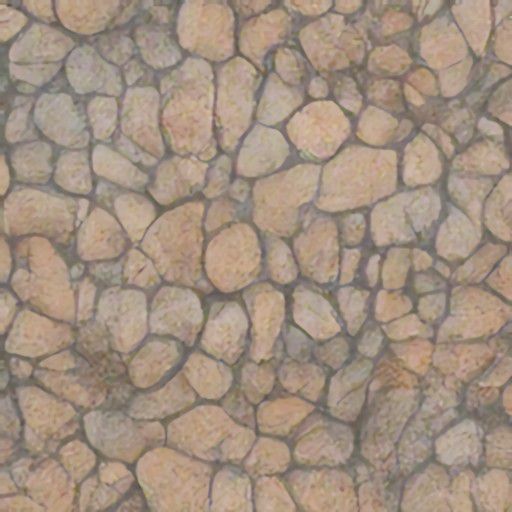}
            \end{minipage}	
            \begin{minipage}{0.13\linewidth}
            \includegraphics[width=\linewidth]{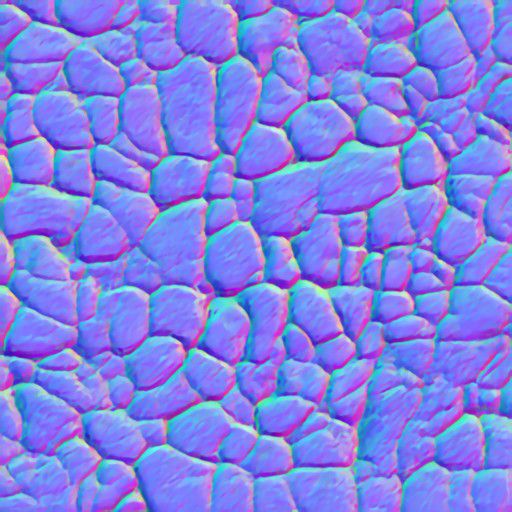}
            \end{minipage}	
            \begin{minipage}{0.13\linewidth}
            \includegraphics[width=\linewidth]{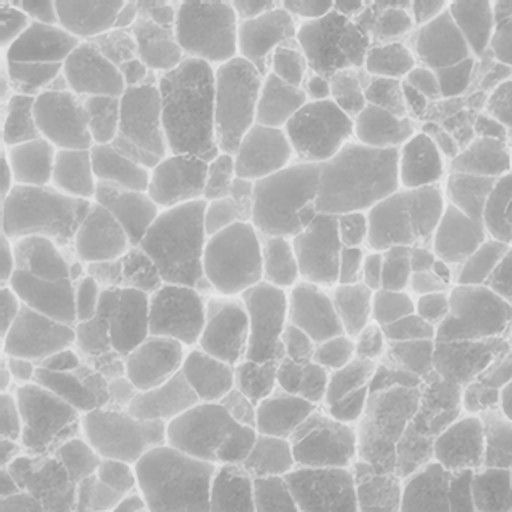}
            \end{minipage}	
            \begin{minipage}{0.13\linewidth}
            \includegraphics[width=\linewidth]{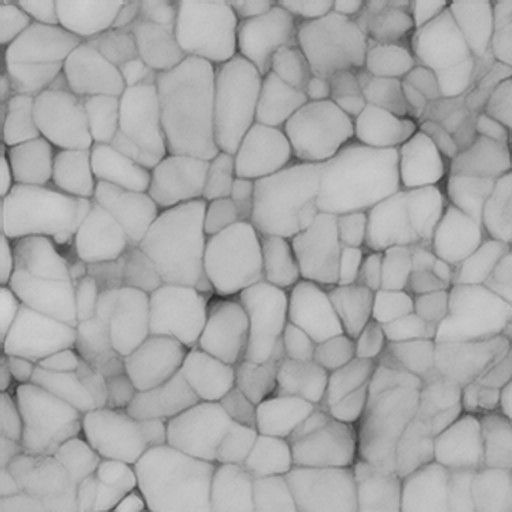}
            \end{minipage}	
            \begin{minipage}{0.13\linewidth}
            \includegraphics[width=\linewidth]{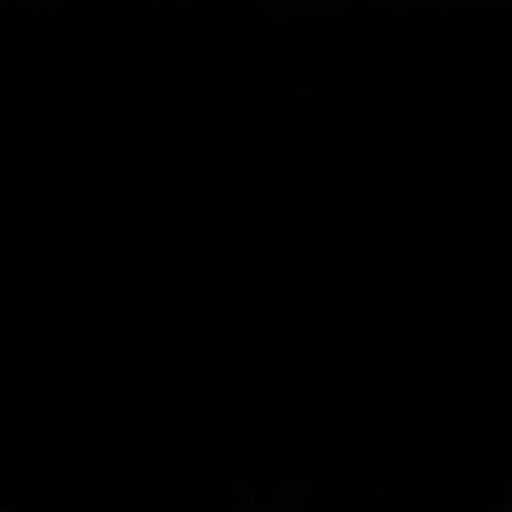}
            \end{minipage}	
        \end{minipage}	
    \end{minipage}	

   \caption{Impact of mask on material generation quality. Here shows the effect of using the mask as an input versus as an output on the quality of generated materials. Each pair of consecutive rows represents the results from the model with the mask as input (top row) and the model with the mask as output (bottom row). The results show that our model can accurately predict masks without a decrease in material quality. The leftmost side of each row is labeled with the text conditioning input used.} 
   \label{fig:ablation_mask}
\end{figure}

As opposed to existing material generation models, our model does not require the target material to cover the entire input image~\cite{vecchio2024controlmat} or manually-created masks~\cite{lopes2024material} to identify the sample of interest. Our model instead outputs a mask along with the generated materials. To assess the impact of generating this mask, we train an alternative model using our two datasets, with a slight modification to the model configuration. We add noise $\epsilon_t$ to the material maps $M$ only, with $x_t = \text{stack}(I, V, M_t)$, leaving the image and mask as non-noised inputs (or $x_t = \text{stack}(V, M_t)$ without $I$), using our adaptation of a video model (as described in Sec. 3.3). The loss is then computed on the material maps $M$ only. As shown in Fig.~\ref{fig:ablation_mask}, we find that our proposed model, which automatically predicts a mask, performs comparably well to this variant requiring the mask as input.

\subsubsection{Input scale} \label{sec:input_scale}
\begin{figure}[htbp!]
    \centering		
    \begin{minipage}{3.4in}
        \begin{minipage}{0.02in}	
            \centering
                \vspace{0.15in}
                \rotatebox{90}{\parbox{1cm}{\centering\tiny \vspace{0.05cm} "Marble tiles"}}
        \end{minipage}	
        \hspace{0.02in}
             \begin{minipage}{3.3in}	
                \centering
                \begin{minipage}{0.13\linewidth}
                    \subcaption*{\tiny Input}
                    \includegraphics[width=\linewidth]{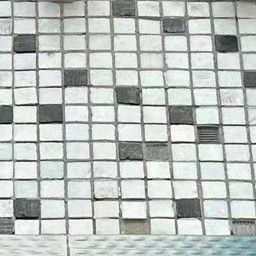}
                \end{minipage}
                \begin{minipage}{0.13\linewidth}
                    \subcaption*{\tiny Albedo}
                    \includegraphics[width=\linewidth]{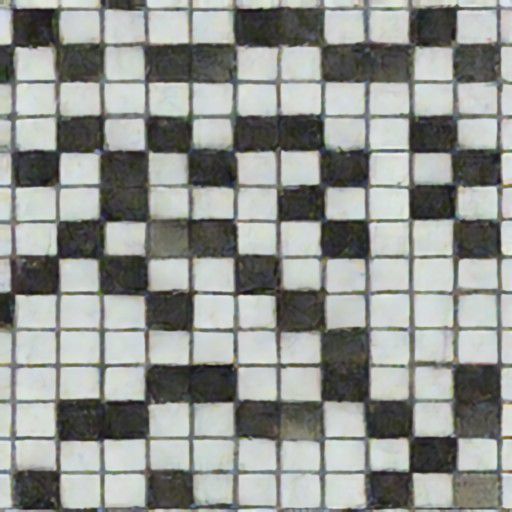}
                \end{minipage}
                \begin{minipage}{0.13\linewidth}
                    \subcaption*{\tiny Normal}
                    \includegraphics[width=\linewidth]{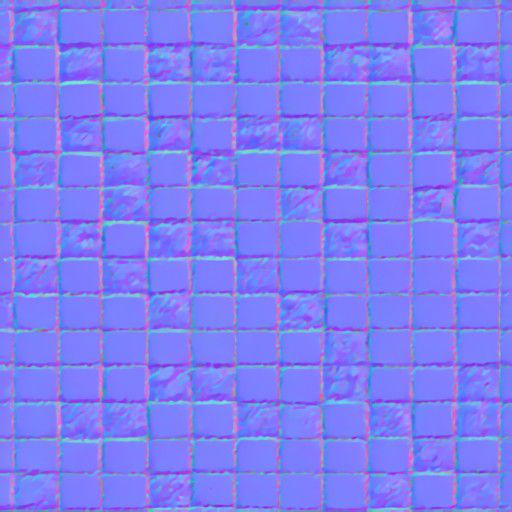}
                \end{minipage}
                \begin{minipage}{0.13\linewidth}
                    \subcaption*{\tiny Roughness}
                    \includegraphics[width=\linewidth]{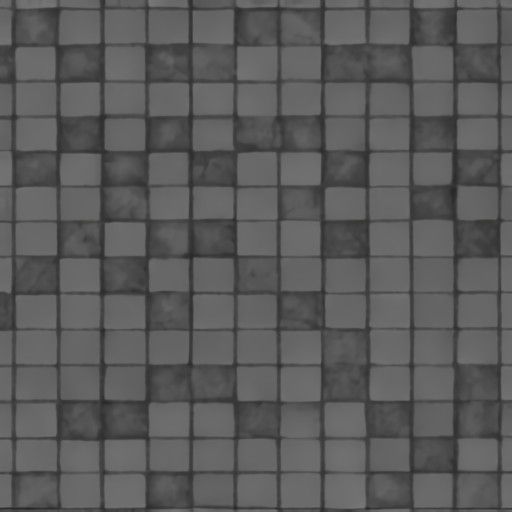}
                \end{minipage}
                \begin{minipage}{0.13\linewidth}
                    \subcaption*{\tiny Height}
                    \includegraphics[width=\linewidth]{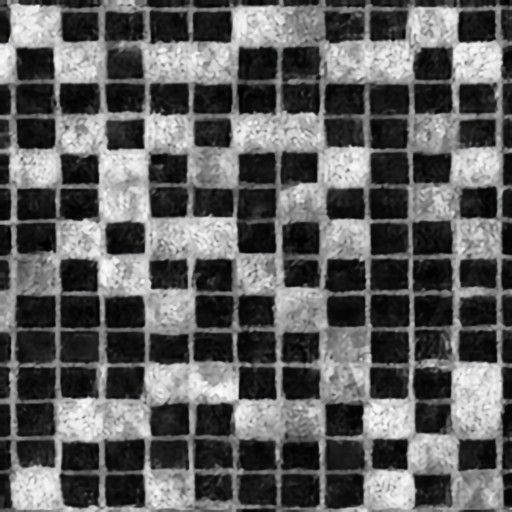}
                \end{minipage}
                \begin{minipage}{0.13\linewidth}
                    \subcaption*{\tiny Metallic}
                    \includegraphics[width=\linewidth]{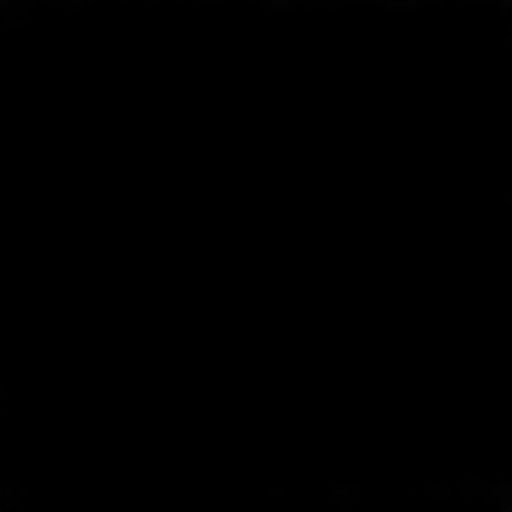}
                \end{minipage}
                \begin{minipage}{0.13\linewidth}
                    \subcaption*{\tiny Render}
                    \includegraphics[width=\linewidth]{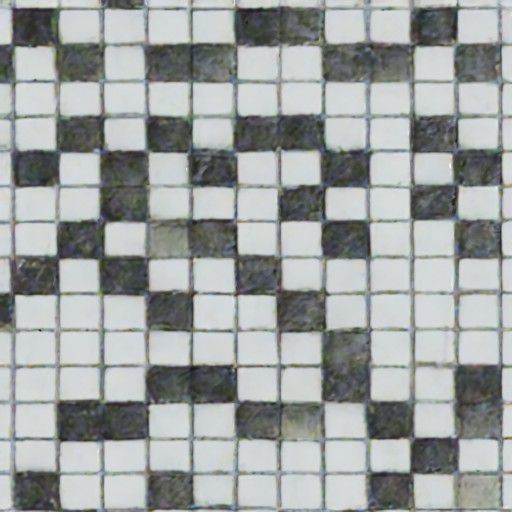}
                \end{minipage}
            \end{minipage}	
        \end{minipage}	

    \begin{minipage}{3.4in}
        \begin{minipage}{0.02in}	
            \centering
                \rotatebox{90}{\parbox{1cm}{\centering\tiny \vspace{0.05cm} "Marble tiles"}}
        \end{minipage}	
        \hspace{0.02in}
         \begin{minipage}{3.3in}	
            \centering
            \begin{minipage}{0.13\linewidth}
            \includegraphics[width=\linewidth]{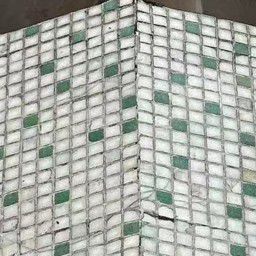}
            \end{minipage}	
            \begin{minipage}{0.13\linewidth}
            \includegraphics[width=\linewidth]{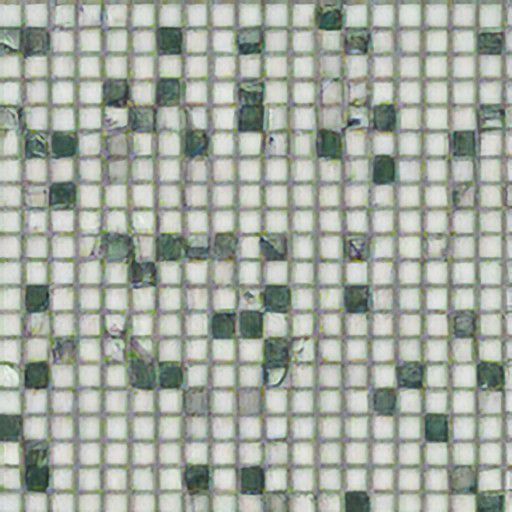}
            \end{minipage}	
            \begin{minipage}{0.13\linewidth}
            \includegraphics[width=\linewidth]{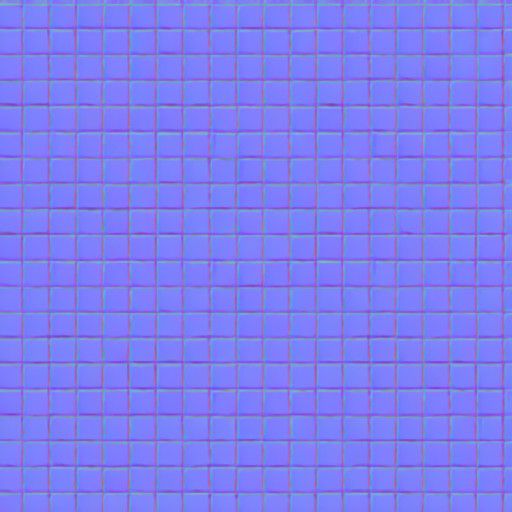}
            \end{minipage}	
            \begin{minipage}{0.13\linewidth}
            \includegraphics[width=\linewidth]{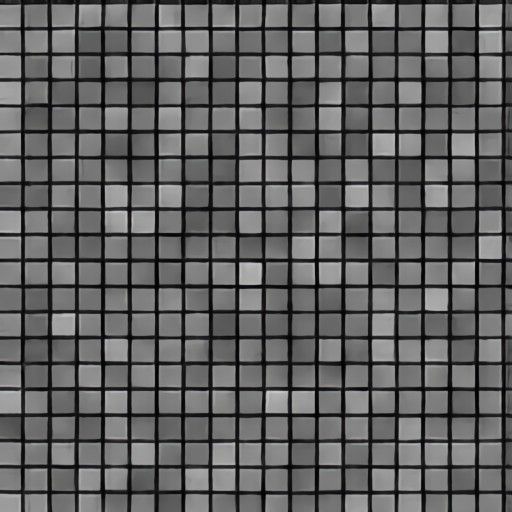}
            \end{minipage}	
            \begin{minipage}{0.13\linewidth}
            \includegraphics[width=\linewidth]{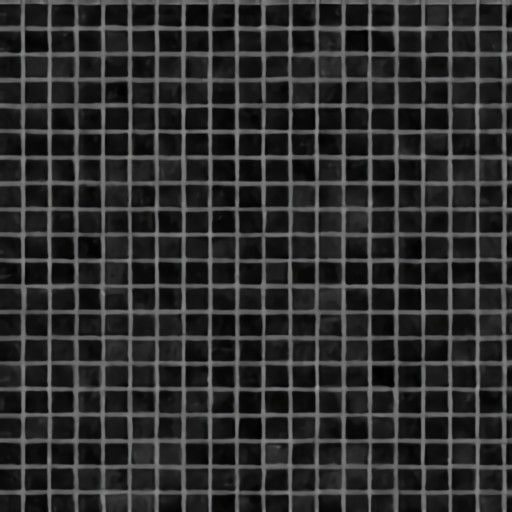}
            \end{minipage}	
            \begin{minipage}{0.13\linewidth}
            \includegraphics[width=\linewidth]{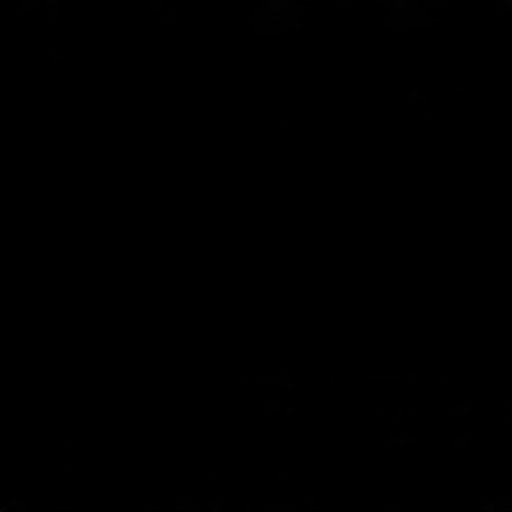}
            \end{minipage}	
            \begin{minipage}{0.13\linewidth}
            \includegraphics[width=\linewidth]{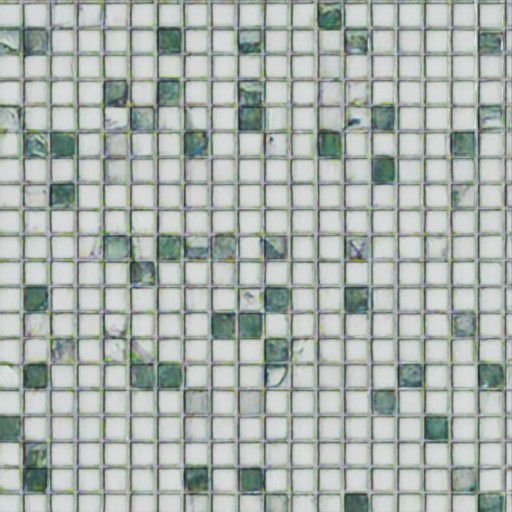}
            \end{minipage}	
        \end{minipage}	
    \end{minipage}	

    \begin{minipage}{3.4in}
        \begin{minipage}{0.02in}	
            \centering
                \rotatebox{90}{\parbox{1cm}{\centering\tiny \vspace{0.05cm} "Marble tiles"}}
        \end{minipage}	
        \hspace{0.02in}
         \begin{minipage}{3.3in}	
            \centering
            \begin{minipage}{0.13\linewidth}
            \includegraphics[width=\linewidth]{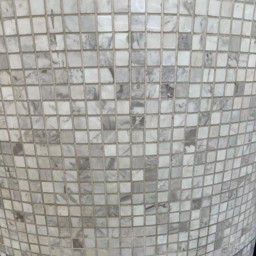}
            \end{minipage}	
            \begin{minipage}{0.13\linewidth}
            \includegraphics[width=\linewidth]{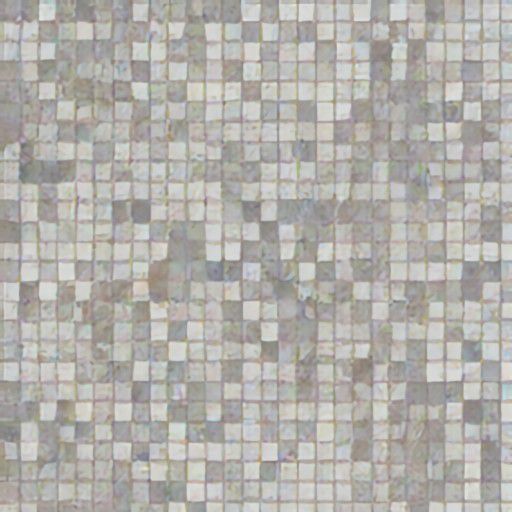}
            \end{minipage}	
            \begin{minipage}{0.13\linewidth}
            \includegraphics[width=\linewidth]{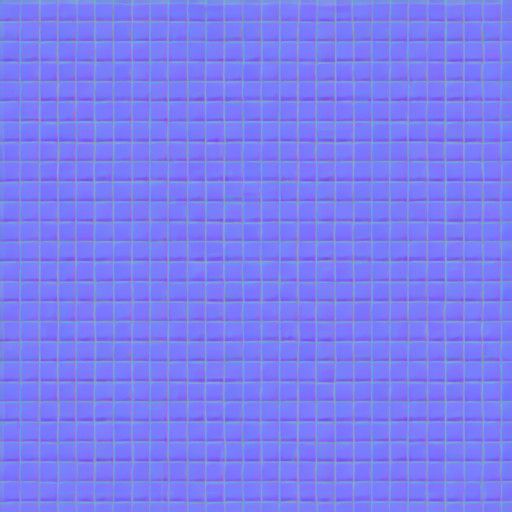}
            \end{minipage}	
            \begin{minipage}{0.13\linewidth}
            \includegraphics[width=\linewidth]{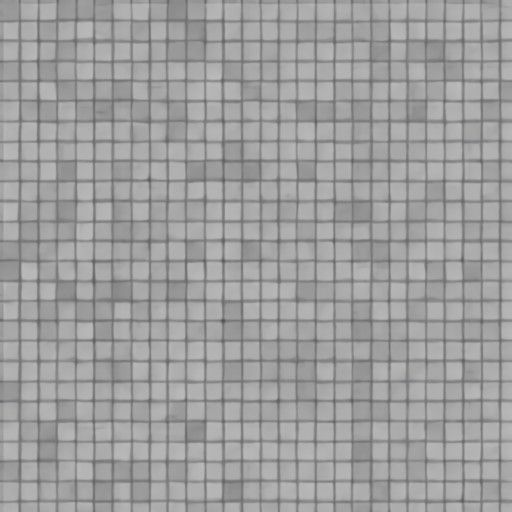}
            \end{minipage}	
            \begin{minipage}{0.13\linewidth}
            \includegraphics[width=\linewidth]{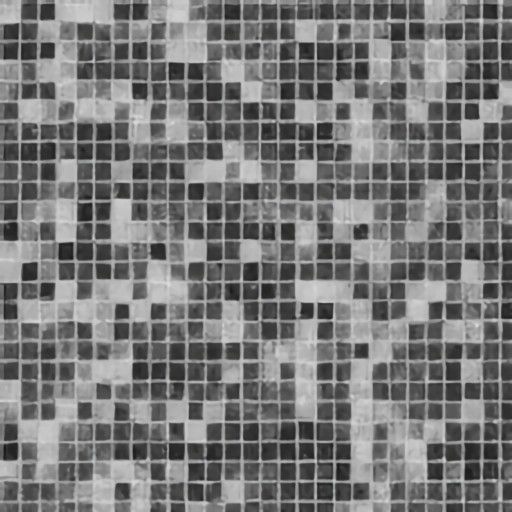}
            \end{minipage}	
            \begin{minipage}{0.13\linewidth}
            \includegraphics[width=\linewidth]{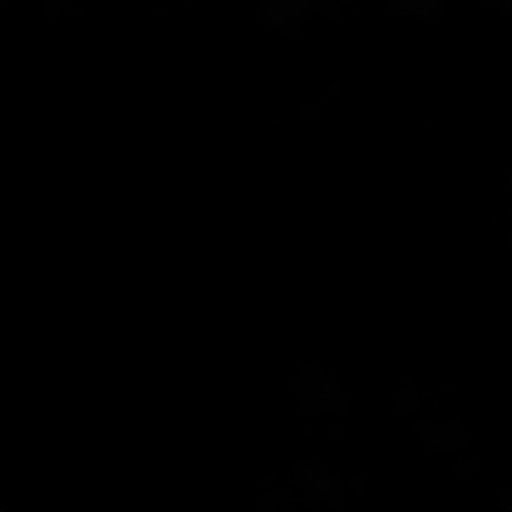}
            \end{minipage}	
            \begin{minipage}{0.13\linewidth}
            \includegraphics[width=\linewidth]{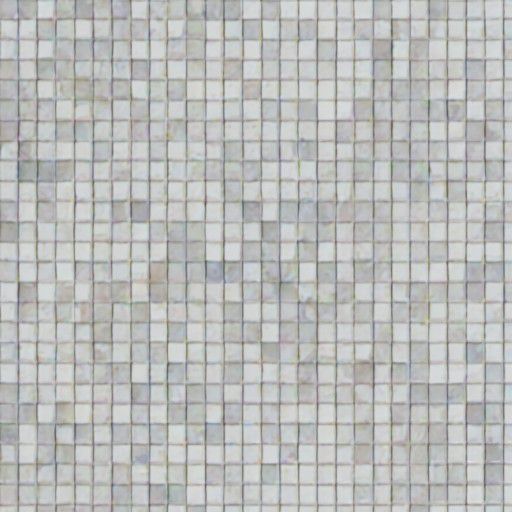}
            \end{minipage}	
        \end{minipage}	
    \end{minipage}	
    
   \caption{Evaluation of our model's adaptability to different input texture scales on real photographs. We can see that our results are generated with a scale matching that of the input.} 
   \label{fig:different_scale}
\end{figure}

Reproducing the texture scale in the input photos is critical for material generation. As we process our training data to align the scales of input images and output material maps (Sec. \ref{sec:dataset}), our model generates scale-matched materials, as shown in Fig.~\ref{fig:different_scale}. We see that our result follows the scale of the input as it increases from top to bottom.

\subsubsection{Evaluations on the Robustness.}\label{sec:robustness}
\begin{figure}[htbp!]
    \centering		
    \begin{minipage}{3.4in}
        \begin{minipage}{0.02in}	
            \centering
                \vspace{0.2in}
                % \rotatebox{90}{\tiny {"Gingham"}}
                \rotatebox{90}{\parbox{1cm}{\centering\tiny \vspace{0.05cm} "Gingham"}}
        \end{minipage}	
        \hspace{0.02in}
             \begin{minipage}{3.3in}	
                \centering
                \begin{minipage}{0.13\linewidth}
                    \subcaption*{\tiny Input}
                    \includegraphics[width=\linewidth]{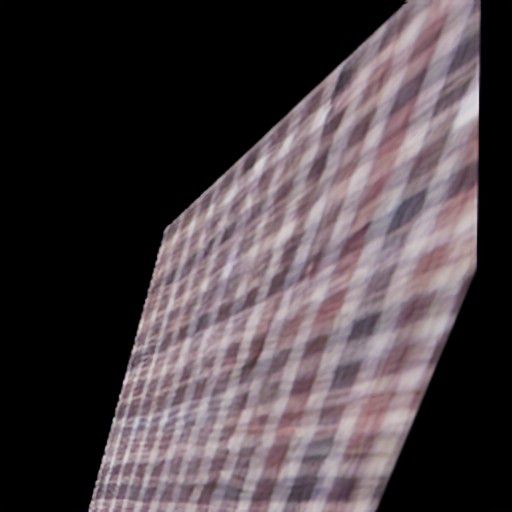}
                \end{minipage}
                \begin{minipage}{0.13\linewidth}
                    \subcaption*{\tiny Albedo}
                    \includegraphics[width=\linewidth]{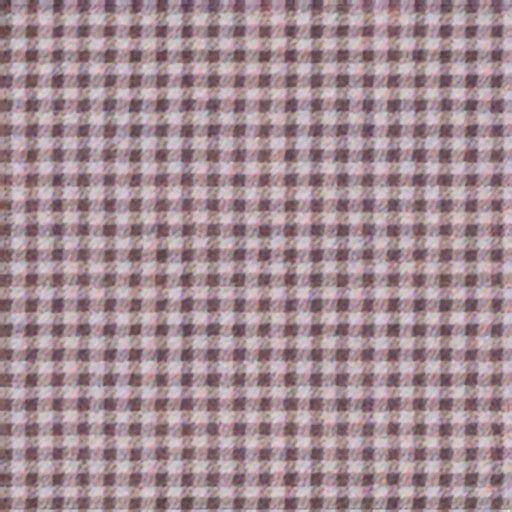}
                \end{minipage}
                \begin{minipage}{0.13\linewidth}
                    \subcaption*{\tiny Normal}
                    \includegraphics[width=\linewidth]{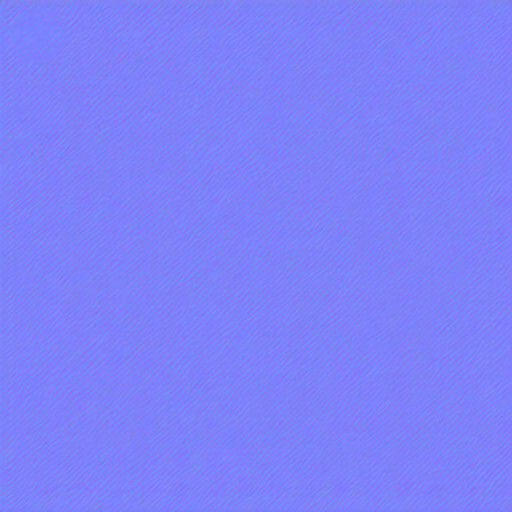}
                \end{minipage}
                \begin{minipage}{0.13\linewidth}
                    \subcaption*{\tiny Roughness}
                    \includegraphics[width=\linewidth]{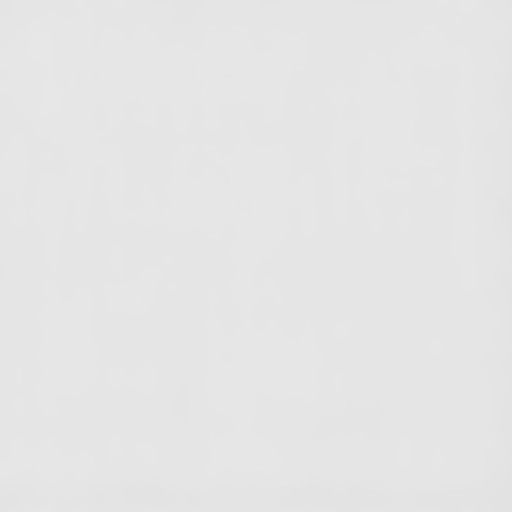}
                \end{minipage}
                \begin{minipage}{0.13\linewidth}
                    \subcaption*{\tiny Height}
                    \includegraphics[width=\linewidth]{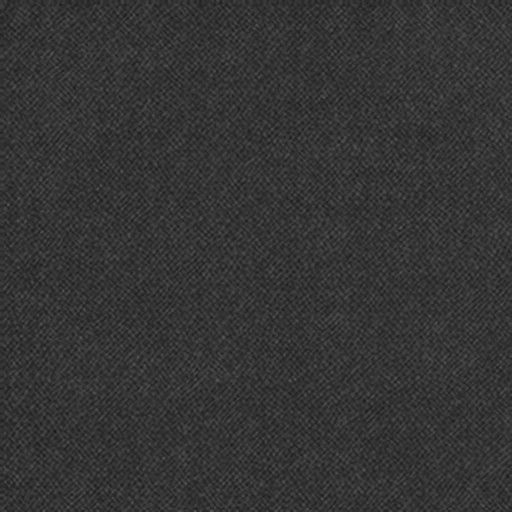}
                \end{minipage}
                \begin{minipage}{0.13\linewidth}
                    \subcaption*{\tiny Metallic}
                    \includegraphics[width=\linewidth]{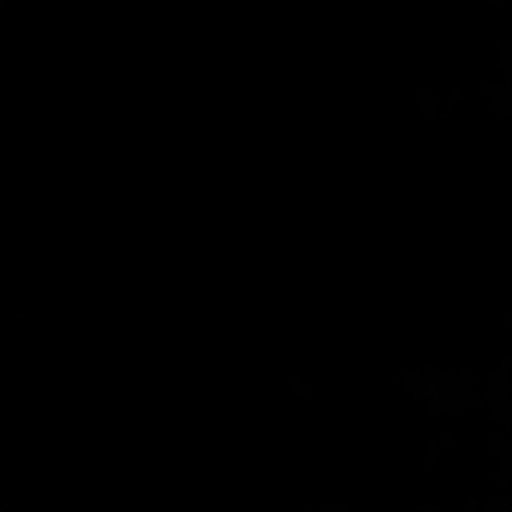}
                \end{minipage}
                \begin{minipage}{0.13\linewidth}
                    \subcaption*{\tiny Render}
                    \includegraphics[width=\linewidth]{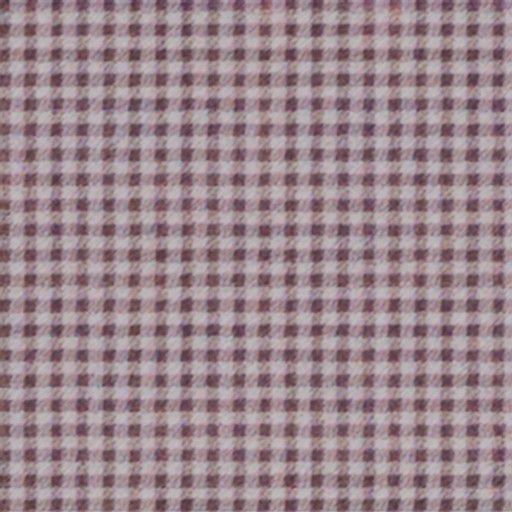}
                \end{minipage}
            \end{minipage}	
        \end{minipage}	

    \begin{minipage}{3.4in}
        \begin{minipage}{0.02in}	
            \centering
                \rotatebox{90}{\parbox{1cm}{\centering\tiny "Marble\vspace{-0.05cm}\\mosaic"}}
        \end{minipage}	
        \hspace{0.02in}
         \begin{minipage}{3.3in}	
            \centering
            \begin{minipage}{0.13\linewidth}
            \includegraphics[width=\linewidth]{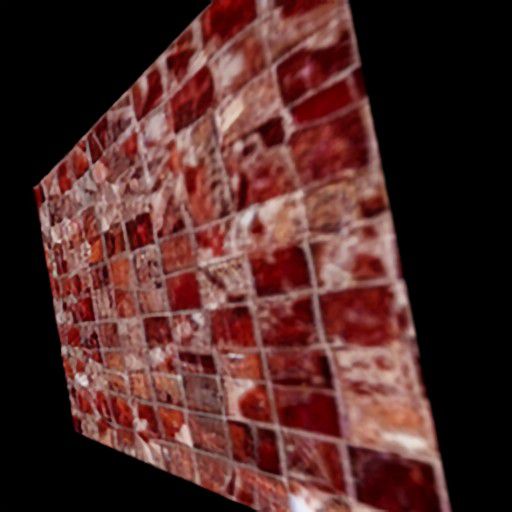}
            \end{minipage}	
            \begin{minipage}{0.13\linewidth}
            \includegraphics[width=\linewidth]{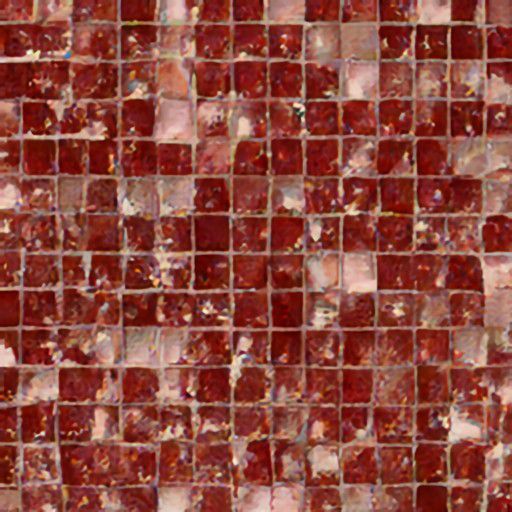}
            \end{minipage}	
            \begin{minipage}{0.13\linewidth}
            \includegraphics[width=\linewidth]{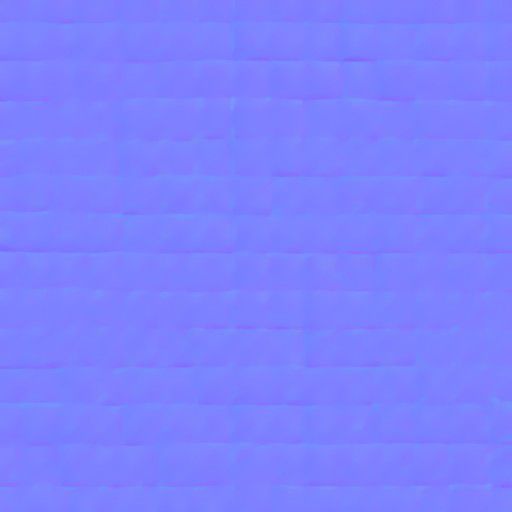}
            \end{minipage}	
            \begin{minipage}{0.13\linewidth}
            \includegraphics[width=\linewidth]{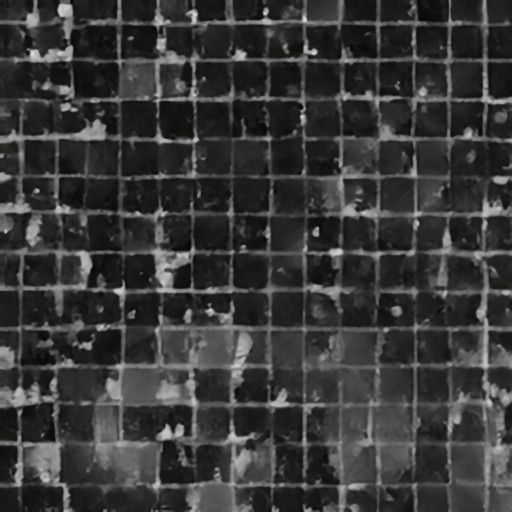}
            \end{minipage}	
            \begin{minipage}{0.13\linewidth}
            \includegraphics[width=\linewidth]{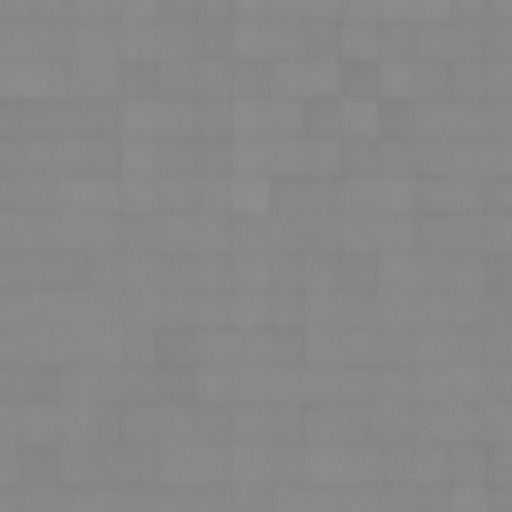}
            \end{minipage}	
            \begin{minipage}{0.13\linewidth}
            \includegraphics[width=\linewidth]{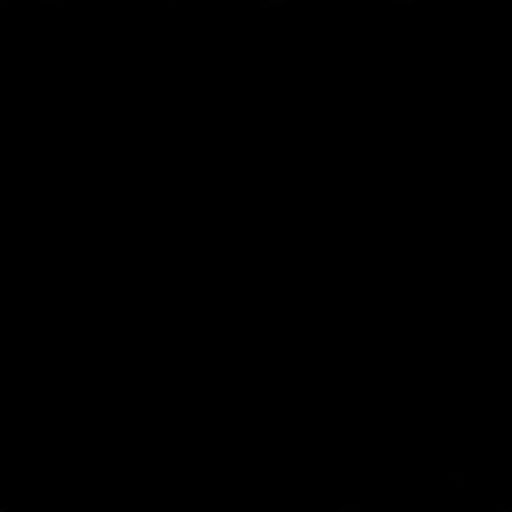}
            \end{minipage}	
            \begin{minipage}{0.13\linewidth}
            \includegraphics[width=\linewidth]{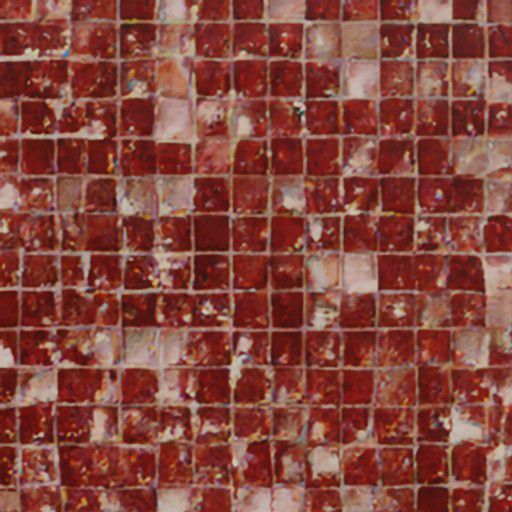}
            \end{minipage}
        \end{minipage}	
    \end{minipage}	

    \begin{minipage}{3.4in}
        \begin{minipage}{0.02in}	
            \centering
                \rotatebox{90}{\parbox{1cm}{\centering\tiny \vspace{0.05cm} "Ice"}}
        \end{minipage}	
        \hspace{0.02in}
         \begin{minipage}{3.3in}	
            \centering
            \begin{minipage}{0.13\linewidth}
            \includegraphics[width=\linewidth]{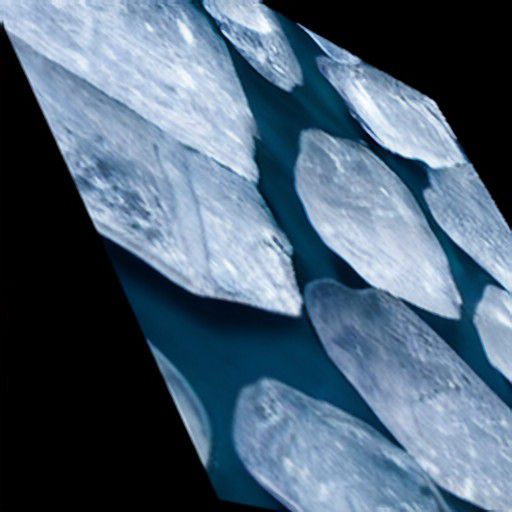}
            \end{minipage}	
            \begin{minipage}{0.13\linewidth}
            \includegraphics[width=\linewidth]{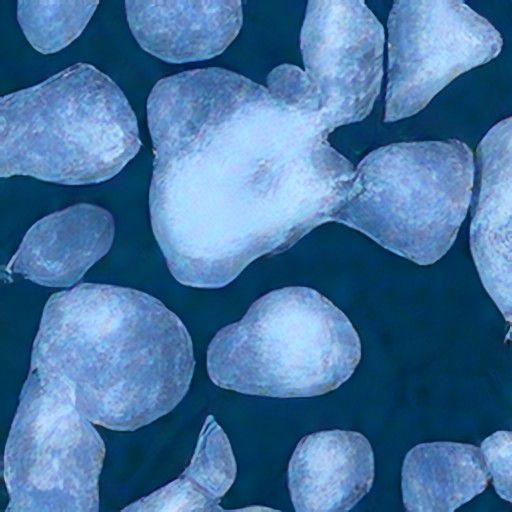}
            \end{minipage}	
            \begin{minipage}{0.13\linewidth}
            \includegraphics[width=\linewidth]{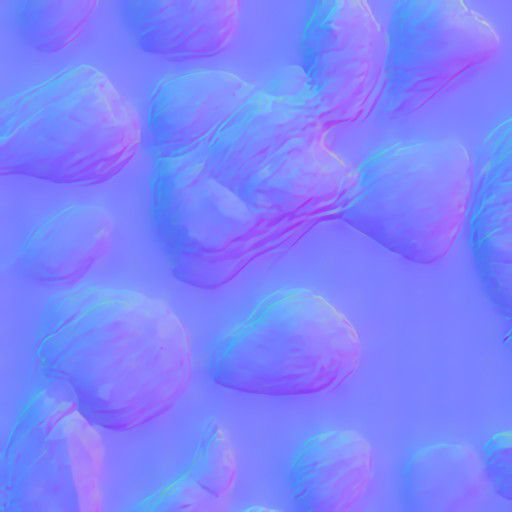}
            \end{minipage}	
            \begin{minipage}{0.13\linewidth}
            \includegraphics[width=\linewidth]{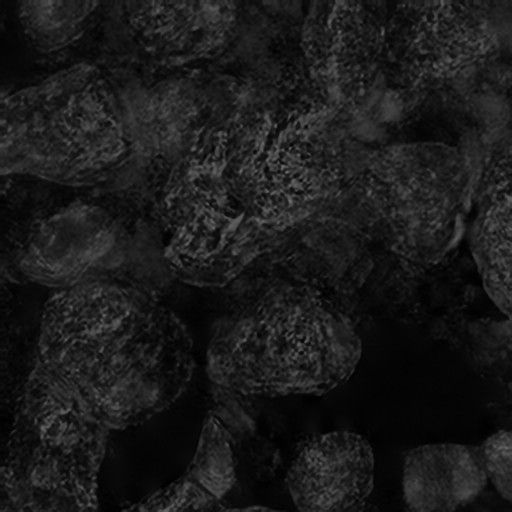}
            \end{minipage}	
            \begin{minipage}{0.13\linewidth}
            \includegraphics[width=\linewidth]{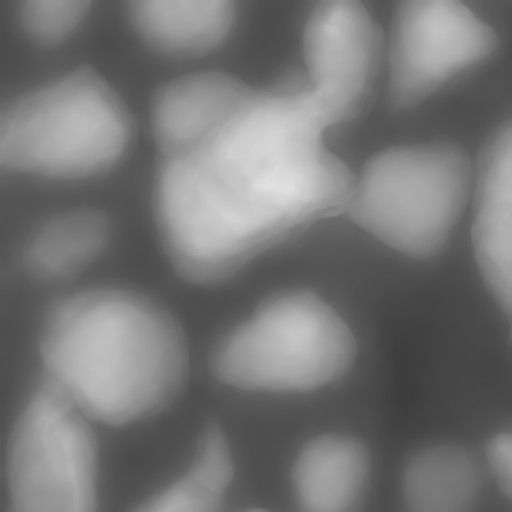}
            \end{minipage}	
            \begin{minipage}{0.13\linewidth}
            \includegraphics[width=\linewidth]{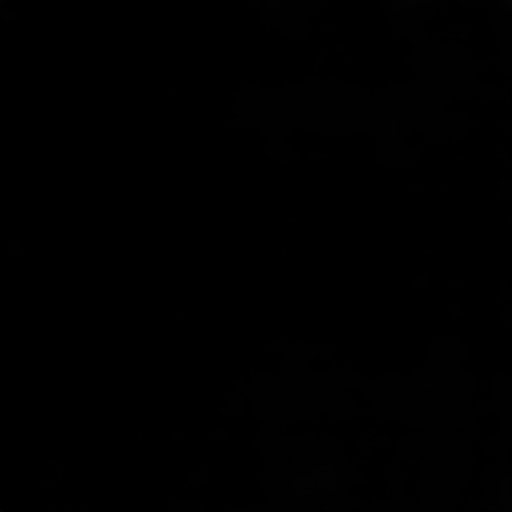}
            \end{minipage}	
            \begin{minipage}{0.13\linewidth}
            \includegraphics[width=\linewidth]{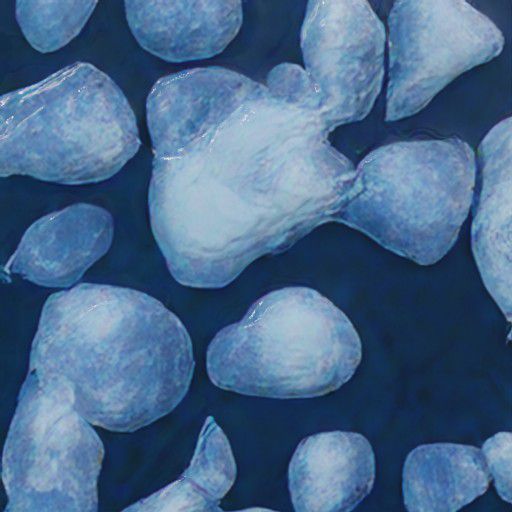}
            \end{minipage}
        \end{minipage}	
    \end{minipage}	
    
    \begin{minipage}{3.4in}
        \begin{minipage}{0.02in}	
            \centering
                \rotatebox{90}{\parbox{1cm}{\centering\tiny \vspace{0.05cm} "Leather"}}
        \end{minipage}	
        \hspace{0.02in}
         \begin{minipage}{3.3in}	
            \centering
            \begin{minipage}{0.13\linewidth}
            \includegraphics[width=\linewidth]{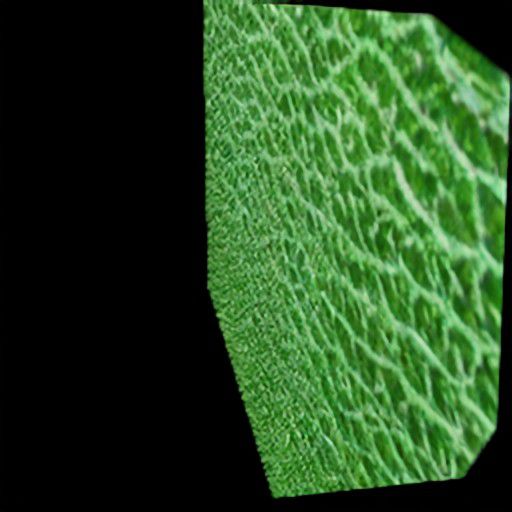}
            \end{minipage}	
            \begin{minipage}{0.13\linewidth}
            \includegraphics[width=\linewidth]{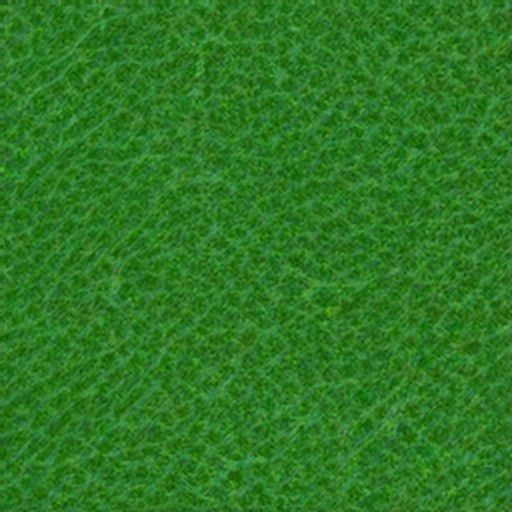}
            \end{minipage}	
            \begin{minipage}{0.13\linewidth}
            \includegraphics[width=\linewidth]{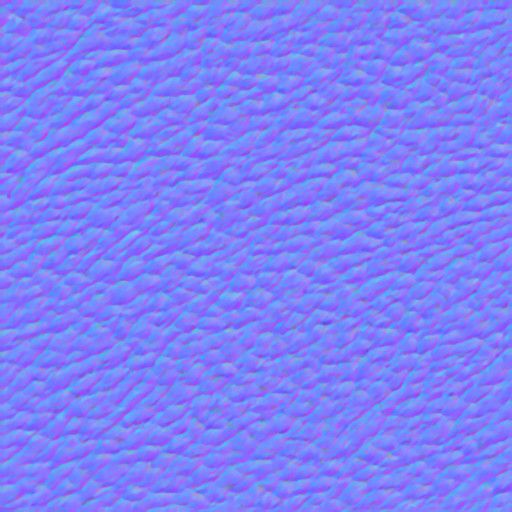}
            \end{minipage}	
            \begin{minipage}{0.13\linewidth}
            \includegraphics[width=\linewidth]{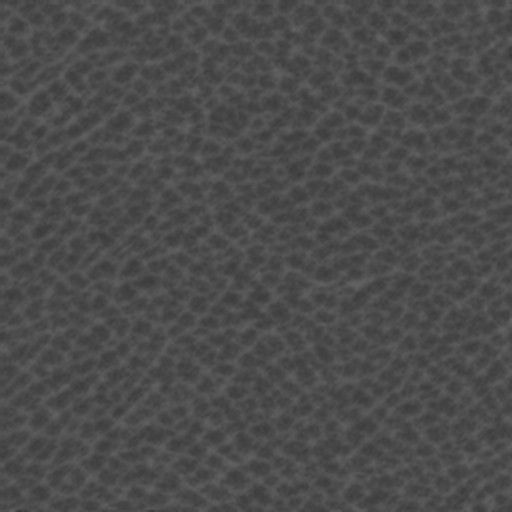}
            \end{minipage}	
            \begin{minipage}{0.13\linewidth}
            \includegraphics[width=\linewidth]{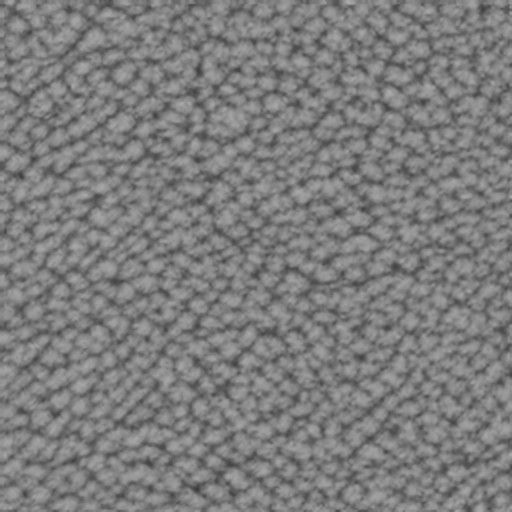}
            \end{minipage}	
            \begin{minipage}{0.13\linewidth}
            \includegraphics[width=\linewidth]{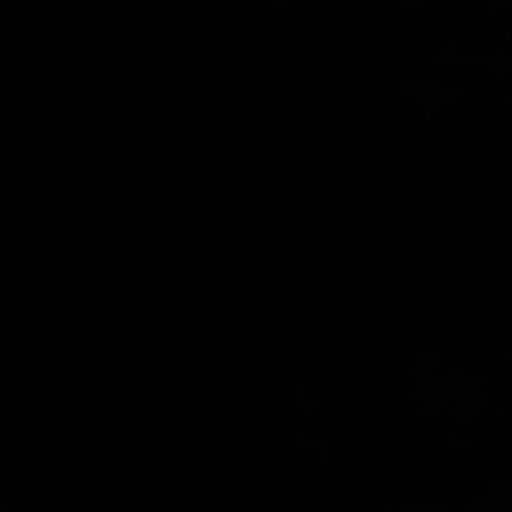}
            \end{minipage}	
            \begin{minipage}{0.13\linewidth}
            \includegraphics[width=\linewidth]{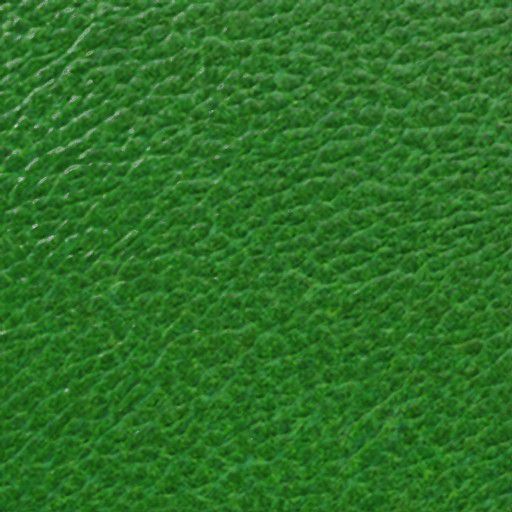}
            \end{minipage}
        \end{minipage}	
    \end{minipage}	

   \caption{Evaluation of our model’s robustness to varying levels of distortion. The first column shows textures transformed with homography and thin plate spline transformations. The following columns present the material maps and the rendering images. The results demonstrate that our model effectively rectifies textures with various patterns and different types of distortion, maintaining high-quality outputs.} 
   \label{fig:distortion}
\end{figure}

To examine the robustness of our model to strong, real-world, distortions, we generate a synthetic test set that use textures from the texture datasets TexSD~\cite{lopes2024material} and follow the texture processing steps outlined by~\citet{hao2023diffusion}. We apply homography transformations~\cite{hartley2003multiple} and thin plate spline transformations~\cite{bookstein1989principal} to the textures. Our results in Fig.~\ref{fig:distortion} show that the model is robust to severe distortions, stretching, and the blurring effects introduced by these transformations. More examples of real photos with distortion or surface geometry diversity can be found in supplemental materials.

\begin{figure}[htbp!]
    \centering		
    \begin{minipage}{3.4in}
        \begin{minipage}{0.02in}	
            \centering
                \vspace{0.1in}
                \rotatebox{90}{\parbox{1cm}{\centering\tiny \vspace{0.05cm} "Wood"}}
        \end{minipage}	
        \hspace{0.02in}
             \begin{minipage}{3.3in}	
                \centering
                \begin{minipage}{0.13\linewidth}
                    \subcaption*{\tiny Input}
                    \includegraphics[width=\linewidth]{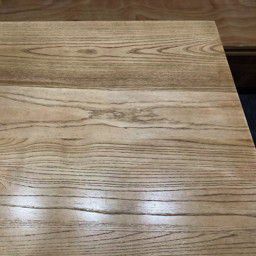}
                \end{minipage}
                \begin{minipage}{0.13\linewidth}
                    \subcaption*{\tiny Albedo}
                    \includegraphics[width=\linewidth]{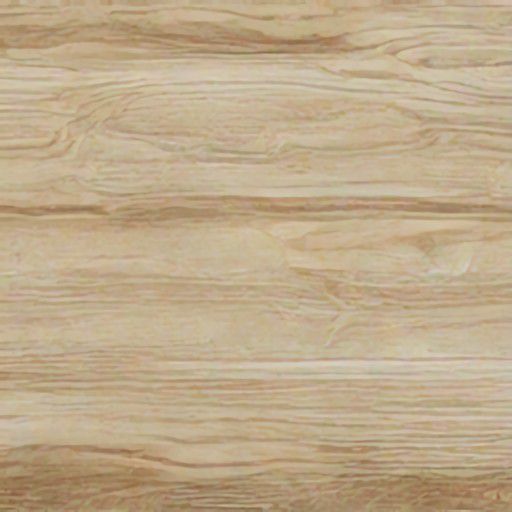}
                \end{minipage}
                \begin{minipage}{0.13\linewidth}
                    \subcaption*{\tiny Normal}
                    \includegraphics[width=\linewidth]{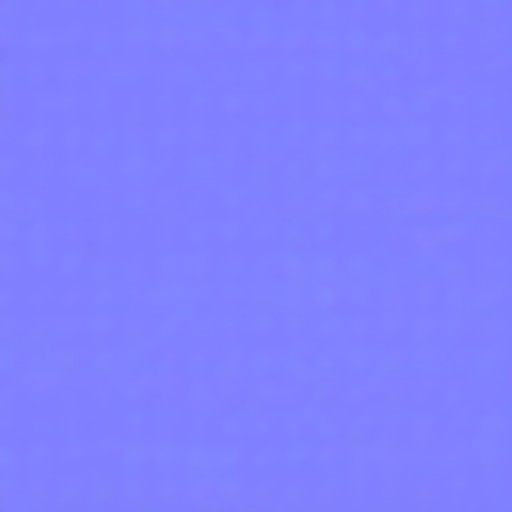}
                \end{minipage}
                \begin{minipage}{0.13\linewidth}
                    \subcaption*{\tiny Roughness}
                    \includegraphics[width=\linewidth]{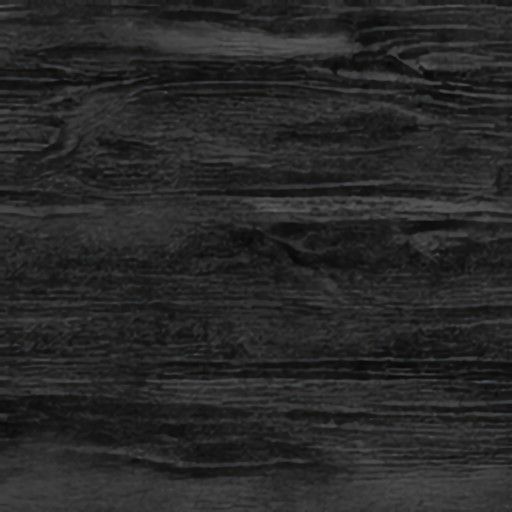}
                \end{minipage}
                \begin{minipage}{0.13\linewidth}
                    \subcaption*{\tiny Height}
                    \includegraphics[width=\linewidth]{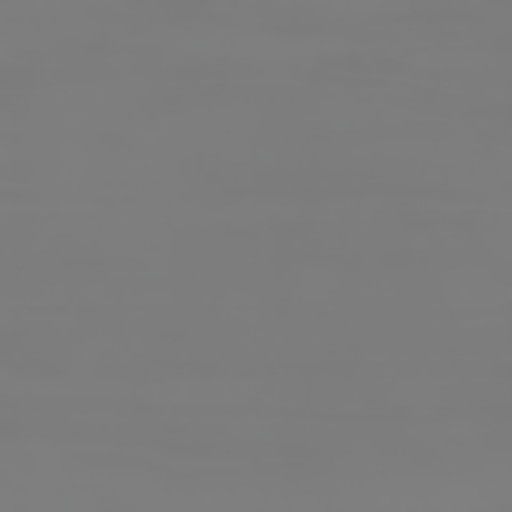}
                \end{minipage}
                \begin{minipage}{0.13\linewidth}
                    \subcaption*{\tiny Metallic}
                    \includegraphics[width=\linewidth]{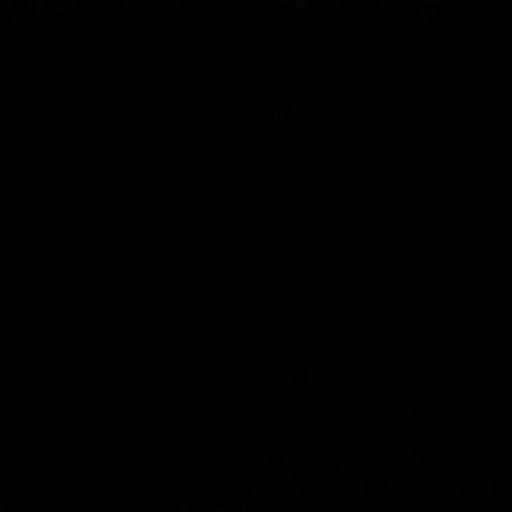}
                \end{minipage}
                \begin{minipage}{0.13\linewidth}
                    \subcaption*{\tiny Render}
                    \includegraphics[width=\linewidth]{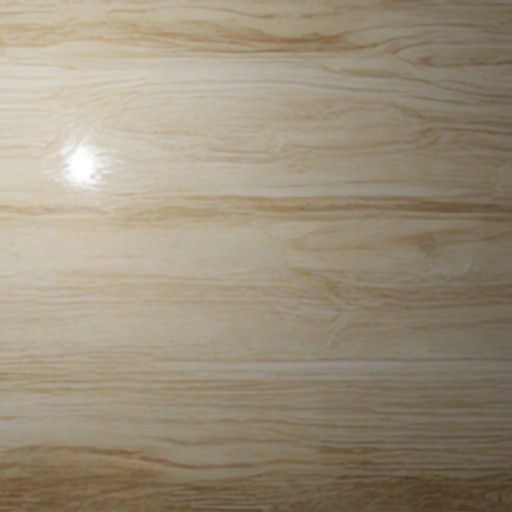}
                \end{minipage}
            \end{minipage}	
        \end{minipage}	

    \begin{minipage}{3.4in}
        \begin{minipage}{0.02in}	
            \centering
                \rotatebox{90}{\parbox{1cm}{\centering\tiny \vspace{0.05cm} "Leather"}}
        \end{minipage}	
        \hspace{0.02in}
         \begin{minipage}{3.3in}	
            \centering
            \begin{minipage}{0.13\linewidth}
            \includegraphics[width=\linewidth]{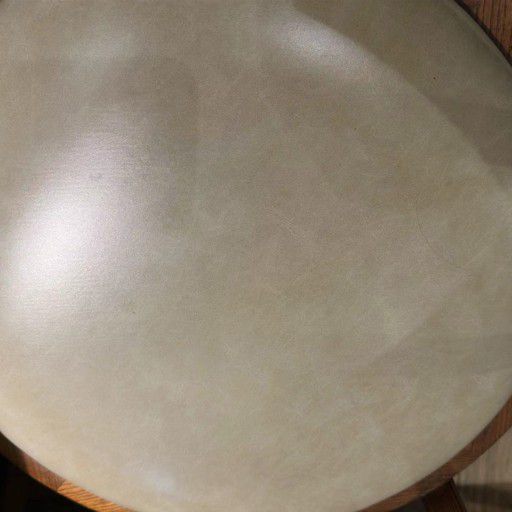}
            \end{minipage}	
            \begin{minipage}{0.13\linewidth}
            \includegraphics[width=\linewidth]{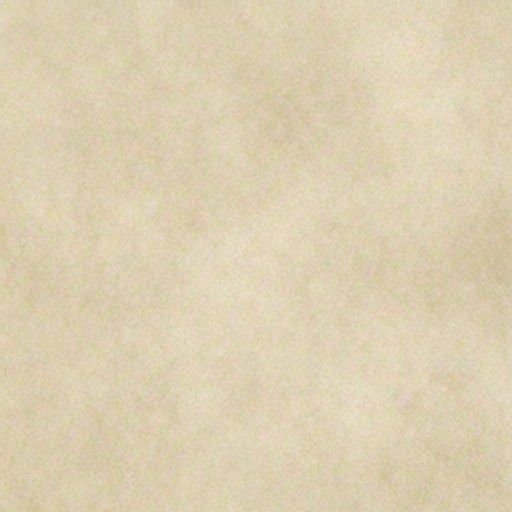}
            \end{minipage}	
            \begin{minipage}{0.13\linewidth}
            \includegraphics[width=\linewidth]{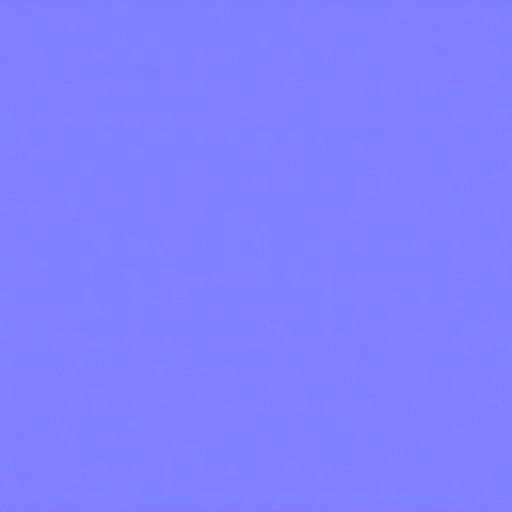}
            \end{minipage}	
            \begin{minipage}{0.13\linewidth}
            \includegraphics[width=\linewidth]{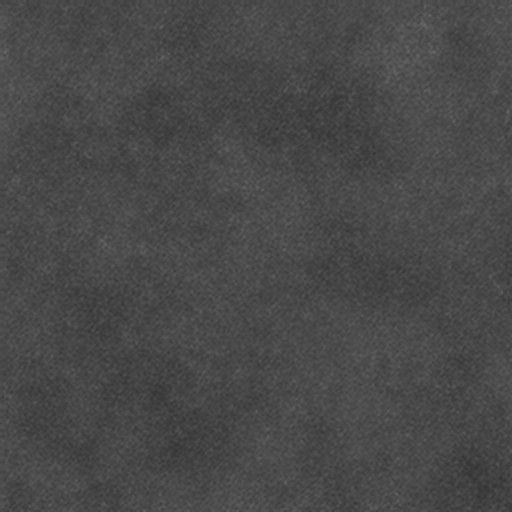}
            \end{minipage}	
            \begin{minipage}{0.13\linewidth}
            \includegraphics[width=\linewidth]{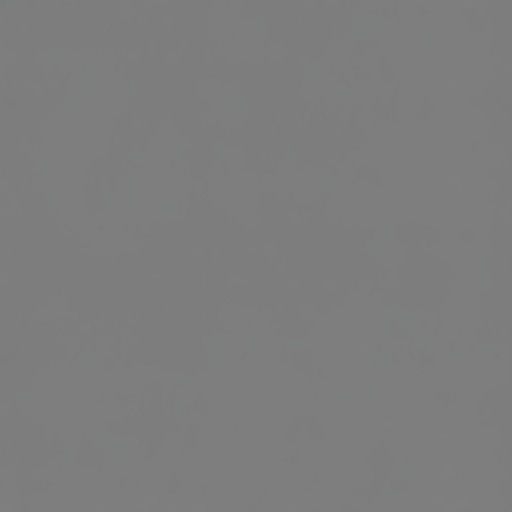}
            \end{minipage}	
            \begin{minipage}{0.13\linewidth}
            \includegraphics[width=\linewidth]{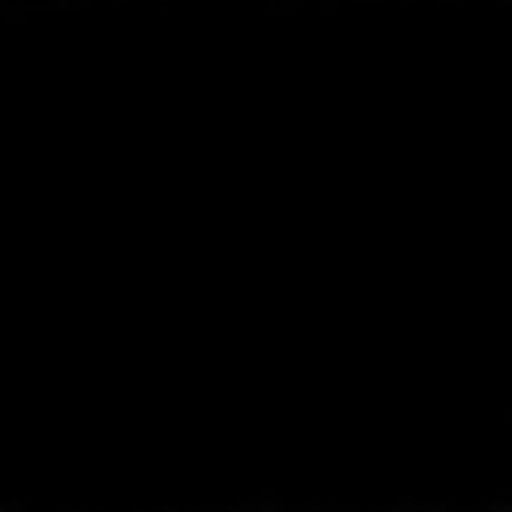}
            \end{minipage}	
            \begin{minipage}{0.13\linewidth}
            \includegraphics[width=\linewidth]{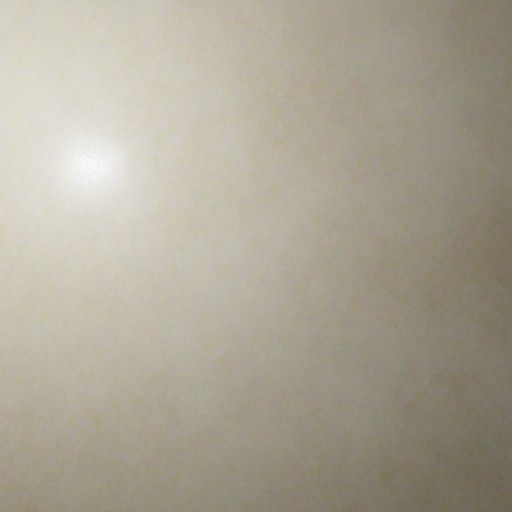}
            \end{minipage}	
        \end{minipage}	
    \end{minipage}	

    \begin{minipage}{3.4in}
        \begin{minipage}{0.02in}	
            \centering
                \rotatebox{90}{\parbox{1cm}{\centering\tiny \vspace{0.05cm} "Marble"}}
        \end{minipage}	
        \hspace{0.02in}
         \begin{minipage}{3.3in}	
            \centering
            \begin{minipage}{0.13\linewidth}
            \includegraphics[width=\linewidth]{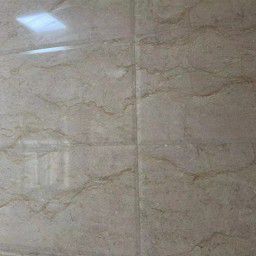}
            \end{minipage}	
            \begin{minipage}{0.13\linewidth}
            \includegraphics[width=\linewidth]{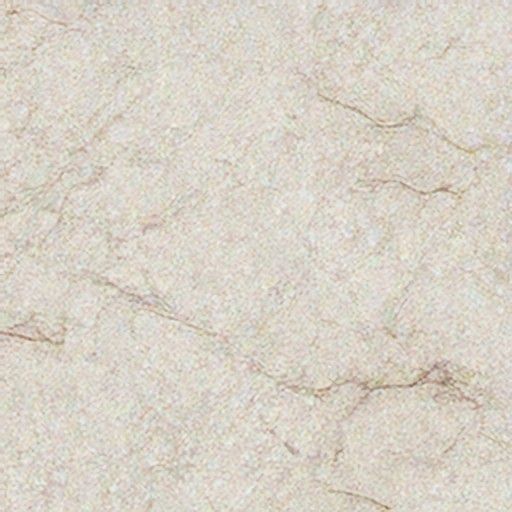}
            \end{minipage}	
            \begin{minipage}{0.13\linewidth}
            \includegraphics[width=\linewidth]{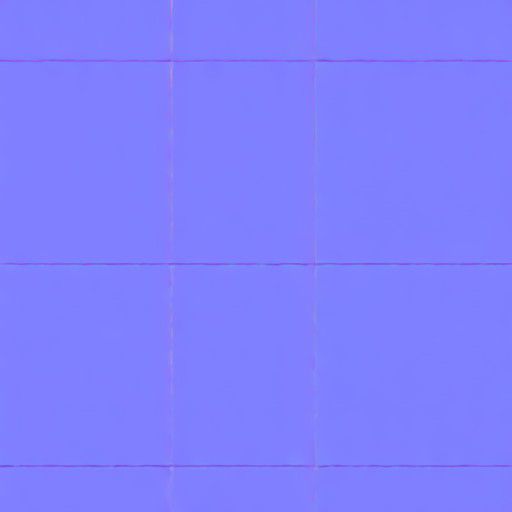}
            \end{minipage}	
            \begin{minipage}{0.13\linewidth}
            \includegraphics[width=\linewidth]{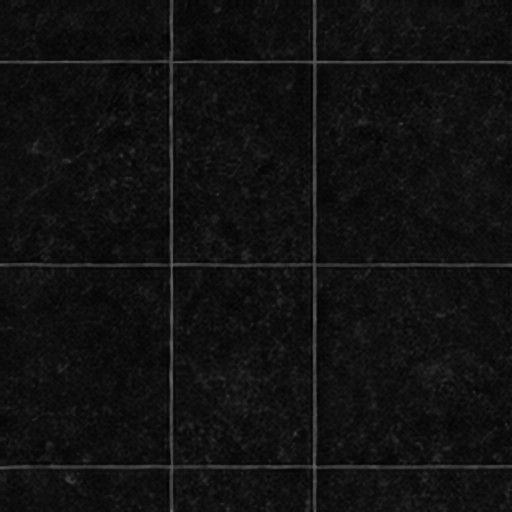}
            \end{minipage}	
            \begin{minipage}{0.13\linewidth}
            \includegraphics[width=\linewidth]{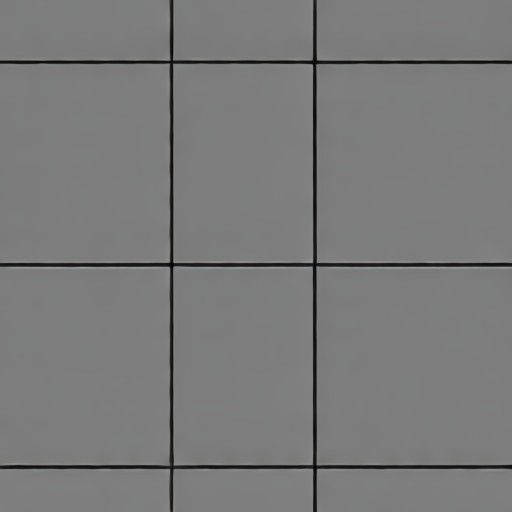}
            \end{minipage}	
            \begin{minipage}{0.13\linewidth}
            \includegraphics[width=\linewidth]{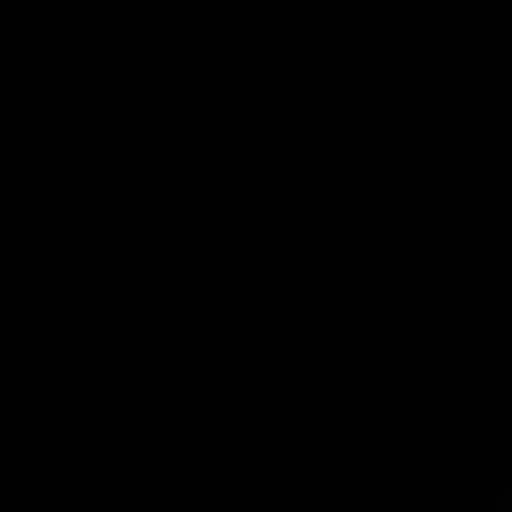}
            \end{minipage}	
            \begin{minipage}{0.13\linewidth}
            \includegraphics[width=\linewidth]{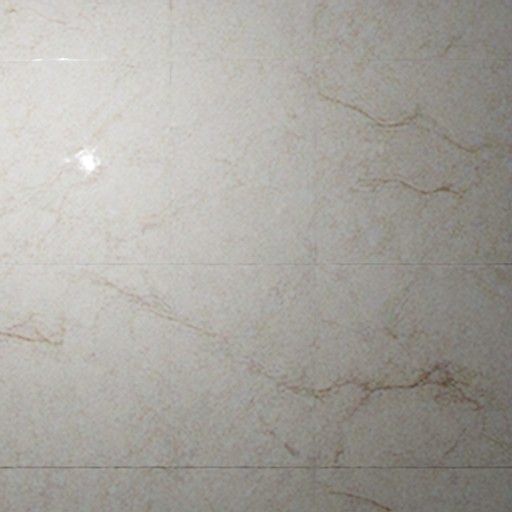}
            \end{minipage}	
        \end{minipage}	
    \end{minipage}		

   \caption{Evaluation of the robustness of our model to lighting and shadow interference. We test scenarios where the input photographs contain point light sources, shadows, or environmental reflections. The generated material maps and rendered images demonstrate the ability of our model to handle these interferences, preserving material quality and accurately representing the input photos. The leftmost side of each row is labeled with the text conditioning input used. }
   \label{fig:different_lights}
\end{figure}

We further evaluate the model's performance when the input image contains specular highlights and shadows in Fig.~\ref{fig:different_lights}. We see that these highlights and shadows in real photos do not ``leak'' into material maps, highlighting the model’s robustness to various lighting conditions.

\begin{figure}[htbp!]
    \centering		
    \begin{minipage}{3.4in}
        \begin{minipage}{0.02in}	
            \centering
                \vspace{0.2in}
                \rotatebox{90}{\parbox{1cm}{\centering\tiny \vspace{0.05cm} "Snake skin"}}
        \end{minipage}	
        \hspace{0.02in}
             \begin{minipage}{3.3in}	
                \centering
                \begin{minipage}{0.13\linewidth}
                    \subcaption*{\tiny Input}
                    \includegraphics[width=\linewidth]{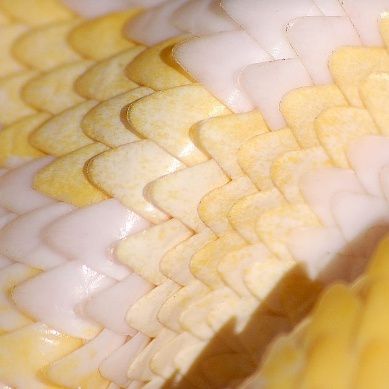}
                \end{minipage}
                \begin{minipage}{0.13\linewidth}
                    \subcaption*{\tiny Albedo}
                    \includegraphics[width=\linewidth]{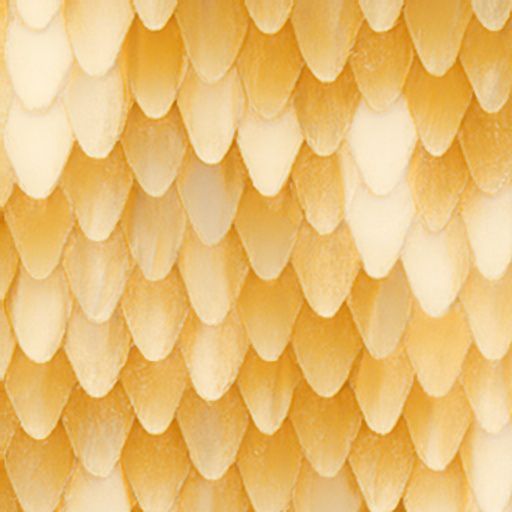}
                \end{minipage}
                \begin{minipage}{0.13\linewidth}
                    \subcaption*{\tiny Normal}
                    \includegraphics[width=\linewidth]{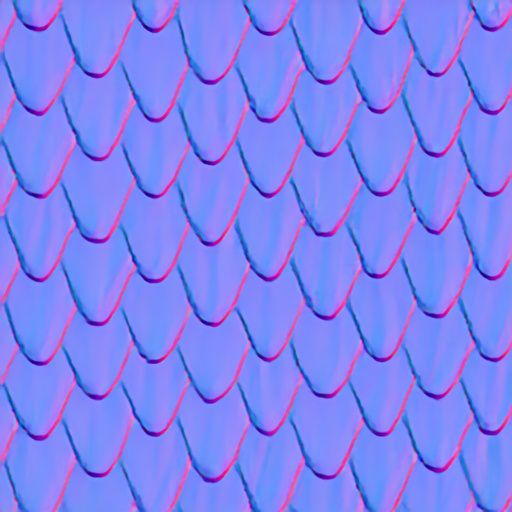}
                \end{minipage}
                \begin{minipage}{0.13\linewidth}
                    \subcaption*{\tiny Roughness}
                    \includegraphics[width=\linewidth]{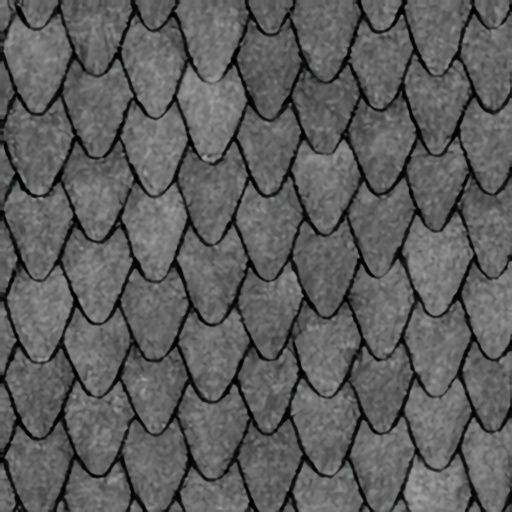}
                \end{minipage}
                \begin{minipage}{0.13\linewidth}
                    \subcaption*{\tiny Height}
                    \includegraphics[width=\linewidth]{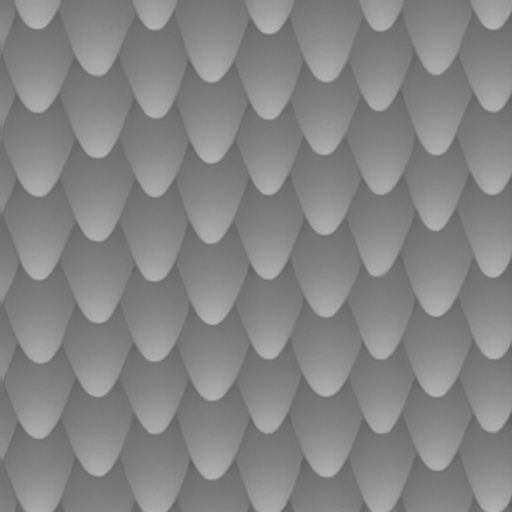}
                \end{minipage}
                \begin{minipage}{0.13\linewidth}
                    \subcaption*{\tiny Metallic}
                    \includegraphics[width=\linewidth]{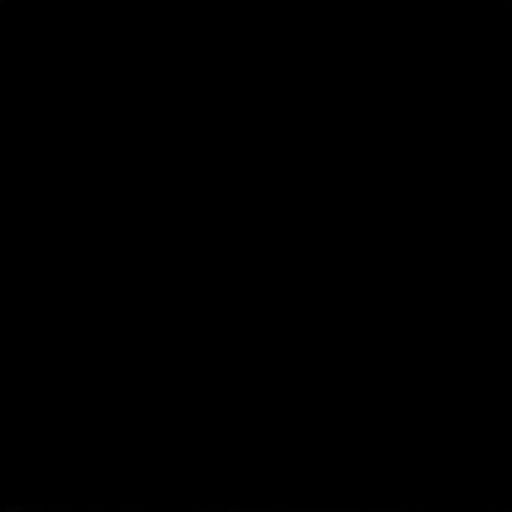}
                \end{minipage}
                \begin{minipage}{0.13\linewidth}
                    \subcaption*{\tiny Render}
                    \includegraphics[width=\linewidth]{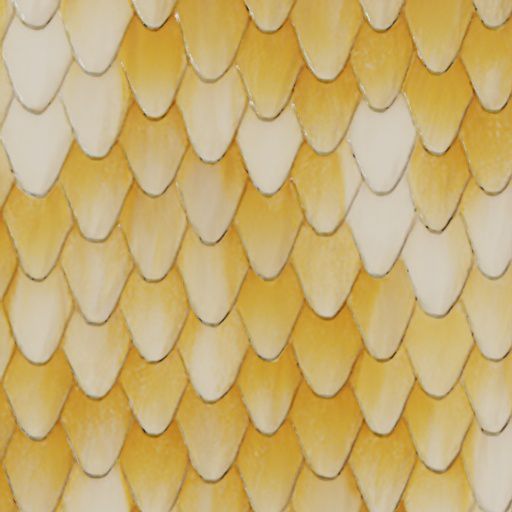}
                \end{minipage}
            \end{minipage}	
        \end{minipage}	

    \begin{minipage}{3.4in}
        \begin{minipage}{0.02in}	
            \centering
                \rotatebox{90}{\parbox{1cm}{\centering\tiny \vspace{0.05cm} "Frog skin"}}
        \end{minipage}	
        \hspace{0.02in}
         \begin{minipage}{3.3in}	
            \centering
            \begin{minipage}{0.13\linewidth}
            \includegraphics[width=\linewidth]{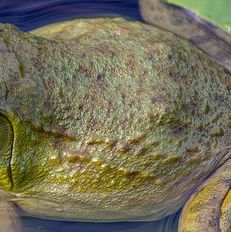}
            \end{minipage}	
            \begin{minipage}{0.13\linewidth}
            \includegraphics[width=\linewidth]{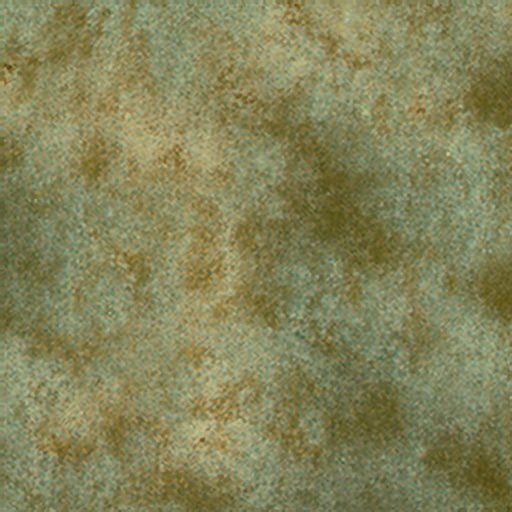}
            \end{minipage}	
            \begin{minipage}{0.13\linewidth}
            \includegraphics[width=\linewidth]{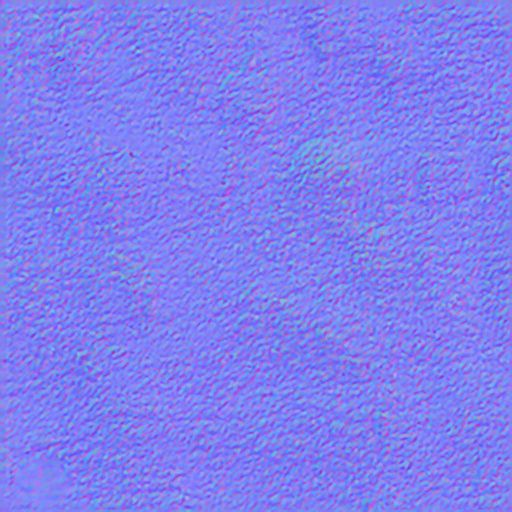}
            \end{minipage}	
            \begin{minipage}{0.13\linewidth}
            \includegraphics[width=\linewidth]{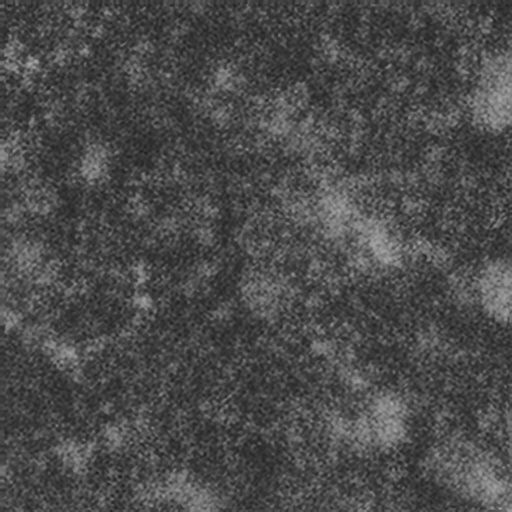}
            \end{minipage}	
            \begin{minipage}{0.13\linewidth}
            \includegraphics[width=\linewidth]{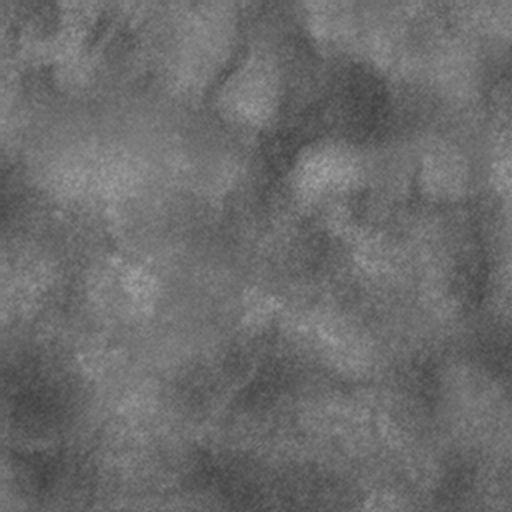}
            \end{minipage}	
            \begin{minipage}{0.13\linewidth}
            \includegraphics[width=\linewidth]{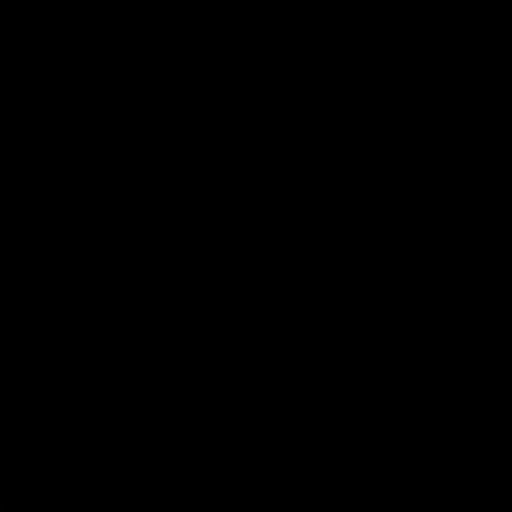}
            \end{minipage}	
            \begin{minipage}{0.13\linewidth}
            \includegraphics[width=\linewidth]{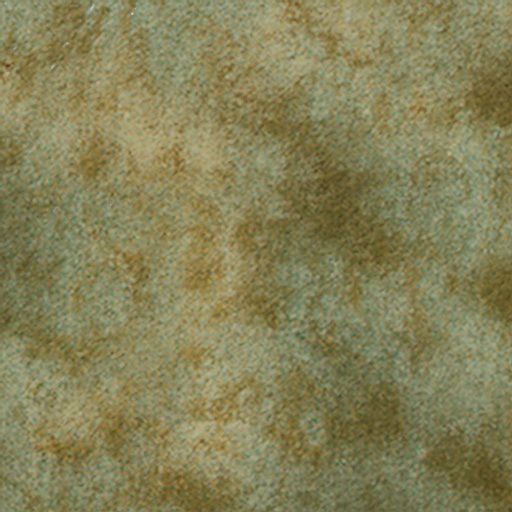}
            \end{minipage}	
        \end{minipage}	
    \end{minipage}	

    \begin{minipage}{3.4in}
        \begin{minipage}{0.02in}	
            \centering
                \rotatebox{90}{\parbox{1cm}{\centering\tiny \vspace{0.05cm} "Tiger fur"}}
        \end{minipage}	
        \hspace{0.02in}
         \begin{minipage}{3.3in}	
            \centering
            \begin{minipage}{0.13\linewidth}
            \includegraphics[width=\linewidth]{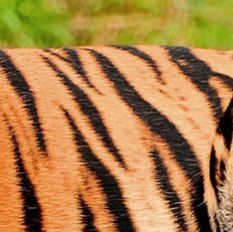}
            \end{minipage}	
            \begin{minipage}{0.13\linewidth}
            \includegraphics[width=\linewidth]{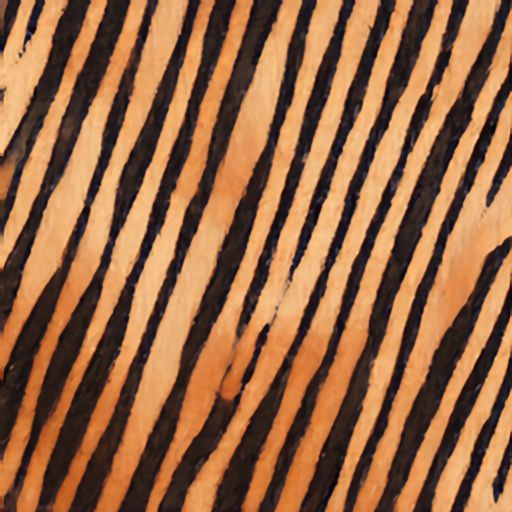}
            \end{minipage}	
            \begin{minipage}{0.13\linewidth}
            \includegraphics[width=\linewidth]{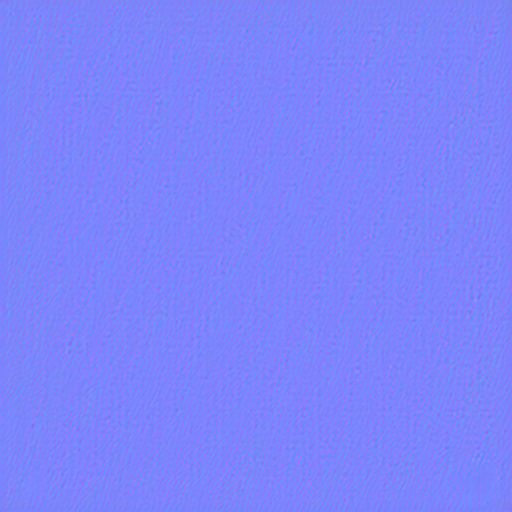}
            \end{minipage}	
            \begin{minipage}{0.13\linewidth}
            \includegraphics[width=\linewidth]{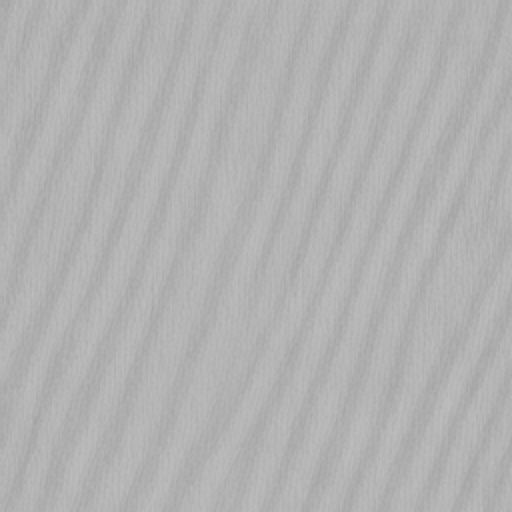}
            \end{minipage}	
            \begin{minipage}{0.13\linewidth}
            \includegraphics[width=\linewidth]{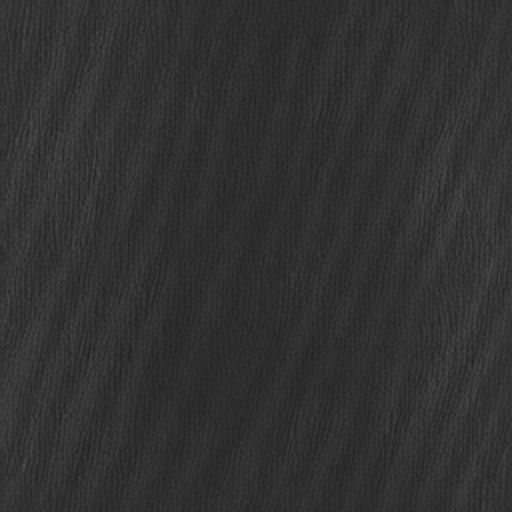}
            \end{minipage}	
            \begin{minipage}{0.13\linewidth}
            \includegraphics[width=\linewidth]{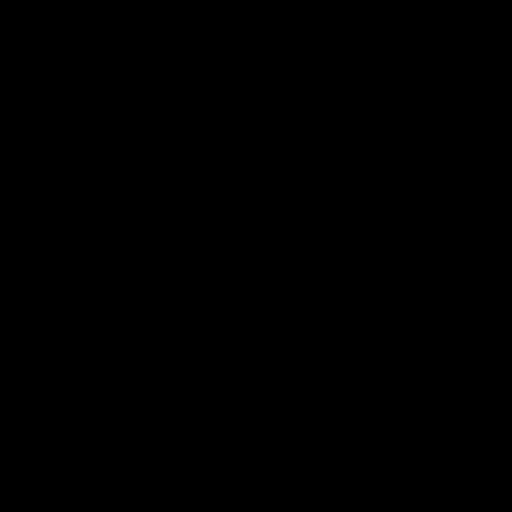}
            \end{minipage}	
            \begin{minipage}{0.13\linewidth}
            \includegraphics[width=\linewidth]{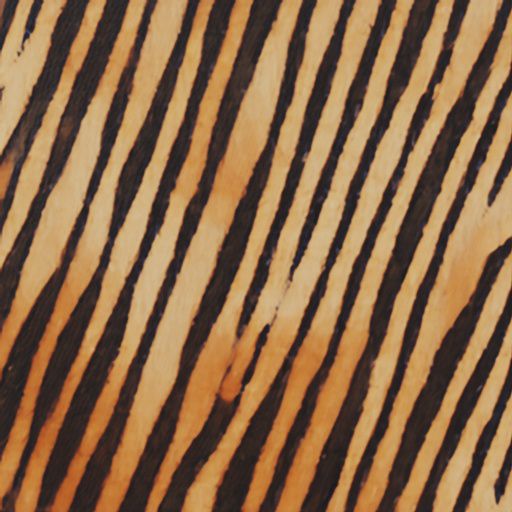}
            \end{minipage}	
        \end{minipage}	
    \end{minipage}	

    \begin{minipage}{3.4in}
        \begin{minipage}{0.02in}	
            \centering
                \rotatebox{90}{\parbox{1cm}{\centering\tiny \vspace{0.05cm} "Cat fur"}}
        \end{minipage}	
        \hspace{0.02in}
         \begin{minipage}{3.3in}	
            \centering
            \begin{minipage}{0.13\linewidth}
            \includegraphics[width=\linewidth]{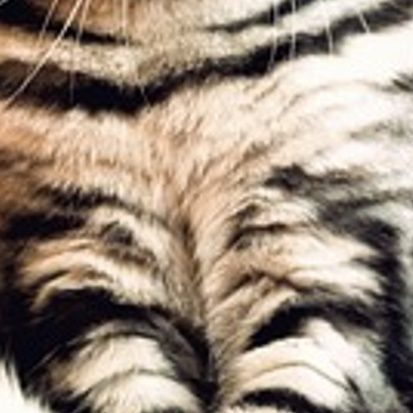}
            \end{minipage}	
            \begin{minipage}{0.13\linewidth}
            \includegraphics[width=\linewidth]{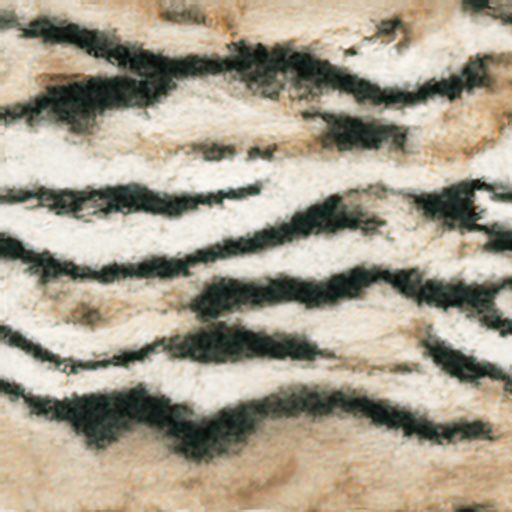}
            \end{minipage}	
            \begin{minipage}{0.13\linewidth}
            \includegraphics[width=\linewidth]{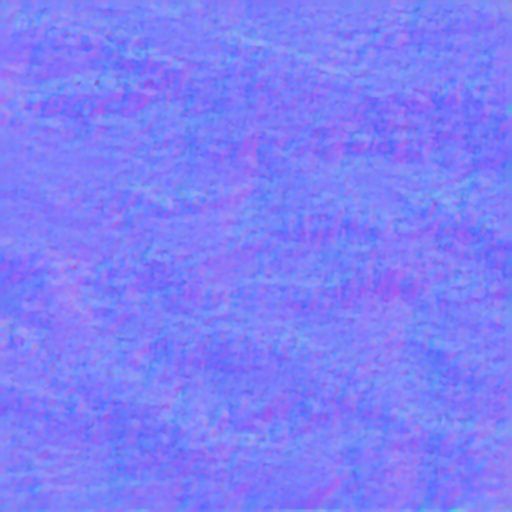}
            \end{minipage}	
            \begin{minipage}{0.13\linewidth}
            \includegraphics[width=\linewidth]{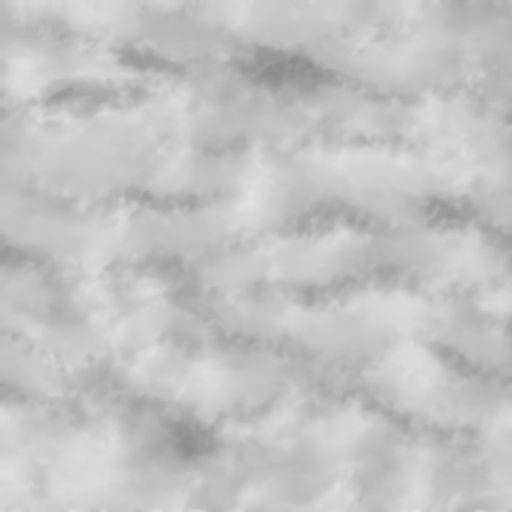}
            \end{minipage}	
            \begin{minipage}{0.13\linewidth}
            \includegraphics[width=\linewidth]{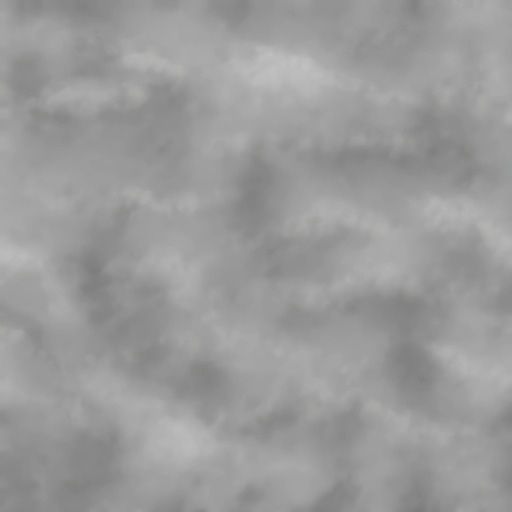}
            \end{minipage}	
            \begin{minipage}{0.13\linewidth}
            \includegraphics[width=\linewidth]{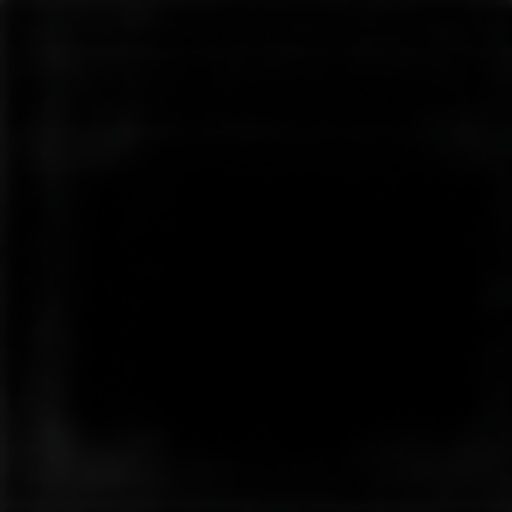}
            \end{minipage}	
            \begin{minipage}{0.13\linewidth}
            \includegraphics[width=\linewidth]{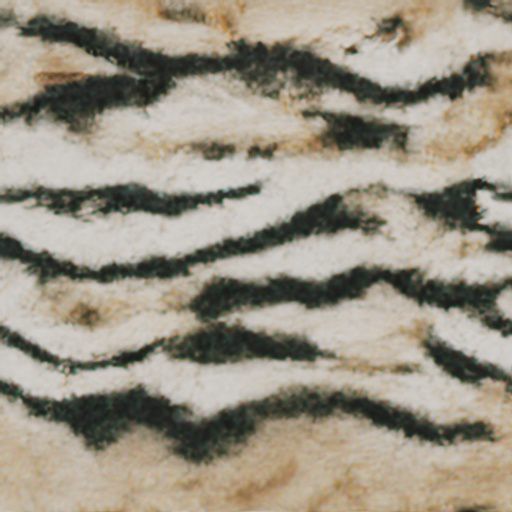}
            \end{minipage}	
        \end{minipage}	
    \end{minipage}	

    \begin{minipage}{3.4in}
        \begin{minipage}{0.02in}	
            \centering
                \rotatebox{90}{\parbox{1cm}{\centering\tiny \vspace{0.05cm} "Strawberry"}}
        \end{minipage}	
        \hspace{0.02in}
         \begin{minipage}{3.3in}	
            \centering
            \begin{minipage}{0.13\linewidth}
            \includegraphics[width=\linewidth]{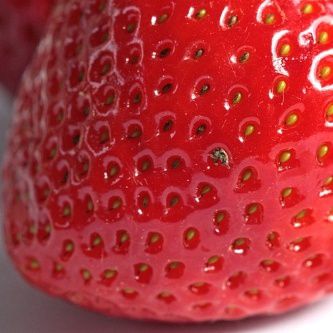}
            \end{minipage}	
            \begin{minipage}{0.13\linewidth}
            \includegraphics[width=\linewidth]{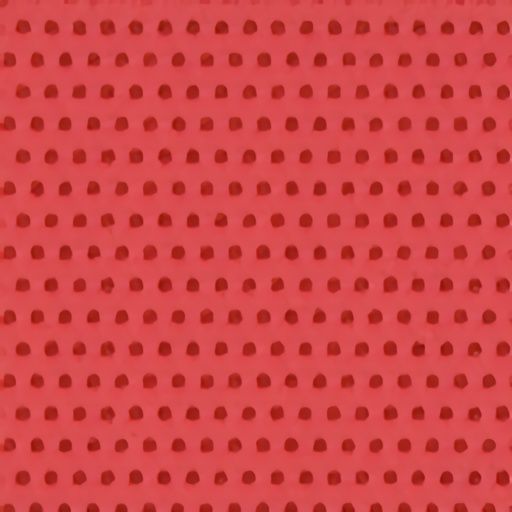}
            \end{minipage}	
            \begin{minipage}{0.13\linewidth}
            \includegraphics[width=\linewidth]{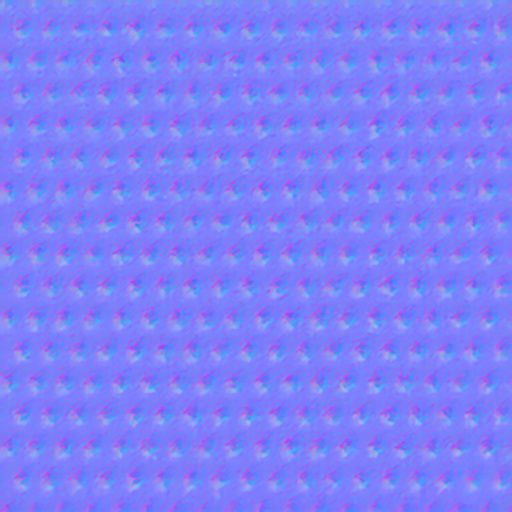}
            \end{minipage}	
            \begin{minipage}{0.13\linewidth}
            \includegraphics[width=\linewidth]{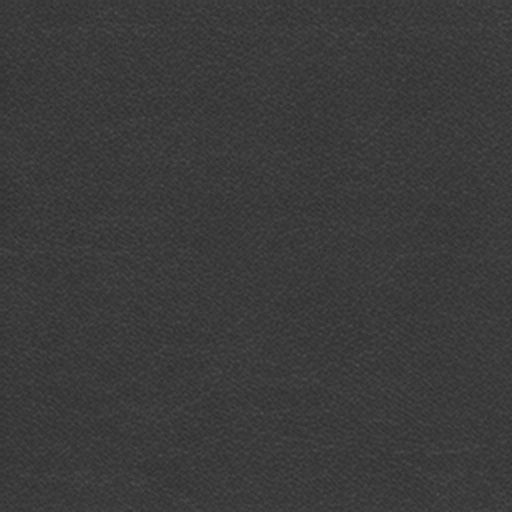}
            \end{minipage}	
            \begin{minipage}{0.13\linewidth}
            \includegraphics[width=\linewidth]{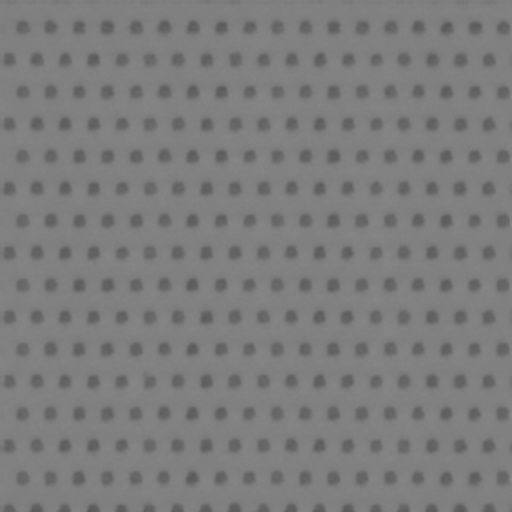}
            \end{minipage}	
            \begin{minipage}{0.13\linewidth}
            \includegraphics[width=\linewidth]{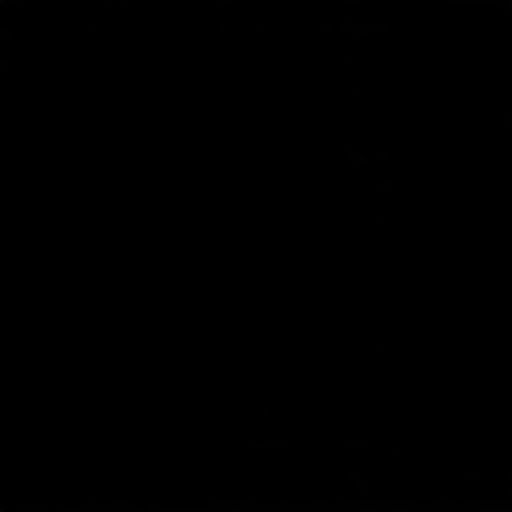}
            \end{minipage}	
            \begin{minipage}{0.13\linewidth}
            \includegraphics[width=\linewidth]{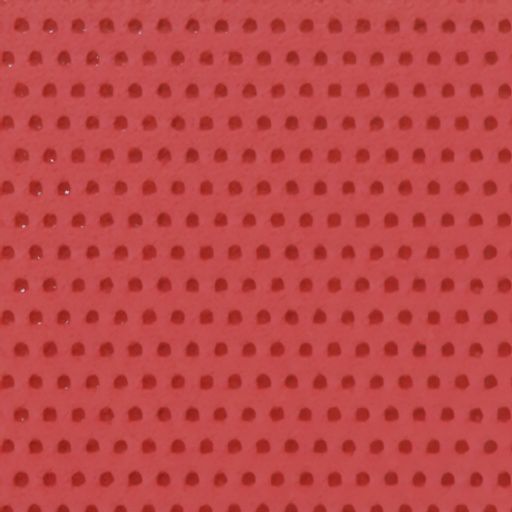}
            \end{minipage}	
        \end{minipage}	
    \end{minipage}	

    \begin{minipage}{3.4in}
        \begin{minipage}{0.02in}	
            \centering
                \rotatebox{90}{\parbox{1cm}{\centering\tiny \vspace{0.05cm} "Leaves"}}
        \end{minipage}	
        \hspace{0.02in}
         \begin{minipage}{3.3in}	
            \centering
            \begin{minipage}{0.13\linewidth}
            \includegraphics[width=\linewidth]{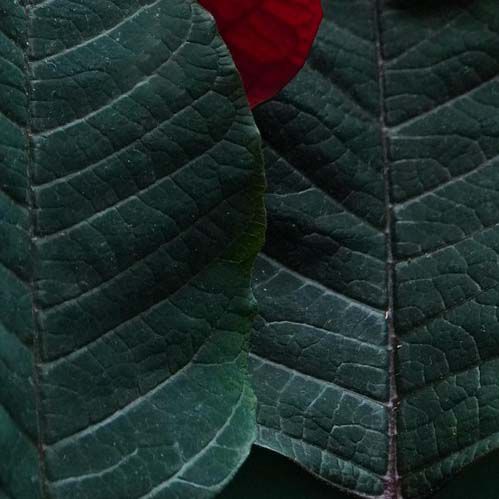}
            \end{minipage}	
            \begin{minipage}{0.13\linewidth}
            \includegraphics[width=\linewidth]{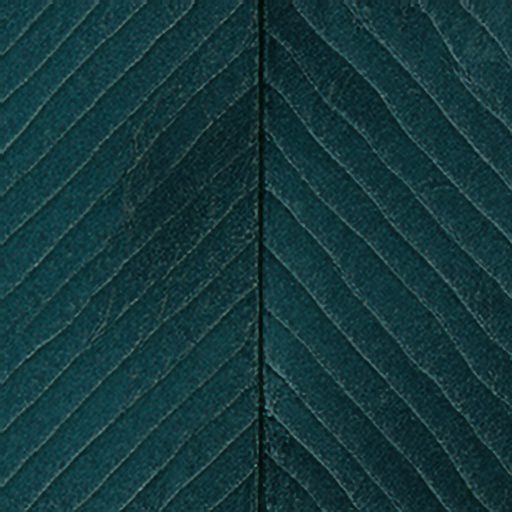}
            \end{minipage}	
            \begin{minipage}{0.13\linewidth}
            \includegraphics[width=\linewidth]{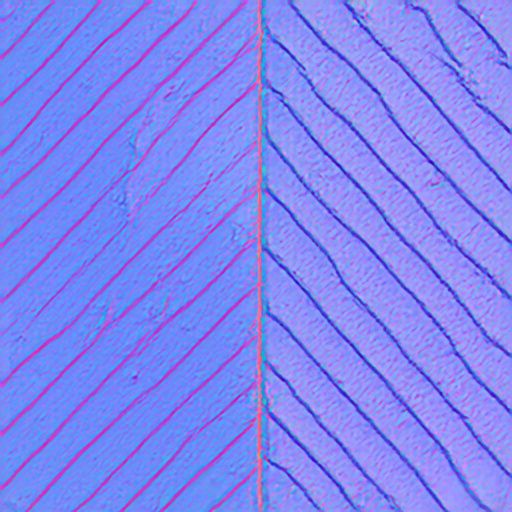}
            \end{minipage}	
            \begin{minipage}{0.13\linewidth}
            \includegraphics[width=\linewidth]{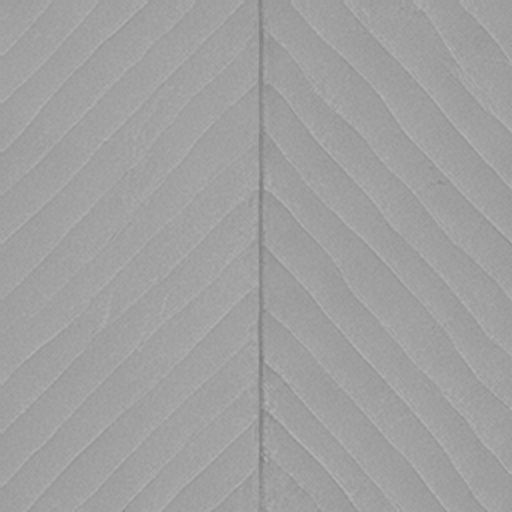}
            \end{minipage}	
            \begin{minipage}{0.13\linewidth}
            \includegraphics[width=\linewidth]{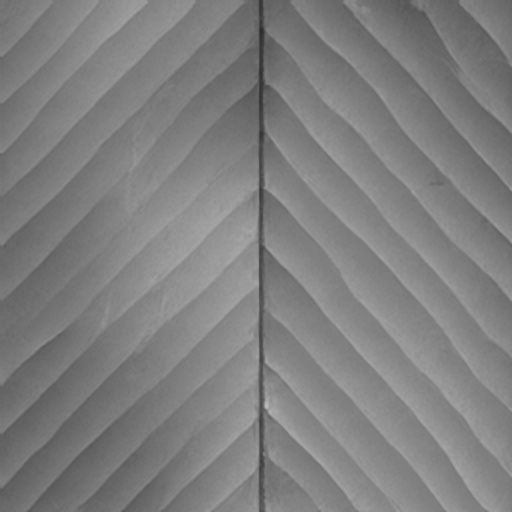}
            \end{minipage}	
            \begin{minipage}{0.13\linewidth}
            \includegraphics[width=\linewidth]{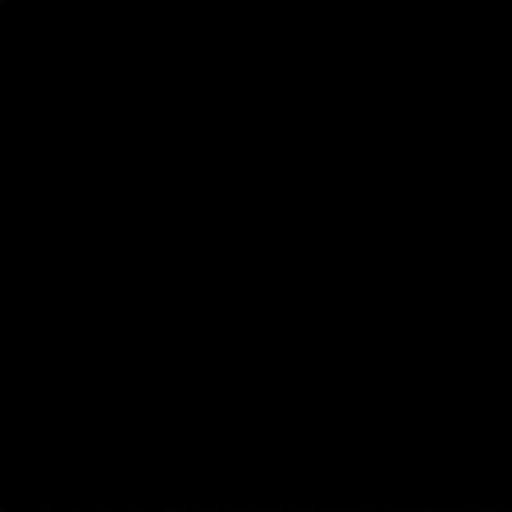}
            \end{minipage}	
            \begin{minipage}{0.13\linewidth}
            \includegraphics[width=\linewidth]{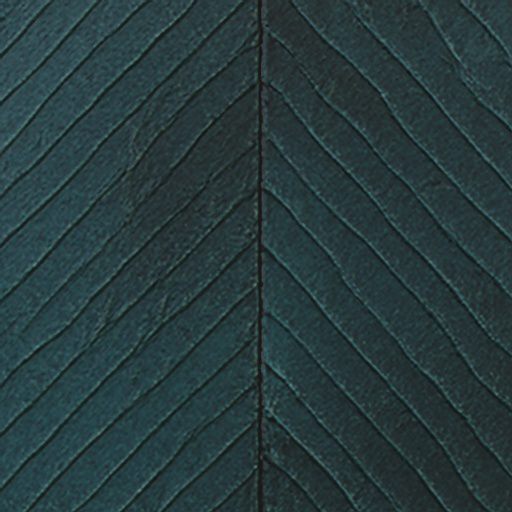}
            \end{minipage}	
        \end{minipage}	
    \end{minipage}	

    \begin{minipage}{3.4in}
        \begin{minipage}{0.02in}	
            \centering
                \rotatebox{90}{\parbox{1cm}{\centering\tiny "Moon\vspace{-0.05cm}\\surface"}}
        \end{minipage}	
        \hspace{0.02in}
         \begin{minipage}{3.3in}	
            \centering
            \begin{minipage}{0.13\linewidth}
            \includegraphics[width=\linewidth]{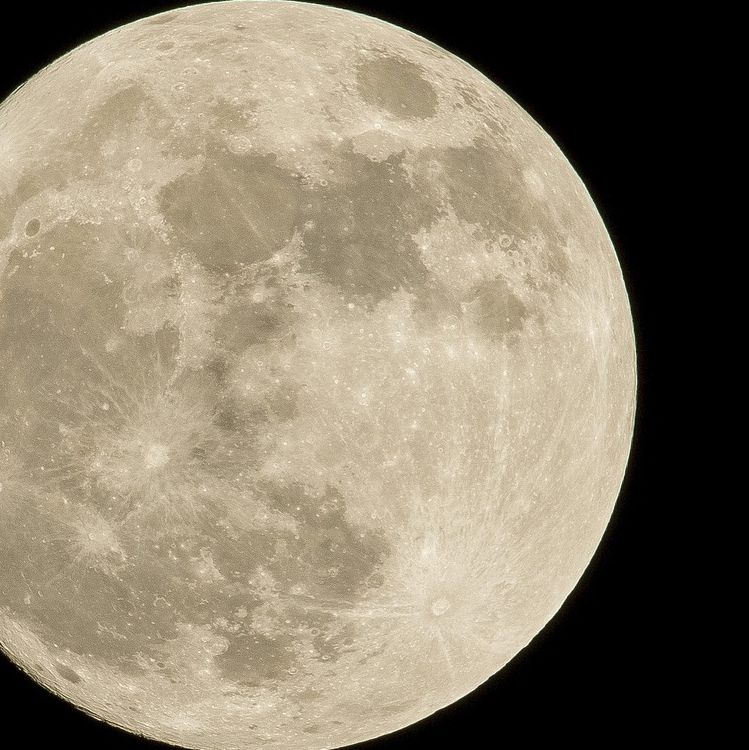}
            \end{minipage}	
            \begin{minipage}{0.13\linewidth}
            \includegraphics[width=\linewidth]{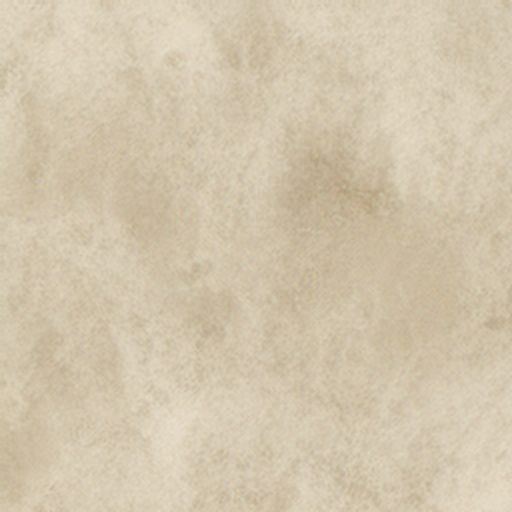}
            \end{minipage}	
            \begin{minipage}{0.13\linewidth}
            \includegraphics[width=\linewidth]{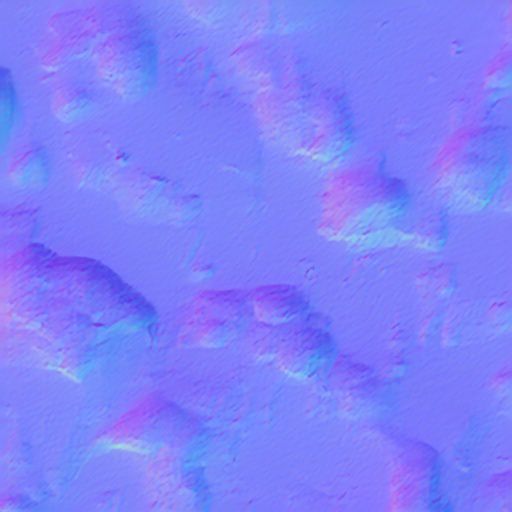}
            \end{minipage}	
            \begin{minipage}{0.13\linewidth}
            \includegraphics[width=\linewidth]{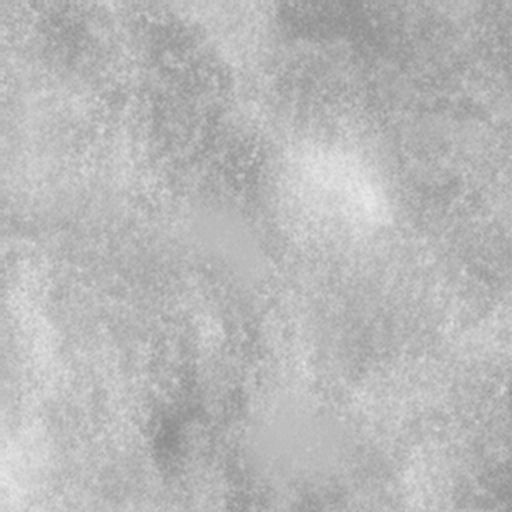}
            \end{minipage}	
            \begin{minipage}{0.13\linewidth}
            \includegraphics[width=\linewidth]{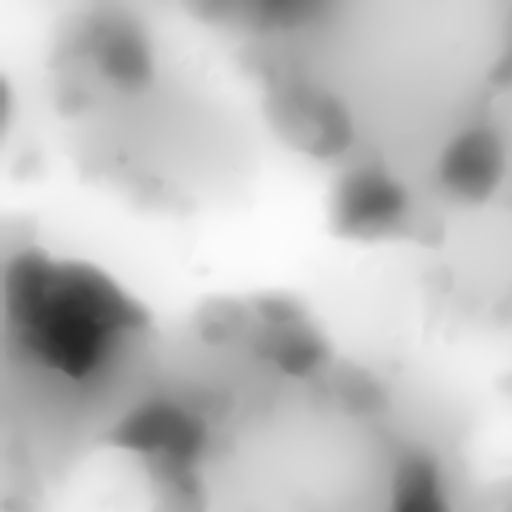}
            \end{minipage}	
            \begin{minipage}{0.13\linewidth}
            \includegraphics[width=\linewidth]{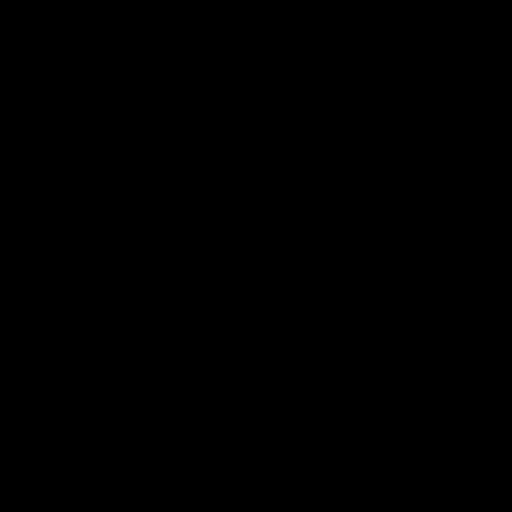}
            \end{minipage}	
            \begin{minipage}{0.13\linewidth}
            \includegraphics[width=\linewidth]{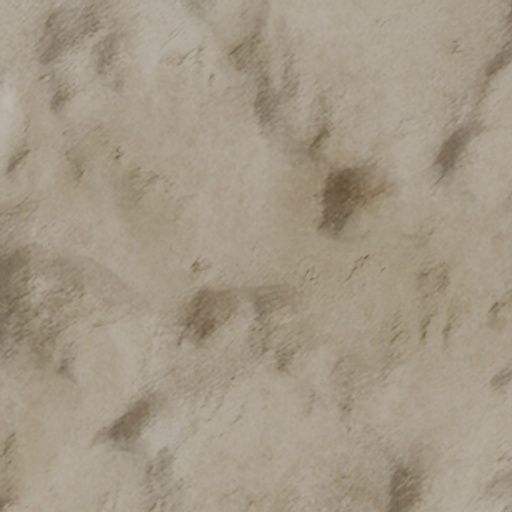}
            \end{minipage}	
        \end{minipage}	
    \end{minipage}	

    \begin{minipage}{3.4in}
        \begin{minipage}{0.02in}	
            \centering
                \rotatebox{90}{\parbox{1cm}{\centering\tiny "Manhole\vspace{-0.05cm}\\cover"}}
        \end{minipage}	
        \hspace{0.02in}
         \begin{minipage}{3.3in}	
            \centering
            \begin{minipage}{0.13\linewidth}
            \includegraphics[width=\linewidth]{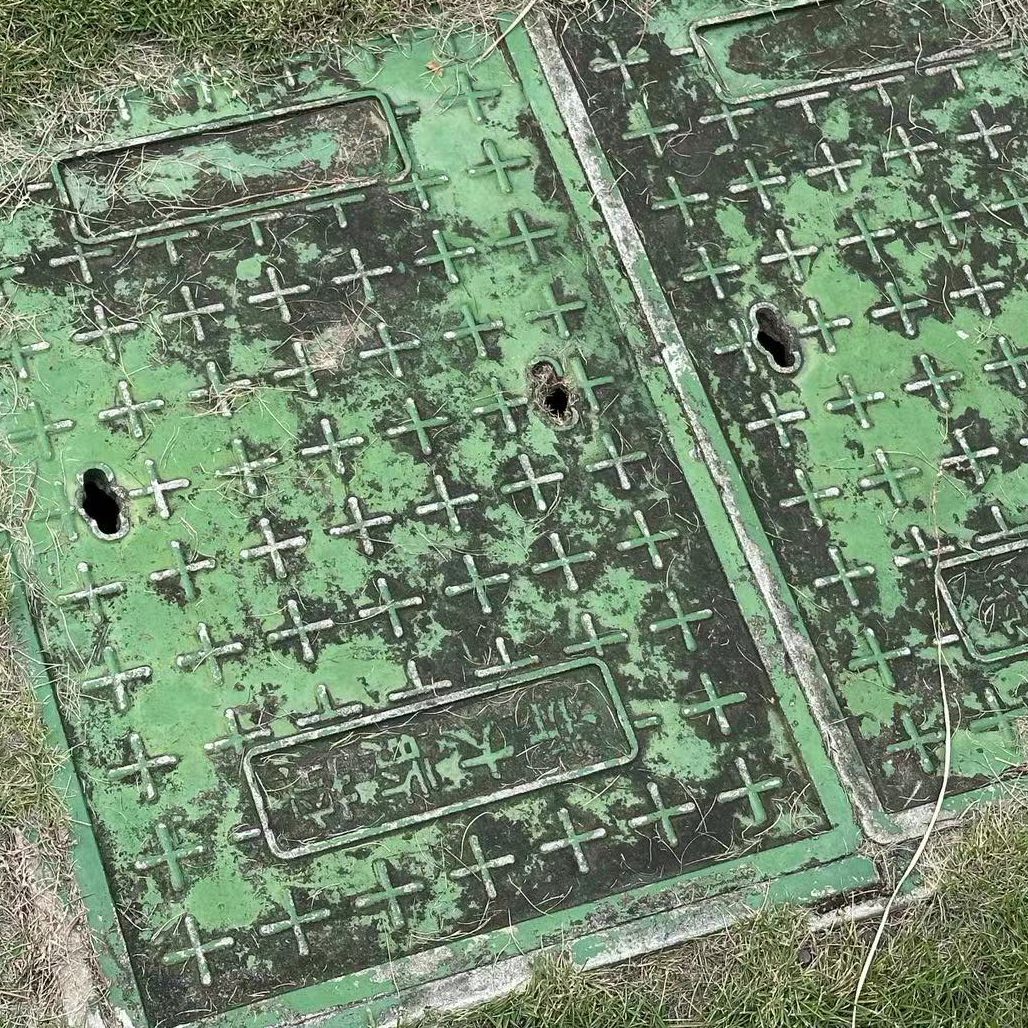}
            \end{minipage}	
            \begin{minipage}{0.13\linewidth}
            \includegraphics[width=\linewidth]{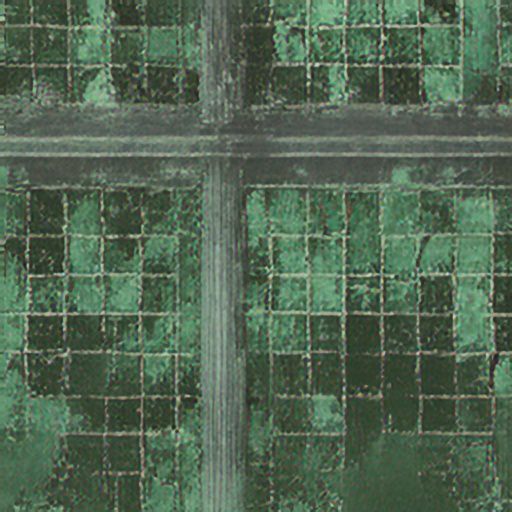}
            \end{minipage}	
            \begin{minipage}{0.13\linewidth}
            \includegraphics[width=\linewidth]{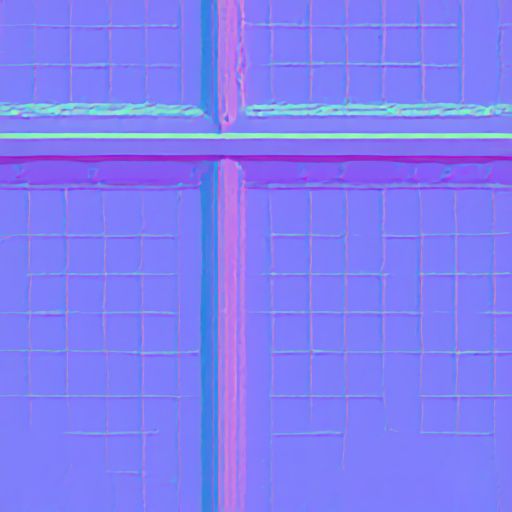}
            \end{minipage}	
            \begin{minipage}{0.13\linewidth}
            \includegraphics[width=\linewidth]{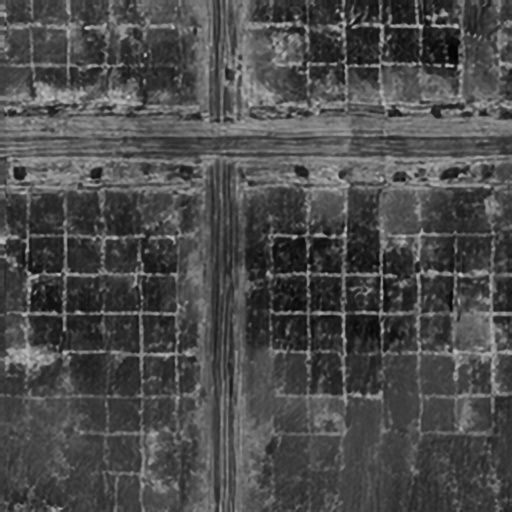}
            \end{minipage}	
            \begin{minipage}{0.13\linewidth}
            \includegraphics[width=\linewidth]{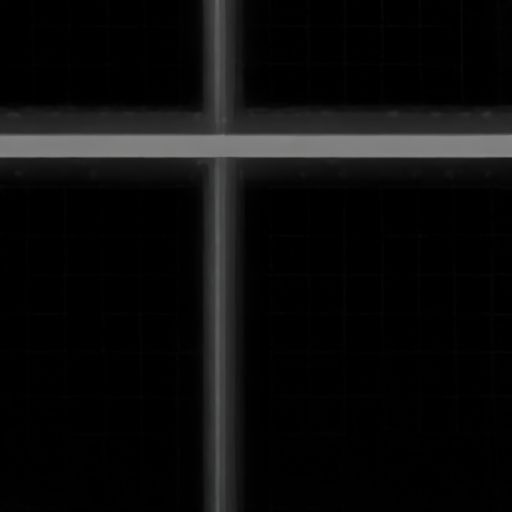}
            \end{minipage}	
            \begin{minipage}{0.13\linewidth}
            \includegraphics[width=\linewidth]{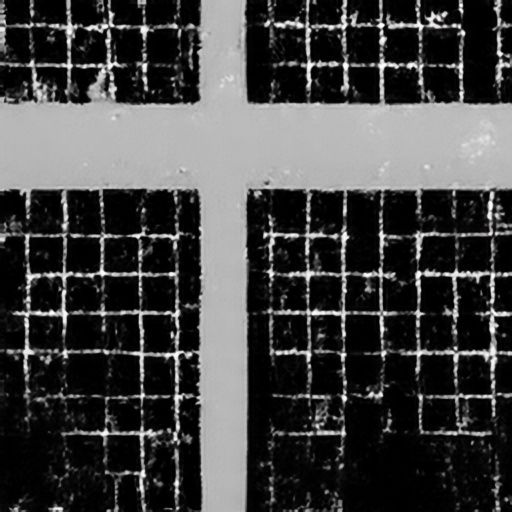}
            \end{minipage}	
            \begin{minipage}{0.13\linewidth}
            \includegraphics[width=\linewidth]{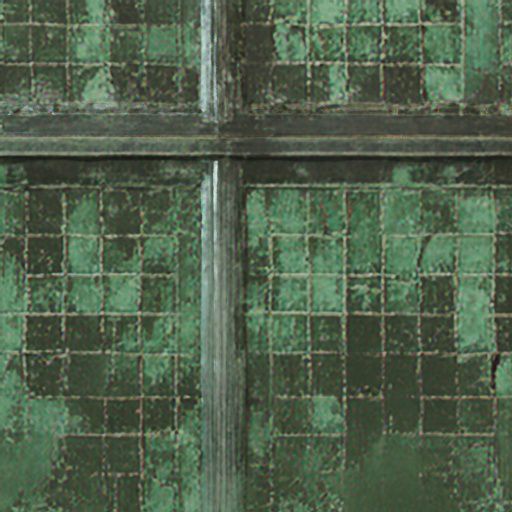}
            \end{minipage}	
        \end{minipage}	
    \end{minipage}

   \caption{Evaluation of the generalization ability of our model to complex patterns. We test on several royalty-free images from Pixabay, including animal and plant appearances as well as abstract textures like the moon surface. The results demonstrate that our model generalizes well beyond typical indoor and outdoor scenes, producing visually plausible outputs for a diverse range of complex surface patterns.}
   \label{fig:complex_pattern}
\end{figure}

To further demonstrate the generalization ability of our model beyond common indoor and outdoor scenes, we test several images sourced from Pixabay\footnote{\url{https://pixabay.com/}}~(royalty-free). The results are shown in Fig.~\ref{fig:complex_pattern}. Notably, for challenging appearances such as animal skin and fur, and plant surfaces, our model can generate visually plausible results.

\subsubsection{Tileable Generation.}\label{sec:tileable}

\begin{figure}[htbp!]
    \centering		
    \begin{minipage}{3.4in}
        \begin{minipage}{0.01in}	
            \centering
                % \vspace{0.1in}
                \rotatebox{90}{\parbox{1cm}{\centering\tiny {}}}
        \end{minipage}	
        \hspace{0.02in}
         \begin{minipage}{3.3in}	
            \centering
            \begin{minipage}{0.13\linewidth}
                \subcaption*{\tiny Input}
            \end{minipage}
            \begin{minipage}{0.26\linewidth}
                \subcaption*{\tiny 512$\times$512}
            \end{minipage}
            \begin{minipage}{0.26\linewidth}
                \subcaption*{\tiny 1024$\times$1024}
            \end{minipage}
            \begin{minipage}{0.26\linewidth}
                \subcaption*{\tiny Render}
            \end{minipage}
        \end{minipage}	
    \end{minipage}	

    \begin{minipage}{3.4in}
        \begin{minipage}{0.02in}	
            \centering
                \vspace{0.1in}
                \rotatebox{90}{\parbox{1cm}{\centering\tiny "checkered \vspace{-0.05cm}\\fabric"}}
        \end{minipage}	
        \hspace{0.02in}
             \begin{minipage}{3.3in}	
                \centering
                \begin{minipage}{0.13\linewidth}
                    % \subcaption*{\tiny Input}
                    \includegraphics[width=\linewidth]{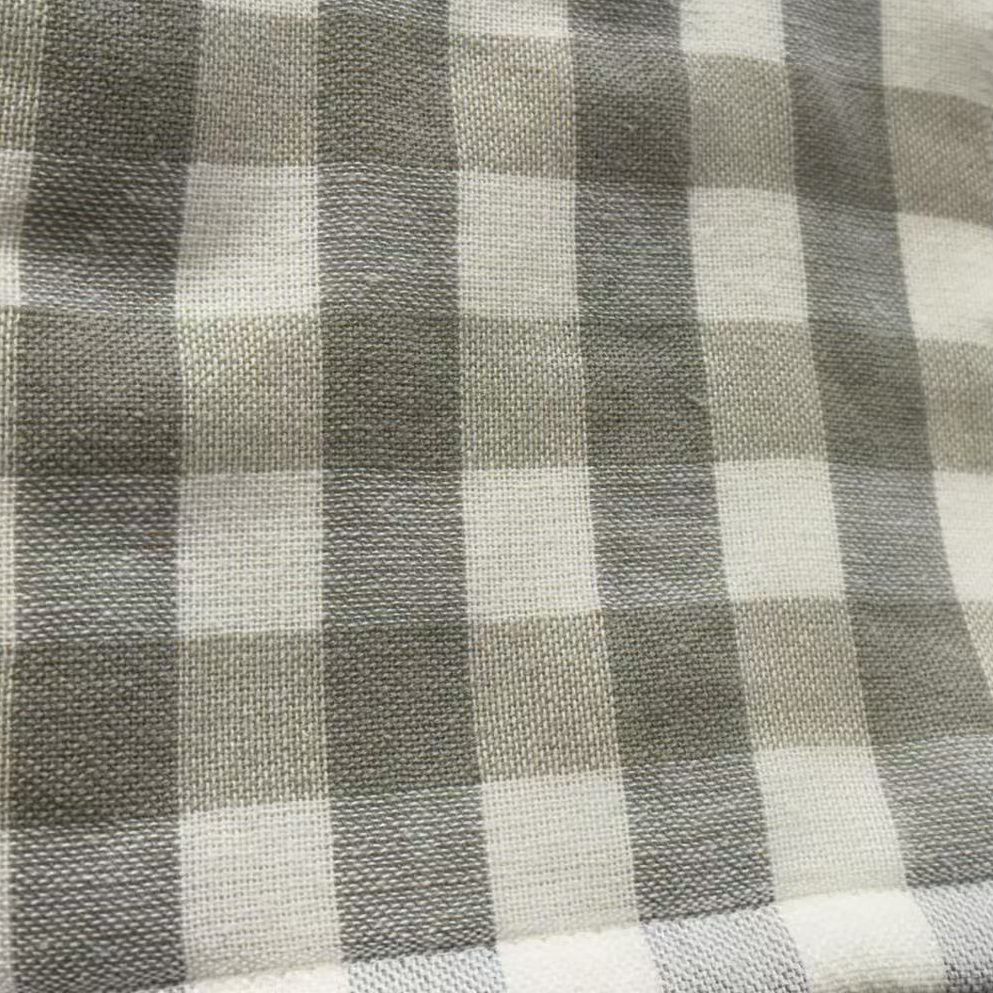}
                \end{minipage}
                \begin{minipage}{0.26\linewidth}
                    % \subcaption*{\tiny Albedo}
                    \includegraphics[width=\linewidth]{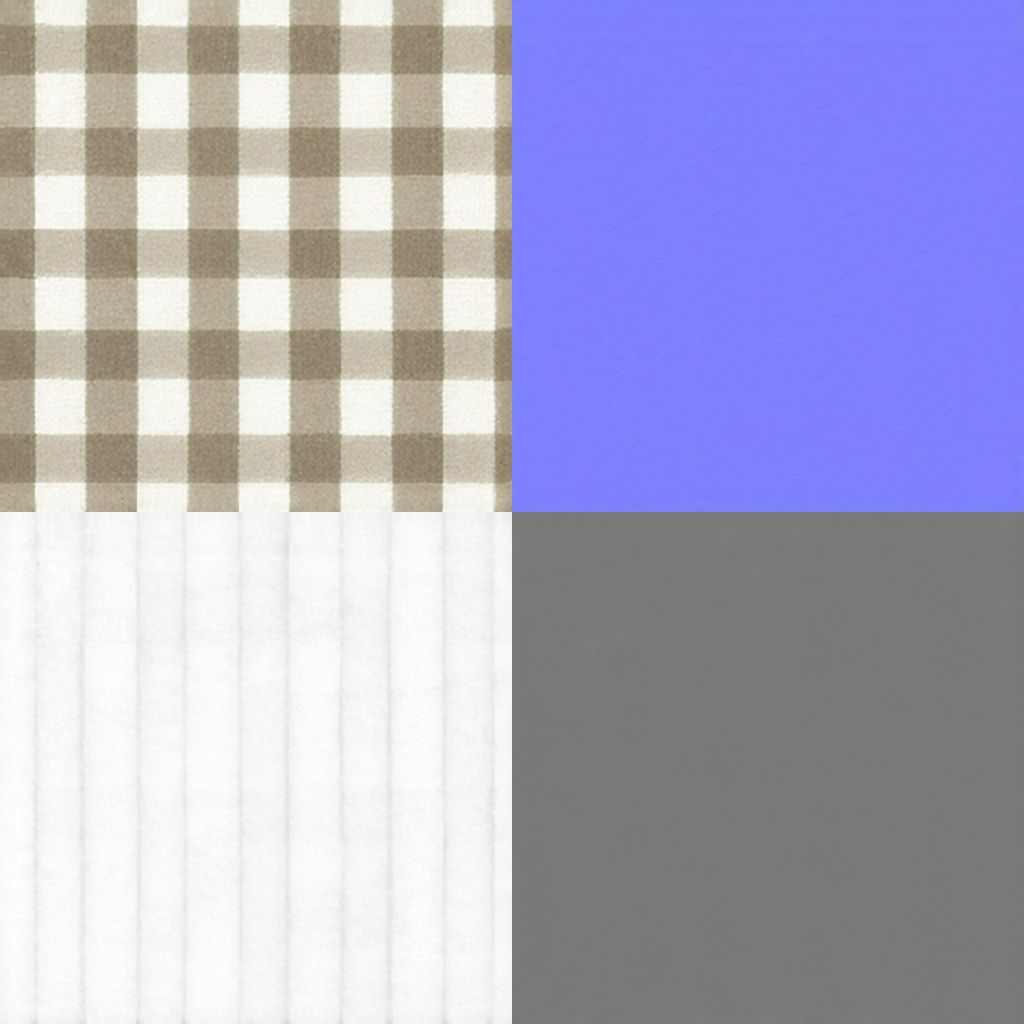}
                \end{minipage}
                \begin{minipage}{0.26\linewidth}
                    % \subcaption*{\tiny Normal}
                    \includegraphics[width=\linewidth]{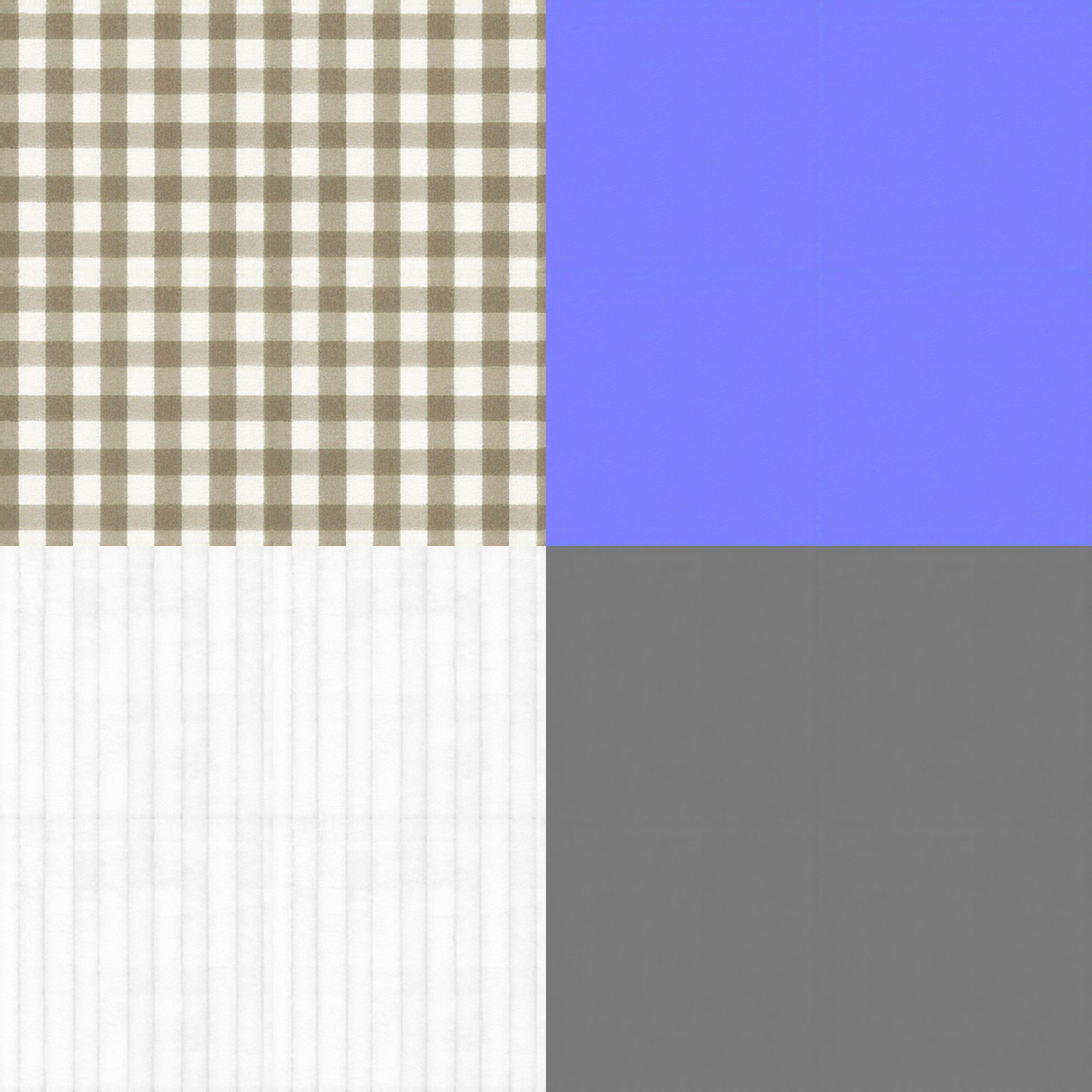}
                \end{minipage}
                \begin{minipage}{0.26\linewidth}
                    % \subcaption*{\tiny Roughness}
                    \includegraphics[width=\linewidth]{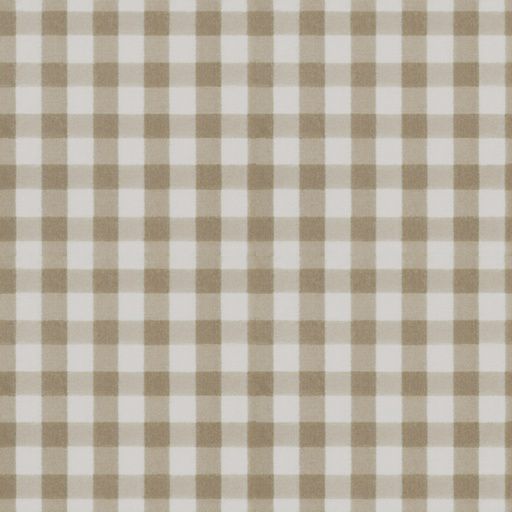}
                \end{minipage}
            \end{minipage}	
        \end{minipage}	

    \begin{minipage}{3.4in}
        \begin{minipage}{0.02in}	
            \centering
                \rotatebox{90}{\parbox{1cm}{\centering\tiny \vspace{0.05cm} "Wood"}}
        \end{minipage}	
        \hspace{0.02in}
         \begin{minipage}{3.3in}	
            \centering
            \begin{minipage}{0.13\linewidth}
            \includegraphics[width=\linewidth]{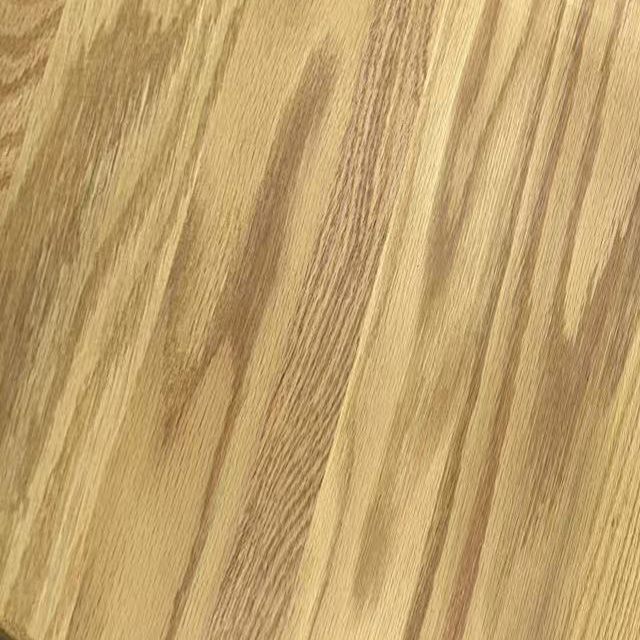}
            \end{minipage}	
            \begin{minipage}{0.26\linewidth}
            \includegraphics[width=\linewidth]{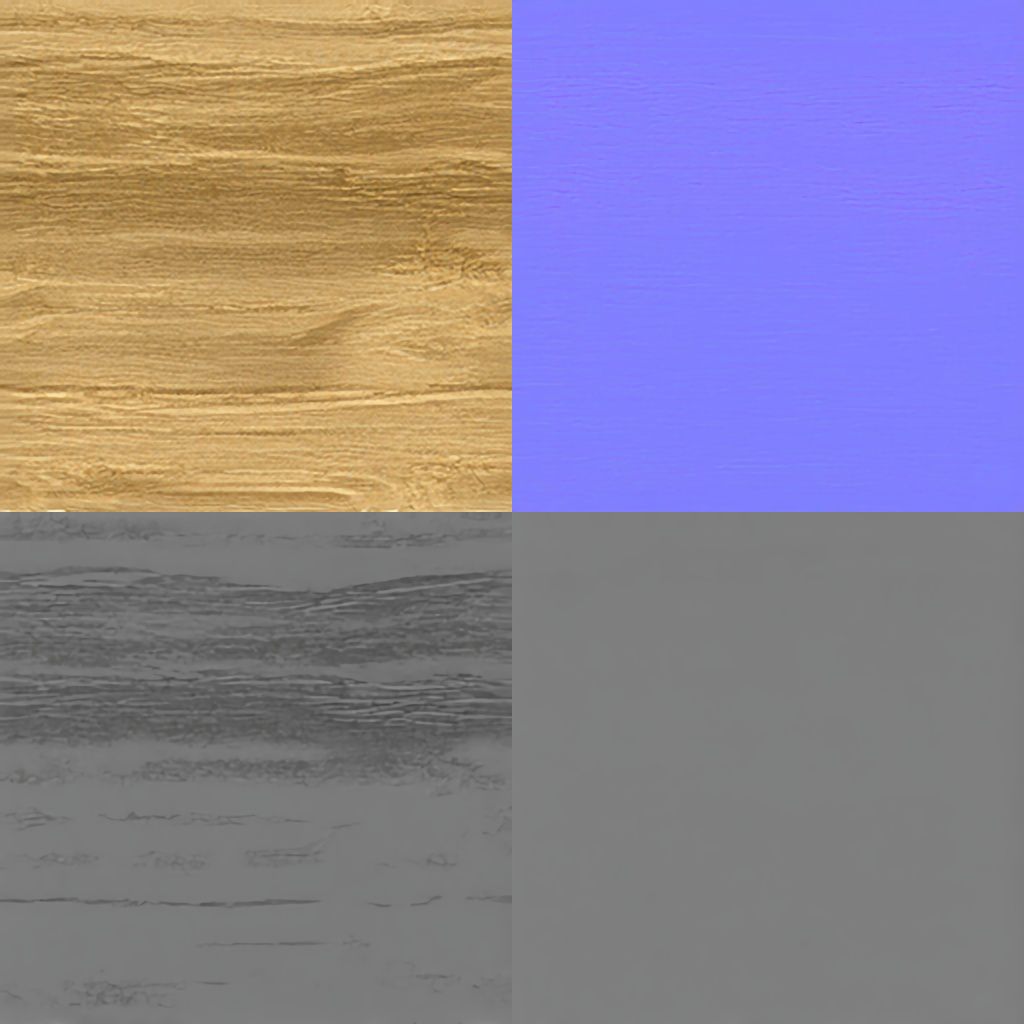}
            \end{minipage}	
            \begin{minipage}{0.26\linewidth}
            \includegraphics[width=\linewidth]{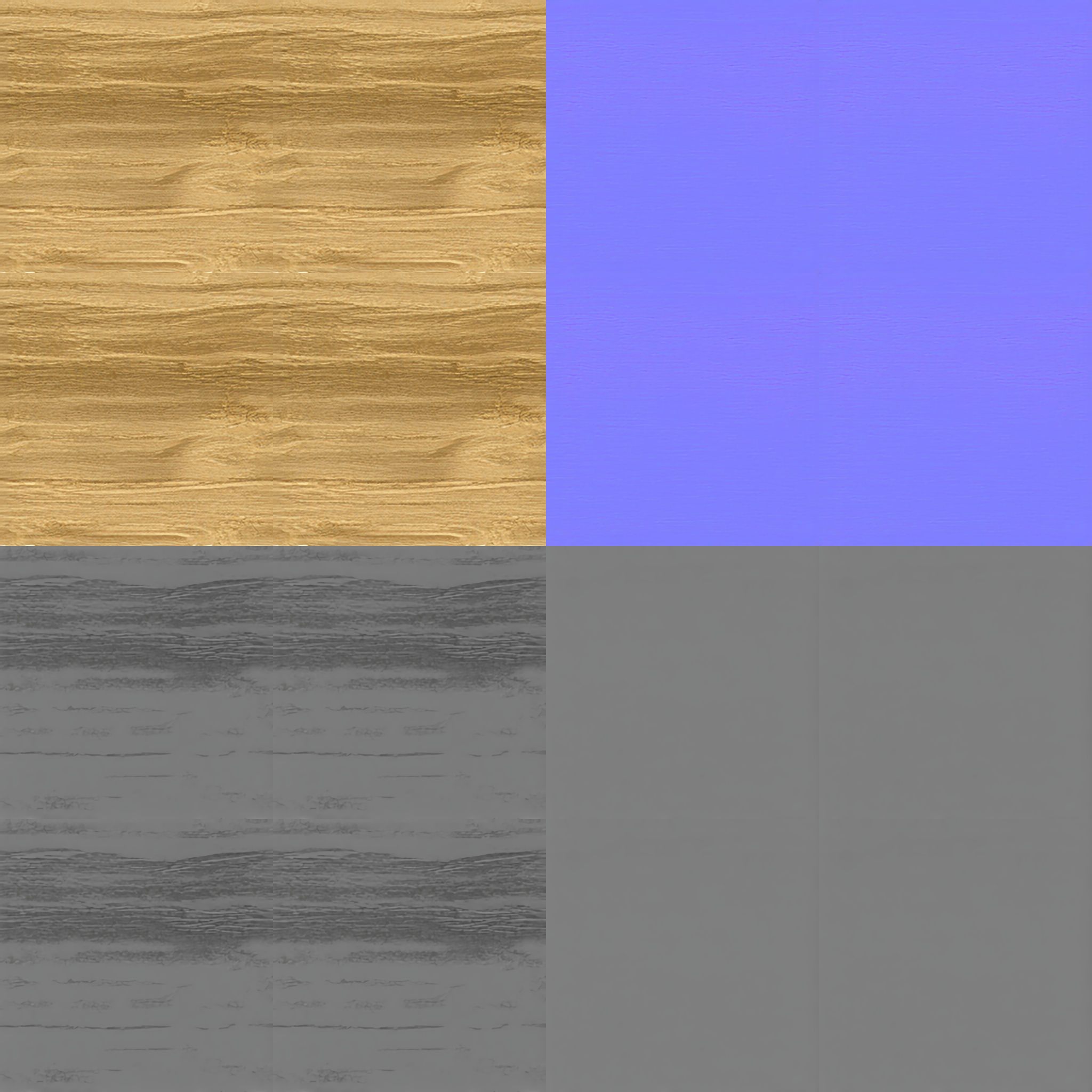}
            \end{minipage}	
            \begin{minipage}{0.26\linewidth}
            \includegraphics[width=\linewidth]{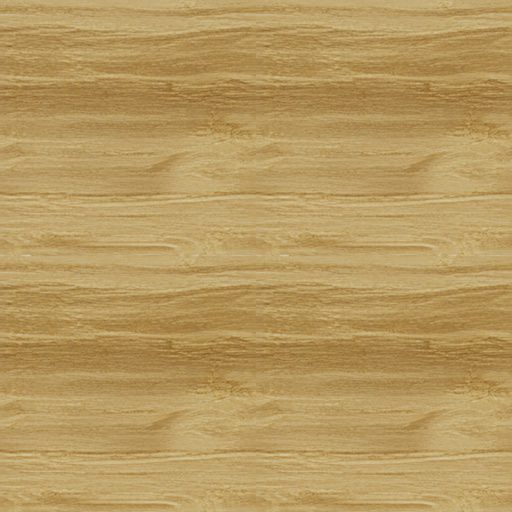}
            \end{minipage}	
        \end{minipage}	
    \end{minipage}	

    \begin{minipage}{3.4in}
        \begin{minipage}{0.02in}	
            \centering
                \rotatebox{90}{\parbox{1cm}{\centering\tiny "Terracotta \vspace{-0.05cm}\\brick wall"}}
        \end{minipage}
        \hspace{0.02in}
         \begin{minipage}{3.3in}	
            \centering
            \begin{minipage}{0.13\linewidth}
            \includegraphics[width=\linewidth]{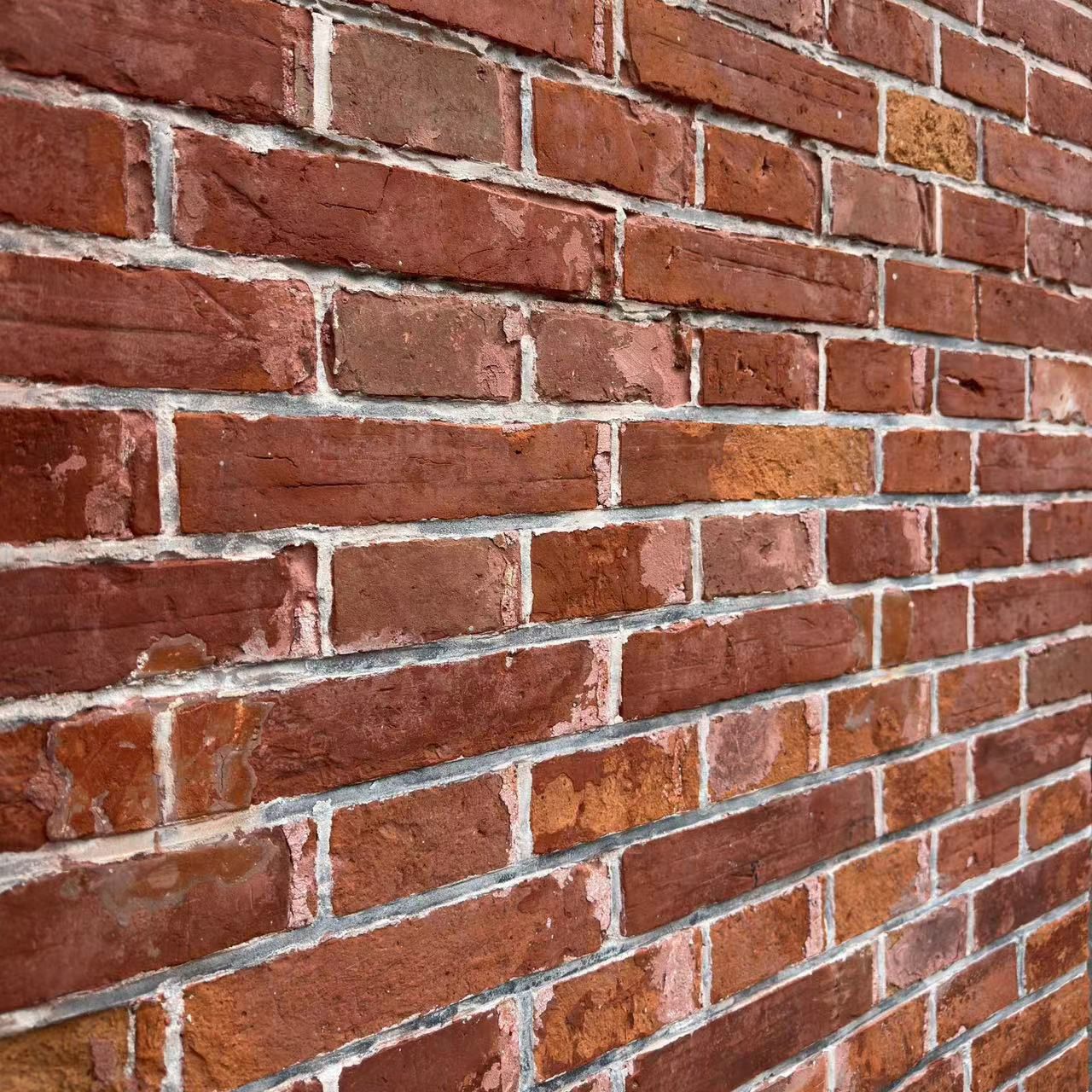}
            \end{minipage}	
            \begin{minipage}{0.26\linewidth}
            \includegraphics[width=\linewidth]{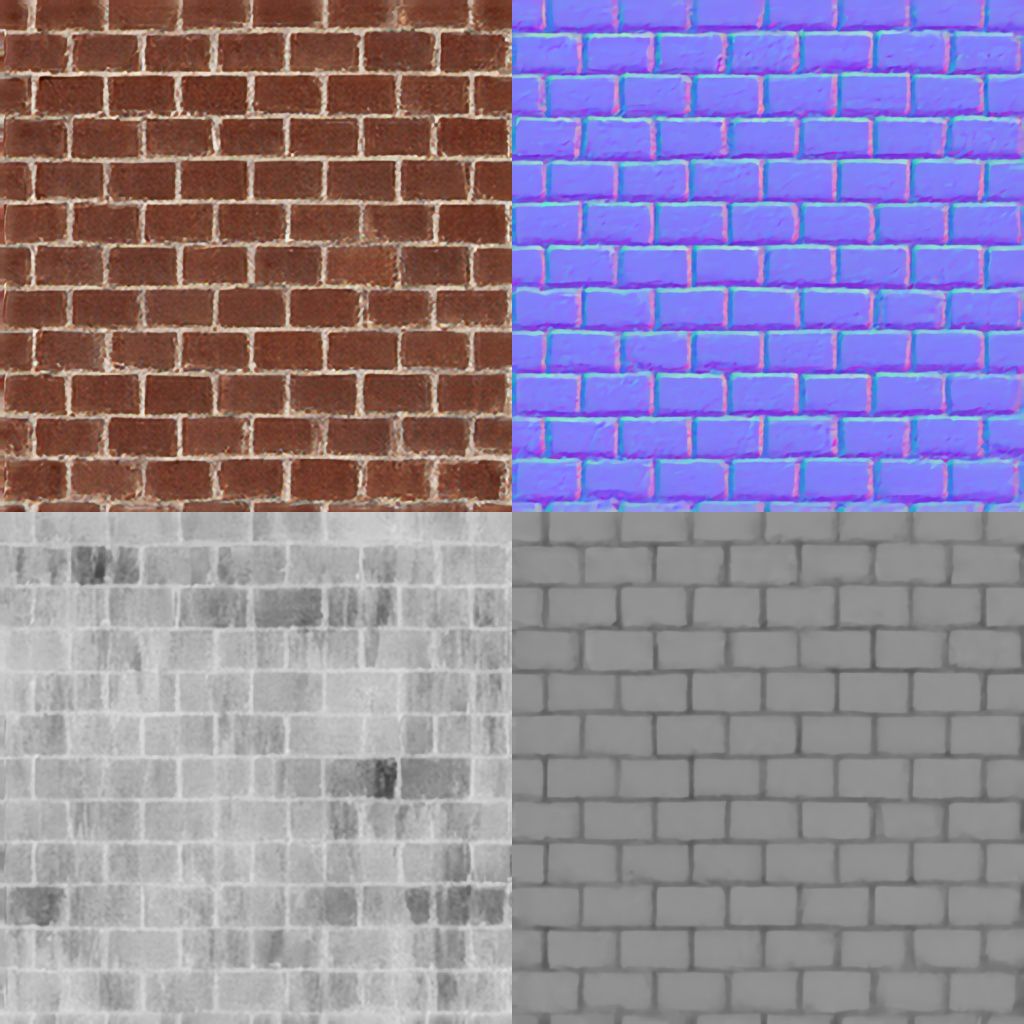}
            \end{minipage}	
            \begin{minipage}{0.26\linewidth}
            \includegraphics[width=\linewidth]{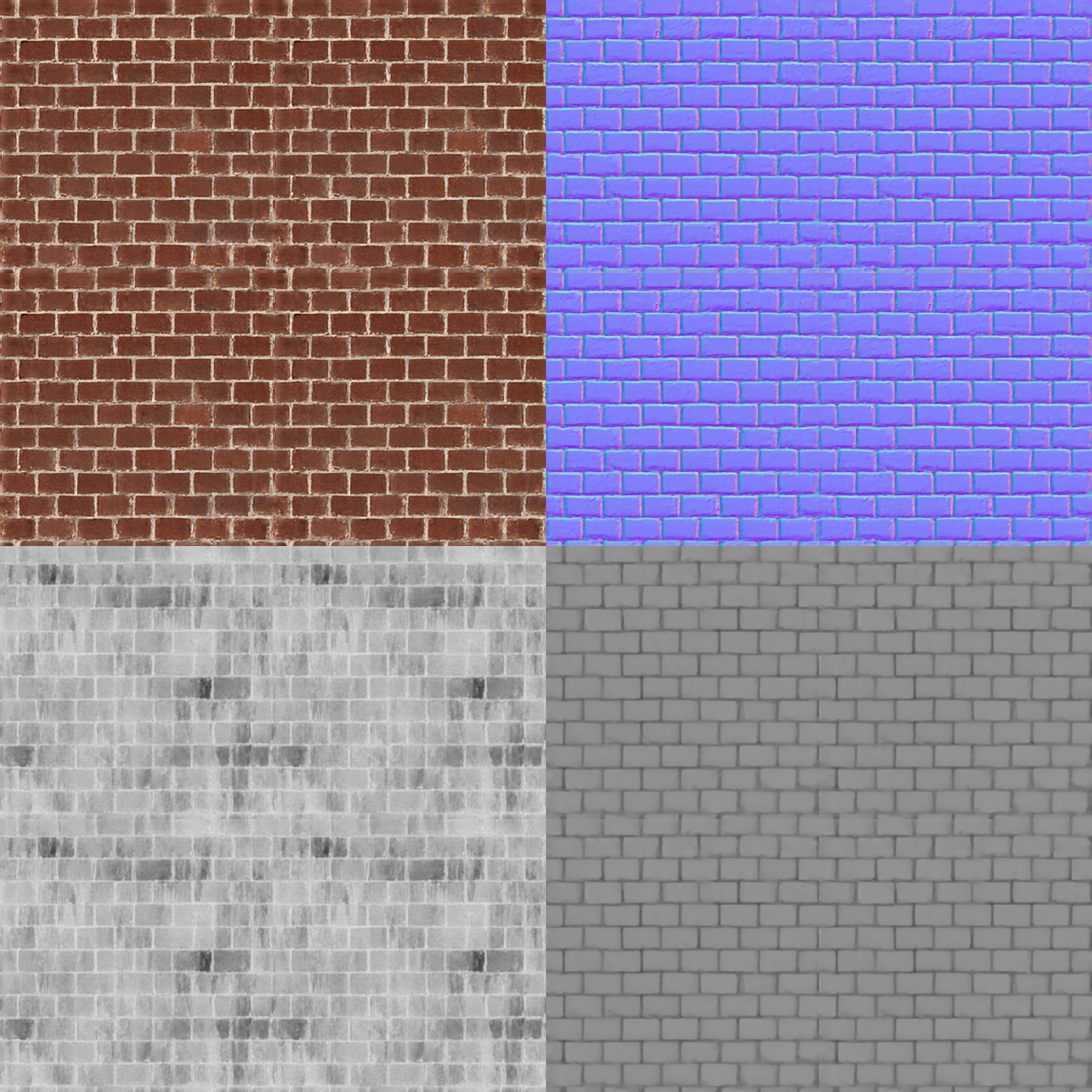}
            \end{minipage}	
            \begin{minipage}{0.26\linewidth}
            \includegraphics[width=\linewidth]{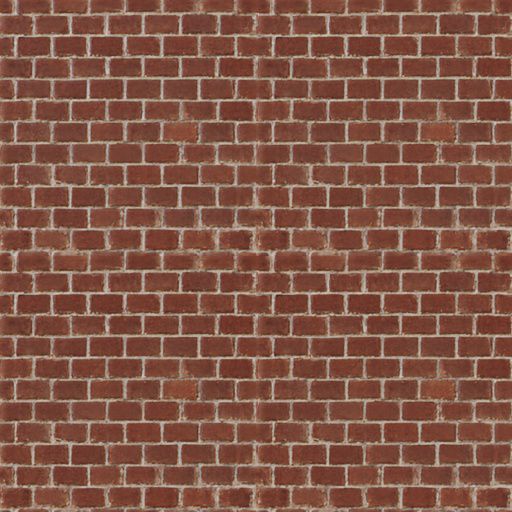}
            \end{minipage}	
        \end{minipage}	
    \end{minipage}	

    \begin{minipage}{3.4in}
        \begin{minipage}{0.02in}	
            \centering
                \rotatebox{90}{\parbox{1cm}{\centering\tiny \vspace{0.05cm} "Stone path"}}
        \end{minipage}	
        \hspace{0.02in}
         \begin{minipage}{3.3in}	
            \centering
            \begin{minipage}{0.13\linewidth}
            \includegraphics[width=\linewidth]{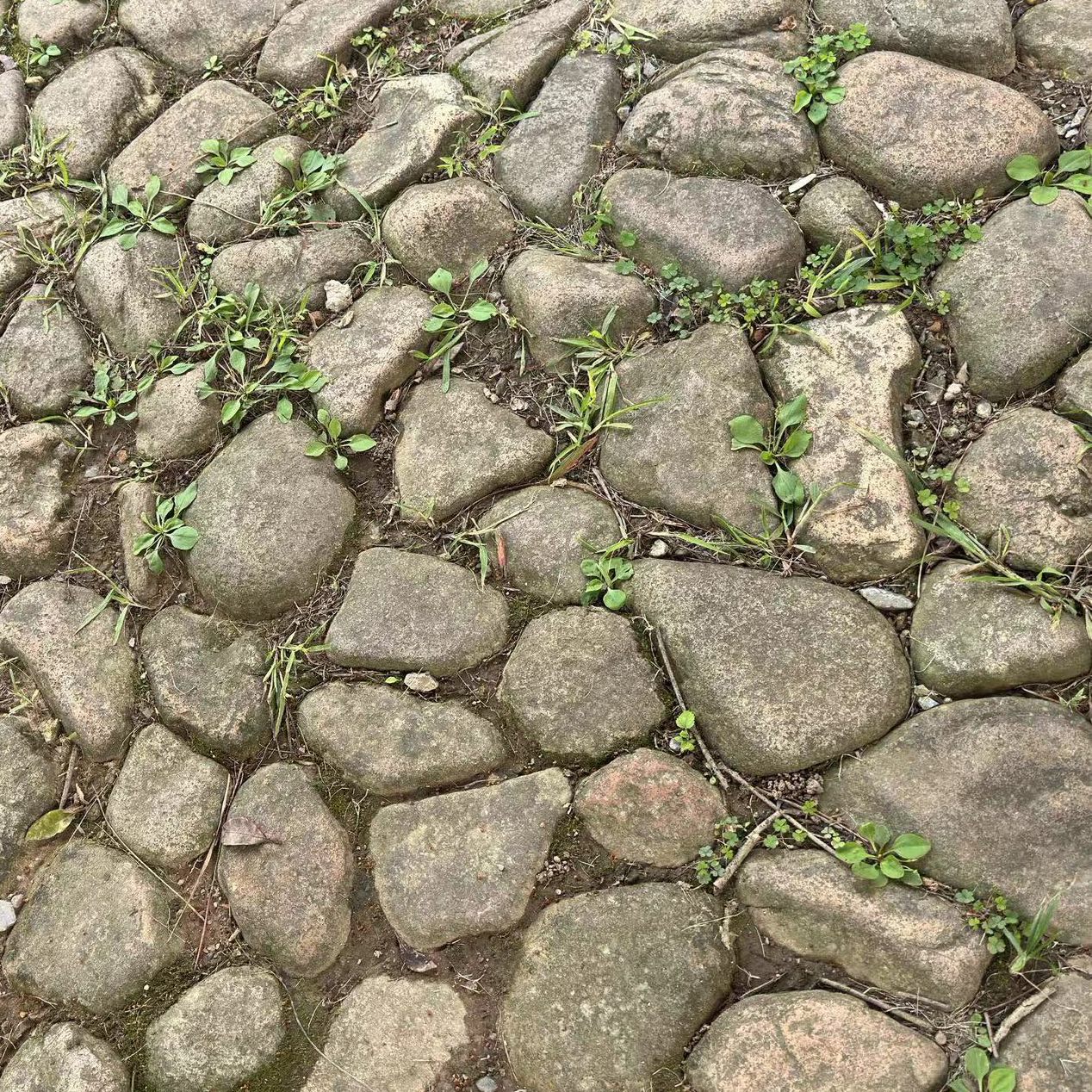}
            \end{minipage}	
            \begin{minipage}{0.26\linewidth}
            \includegraphics[width=\linewidth]{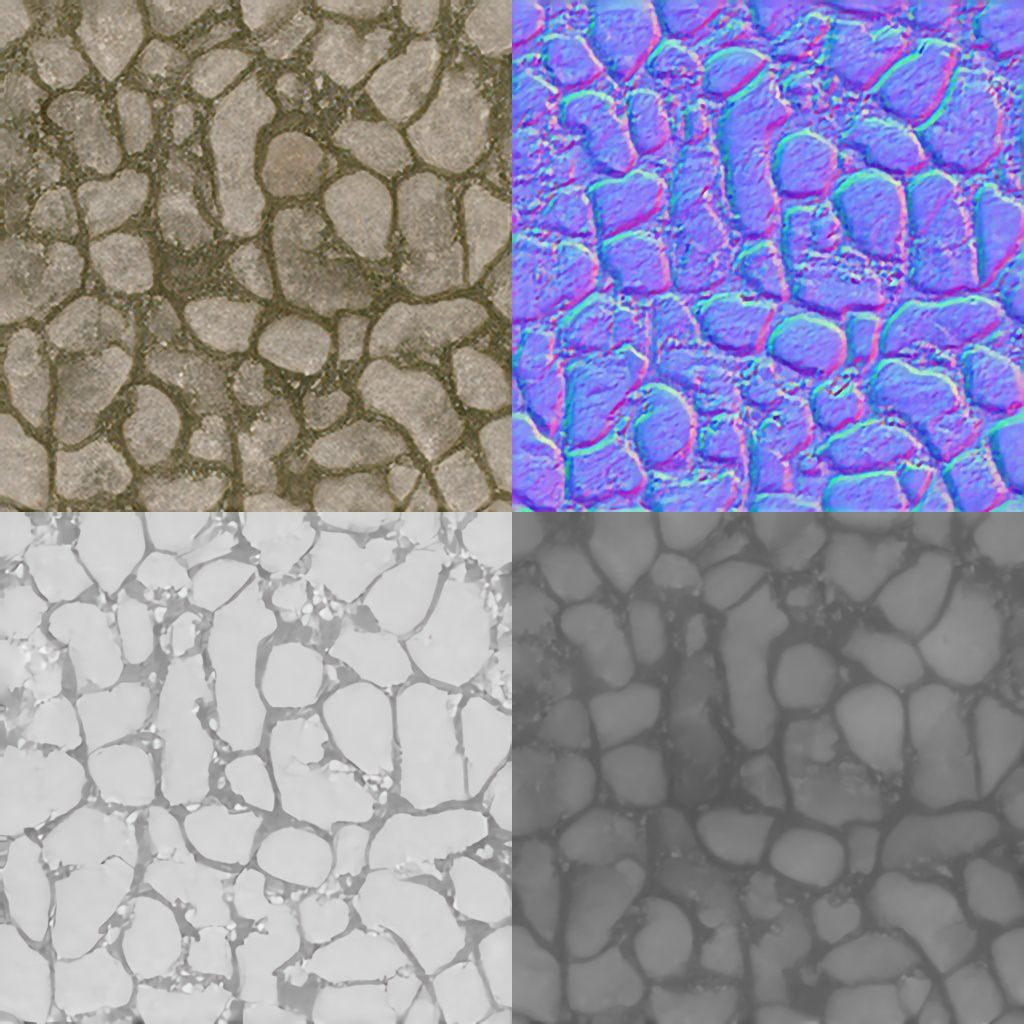}
            \end{minipage}	
            \begin{minipage}{0.26\linewidth}
            \includegraphics[width=\linewidth]{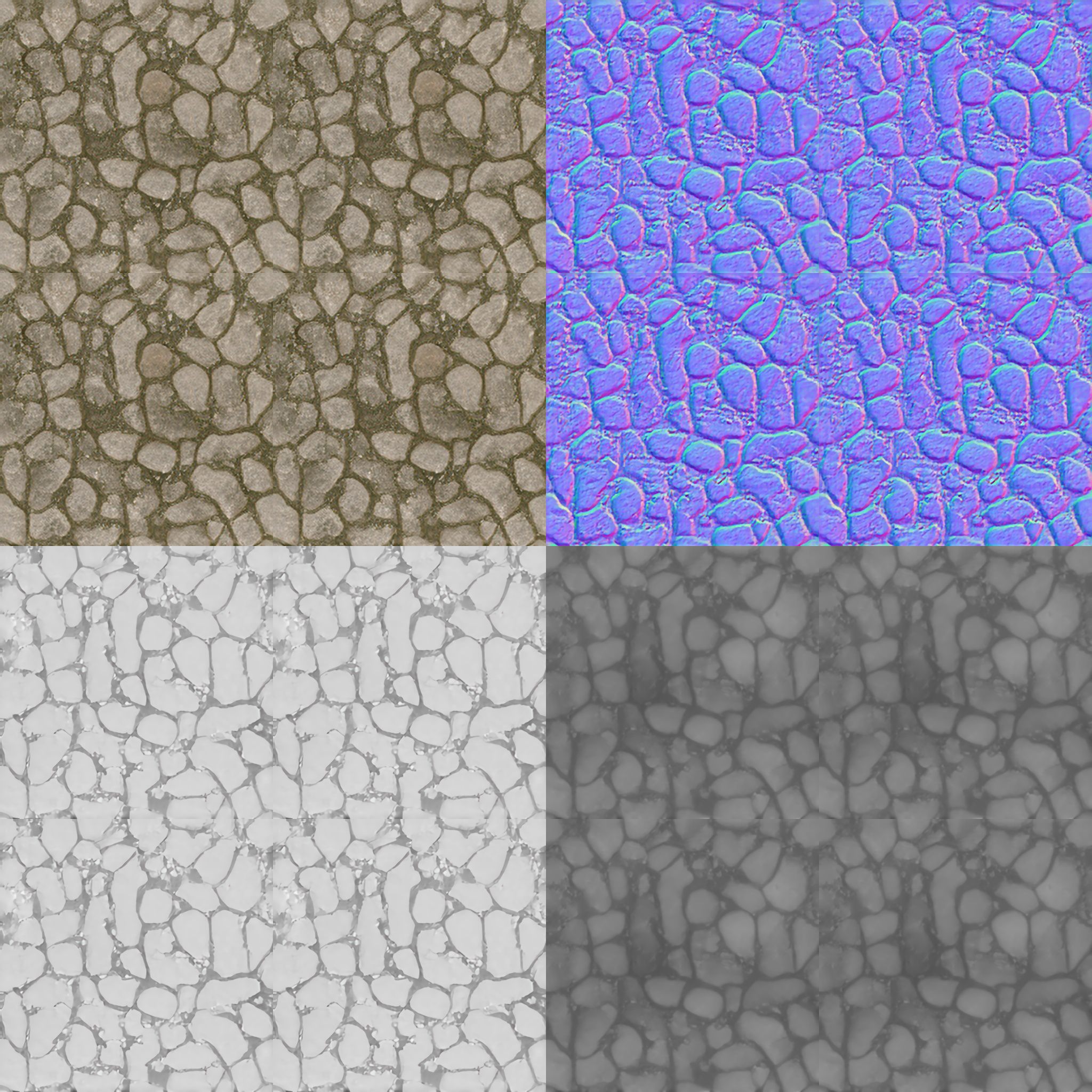}
            \end{minipage}	
            \begin{minipage}{0.26\linewidth}
            \includegraphics[width=\linewidth]{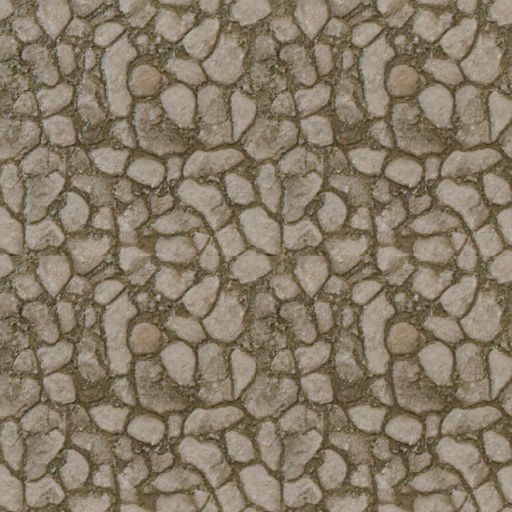}
            \end{minipage}	
        \end{minipage}	
    \end{minipage}	

   \caption{Evaluation of tileable material generation. During inference, we can employ the noise rolling strategy~\cite{vecchio2024controlmat} to enhance the seamlessness of the generated results. The first column shows the input images, followed by the 512$\times$512 material maps. The third column displays 1024$\times$1024 tiling results obtained by stitching the 512-resolution outputs, and the final column presents the rendering results using the 1024-resolution maps. Maps are ordered as albedo, normal, roughness, and height, from top to bottom and left to right. Metallic maps are omitted as they are black in these cases.}
   \label{fig:tileable}
\end{figure}

Although our model is not explicitly trained to produce tileable outputs, we can easily incorporate noise rolling, a test-time technique proposed in ControlMat~\cite{vecchio2024controlmat}, to generate seamless material maps without any re-training. We present several successful examples of applying this technique in Fig.~\ref{fig:tileable}.

\section{Limitations}
\begin{figure}[htbp!]
    \centering		
    \begin{minipage}{3.4in}
        \begin{minipage}{0.02in}	
            \centering
                \vspace{0.2in}
                % \rotatebox{90}{\tiny {"Concrete pavement"}}
                \rotatebox{90}{\parbox{1cm}{\centering\tiny "Concrete\vspace{-0.05cm}\\pavement"}}
        \end{minipage}	
        \hspace{0.02in}
             \begin{minipage}{3.3in}	
                \centering
                \begin{minipage}{0.13\linewidth}
                    \subcaption*{\tiny Input}
                    \includegraphics[width=\linewidth]{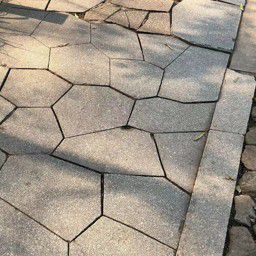}
                \end{minipage}
                \begin{minipage}{0.13\linewidth}
                    \subcaption*{\tiny Albedo}
                    \includegraphics[width=\linewidth]{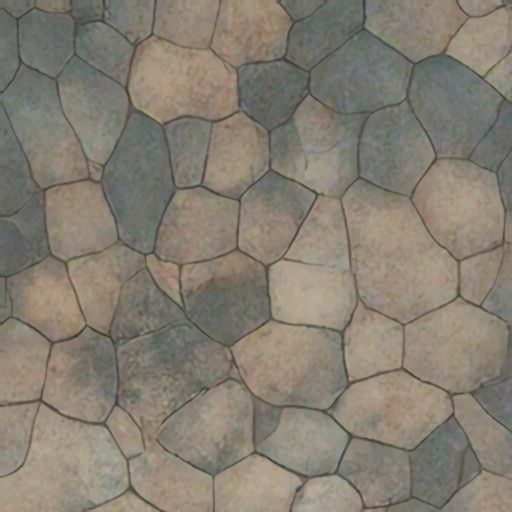}
                \end{minipage}
                \begin{minipage}{0.13\linewidth}
                    \subcaption*{\tiny Normal}
                    \includegraphics[width=\linewidth]{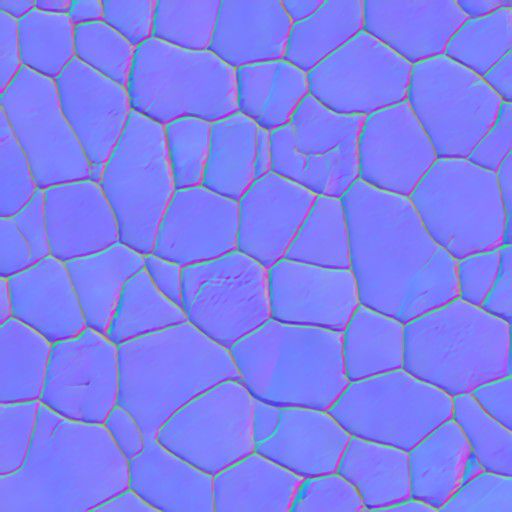}
                \end{minipage}
                \begin{minipage}{0.13\linewidth}
                    \subcaption*{\tiny Roughness}
                    \includegraphics[width=\linewidth]{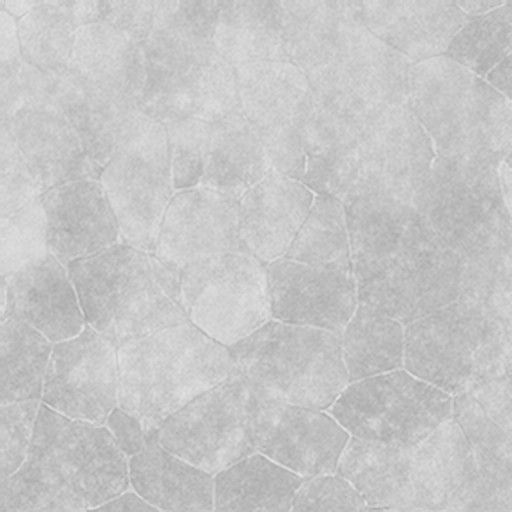}
                \end{minipage}
                \begin{minipage}{0.13\linewidth}
                    \subcaption*{\tiny Height}
                    \includegraphics[width=\linewidth]{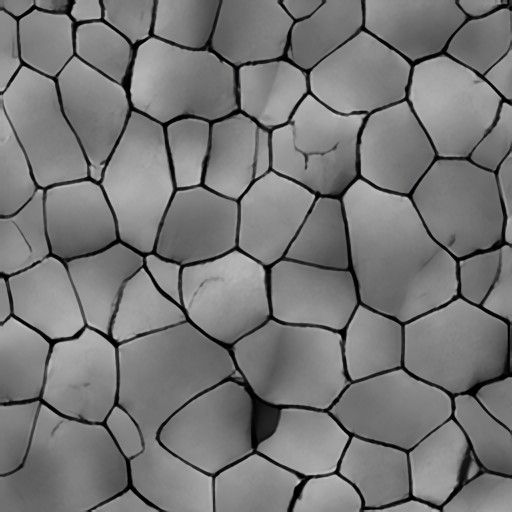}
                \end{minipage}
                \begin{minipage}{0.13\linewidth}
                    \subcaption*{\tiny Metallic}
                    \includegraphics[width=\linewidth]{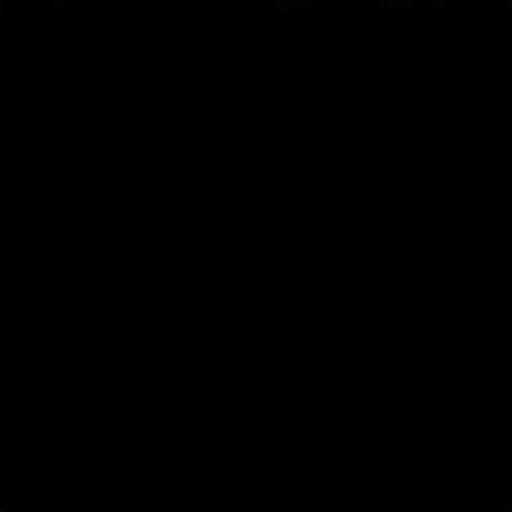}
                \end{minipage}
                \begin{minipage}{0.13\linewidth}
                    \subcaption*{\tiny Render}
                    \includegraphics[width=\linewidth]{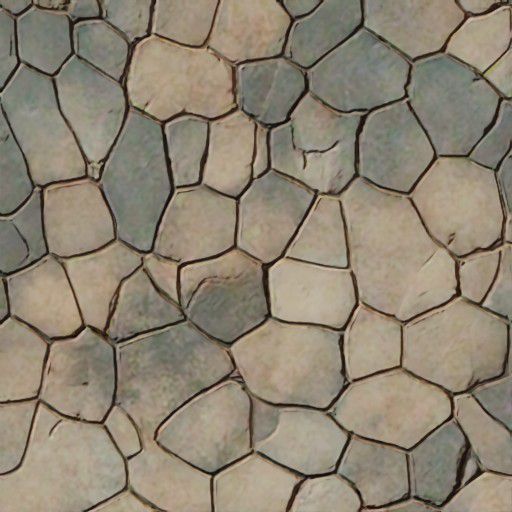}
                \end{minipage}
            \end{minipage}	
        \end{minipage}	

    \begin{minipage}{3.4in}
        \begin{minipage}{0.02in}	
            \centering
                \rotatebox{90}{\parbox{1cm}{\centering\tiny "Perforated\vspace{-0.05cm}\\  metal"}}
        \end{minipage}	
        \hspace{0.02in}
         \begin{minipage}{3.3in}	
            \centering
            \begin{minipage}{0.13\linewidth}
            \includegraphics[width=\linewidth]{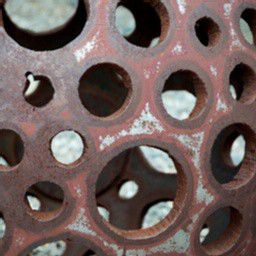}
            \end{minipage}	
            \begin{minipage}{0.13\linewidth}
            \includegraphics[width=\linewidth]{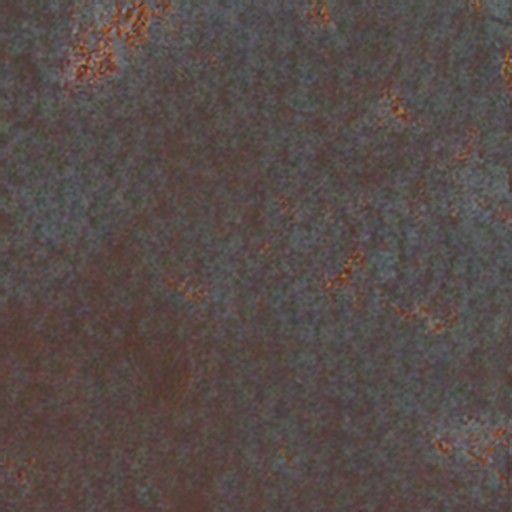}
            \end{minipage}	
            \begin{minipage}{0.13\linewidth}
            \includegraphics[width=\linewidth]{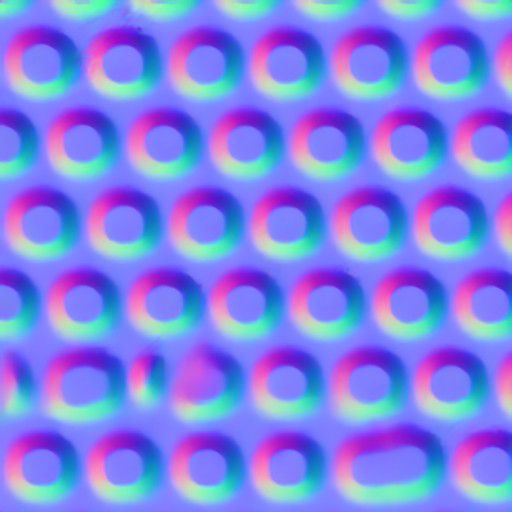}
            \end{minipage}	
            \begin{minipage}{0.13\linewidth}
            \includegraphics[width=\linewidth]{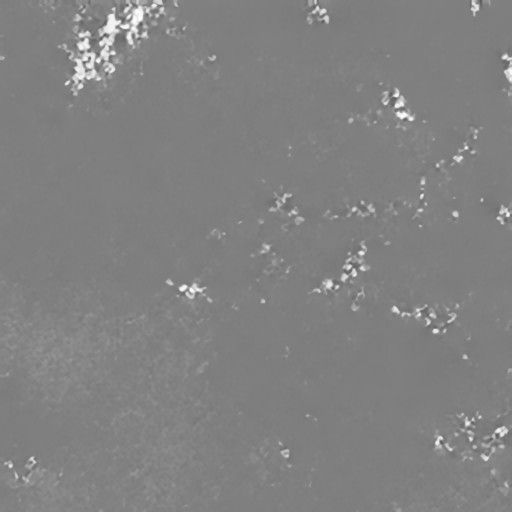}
            \end{minipage}	
            \begin{minipage}{0.13\linewidth}
            \includegraphics[width=\linewidth]{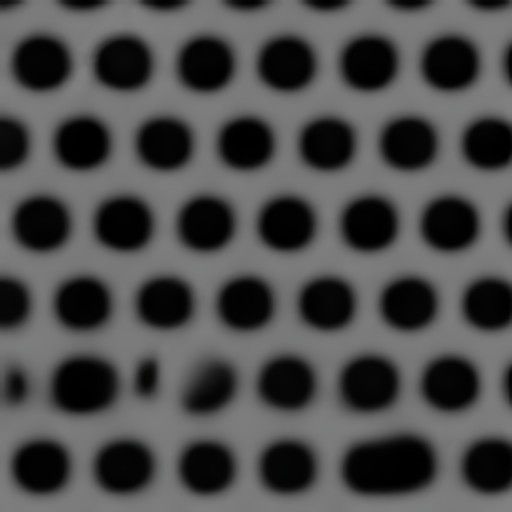}
            \end{minipage}	
            \begin{minipage}{0.13\linewidth}
            \includegraphics[width=\linewidth]{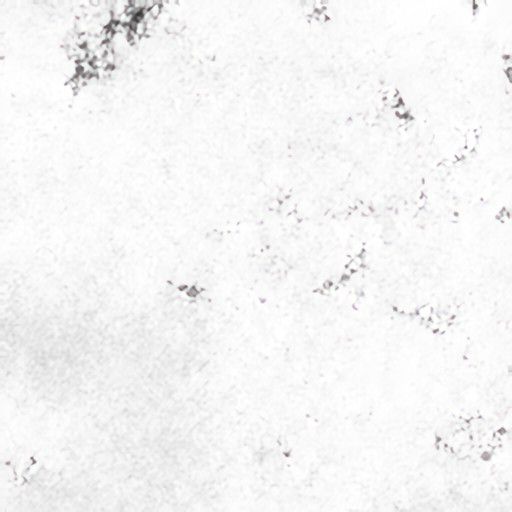}
            \end{minipage}	
            \begin{minipage}{0.13\linewidth}
            \includegraphics[width=\linewidth]{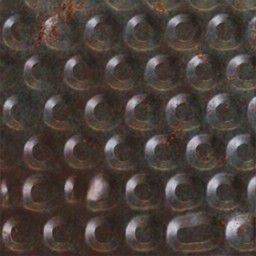}
            \end{minipage}
        \end{minipage}	
    \end{minipage}	

    \begin{minipage}{3.4in}
        \begin{minipage}{0.02in}	
            \centering
                \rotatebox{90}{\parbox{1cm}{\centering\tiny \vspace{0.05cm} "Towel"}}
        \end{minipage}	
        \hspace{0.02in}
         \begin{minipage}{3.3in}	
            \centering
            \begin{minipage}{0.13\linewidth}
            \includegraphics[width=\linewidth]{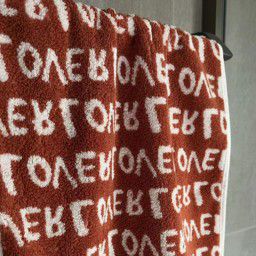}
            \end{minipage}	
            \begin{minipage}{0.13\linewidth}
            \includegraphics[width=\linewidth]{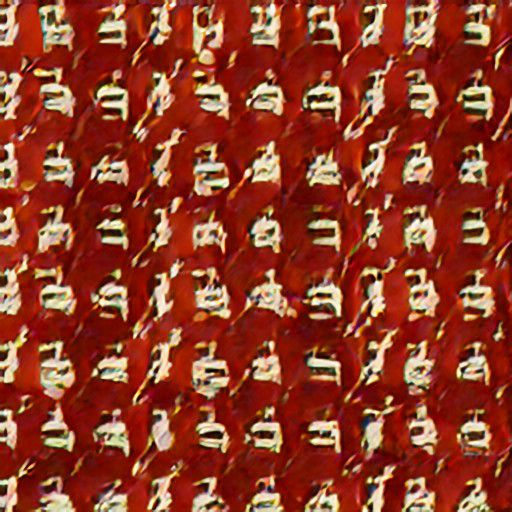}
            \end{minipage}	
            \begin{minipage}{0.13\linewidth}
            \includegraphics[width=\linewidth]{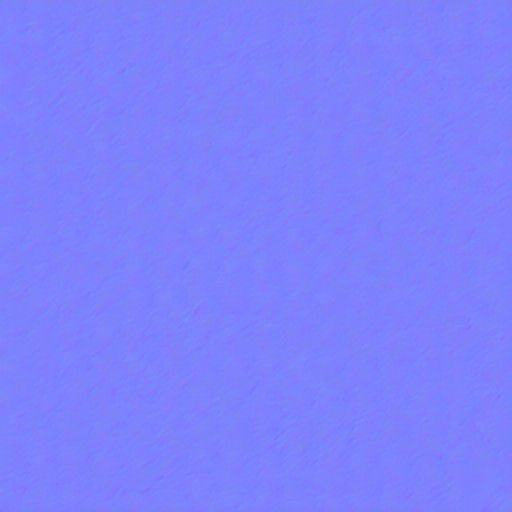}
            \end{minipage}	
            \begin{minipage}{0.13\linewidth}
            \includegraphics[width=\linewidth]{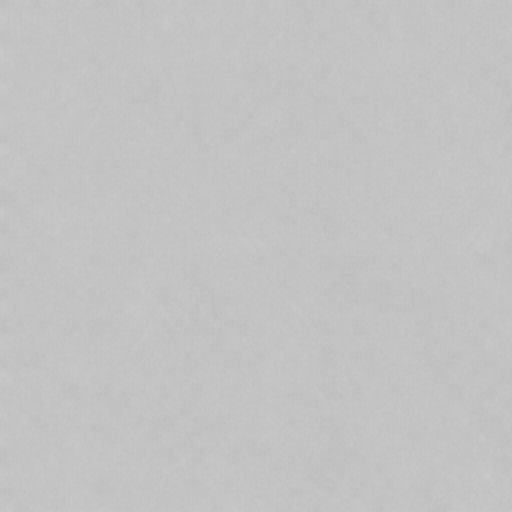}
            \end{minipage}	
            \begin{minipage}{0.13\linewidth}
            \includegraphics[width=\linewidth]{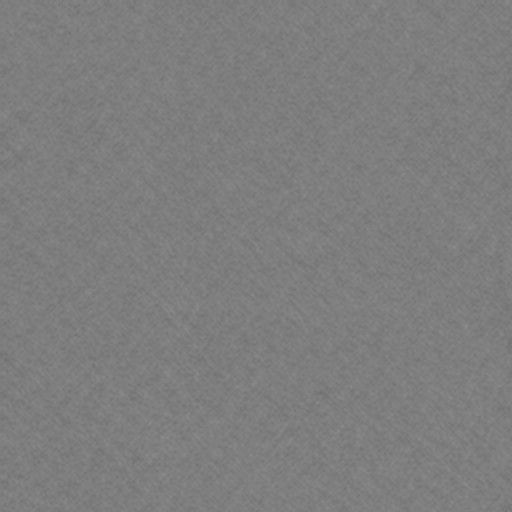}
            \end{minipage}	
            \begin{minipage}{0.13\linewidth}
            \includegraphics[width=\linewidth]{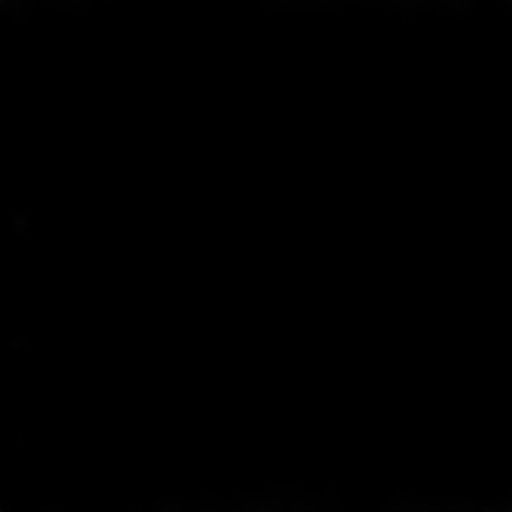}
            \end{minipage}	
            \begin{minipage}{0.13\linewidth}
            \includegraphics[width=\linewidth]{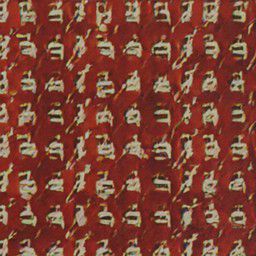}
            \end{minipage}
        \end{minipage}	
    \end{minipage}	

    \begin{minipage}{3.4in}
        \begin{minipage}{0.02in}	
            \centering
                \rotatebox{90}{\parbox{1cm}{\centering\tiny \vspace{0.05cm} "Glass"}}
        \end{minipage}	
        \hspace{0.02in}
         \begin{minipage}{3.3in}	
            \centering
            \begin{minipage}{0.13\linewidth}
            \includegraphics[width=\linewidth]{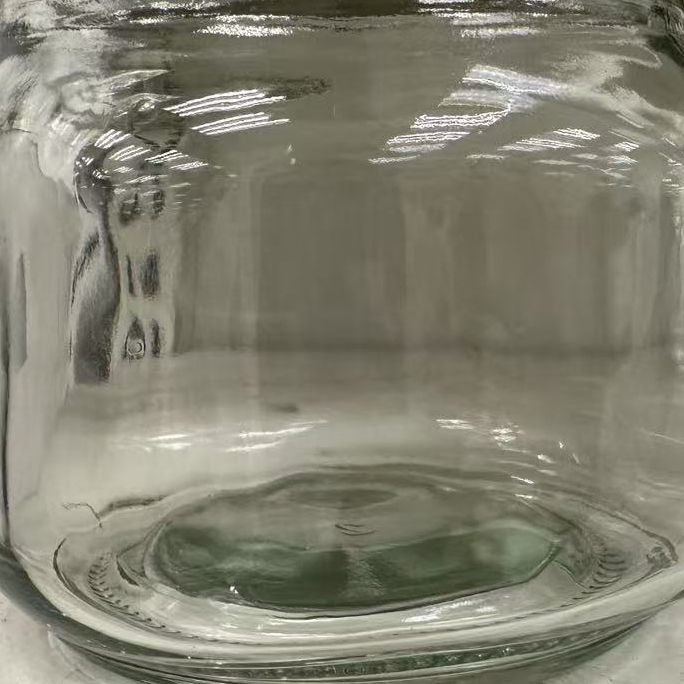}
            \end{minipage}	
            \begin{minipage}{0.13\linewidth}
            \includegraphics[width=\linewidth]{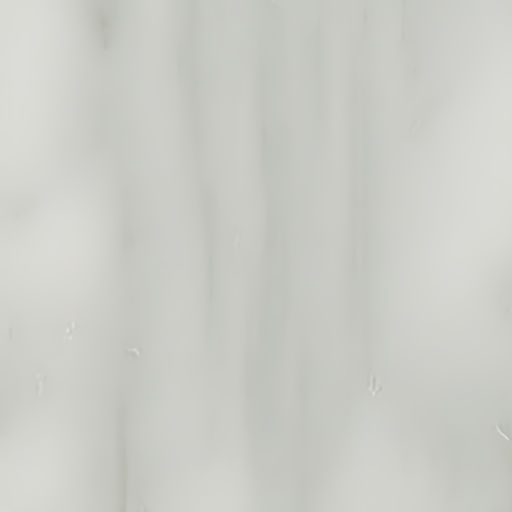}
            \end{minipage}	
            \begin{minipage}{0.13\linewidth}
            \includegraphics[width=\linewidth]{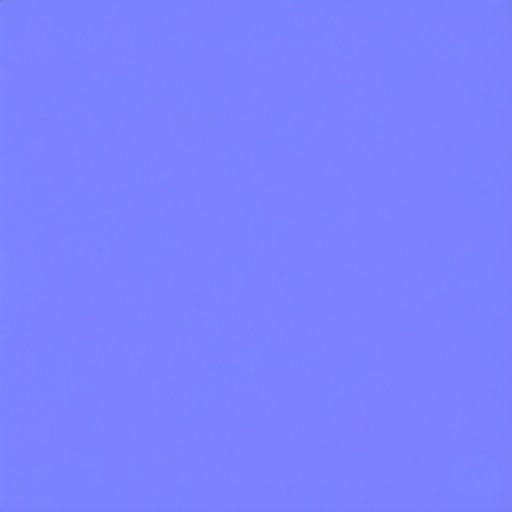}
            \end{minipage}	
            \begin{minipage}{0.13\linewidth}
            \includegraphics[width=\linewidth]{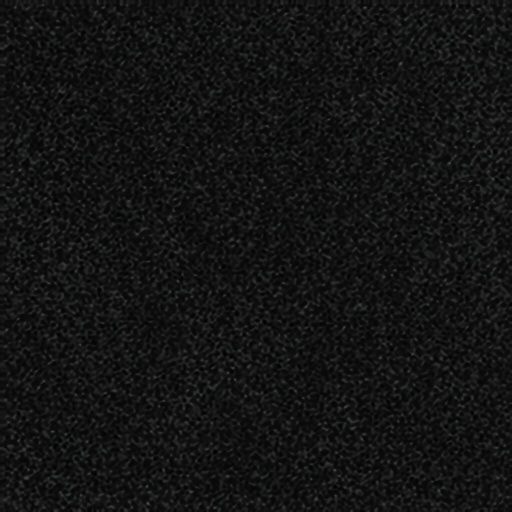}
            \end{minipage}	
            \begin{minipage}{0.13\linewidth}
            \includegraphics[width=\linewidth]{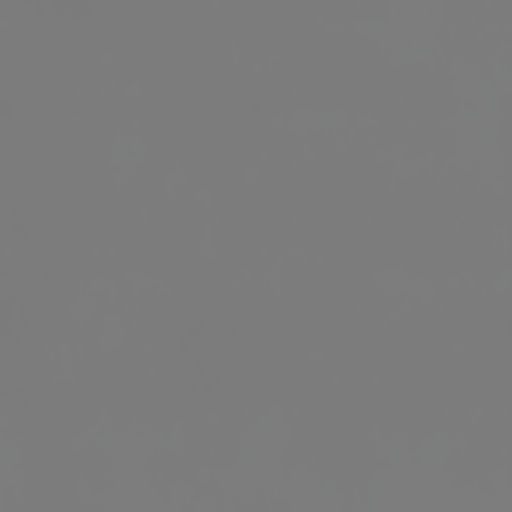}
            \end{minipage}	
            \begin{minipage}{0.13\linewidth}
            \includegraphics[width=\linewidth]{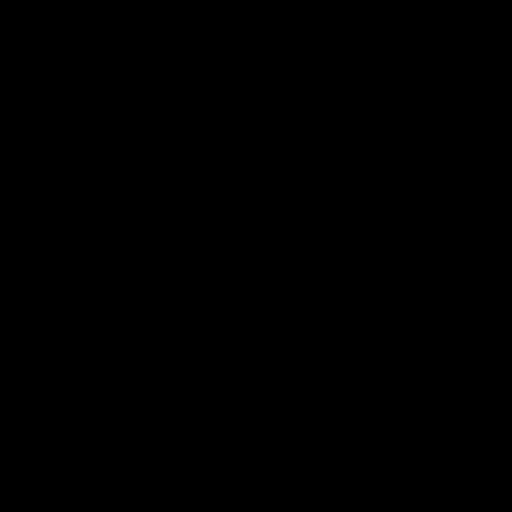}
            \end{minipage}	
            \begin{minipage}{0.13\linewidth}
            \includegraphics[width=\linewidth]{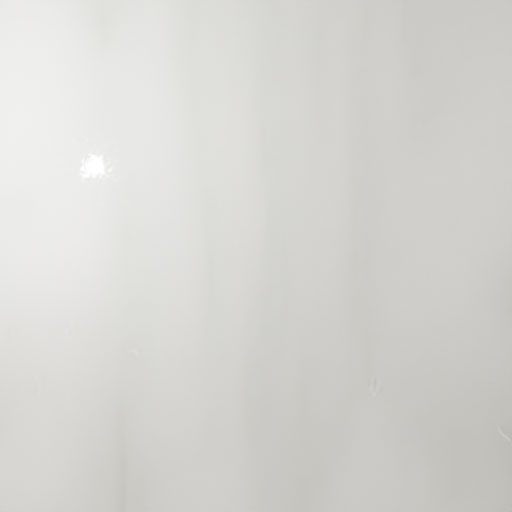}
            \end{minipage}
        \end{minipage}	
    \end{minipage}	

    \begin{minipage}{3.4in}
        \begin{minipage}{0.02in}	
            \centering
                \rotatebox{90}{\parbox{1cm}{\centering\tiny \vspace{0.05cm} "Sheer\vspace{-0.05cm}\\  curtain "}}
        \end{minipage}	
        \hspace{0.02in}
         \begin{minipage}{3.3in}	
            \centering
            \begin{minipage}{0.13\linewidth}
            \includegraphics[width=\linewidth]{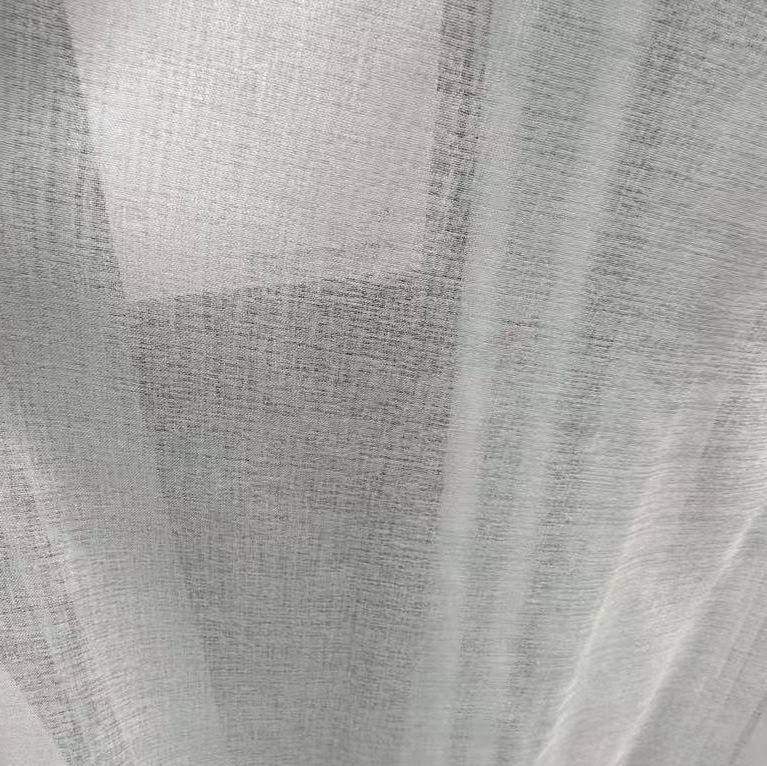}
            \end{minipage}	
            \begin{minipage}{0.13\linewidth}
            \includegraphics[width=\linewidth]{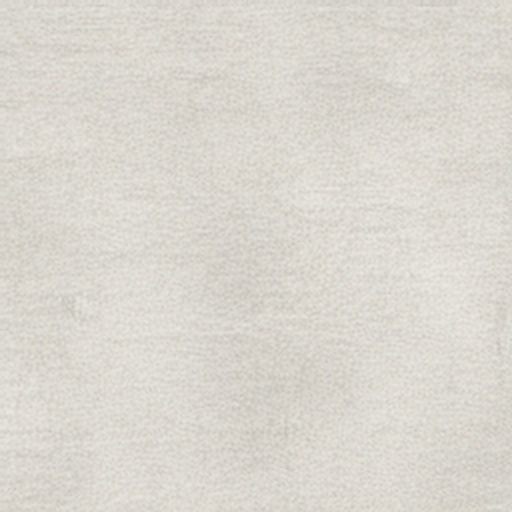}
            \end{minipage}	
            \begin{minipage}{0.13\linewidth}
            \includegraphics[width=\linewidth]{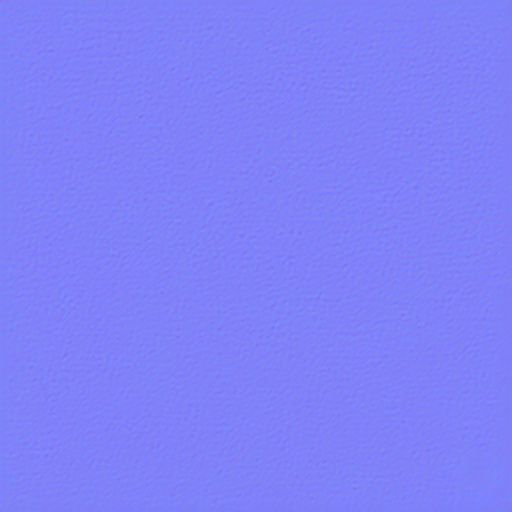}
            \end{minipage}	
            \begin{minipage}{0.13\linewidth}
            \includegraphics[width=\linewidth]{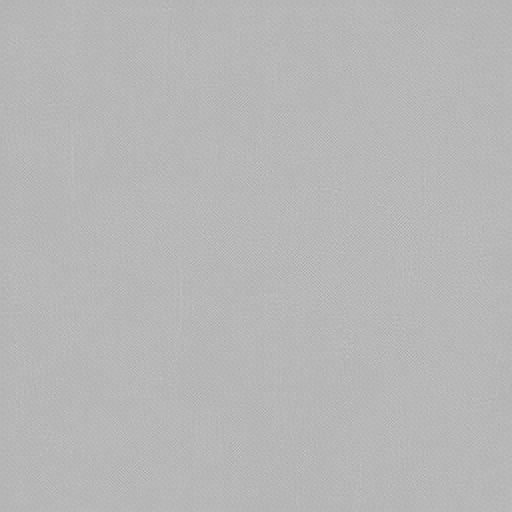}
            \end{minipage}	
            \begin{minipage}{0.13\linewidth}
            \includegraphics[width=\linewidth]{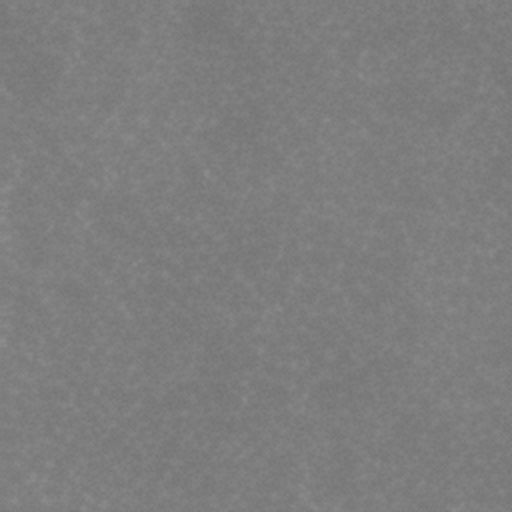}
            \end{minipage}	
            \begin{minipage}{0.13\linewidth}
            \includegraphics[width=\linewidth]{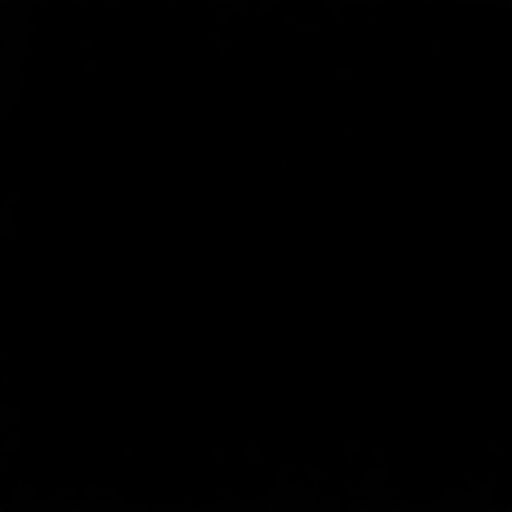}
            \end{minipage}	
            \begin{minipage}{0.13\linewidth}
            \includegraphics[width=\linewidth]{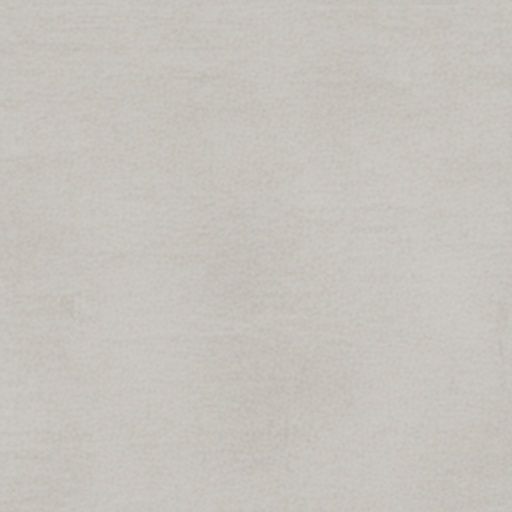}
            \end{minipage}
        \end{minipage}	
    \end{minipage}

   \caption{Limitations. We show limitations of our model, such as complex lighting and shadows in the first row, materials with perforations in the second row, structurally significant elements like text in the third row, and effects involving transparency and translucency in the last two rows.}
   \label{fig:limitations}
\end{figure}

Despite strong generation capacity, our model may still encounter challenging inputs, as shown in Fig.~\ref{fig:limitations}. In the first row we show an example where our model confuses shading and albedo variation. Our model may also have difficulty handling materials with cutouts or holes, since it does not produce opacity maps as outputs. Also, preserving semantically meaningful patterns, such as text, is a remaining challenge in our approach. Furthermore, our current model is not capable of handling transparency and refraction effects. In the third row of Fig.~\ref{fig:distortion}, we present an example of generating "ice", where the model tries to approximates the appearance using the albedo map. The fourth and fifth rows of Fig.~\ref{fig:limitations} illustrate more challenging cases, in which the model fails to reproduce the visual characteristics of transparent and translucent materials due to the lack of relevant training data.

\begin{figure}[H]
    \centering		
    \begin{minipage}{3.4in}
        \centering
        \begin{minipage}{0.24\linewidth}
        \includegraphics[width=\linewidth]{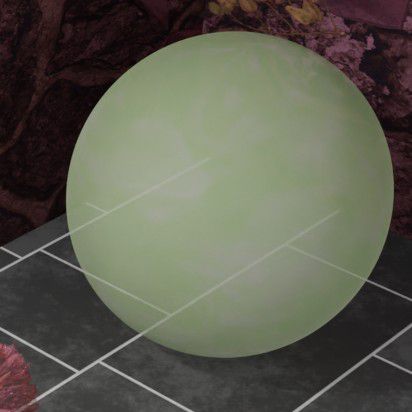}
        \end{minipage}	
        \begin{minipage}{0.24\linewidth}
        \includegraphics[width=\linewidth]{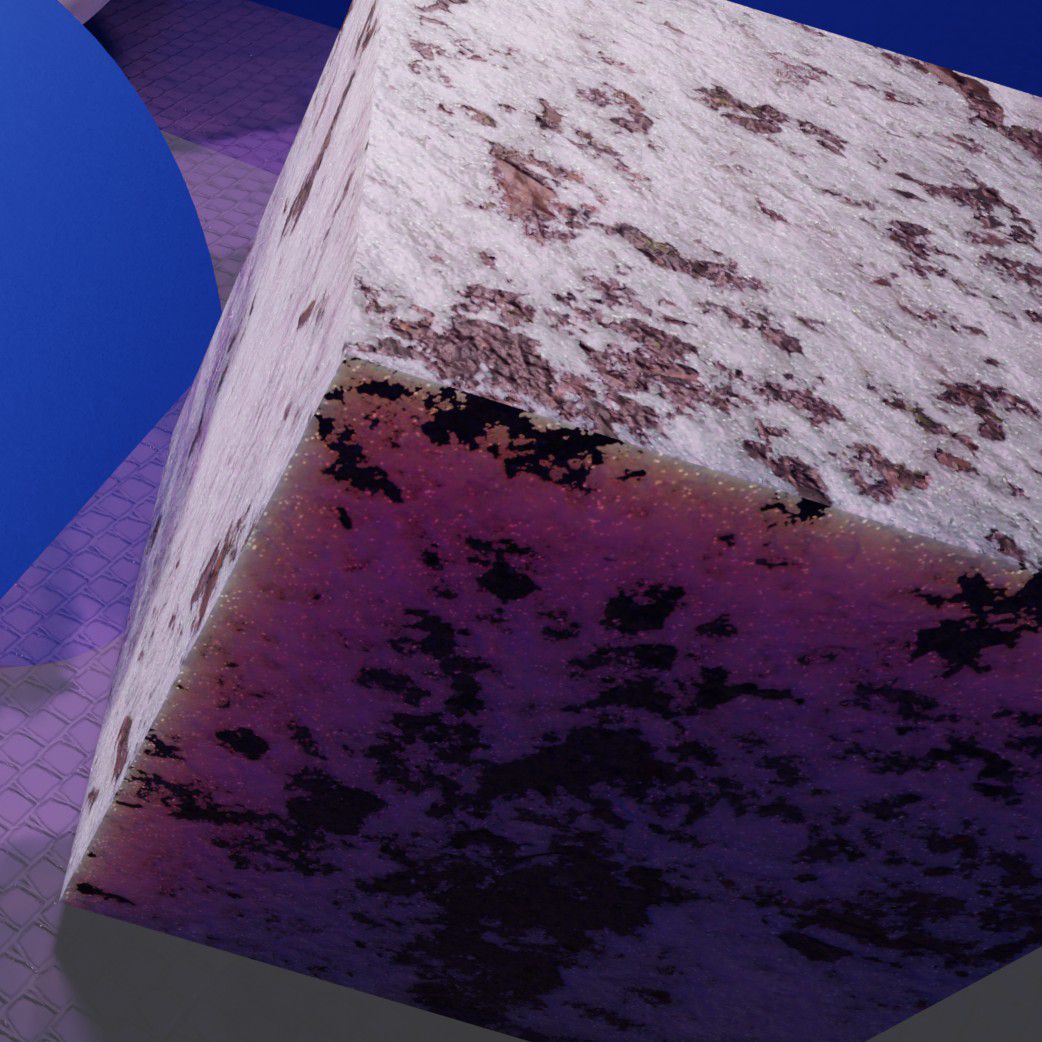}
        \end{minipage}	
        \begin{minipage}{0.24\linewidth}
        \includegraphics[width=\linewidth]{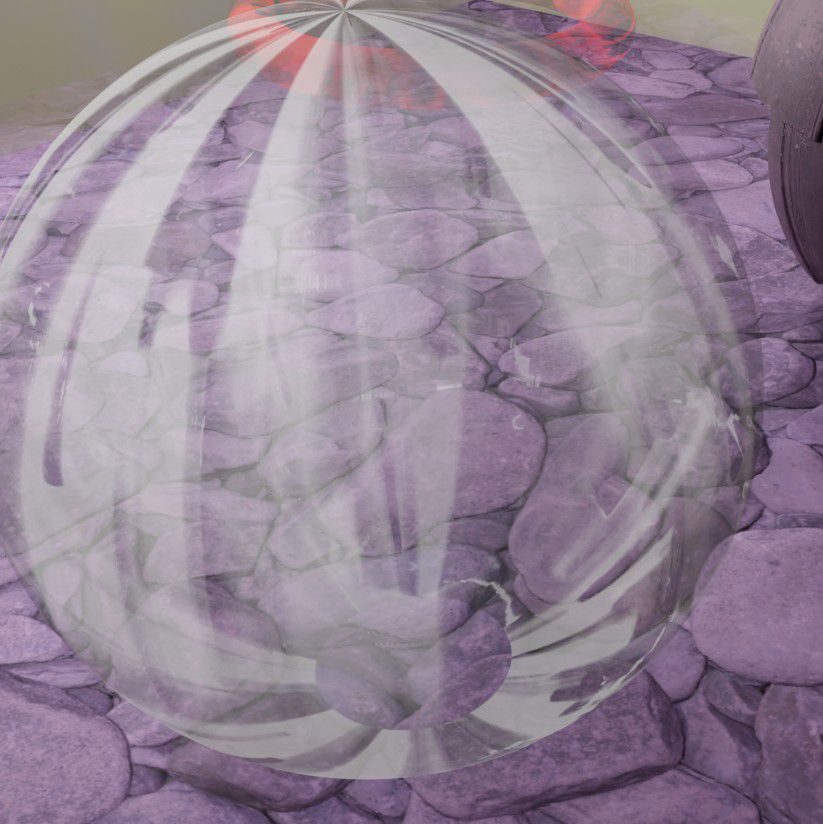}
        \end{minipage}	
        \begin{minipage}{0.24\linewidth}
        \includegraphics[width=\linewidth]{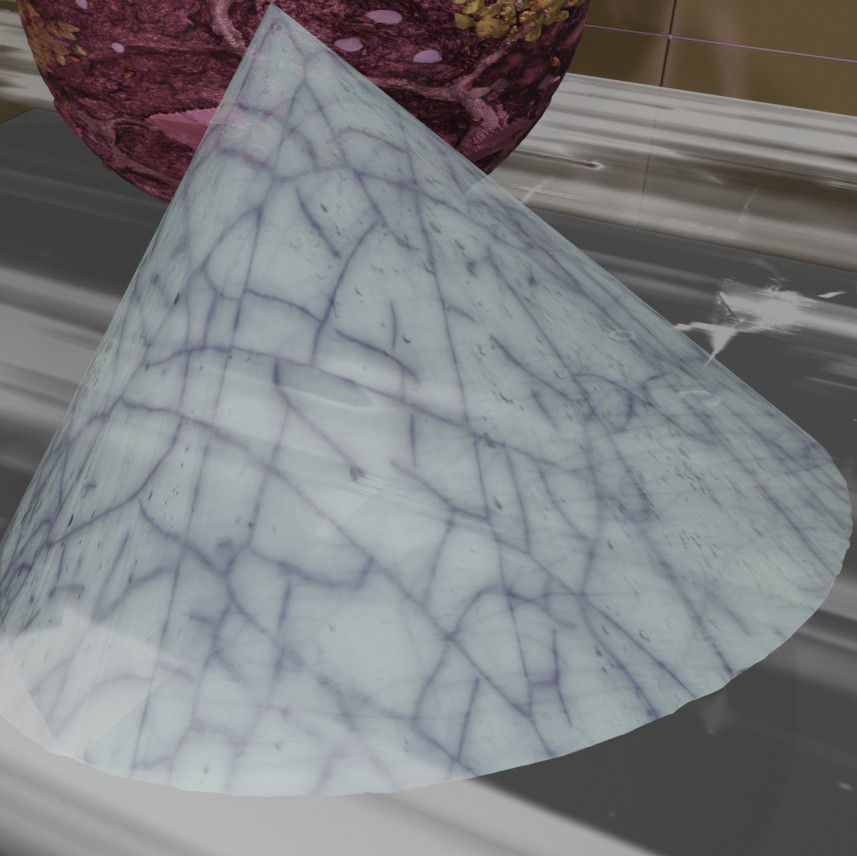}
        \end{minipage}	
    \end{minipage}	

   \caption{Future work. These images are obtained using the method described in Sec.~\ref{sec:dataset}, with the addition of a translucency map during rendering. We consider such complex scenarios as future work.}
   \label{fig:future_work}
\end{figure}

To improve the model’s capacity for handling complex appearance effects such as transparency, translucency, and detailed BRDF components, one promising direction is to incorporate more advanced material maps (e.g., specular, coat, and subsurface scattering) into the training data, illustrated in Fig.~\ref{fig:future_work}. As outlined in Sec.~\ref{sec:our_model}, our architecture is flexible and can be extended to generate additional output channels corresponding to these effects. We leave this extension for future work, once sufficiently large-scale datasets with such complex annotations become available.

\section{Conclusion}
We present a generative model for high-quality material synthesis from text prompts and/or crops of natural images by finetuning a pretrained text-to-video generative model, which provides strong prior knowledge. The flexible video DiT architecture lets us adjust the model for multi-channel material generation. We show extensive evaluation on both synthetic and real examples and conduct systematic ablation studies and test on robustness. We believe that our re-purposing of a video model for multi-channel generation opens an interesting avenue for other domain which require the generation of additional channels, such as intrinsic decomposition~\cite{vecchio2024controlmat}.

% \clearpage
% \setcounter{page}{1}
% \maketitlesupplementary
\appendix
\section{More synthetic examples}

\begin{figure*}[t]
    \centering		
    \begin{minipage}{7.0in}
         \begin{minipage}{7.0in}	
            \begin{minipage}{0.16\linewidth}
            \centering
            \subcaption*{\small Ground-Truth}
            \end{minipage}
            \begin{minipage}{0.16\linewidth}
            \centering
            \subcaption*{\small Ours}
            \end{minipage}
            \begin{minipage}{0.16\linewidth}
            \centering
            \subcaption*{\small Material Palette}
            \end{minipage}
            \hspace{0.05in}
            \begin{minipage}{0.16\linewidth}
            \centering
            \subcaption*{\small Ground-Truth}
            \end{minipage}
            \begin{minipage}{0.16\linewidth}
            \centering
            \subcaption*{\small Ours}
            \end{minipage}	
            \begin{minipage}{0.16\linewidth}
            \centering
            \subcaption*{\small Material Palette}
            \end{minipage}
        \end{minipage}	
    \end{minipage}		
    
    \begin{minipage}{7.0in}
         \begin{minipage}{7.0in}	
            \centering
            \includegraphics[width=\linewidth]{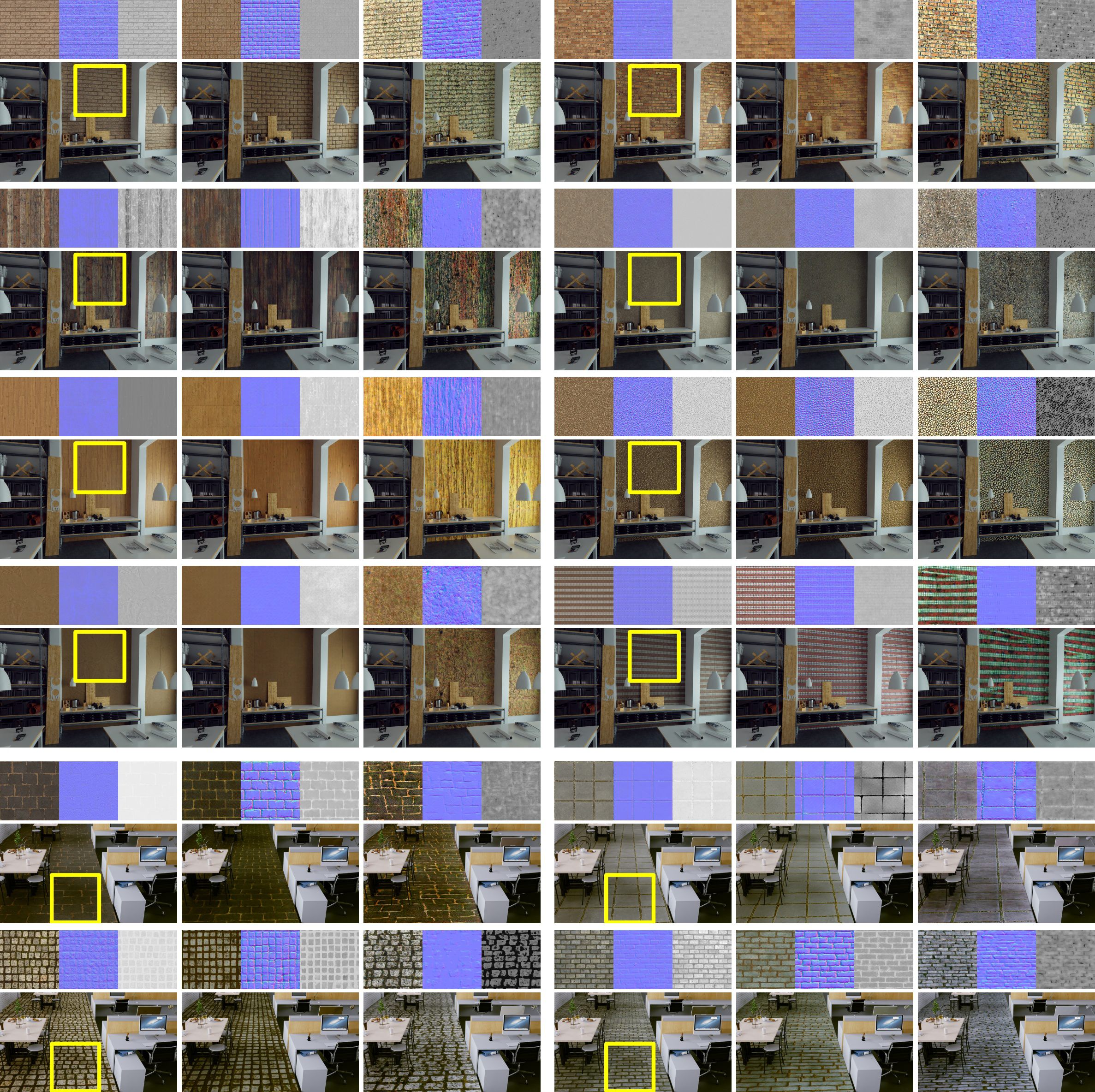}
        \end{minipage}	
    \end{minipage}	
    \begin{tikzpicture}
        \node (txt) at (0,0) {};
        \begin{scope}[shift={(txt.east)}, overlay, remember picture]
            \draw[dashed, color=darkred,line width=0.1pt] (0.06cm,0.1cm) -- ++(0,18.0cm);
        \end{scope}
    \end{tikzpicture}
    \begin{tikzpicture}
    \node (txt) at (0,0) {};
        \begin{scope}[shift={(txt.east)}, overlay, remember picture]
            \draw[dashed, color=darkred, line width=0.1pt] (-9cm, 5.6cm) -- ++(17.7cm, 0);
        \end{scope}
    \end{tikzpicture}
   \caption{More results of comparisons with Material palette~\cite{lopes2024material} on synthetic dataset (Sec. 4.1) for material extraction. Each row contains two sets of comparisons. In each set, the first column shows the ground truth material maps from PolyHaven, with the rendered scene below. The yellow square area indicates the crop used as the input for both models. The second and third columns show the material maps extracted by our model and Material Palette, along with the re-rendered images. We can see that our approach better matches the Ground-Truth appearance.} 
   \label{fig:more_synthetic_results_1}
\end{figure*}

\begin{figure*}[t]
    \centering		
    \begin{minipage}{7.0in}
         \begin{minipage}{7.0in}	
            \begin{minipage}{0.16\linewidth}
            \centering
            \subcaption*{\small Ground-Truth}
            \end{minipage}
            \begin{minipage}{0.16\linewidth}
            \centering
            \subcaption*{\small Ours}
            \end{minipage}
            \begin{minipage}{0.16\linewidth}
            \centering
            \subcaption*{\small Material Palette}
            \end{minipage}
            \hspace{0.05in}
            \begin{minipage}{0.16\linewidth}
            \centering
            \subcaption*{\small Ground-Truth}
            \end{minipage}
            \begin{minipage}{0.16\linewidth}
            \centering
            \subcaption*{\small Ours}
            \end{minipage}	
            \begin{minipage}{0.16\linewidth}
            \centering
            \subcaption*{\small Material Palette}
            \end{minipage}
        \end{minipage}	
    \end{minipage}		
    
    \begin{minipage}{7.0in}
         \begin{minipage}{7.0in}	
            \centering
            \includegraphics[width=\linewidth]{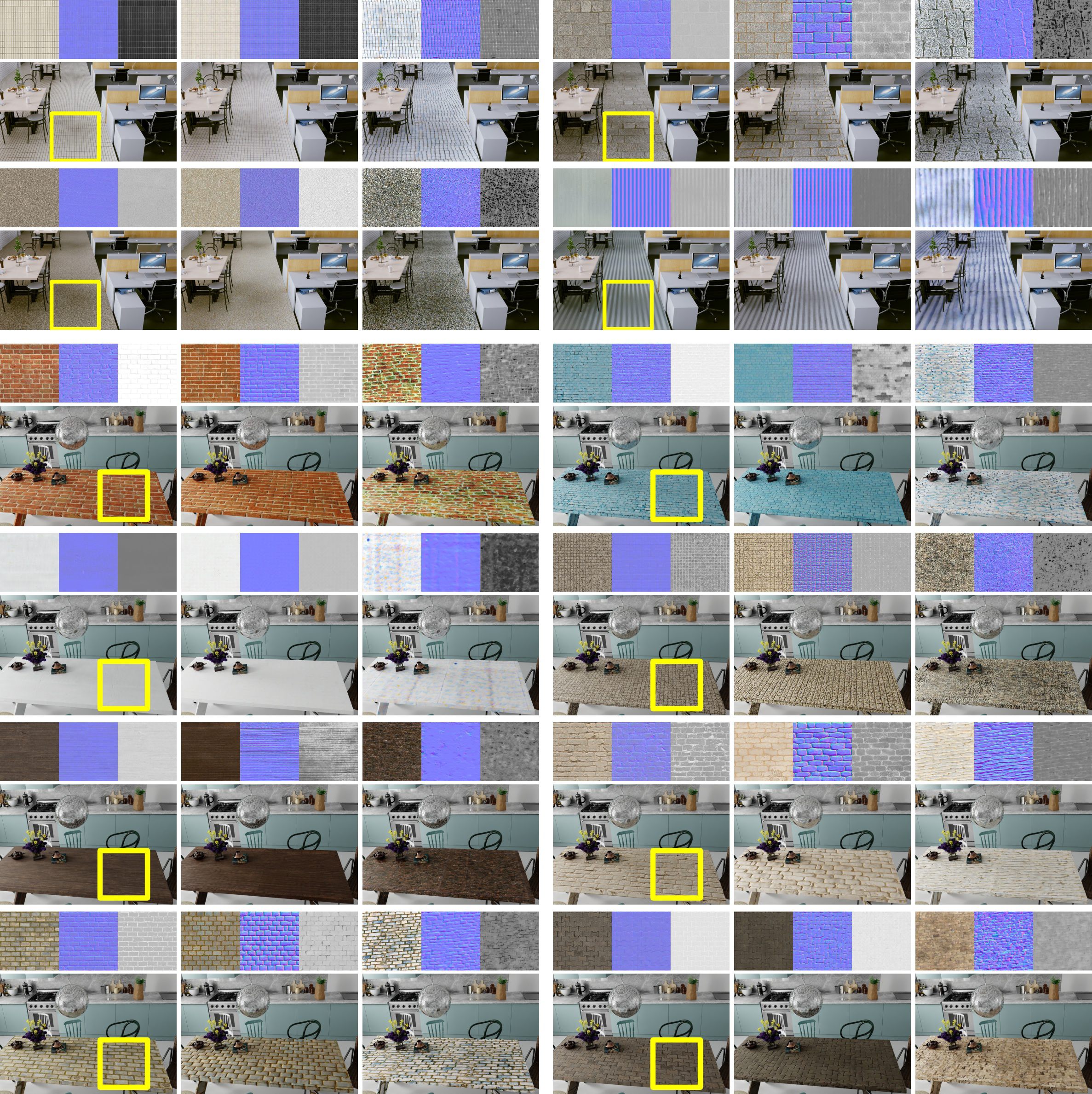}
        \end{minipage}	
    \end{minipage}	
    \begin{tikzpicture}
        \node (txt) at (0,0) {};
        \begin{scope}[shift={(txt.east)}, overlay, remember picture]
            \draw[dashed, color=darkred,line width=0.1pt] (0.06cm,0.1cm) -- ++(0,18.0cm);
        \end{scope}
    \end{tikzpicture}
    \begin{tikzpicture}
    \node (txt) at (0,0) {};
        \begin{scope}[shift={(txt.east)}, overlay, remember picture]
            \draw[dashed, color=darkred, line width=0.1pt] (-9cm, 12.5cm) -- ++(17.7cm, 0);
        \end{scope}
    \end{tikzpicture}
   \caption{More results of comparisons with Material palette~\cite{lopes2024material} on synthetic dataset (Sec. 4.1) for material extraction. Each row contains two sets of comparisons. In each set, the first column shows the ground truth material maps from PolyHaven, with the rendered scene below. The yellow square area indicates the crop used as the input for both models. The second and third columns show the material maps extracted by our model and Material Palette, along with the re-rendered images. We can see that our approach better matches the Ground-Truth appearance.} 
   \label{fig:more_synthetic_results_2}
\end{figure*}

We show qualitative and quantitative comparisons with Material Palette~\cite{lopes2024material} on material extraction in Sec. 4.2 % ~\ref{sec:image_conditioned_generation} 
of the main paper. We provide more visual comparisons in Fig.~\ref{fig:more_synthetic_results_1} and Fig.~\ref{fig:more_synthetic_results_2}. Since Material Palette generates only three material maps (albedo, normal, and roughness), we present results for these channels, along with the re-rendered images.

\section{More Real Examples}
\begin{figure*}[h]
    \centering
    \begin{minipage}{7.0in}
        \begin{minipage}{7.0in}
            \centering
            \hspace{0.02in}
            \begin{minipage}{0.103\linewidth}
            \centering
            \subcaption*{\small  Photo}
            \end{minipage}
            \begin{minipage}{0.093\linewidth}
            \centering
            \subcaption*{\small Input}
            \end{minipage}
            \begin{minipage}{0.093\linewidth}
            \centering
            \subcaption*{\small Albedo}
            \end{minipage}
            \begin{minipage}{0.093\linewidth}
            \centering
            \subcaption*{\small Normal}
            \end{minipage}
            \begin{minipage}{0.093\linewidth}
            \centering
            \subcaption*{\small Roughness}
            \end{minipage}
            \begin{minipage}{0.093\linewidth}
            \centering
            \subcaption*{\small Height}
            \end{minipage}
            \begin{minipage}{0.093\linewidth}
            \centering
            \subcaption*{\small Metallic}
            \end{minipage}
            \begin{minipage}{0.093\linewidth}
            \centering
            \subcaption*{\small Render 1}
            \end{minipage}
            \begin{minipage}{0.093\linewidth}
            \centering
            \subcaption*{\small Render 2}
            \end{minipage}
            \begin{minipage}{0.093\linewidth}
            \centering
            \subcaption*{\small Mask}
            \end{minipage}
        \end{minipage}
    \end{minipage}

    \begin{minipage}{7.0in}
        \begin{minipage}{7.0in}
            \centering
            \includegraphics[width=\linewidth]{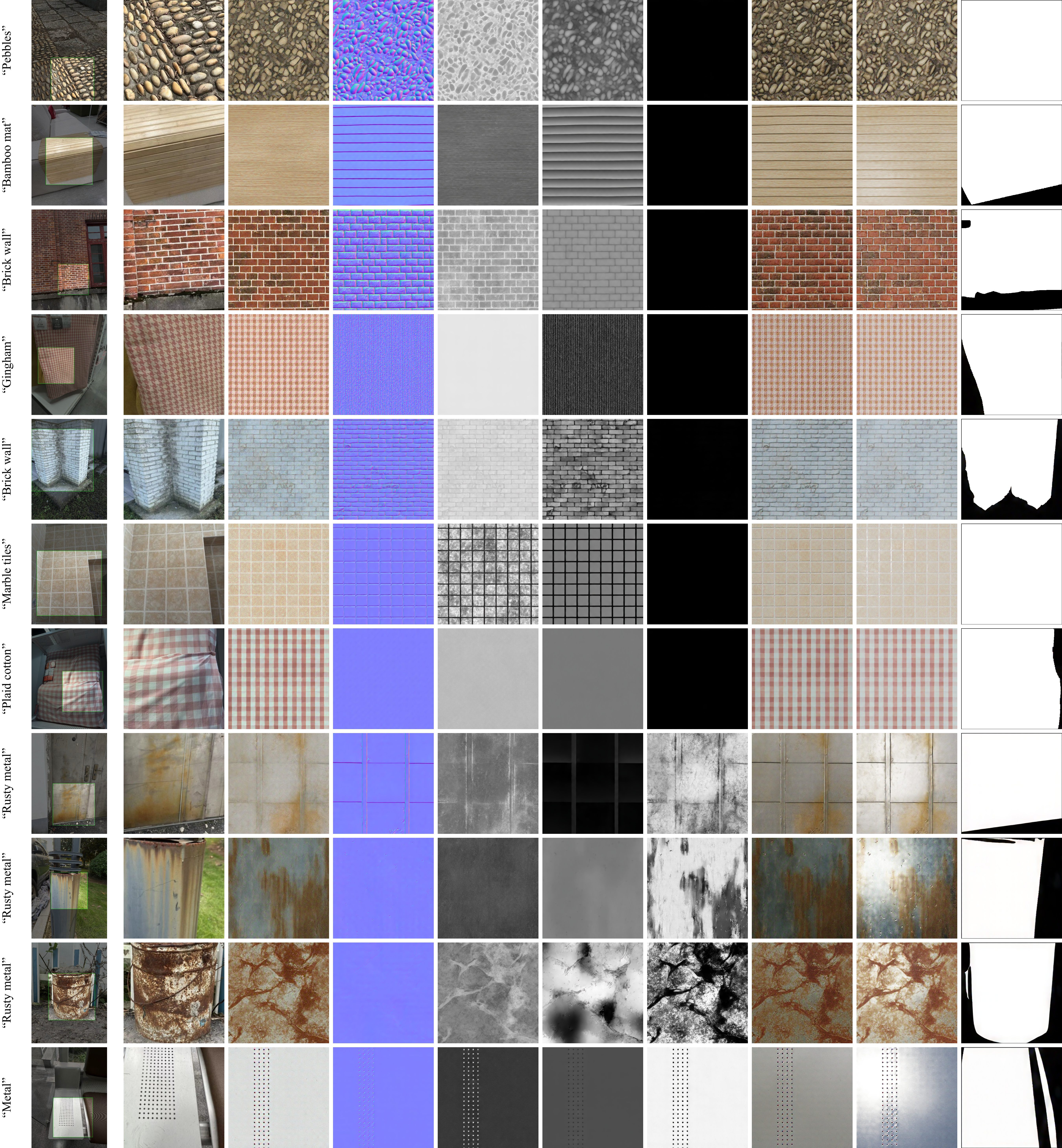}
        \end{minipage}
    \end{minipage}

    \caption{More results of our method on material extraction for real photographs. The first column shows real photographs captured by smartphones, before cropping (as described in Sec. 3.2). The box indicates the cropped area, which is the image actually used as input for the model in the second column. Third to the ninth columns show the generated material maps and rendering under two environment maps. The last column shows the mask of the dominant material location automatically predicted by our model. We conduct tests on diverse material types in both indoor and outdoor scenes, demonstrating the generalization capability of our model. The leftmost side of each row is labeled with the text conditioning input used.} 
   \label{fig:real_material_supp_1}
\end{figure*}

\begin{figure*}[h]
    \centering
    \begin{minipage}{7.0in}
        \begin{minipage}{7.0in}
            \centering
            \hspace{0.02in}
            \begin{minipage}{0.103\linewidth}
            \centering
            \subcaption*{\small  Photo}
            \end{minipage}
            \begin{minipage}{0.093\linewidth}
            \centering
            \subcaption*{\small Input}
            \end{minipage}
            \begin{minipage}{0.093\linewidth}
            \centering
            \subcaption*{\small Albedo}
            \end{minipage}
            \begin{minipage}{0.093\linewidth}
            \centering
            \subcaption*{\small Normal}
            \end{minipage}
            \begin{minipage}{0.093\linewidth}
            \centering
            \subcaption*{\small Roughness}
            \end{minipage}
            \begin{minipage}{0.093\linewidth}
            \centering
            \subcaption*{\small Height}
            \end{minipage}
            \begin{minipage}{0.093\linewidth}
            \centering
            \subcaption*{\small Metallic}
            \end{minipage}
            \begin{minipage}{0.093\linewidth}
            \centering
            \subcaption*{\small Render 1}
            \end{minipage}
            \begin{minipage}{0.093\linewidth}
            \centering
            \subcaption*{\small Render 2}
            \end{minipage}
            \begin{minipage}{0.093\linewidth}
            \centering
            \subcaption*{\small Mask}
            \end{minipage}
        \end{minipage}
    \end{minipage}
    
    \begin{minipage}{7.0in}
        \begin{minipage}{7.0in}
            \centering
            \includegraphics[width=\linewidth]{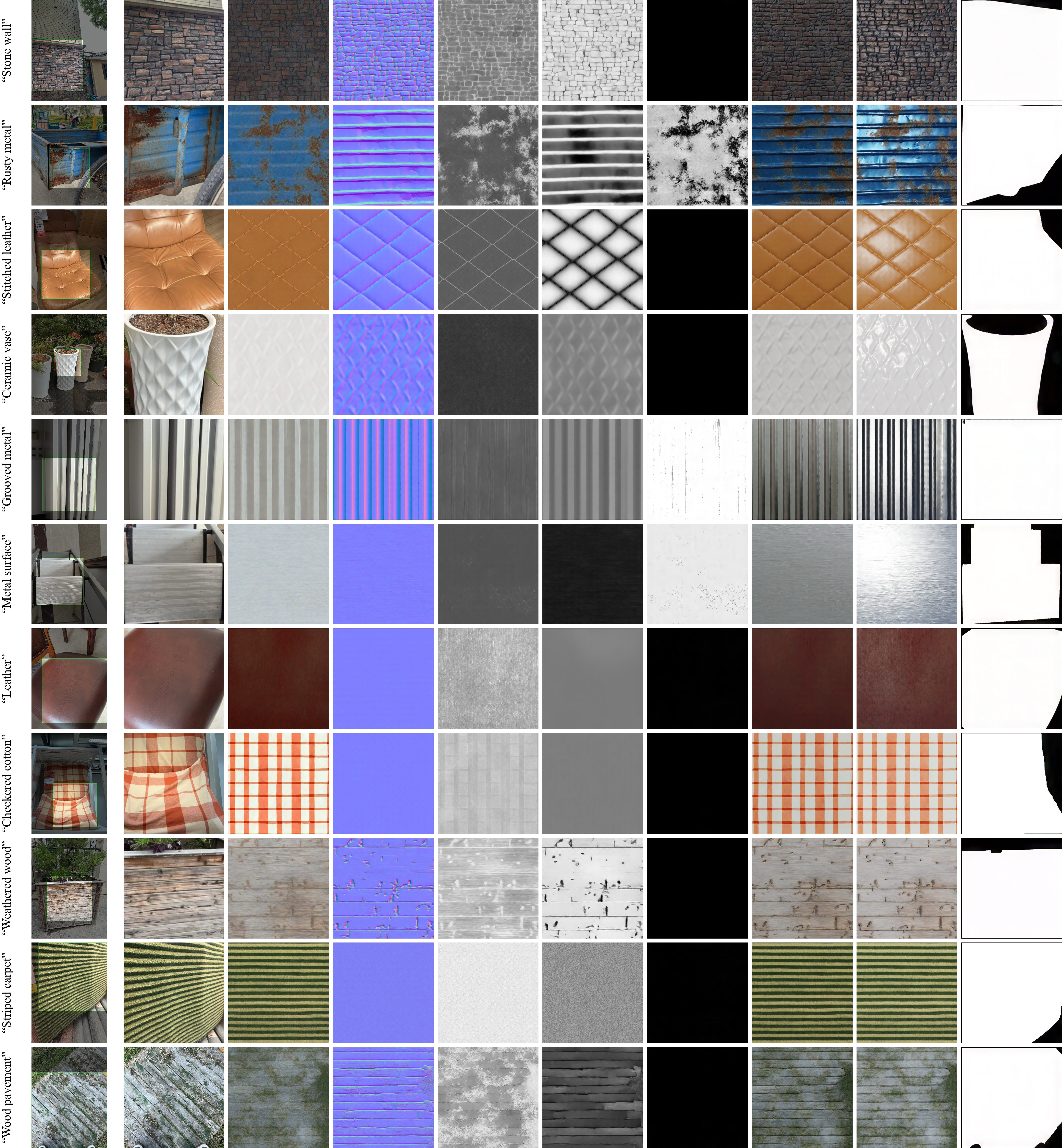}
        \end{minipage}
    \end{minipage}

    \caption{More results of our method on material extraction for real photographs. The first column shows real photographs captured by smartphones, before cropping (as described in Sec. 3.2). The box indicates the cropped area, which is the image actually used as input for the model in the second column. Third to the ninth columns show the generated material maps and rendering under two environment maps. The last column shows the mask of the dominant material location automatically predicted by our model. We conduct tests on diverse material types in both indoor and outdoor scenes, demonstrating the generalization capability of our model. The leftmost side of each row is labeled with the text conditioning input used.} 
   \label{fig:real_material_supp_2}
\end{figure*}

\begin{figure*}[h]
    \centering
    \begin{minipage}{7.0in}
        \begin{minipage}{7.0in}
            \centering
            \hspace{0.02in}
            \begin{minipage}{0.103\linewidth}
            \centering
            \subcaption*{\small  Photo}
            \end{minipage}
            \begin{minipage}{0.093\linewidth}
            \centering
            \subcaption*{\small Input}
            \end{minipage}
            \begin{minipage}{0.093\linewidth}
            \centering
            \subcaption*{\small Albedo}
            \end{minipage}
            \begin{minipage}{0.093\linewidth}
            \centering
            \subcaption*{\small Normal}
            \end{minipage}
            \begin{minipage}{0.093\linewidth}
            \centering
            \subcaption*{\small Roughness}
            \end{minipage}
            \begin{minipage}{0.093\linewidth}
            \centering
            \subcaption*{\small Height}
            \end{minipage}
            \begin{minipage}{0.093\linewidth}
            \centering
            \subcaption*{\small Metallic}
            \end{minipage}
            \begin{minipage}{0.093\linewidth}
            \centering
            \subcaption*{\small Render 1}
            \end{minipage}
            \begin{minipage}{0.093\linewidth}
            \centering
            \subcaption*{\small Render 2}
            \end{minipage}
            \begin{minipage}{0.093\linewidth}
            \centering
            \subcaption*{\small Mask}
            \end{minipage}
        \end{minipage}
    \end{minipage}

    \begin{minipage}{7.0in}
        \begin{minipage}{7.0in}
            \centering
            \includegraphics[width=\linewidth]{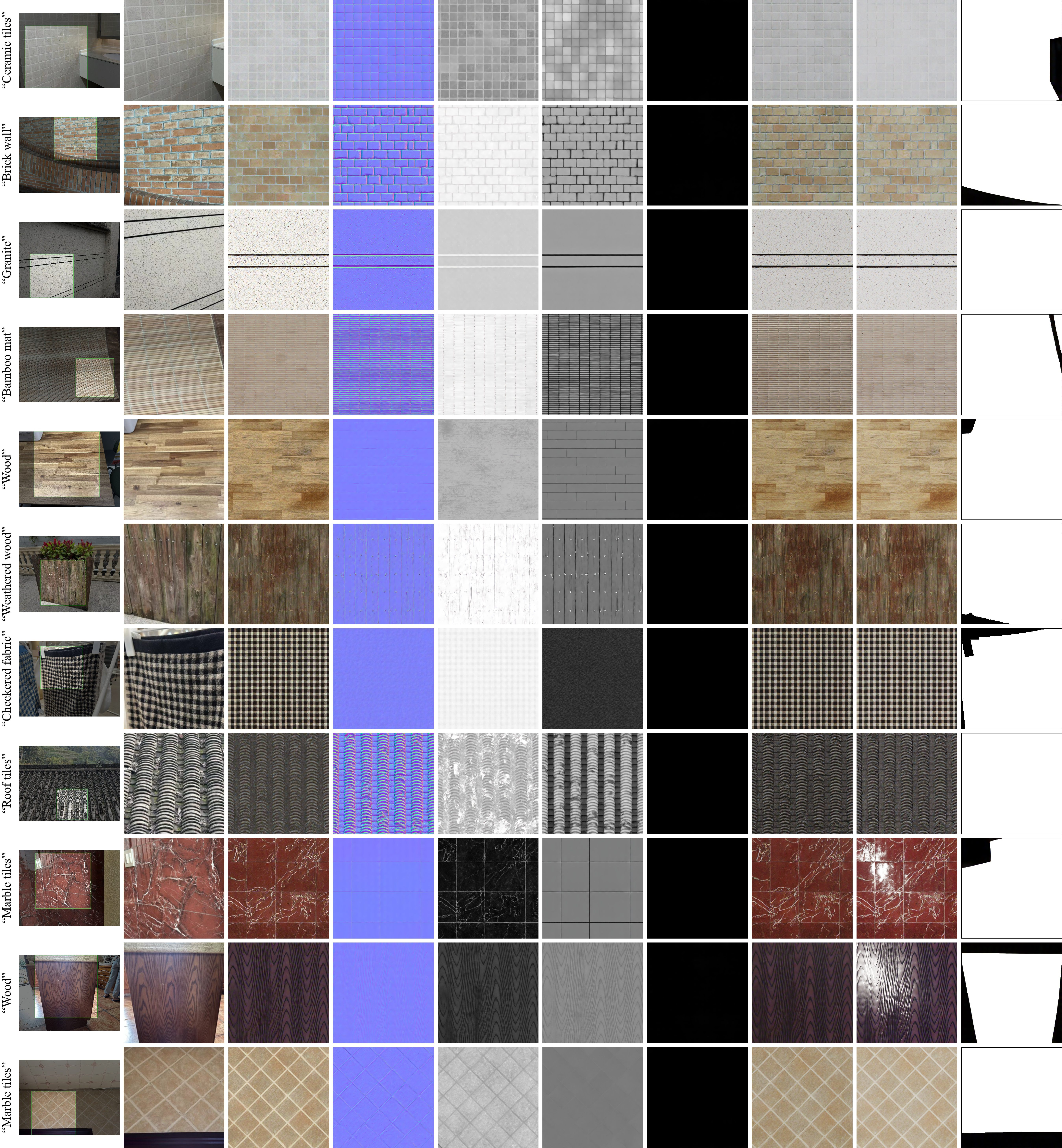}
        \end{minipage}
    \end{minipage}

    \caption{More results of our method on material extraction for real photographs. The first column shows real photographs captured by smartphones, before cropping (as described in Sec. 3.2). The box indicates the cropped area, which is the image actually used as input for the model in the second column. Third to the ninth columns show the generated material maps and rendering under two environment maps. The last column shows the mask of the dominant material location automatically predicted by our model. We conduct tests on diverse material types in both indoor and outdoor scenes, demonstrating the generalization capability of our model. The leftmost side of each row is labeled with the text conditioning input used.} 
   \label{fig:real_material_supp_3}
\end{figure*}

\begin{figure*}[h]
    \centering
    \begin{minipage}{7.0in}
        \begin{minipage}{7.0in}
            \centering
            \hspace{0.02in}
            \begin{minipage}{0.103\linewidth}
            \centering
            \subcaption*{\small  Photo}
            \end{minipage}
            \begin{minipage}{0.093\linewidth}
            \centering
            \subcaption*{\small Input}
            \end{minipage}
            \begin{minipage}{0.093\linewidth}
            \centering
            \subcaption*{\small Albedo}
            \end{minipage}
            \begin{minipage}{0.093\linewidth}
            \centering
            \subcaption*{\small Normal}
            \end{minipage}
            \begin{minipage}{0.093\linewidth}
            \centering
            \subcaption*{\small Roughness}
            \end{minipage}
            \begin{minipage}{0.093\linewidth}
            \centering
            \subcaption*{\small Height}
            \end{minipage}
            \begin{minipage}{0.093\linewidth}
            \centering
            \subcaption*{\small Metallic}
            \end{minipage}
            \begin{minipage}{0.093\linewidth}
            \centering
            \subcaption*{\small Render 1}
            \end{minipage}
            \begin{minipage}{0.093\linewidth}
            \centering
            \subcaption*{\small Render 2}
            \end{minipage}
            \begin{minipage}{0.093\linewidth}
            \centering
            \subcaption*{\small Mask}
            \end{minipage}
        \end{minipage}
    \end{minipage}

    \begin{minipage}{7.0in}
        \begin{minipage}{7.0in}
            \centering
            \includegraphics[width=\linewidth]{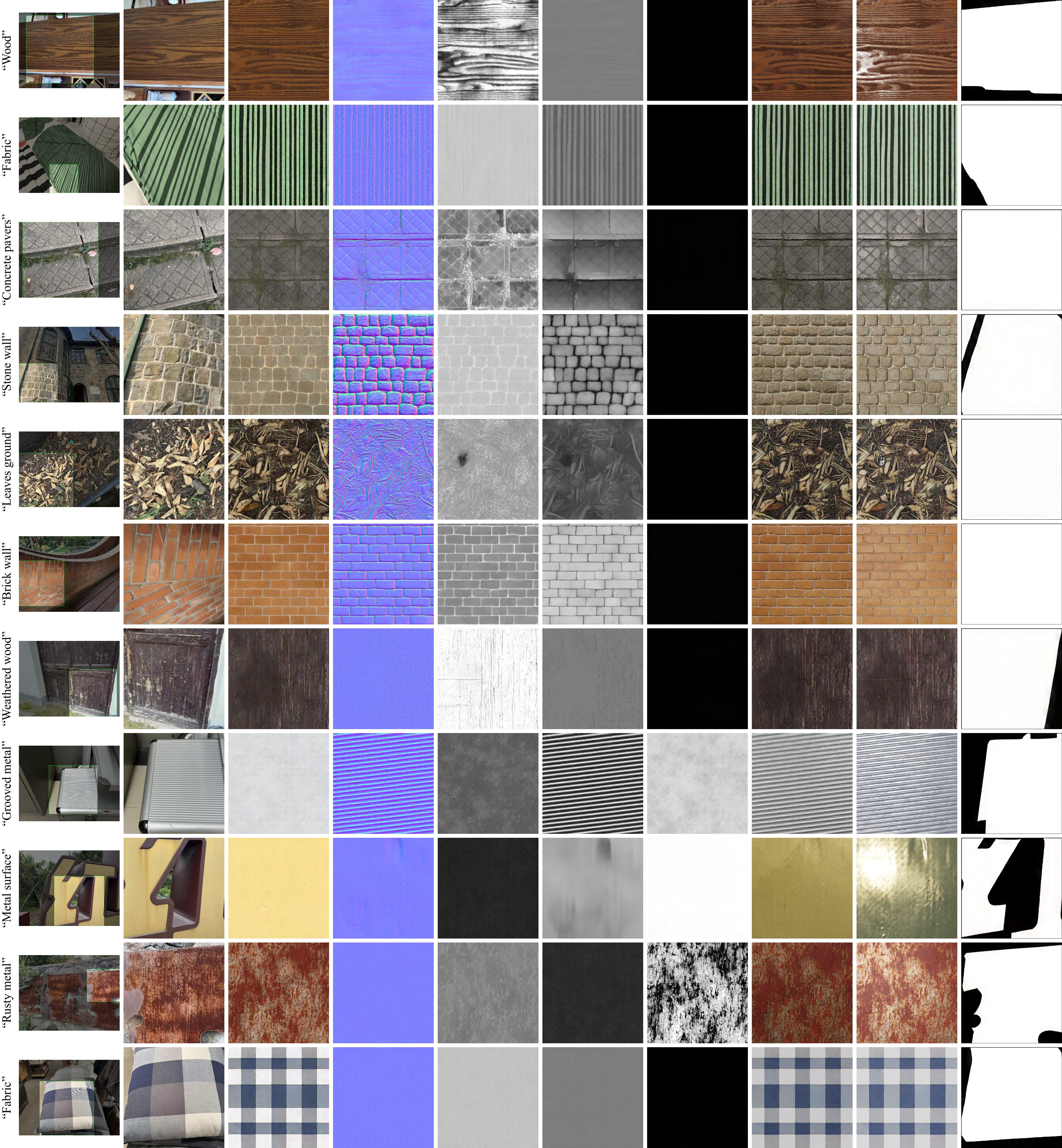}
        \end{minipage}
    \end{minipage}

    \caption{More results of our method on material extraction for real photographs. The first column shows real photographs captured by smartphones, before cropping (as described in Sec. 3.2). The box indicates the cropped area, which is the image actually used as input for the model in the second column. Third to the ninth columns show the generated material maps and rendering under two environment maps. The last column shows the mask of the dominant material location automatically predicted by our model. We conduct tests on diverse material types in both indoor and outdoor scenes, demonstrating the generalization capability of our model. The leftmost side of each row is labeled with the text conditioning input used.} 
   \label{fig:real_material_supp_4}
\end{figure*}

\begin{figure*}[h]
    \centering
    \begin{minipage}{7.0in}
        \begin{minipage}{7.0in}
            \centering
            \hspace{0.02in}
            \begin{minipage}{0.103\linewidth}
            \centering
            \subcaption*{\small  Photo}
            \end{minipage}
            \begin{minipage}{0.093\linewidth}
            \centering
            \subcaption*{\small Input}
            \end{minipage}
            \begin{minipage}{0.093\linewidth}
            \centering
            \subcaption*{\small Albedo}
            \end{minipage}
            \begin{minipage}{0.093\linewidth}
            \centering
            \subcaption*{\small Normal}
            \end{minipage}
            \begin{minipage}{0.093\linewidth}
            \centering
            \subcaption*{\small Roughness}
            \end{minipage}
            \begin{minipage}{0.093\linewidth}
            \centering
            \subcaption*{\small Height}
            \end{minipage}
            \begin{minipage}{0.093\linewidth}
            \centering
            \subcaption*{\small Metallic}
            \end{minipage}
            \begin{minipage}{0.093\linewidth}
            \centering
            \subcaption*{\small Render 1}
            \end{minipage}
            \begin{minipage}{0.093\linewidth}
            \centering
            \subcaption*{\small Render 2}
            \end{minipage}
            \begin{minipage}{0.093\linewidth}
            \centering
            \subcaption*{\small Mask}
            \end{minipage}
        \end{minipage}
    \end{minipage}

    \begin{minipage}{7.0in}
        \begin{minipage}{7.0in}
            \centering
            \includegraphics[width=\linewidth]{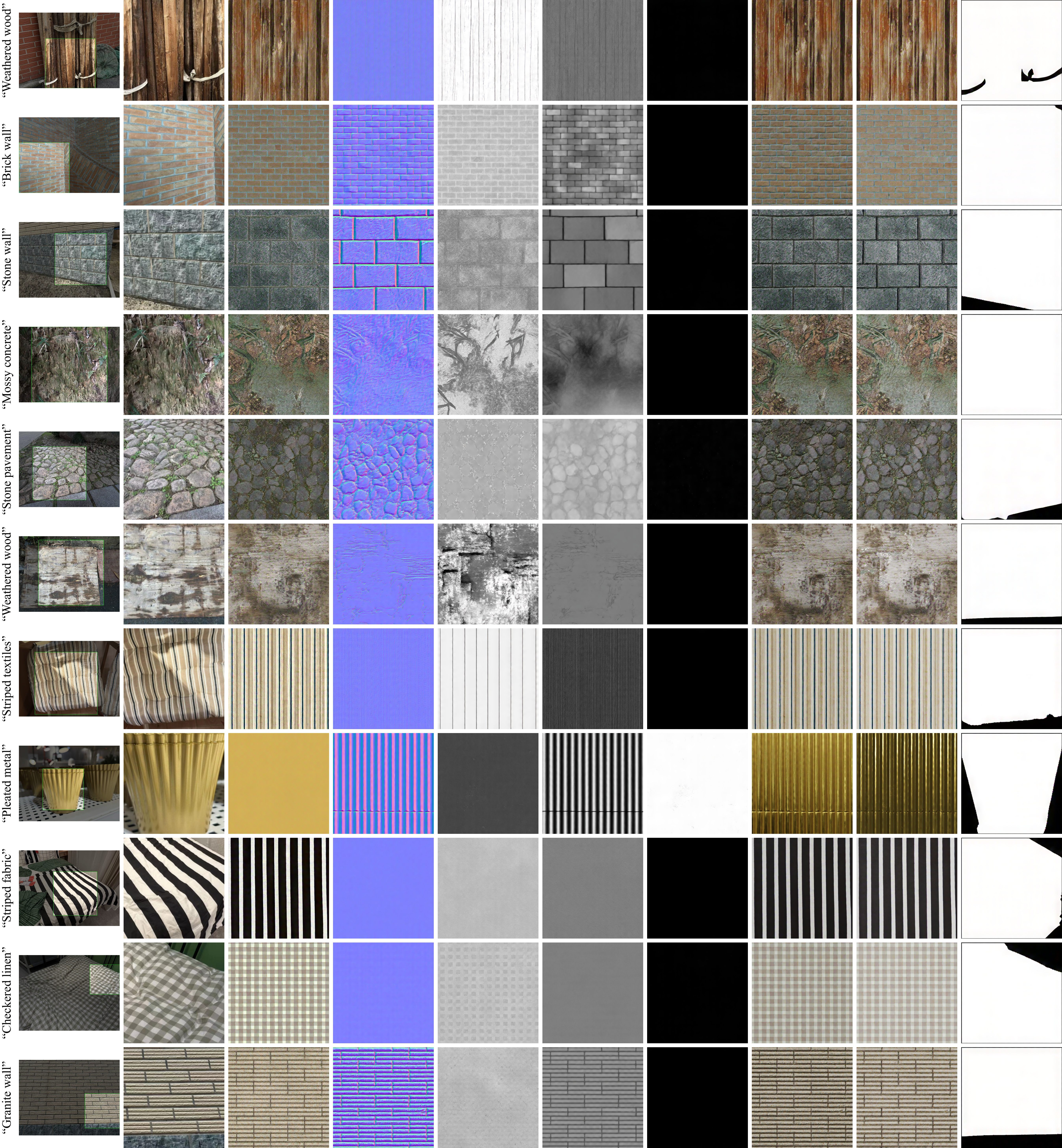}
        \end{minipage}
    \end{minipage}

    \caption{More results of our method on material extraction for real photographs. The first column shows real photographs captured by smartphones, before cropping (as described in Sec. 3.2). The box indicates the cropped area, which is the image actually used as input for the model in the second column. Third to the ninth columns show the generated material maps and rendering under two environment maps. The last column shows the mask of the dominant material location automatically predicted by our model. We conduct tests on diverse material types in both indoor and outdoor scenes, demonstrating the generalization capability of our model. The leftmost side of each row is labeled with the text conditioning input used.} 
   \label{fig:real_material_supp_5}
\end{figure*}

\begin{figure*}[h]
    \centering
    \begin{minipage}{7.0in}
        \begin{minipage}{7.0in}
            \centering
            \hspace{0.02in}
            \begin{minipage}{0.103\linewidth}
            \centering
            \subcaption*{\small  Photo}
            \end{minipage}
            \begin{minipage}{0.093\linewidth}
            \centering
            \subcaption*{\small Input}
            \end{minipage}
            \begin{minipage}{0.093\linewidth}
            \centering
            \subcaption*{\small Albedo}
            \end{minipage}
            \begin{minipage}{0.093\linewidth}
            \centering
            \subcaption*{\small Normal}
            \end{minipage}
            \begin{minipage}{0.093\linewidth}
            \centering
            \subcaption*{\small Roughness}
            \end{minipage}
            \begin{minipage}{0.093\linewidth}
            \centering
            \subcaption*{\small Height}
            \end{minipage}
            \begin{minipage}{0.093\linewidth}
            \centering
            \subcaption*{\small Metallic}
            \end{minipage}
            \begin{minipage}{0.093\linewidth}
            \centering
            \subcaption*{\small Render 1}
            \end{minipage}
            \begin{minipage}{0.093\linewidth}
            \centering
            \subcaption*{\small Render 2}
            \end{minipage}
            \begin{minipage}{0.093\linewidth}
            \centering
            \subcaption*{\small Mask}
            \end{minipage}
        \end{minipage}
    \end{minipage}

    \begin{minipage}{7.0in}
        \begin{minipage}{7.0in}
            \centering
            \includegraphics[width=\linewidth]{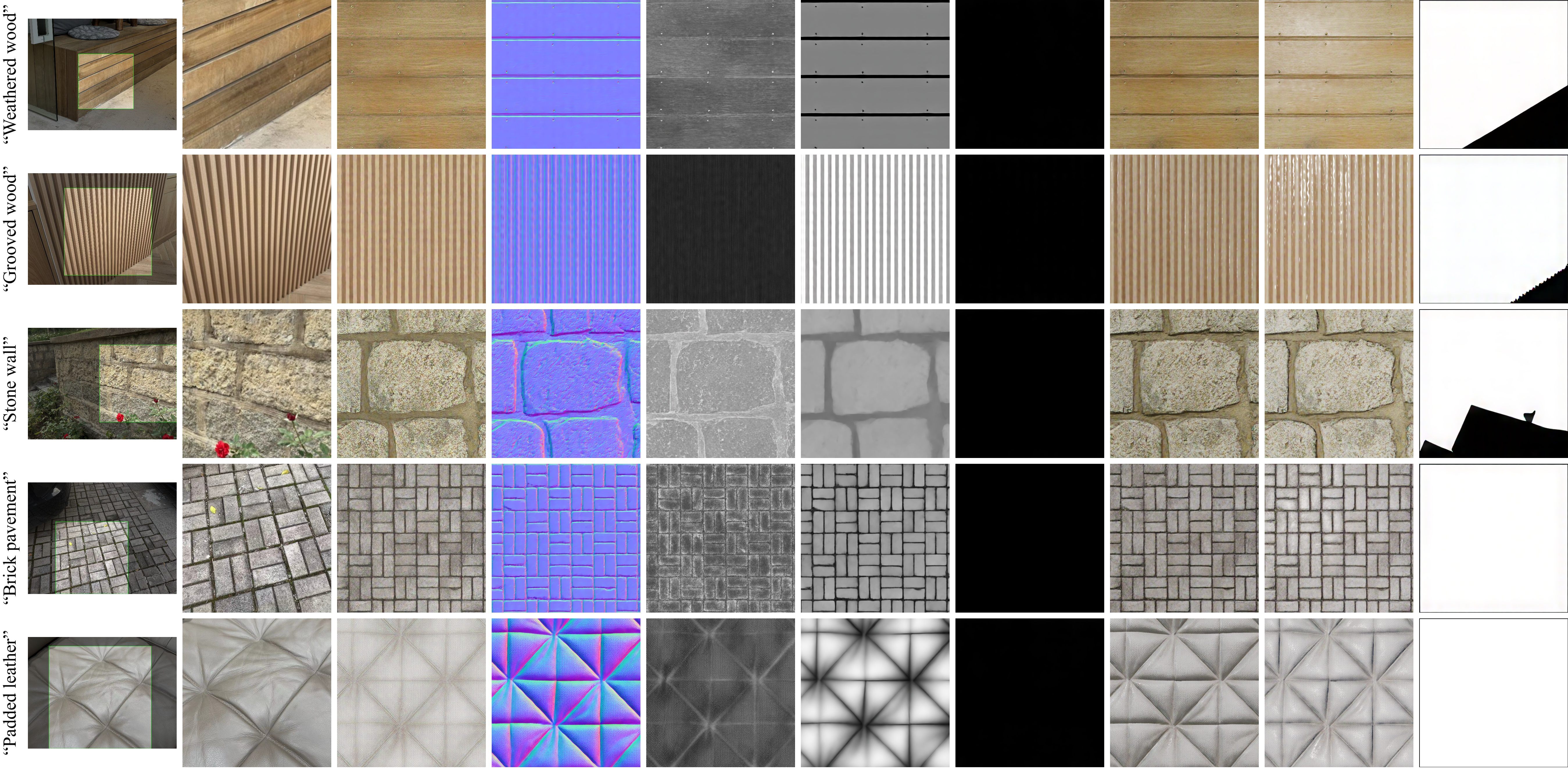}
        \end{minipage}
    \end{minipage}

    \caption{More results of our method on material extraction for real photographs. The first column shows real photographs captured by smartphones, before cropping (as described in Sec. 3.2). The box indicates the cropped area, which is the image actually used as input for the model in the second column. Third to the ninth columns show the generated material maps and rendering under two environment maps. The last column shows the mask of the dominant material location automatically predicted by our model. We conduct tests on diverse material types in both indoor and outdoor scenes, demonstrating the generalization capability of our model. The leftmost side of each row is labeled with the text conditioning input used.} 
   \label{fig:real_material_supp_6}
\end{figure*}

In Fig.~\ref{fig:real_material_supp_1}, Fig.~\ref{fig:real_material_supp_2}, Fig.~\ref{fig:real_material_supp_3}, Fig.~\ref{fig:real_material_supp_4},  Fig.~\ref{fig:real_material_supp_5} and Fig.~\ref{fig:real_material_supp_6}, we show additional examples of material extraction using  our model, along with the uncropped original images from our real photographs evaluation dataset (Sec. 3.2). % ~\ref{sec:dataset}). 
These examples include various indoor and outdoor materials captured under complex real-world lighting conditions. Our model generalizes well to real photos, producing renderings that are visually similar to the photographs and providing accurate masks, demonstrating our model's generalization capabilities.

\bibliographystyle{ACM-Reference-Format}
\bibliography{main}


\begin{thebibliography}{60}

%%% the Web address.

\ifx \showCODEN    \undefined \def \showCODEN     #1{\unskip}     \fi
\ifx \showDOI      \undefined \def \showDOI       #1{#1}\fi
\ifx \showISBNx    \undefined \def \showISBNx     #1{\unskip}     \fi
\ifx \showISBNxiii \undefined \def \showISBNxiii  #1{\unskip}     \fi
\ifx \showISSN     \undefined \def \showISSN      #1{\unskip}     \fi
\ifx \showLCCN     \undefined \def \showLCCN      #1{\unskip}     \fi
\ifx \shownote     \undefined \def \shownote      #1{#1}          \fi
\ifx \showarticletitle \undefined \def \showarticletitle #1{#1}   \fi
\ifx \showURL      \undefined \def \showURL       {\relax}        \fi
% invisible to TeX
\providecommand\bibfield[2]{#2}
\providecommand\bibinfo[2]{#2}
\providecommand\natexlab[1]{#1}
\providecommand\showeprint[2][]{arXiv:#2}

\bibitem[Adobe(2024)]%
        {SubstanceAsset}
\bibfield{author}{\bibinfo{person}{Adobe}.} \bibinfo{year}{2024}\natexlab{}.
\newblock \bibinfo{title}{Adobe Substance3D Asset}.
\newblock
\newblock
\newblock
\shownote{https://substance3d.adobe.com/assets}.


\bibitem[Blattmann et~al\mbox{.}(2023a)]%
        {blattmann2023stable}
\bibfield{author}{\bibinfo{person}{Andreas Blattmann}, \bibinfo{person}{Tim Dockhorn}, \bibinfo{person}{Sumith Kulal}, \bibinfo{person}{Daniel Mendelevitch}, \bibinfo{person}{Maciej Kilian}, \bibinfo{person}{Dominik Lorenz}, \bibinfo{person}{Yam Levi}, \bibinfo{person}{Zion English}, \bibinfo{person}{Vikram Voleti}, \bibinfo{person}{Adam Letts}, {et~al\mbox{.}}} \bibinfo{year}{2023}\natexlab{a}.
\newblock \showarticletitle{Stable video diffusion: Scaling latent video diffusion models to large datasets}.
\newblock \bibinfo{journal}{\emph{arXiv preprint arXiv:2311.15127}} (\bibinfo{year}{2023}).
\newblock


\bibitem[Blattmann et~al\mbox{.}(2023b)]%
        {blattmann2023align}
\bibfield{author}{\bibinfo{person}{Andreas Blattmann}, \bibinfo{person}{Robin Rombach}, \bibinfo{person}{Huan Ling}, \bibinfo{person}{Tim Dockhorn}, \bibinfo{person}{Seung~Wook Kim}, \bibinfo{person}{Sanja Fidler}, {and} \bibinfo{person}{Karsten Kreis}.} \bibinfo{year}{2023}\natexlab{b}.
\newblock \showarticletitle{Align your latents: High-resolution video synthesis with latent diffusion models}. In \bibinfo{booktitle}{\emph{Proceedings of the IEEE/CVF Conference on Computer Vision and Pattern Recognition}}. \bibinfo{pages}{22563--22575}.
\newblock


\bibitem[{Blender Community}(2018)]%
        {blender2018}
\bibfield{author}{\bibinfo{person}{{Blender Community}}.} \bibinfo{year}{2018}\natexlab{}.
\newblock \bibinfo{title}{Blender - A 3D Modelling and Rendering Package}.
\newblock \bibinfo{howpublished}{\url{http://www.blender.org}}.
\newblock
\newblock
\shownote{Accessed: 2024-11-05}.


\bibitem[Bookstein(1989)]%
        {bookstein1989principal}
\bibfield{author}{\bibinfo{person}{Fred~L. Bookstein}.} \bibinfo{year}{1989}\natexlab{}.
\newblock \showarticletitle{Principal warps: Thin-plate splines and the decomposition of deformations}.
\newblock \bibinfo{journal}{\emph{IEEE Transactions on pattern analysis and machine intelligence}} \bibinfo{volume}{11}, \bibinfo{number}{6} (\bibinfo{year}{1989}), \bibinfo{pages}{567--585}.
\newblock


\bibitem[Brooks et~al\mbox{.}(2024)]%
        {videoworldsimulators2024}
\bibfield{author}{\bibinfo{person}{Tim Brooks}, \bibinfo{person}{Bill Peebles}, \bibinfo{person}{Connor Holmes}, \bibinfo{person}{Will DePue}, \bibinfo{person}{Yufei Guo}, \bibinfo{person}{Li Jing}, \bibinfo{person}{David Schnurr}, \bibinfo{person}{Joe Taylor}, \bibinfo{person}{Troy Luhman}, \bibinfo{person}{Eric Luhman}, \bibinfo{person}{Clarence Ng}, \bibinfo{person}{Ricky Wang}, {and} \bibinfo{person}{Aditya Ramesh}.} \bibinfo{year}{2024}\natexlab{}.
\newblock \showarticletitle{Video generation models as world simulators}.
\newblock  (\bibinfo{year}{2024}).
\newblock
\urldef\tempurl%
\url{https://openai.com/research/video-generation-models-as-world-simulators}
\showURL{%
\tempurl}


\bibitem[Cohen and Greenberg(1985)]%
        {cohen1985hemi}
\bibfield{author}{\bibinfo{person}{Michael~F Cohen} {and} \bibinfo{person}{Donald~P Greenberg}.} \bibinfo{year}{1985}\natexlab{}.
\newblock \showarticletitle{The hemi-cube: A radiosity solution for complex environments}.
\newblock \bibinfo{journal}{\emph{ACM Siggraph Computer Graphics}} \bibinfo{volume}{19}, \bibinfo{number}{3} (\bibinfo{year}{1985}), \bibinfo{pages}{31--40}.
\newblock


\bibitem[Deschaintre et~al\mbox{.}(2018)]%
        {Deschaintre2018}
\bibfield{author}{\bibinfo{person}{Valentin Deschaintre}, \bibinfo{person}{Miika Aittala}, \bibinfo{person}{Fredo Durand}, \bibinfo{person}{George Drettakis}, {and} \bibinfo{person}{Adrien Bousseau}.} \bibinfo{year}{2018}\natexlab{}.
\newblock \showarticletitle{Single-image SVBRDF Capture with a Rendering-aware Deep Network}.
\newblock \bibinfo{journal}{\emph{ACM Trans. Graph.}} \bibinfo{volume}{37}, \bibinfo{number}{4} (\bibinfo{year}{2018}), \bibinfo{pages}{128:1--128:15}.
\newblock


\bibitem[Deschaintre et~al\mbox{.}(2019)]%
        {deschaintre2019flexible}
\bibfield{author}{\bibinfo{person}{Valentin Deschaintre}, \bibinfo{person}{Miika Aittala}, \bibinfo{person}{Fr{\'e}do Durand}, \bibinfo{person}{George Drettakis}, {and} \bibinfo{person}{Adrien Bousseau}.} \bibinfo{year}{2019}\natexlab{}.
\newblock \showarticletitle{Flexible svbrdf capture with a multi-image deep network}. In \bibinfo{booktitle}{\emph{Computer graphics forum}}, Vol.~\bibinfo{volume}{38}. Wiley Online Library, \bibinfo{pages}{1--13}.
\newblock


\bibitem[Deschaintre et~al\mbox{.}(2020)]%
        {DDB20}
\bibfield{author}{\bibinfo{person}{Valentin Deschaintre}, \bibinfo{person}{George Drettakis}, {and} \bibinfo{person}{Adrien Bousseau}.} \bibinfo{year}{2020}\natexlab{}.
\newblock \showarticletitle{Guided Fine-Tuning for Large-Scale Material Transfer}.
\newblock \bibinfo{journal}{\emph{Computer Graphics Forum (Proceedings of the Eurographics Symposium on Rendering)}} \bibinfo{volume}{39}, \bibinfo{number}{4} (\bibinfo{year}{2020}).
\newblock


\bibitem[Dhariwal and Nichol(2021)]%
        {dhariwal2021diffusion}
\bibfield{author}{\bibinfo{person}{Prafulla Dhariwal} {and} \bibinfo{person}{Alexander Nichol}.} \bibinfo{year}{2021}\natexlab{}.
\newblock \showarticletitle{Diffusion models beat gans on image synthesis}.
\newblock \bibinfo{journal}{\emph{Advances in neural information processing systems}}  \bibinfo{volume}{34} (\bibinfo{year}{2021}), \bibinfo{pages}{8780--8794}.
\newblock


\bibitem[Esser et~al\mbox{.}(2024)]%
        {esser2024scaling}
\bibfield{author}{\bibinfo{person}{Patrick Esser}, \bibinfo{person}{Sumith Kulal}, \bibinfo{person}{Andreas Blattmann}, \bibinfo{person}{Rahim Entezari}, \bibinfo{person}{Jonas M{\"u}ller}, \bibinfo{person}{Harry Saini}, \bibinfo{person}{Yam Levi}, \bibinfo{person}{Dominik Lorenz}, \bibinfo{person}{Axel Sauer}, \bibinfo{person}{Frederic Boesel}, {et~al\mbox{.}}} \bibinfo{year}{2024}\natexlab{}.
\newblock \showarticletitle{Scaling rectified flow transformers for high-resolution image synthesis}. In \bibinfo{booktitle}{\emph{Forty-first International Conference on Machine Learning}}.
\newblock


\bibitem[Esser et~al\mbox{.}(2021)]%
        {esser2021taming}
\bibfield{author}{\bibinfo{person}{Patrick Esser}, \bibinfo{person}{Robin Rombach}, {and} \bibinfo{person}{Bjorn Ommer}.} \bibinfo{year}{2021}\natexlab{}.
\newblock \showarticletitle{Taming transformers for high-resolution image synthesis}. In \bibinfo{booktitle}{\emph{Proceedings of the IEEE/CVF conference on computer vision and pattern recognition}}. \bibinfo{pages}{12873--12883}.
\newblock


\bibitem[{Evermotion}(2021)]%
        {evermotion2021archinterior}
\bibfield{author}{\bibinfo{person}{{Evermotion}}.} \bibinfo{year}{2021}\natexlab{}.
\newblock \bibinfo{title}{Archinteriors Collection}.
\newblock
\newblock
\urldef\tempurl%
\url{https://evermotion.org/shop/cat/397/archinteriors}
\showURL{%
\tempurl}
\newblock
\shownote{Accessed: 2024-11-05}.


\bibitem[Guarnera et~al\mbox{.}(2016)]%
        {Guarnera16}
\bibfield{author}{\bibinfo{person}{D. Guarnera}, \bibinfo{person}{G.C. Guarnera}, \bibinfo{person}{A. Ghosh}, \bibinfo{person}{C. Denk}, {and} \bibinfo{person}{M. Glencross}.} \bibinfo{year}{2016}\natexlab{}.
\newblock \showarticletitle{BRDF Representation and Acquisition}.
\newblock \bibinfo{journal}{\emph{Computer Graphics Forum}} \bibinfo{volume}{35}, \bibinfo{number}{2} (\bibinfo{year}{2016}), \bibinfo{pages}{625--650}.
\newblock
\urldef\tempurl%
\url{https://doi.org/10.1111/cgf.12867}
\showDOI{\tempurl}


\bibitem[Guerrero et~al\mbox{.}(2022)]%
        {guerrero2022matformer}
\bibfield{author}{\bibinfo{person}{Paul Guerrero}, \bibinfo{person}{Milos Hasan}, \bibinfo{person}{Kalyan Sunkavalli}, \bibinfo{person}{Radomir Mech}, \bibinfo{person}{Tamy Boubekeur}, {and} \bibinfo{person}{Niloy Mitra}.} \bibinfo{year}{2022}\natexlab{}.
\newblock \showarticletitle{MatFormer: A Generative Model for Procedural Materials}.
\newblock \bibinfo{journal}{\emph{ACM Trans. Graph.}} \bibinfo{volume}{41}, \bibinfo{number}{4}, Article \bibinfo{articleno}{46} (\bibinfo{year}{2022}).
\newblock
\urldef\tempurl%
\url{https://doi.org/10.1145/3528223.3530173}
\showDOI{\tempurl}


\bibitem[Guerrero-Viu et~al\mbox{.}(2024)]%
        {guerrero2024texsliders}
\bibfield{author}{\bibinfo{person}{Julia Guerrero-Viu}, \bibinfo{person}{Milos Hasan}, \bibinfo{person}{Arthur Roullier}, \bibinfo{person}{Midhun Harikumar}, \bibinfo{person}{Yiwei Hu}, \bibinfo{person}{Paul Guerrero}, \bibinfo{person}{Diego Gutierrez}, \bibinfo{person}{Belen Masia}, {and} \bibinfo{person}{Valentin Deschaintre}.} \bibinfo{year}{2024}\natexlab{}.
\newblock \showarticletitle{Texsliders: Diffusion-based texture editing in clip space}. In \bibinfo{booktitle}{\emph{ACM SIGGRAPH 2024 Conference Papers}}. \bibinfo{pages}{1--11}.
\newblock


\bibitem[Guo et~al\mbox{.}(2021)]%
        {Guo2021}
\bibfield{author}{\bibinfo{person}{Jie Guo}, \bibinfo{person}{Shuichang Lai}, \bibinfo{person}{Chengzhi Tao}, \bibinfo{person}{Yuelong Cai}, \bibinfo{person}{Lei Wang}, \bibinfo{person}{Yanwen Guo}, {and} \bibinfo{person}{Ling-Qi Yan}.} \bibinfo{year}{2021}\natexlab{}.
\newblock \showarticletitle{Highlight-Aware Two-Stream Network for Single-Image SVBRDF Acquisition}.
\newblock \bibinfo{journal}{\emph{ACM Trans. Graph.}} \bibinfo{volume}{40}, \bibinfo{number}{4}, Article \bibinfo{articleno}{123} (\bibinfo{date}{jul} \bibinfo{year}{2021}), \bibinfo{numpages}{14}~pages.
\newblock
\showISSN{0730-0301}
\urldef\tempurl%
\url{https://doi.org/10.1145/3450626.3459854}
\showDOI{\tempurl}


\bibitem[Guo et~al\mbox{.}(2020)]%
        {Guo2020}
\bibfield{author}{\bibinfo{person}{Yu Guo}, \bibinfo{person}{Cameron Smith}, \bibinfo{person}{Milo\v{s} Ha\v{s}an}, \bibinfo{person}{Kalyan Sunkavalli}, {and} \bibinfo{person}{Shuang Zhao}.} \bibinfo{year}{2020}\natexlab{}.
\newblock \showarticletitle{MaterialGAN: Reflectance Capture using a Generative SVBRDF Model}.
\newblock \bibinfo{journal}{\emph{ACM Trans. Graph.}} \bibinfo{volume}{39}, \bibinfo{number}{6} (\bibinfo{year}{2020}), \bibinfo{pages}{254:1--254:13}.
\newblock


\bibitem[Hao et~al\mbox{.}(2023)]%
        {hao2023diffusion}
\bibfield{author}{\bibinfo{person}{Guoqing Hao}, \bibinfo{person}{Satoshi Iizuka}, \bibinfo{person}{Kensho Hara}, \bibinfo{person}{Edgar Simo-Serra}, \bibinfo{person}{Hirokatsu Kataoka}, {and} \bibinfo{person}{Kazuhiro Fukui}.} \bibinfo{year}{2023}\natexlab{}.
\newblock \showarticletitle{Diffusion-based Holistic Texture Rectification and Synthesis}. In \bibinfo{booktitle}{\emph{SIGGRAPH Asia 2023 Conference Papers}}. \bibinfo{pages}{1--11}.
\newblock


\bibitem[Hartley and Zisserman(2003)]%
        {hartley2003multiple}
\bibfield{author}{\bibinfo{person}{Richard Hartley} {and} \bibinfo{person}{Andrew Zisserman}.} \bibinfo{year}{2003}\natexlab{}.
\newblock \bibinfo{booktitle}{\emph{Multiple view geometry in computer vision}}.
\newblock \bibinfo{publisher}{Cambridge university press}.
\newblock


\bibitem[Ho et~al\mbox{.}(2020)]%
        {ho2020denoising}
\bibfield{author}{\bibinfo{person}{Jonathan Ho}, \bibinfo{person}{Ajay Jain}, {and} \bibinfo{person}{Pieter Abbeel}.} \bibinfo{year}{2020}\natexlab{}.
\newblock \showarticletitle{Denoising diffusion probabilistic models}.
\newblock \bibinfo{journal}{\emph{Advances in neural information processing systems}}  \bibinfo{volume}{33} (\bibinfo{year}{2020}), \bibinfo{pages}{6840--6851}.
\newblock


\bibitem[Hu et~al\mbox{.}(2021)]%
        {hu2021lora}
\bibfield{author}{\bibinfo{person}{Edward~J Hu}, \bibinfo{person}{Yelong Shen}, \bibinfo{person}{Phillip Wallis}, \bibinfo{person}{Zeyuan Allen-Zhu}, \bibinfo{person}{Yuanzhi Li}, \bibinfo{person}{Shean Wang}, \bibinfo{person}{Lu Wang}, {and} \bibinfo{person}{Weizhu Chen}.} \bibinfo{year}{2021}\natexlab{}.
\newblock \showarticletitle{Lora: Low-rank adaptation of large language models}.
\newblock \bibinfo{journal}{\emph{arXiv preprint arXiv:2106.09685}} (\bibinfo{year}{2021}).
\newblock


\bibitem[Hu et~al\mbox{.}(2023)]%
        {hu2023generating}
\bibfield{author}{\bibinfo{person}{Yiwei Hu}, \bibinfo{person}{Paul Guerrero}, \bibinfo{person}{Milos Hasan}, \bibinfo{person}{Holly Rushmeier}, {and} \bibinfo{person}{Valentin Deschaintre}.} \bibinfo{year}{2023}\natexlab{}.
\newblock \showarticletitle{Generating Procedural Materials from Text or Image Prompts}. In \bibinfo{booktitle}{\emph{ACM SIGGRAPH 2023 Conference Proceedings}}. \bibinfo{pages}{1--11}.
\newblock


\bibitem[Kingma and Welling(2022)]%
        {kingma2022autoencoding}
\bibfield{author}{\bibinfo{person}{Diederik~P Kingma} {and} \bibinfo{person}{Max Welling}.} \bibinfo{year}{2022}\natexlab{}.
\newblock \bibinfo{title}{Auto-Encoding Variational Bayes}.
\newblock
\newblock
\showeprint[arxiv]{1312.6114}~[stat.ML]
\urldef\tempurl%
\url{https://arxiv.org/abs/1312.6114}
\showURL{%
\tempurl}


\bibitem[Liu et~al\mbox{.}(2023)]%
        {liu2023hyperhuman}
\bibfield{author}{\bibinfo{person}{Xian Liu}, \bibinfo{person}{Jian Ren}, \bibinfo{person}{Aliaksandr Siarohin}, \bibinfo{person}{Ivan Skorokhodov}, \bibinfo{person}{Yanyu Li}, \bibinfo{person}{Dahua Lin}, \bibinfo{person}{Xihui Liu}, \bibinfo{person}{Ziwei Liu}, {and} \bibinfo{person}{Sergey Tulyakov}.} \bibinfo{year}{2023}\natexlab{}.
\newblock \showarticletitle{Hyperhuman: Hyper-realistic human generation with latent structural diffusion}.
\newblock \bibinfo{journal}{\emph{arXiv preprint arXiv:2310.08579}} (\bibinfo{year}{2023}).
\newblock


\bibitem[Lopes et~al\mbox{.}(2024)]%
        {lopes2024material}
\bibfield{author}{\bibinfo{person}{Ivan Lopes}, \bibinfo{person}{Fabio Pizzati}, {and} \bibinfo{person}{Raoul de Charette}.} \bibinfo{year}{2024}\natexlab{}.
\newblock \showarticletitle{Material palette: Extraction of materials from a single image}. In \bibinfo{booktitle}{\emph{Proceedings of the IEEE/CVF Conference on Computer Vision and Pattern Recognition}}. \bibinfo{pages}{4379--4388}.
\newblock


\bibitem[Lu et~al\mbox{.}(2009)]%
        {Lu2009Dominant}
\bibfield{author}{\bibinfo{person}{Jianye Lu}, \bibinfo{person}{Julie Dorsey}, {and} \bibinfo{person}{Holly Rushmeier}.} \bibinfo{year}{2009}\natexlab{}.
\newblock \showarticletitle{Dominant Texture and Diffusion Distance Manifolds}.
\newblock \bibinfo{journal}{\emph{Computer Graphics Forum}} \bibinfo{volume}{28}, \bibinfo{number}{2} (\bibinfo{year}{2009}), \bibinfo{pages}{667--676}.
\newblock
\urldef\tempurl%
\url{https://doi.org/10.1111/j.1467-8659.2009.01407.x}
\showDOI{\tempurl}
\showeprint{https://onlinelibrary.wiley.com/doi/pdf/10.1111/j.1467-8659.2009.01407.x}


\bibitem[Ma et~al\mbox{.}(2023)]%
        {ma2023opensvbrdf}
\bibfield{author}{\bibinfo{person}{Xiaohe Ma}, \bibinfo{person}{Xianmin Xu}, \bibinfo{person}{Leyao Zhang}, \bibinfo{person}{Kun Zhou}, {and} \bibinfo{person}{Hongzhi Wu}.} \bibinfo{year}{2023}\natexlab{}.
\newblock \showarticletitle{OpenSVBRDF: a database of measured spatially-varying reflectance}.
\newblock \bibinfo{journal}{\emph{ACM Transactions on Graphics (TOG)}} \bibinfo{volume}{42}, \bibinfo{number}{6} (\bibinfo{year}{2023}), \bibinfo{pages}{1--14}.
\newblock


\bibitem[Martin et~al\mbox{.}(2022)]%
        {Martin2022}
\bibfield{author}{\bibinfo{person}{Rosalie Martin}, \bibinfo{person}{Arthur Roullier}, \bibinfo{person}{Romain Rouffet}, \bibinfo{person}{Adrien Kaiser}, {and} \bibinfo{person}{Tamy Boubekeur}.} \bibinfo{year}{2022}\natexlab{}.
\newblock \showarticletitle{MaterIA: Single Image High-Resolution Material Capture in the Wild}.
\newblock \bibinfo{journal}{\emph{Computer Graphics Forum}} \bibinfo{volume}{41}, \bibinfo{number}{2} (\bibinfo{year}{2022}), \bibinfo{pages}{163--177}.
\newblock
\urldef\tempurl%
\url{https://doi.org/10.1111/cgf.14466}
\showDOI{\tempurl}
\showeprint{https://onlinelibrary.wiley.com/doi/pdf/10.1111/cgf.14466}


\bibitem[McAuley et~al\mbox{.}(2012)]%
        {disneybrdf}
\bibfield{author}{\bibinfo{person}{Stephen McAuley}, \bibinfo{person}{Stephen Hill}, \bibinfo{person}{Naty Hoffman}, \bibinfo{person}{Yoshiharu Gotanda}, \bibinfo{person}{Brian Smits}, \bibinfo{person}{Brent Burley}, {and} \bibinfo{person}{Adam Martinez}.} \bibinfo{year}{2012}\natexlab{}.
\newblock \showarticletitle{Practical physically-based shading in film and game production}. In \bibinfo{booktitle}{\emph{ACM SIGGRAPH 2012 Courses}} (Los Angeles, California) \emph{(\bibinfo{series}{SIGGRAPH '12})}. \bibinfo{publisher}{Association for Computing Machinery}, \bibinfo{address}{New York, NY, USA}, Article \bibinfo{articleno}{10}, \bibinfo{numpages}{7}~pages.
\newblock
\showISBNx{9781450316781}
\urldef\tempurl%
\url{https://doi.org/10.1145/2343483.2343493}
\showDOI{\tempurl}


\bibitem[Meng et~al\mbox{.}(2021)]%
        {meng2021sdedit}
\bibfield{author}{\bibinfo{person}{Chenlin Meng}, \bibinfo{person}{Yutong He}, \bibinfo{person}{Yang Song}, \bibinfo{person}{Jiaming Song}, \bibinfo{person}{Jiajun Wu}, \bibinfo{person}{Jun-Yan Zhu}, {and} \bibinfo{person}{Stefano Ermon}.} \bibinfo{year}{2021}\natexlab{}.
\newblock \showarticletitle{Sdedit: Guided image synthesis and editing with stochastic differential equations}.
\newblock \bibinfo{journal}{\emph{arXiv preprint arXiv:2108.01073}} (\bibinfo{year}{2021}).
\newblock


\bibitem[Nichol et~al\mbox{.}(2021)]%
        {nichol2021glide}
\bibfield{author}{\bibinfo{person}{Alex Nichol}, \bibinfo{person}{Prafulla Dhariwal}, \bibinfo{person}{Aditya Ramesh}, \bibinfo{person}{Pranav Shyam}, \bibinfo{person}{Pamela Mishkin}, \bibinfo{person}{Bob McGrew}, \bibinfo{person}{Ilya Sutskever}, {and} \bibinfo{person}{Mark Chen}.} \bibinfo{year}{2021}\natexlab{}.
\newblock \showarticletitle{Glide: Towards photorealistic image generation and editing with text-guided diffusion models}.
\newblock \bibinfo{journal}{\emph{arXiv preprint arXiv:2112.10741}} (\bibinfo{year}{2021}).
\newblock


\bibitem[Niu(2023)]%
        {waifu2x-ncnn-vulkan}
\bibfield{author}{\bibinfo{person}{Hui Niu}.} \bibinfo{year}{2023}\natexlab{}.
\newblock \bibinfo{title}{waifu2x-ncnn-vulkan}.
\newblock
\newblock
\urldef\tempurl%
\url{https://github.com/nihui/waifu2x-ncnn-vulkan}
\showURL{%
\tempurl}
\newblock
\shownote{GitHub repository}.


\bibitem[Peebles and Xie(2023)]%
        {peebles2023scalable}
\bibfield{author}{\bibinfo{person}{William Peebles} {and} \bibinfo{person}{Saining Xie}.} \bibinfo{year}{2023}\natexlab{}.
\newblock \showarticletitle{Scalable diffusion models with transformers}. In \bibinfo{booktitle}{\emph{Proceedings of the IEEE/CVF International Conference on Computer Vision}}. \bibinfo{pages}{4195--4205}.
\newblock


\bibitem[Podell et~al\mbox{.}(2023)]%
        {podell2023sdxl}
\bibfield{author}{\bibinfo{person}{Dustin Podell}, \bibinfo{person}{Zion English}, \bibinfo{person}{Kyle Lacey}, \bibinfo{person}{Andreas Blattmann}, \bibinfo{person}{Tim Dockhorn}, \bibinfo{person}{Jonas M{\"u}ller}, \bibinfo{person}{Joe Penna}, {and} \bibinfo{person}{Robin Rombach}.} \bibinfo{year}{2023}\natexlab{}.
\newblock \showarticletitle{Sdxl: Improving latent diffusion models for high-resolution image synthesis}.
\newblock \bibinfo{journal}{\emph{arXiv preprint arXiv:2307.01952}} (\bibinfo{year}{2023}).
\newblock


\bibitem[Radford et~al\mbox{.}(2021)]%
        {radford2021learning}
\bibfield{author}{\bibinfo{person}{Alec Radford}, \bibinfo{person}{Jong~Wook Kim}, \bibinfo{person}{Chris Hallacy}, \bibinfo{person}{Aditya Ramesh}, \bibinfo{person}{Gabriel Goh}, \bibinfo{person}{Sandhini Agarwal}, \bibinfo{person}{Girish Sastry}, \bibinfo{person}{Amanda Askell}, \bibinfo{person}{Pamela Mishkin}, \bibinfo{person}{Jack Clark}, {et~al\mbox{.}}} \bibinfo{year}{2021}\natexlab{}.
\newblock \showarticletitle{Learning transferable visual models from natural language supervision}. In \bibinfo{booktitle}{\emph{International conference on machine learning}}. PMLR, \bibinfo{pages}{8748--8763}.
\newblock


\bibitem[Ramesh et~al\mbox{.}(2022)]%
        {ramesh2022hierarchical}
\bibfield{author}{\bibinfo{person}{Aditya Ramesh}, \bibinfo{person}{Prafulla Dhariwal}, \bibinfo{person}{Alex Nichol}, \bibinfo{person}{Casey Chu}, {and} \bibinfo{person}{Mark Chen}.} \bibinfo{year}{2022}\natexlab{}.
\newblock \showarticletitle{Hierarchical text-conditional image generation with clip latents}.
\newblock \bibinfo{journal}{\emph{arXiv preprint arXiv:2204.06125}} \bibinfo{volume}{1}, \bibinfo{number}{2} (\bibinfo{year}{2022}), \bibinfo{pages}{3}.
\newblock


\bibitem[Rombach et~al\mbox{.}(2022)]%
        {rombach2022high}
\bibfield{author}{\bibinfo{person}{Robin Rombach}, \bibinfo{person}{Andreas Blattmann}, \bibinfo{person}{Dominik Lorenz}, \bibinfo{person}{Patrick Esser}, {and} \bibinfo{person}{Bj{\"o}rn Ommer}.} \bibinfo{year}{2022}\natexlab{}.
\newblock \showarticletitle{High-resolution image synthesis with latent diffusion models}. In \bibinfo{booktitle}{\emph{Proceedings of the IEEE/CVF conference on computer vision and pattern recognition}}. \bibinfo{pages}{10684--10695}.
\newblock


\bibitem[Ruiz et~al\mbox{.}(2023)]%
        {ruiz2023dreambooth}
\bibfield{author}{\bibinfo{person}{Nataniel Ruiz}, \bibinfo{person}{Yuanzhen Li}, \bibinfo{person}{Varun Jampani}, \bibinfo{person}{Yael Pritch}, \bibinfo{person}{Michael Rubinstein}, {and} \bibinfo{person}{Kfir Aberman}.} \bibinfo{year}{2023}\natexlab{}.
\newblock \showarticletitle{Dreambooth: Fine tuning text-to-image diffusion models for subject-driven generation}. In \bibinfo{booktitle}{\emph{Proceedings of the IEEE/CVF conference on computer vision and pattern recognition}}. \bibinfo{pages}{22500--22510}.
\newblock


\bibitem[Sharma et~al\mbox{.}(2023)]%
        {Sharma2023materialistic}
\bibfield{author}{\bibinfo{person}{Prafull Sharma}, \bibinfo{person}{Julien Philip}, \bibinfo{person}{Micha\"{e}l Gharbi}, \bibinfo{person}{Bill Freeman}, \bibinfo{person}{Fredo Durand}, {and} \bibinfo{person}{Valentin Deschaintre}.} \bibinfo{year}{2023}\natexlab{}.
\newblock \showarticletitle{Materialistic: Selecting Similar Materials in Images}.
\newblock \bibinfo{journal}{\emph{ACM Trans. Graph.}} \bibinfo{volume}{42}, \bibinfo{number}{4}, Article \bibinfo{articleno}{154} (\bibinfo{date}{jul} \bibinfo{year}{2023}), \bibinfo{numpages}{14}~pages.
\newblock
\showISSN{0730-0301}
\urldef\tempurl%
\url{https://doi.org/10.1145/3592390}
\showDOI{\tempurl}


\bibitem[Shi et~al\mbox{.}(2020a)]%
        {Shi2020}
\bibfield{author}{\bibinfo{person}{Liang Shi}, \bibinfo{person}{Beichen Li}, \bibinfo{person}{Milo{\v s} Ha{\v s}an}, \bibinfo{person}{Kalyan Sunkavalli}, \bibinfo{person}{Tamy Boubekeur}, \bibinfo{person}{Radomir Mech}, {and} \bibinfo{person}{Wojciech Matusik}.} \bibinfo{year}{2020}\natexlab{a}.
\newblock \showarticletitle{MATch: Differentiable Material Graphs for Procedural Material Capture}.
\newblock \bibinfo{journal}{\emph{ACM Trans. Graph.}} \bibinfo{volume}{39}, \bibinfo{number}{6} (\bibinfo{date}{Dec.} \bibinfo{year}{2020}), \bibinfo{pages}{1--15}.
\newblock


\bibitem[Shi et~al\mbox{.}(2020b)]%
        {match}
\bibfield{author}{\bibinfo{person}{Liang Shi}, \bibinfo{person}{Beichen Li}, \bibinfo{person}{Milo{\v s} Ha{\v s}an}, \bibinfo{person}{Kalyan Sunkavalli}, \bibinfo{person}{Tamy Boubekeur}, \bibinfo{person}{Radomir Mech}, {and} \bibinfo{person}{Wojciech Matusik}.} \bibinfo{year}{2020}\natexlab{b}.
\newblock \showarticletitle{MATch: Differentiable Material Graphs for Procedural Material Capture}.
\newblock \bibinfo{journal}{\emph{ACM Trans. Graph.}} \bibinfo{volume}{39}, \bibinfo{number}{6} (\bibinfo{date}{Dec.} \bibinfo{year}{2020}), \bibinfo{pages}{1--15}.
\newblock


\bibitem[Sohl-Dickstein et~al\mbox{.}(2015)]%
        {sohl2015deep}
\bibfield{author}{\bibinfo{person}{Jascha Sohl-Dickstein}, \bibinfo{person}{Eric Weiss}, \bibinfo{person}{Niru Maheswaranathan}, {and} \bibinfo{person}{Surya Ganguli}.} \bibinfo{year}{2015}\natexlab{}.
\newblock \showarticletitle{Deep unsupervised learning using nonequilibrium thermodynamics}. In \bibinfo{booktitle}{\emph{International conference on machine learning}}. PMLR, \bibinfo{pages}{2256--2265}.
\newblock


\bibitem[Song et~al\mbox{.}(2020)]%
        {song2020denoising}
\bibfield{author}{\bibinfo{person}{Jiaming Song}, \bibinfo{person}{Chenlin Meng}, {and} \bibinfo{person}{Stefano Ermon}.} \bibinfo{year}{2020}\natexlab{}.
\newblock \showarticletitle{Denoising Diffusion Implicit Models}.
\newblock \bibinfo{journal}{\emph{arXiv:2010.02502}} (\bibinfo{date}{October} \bibinfo{year}{2020}).
\newblock
\urldef\tempurl%
\url{https://arxiv.org/abs/2010.02502}
\showURL{%
\tempurl}


\bibitem[Song and Ermon(2019)]%
        {song2019generative}
\bibfield{author}{\bibinfo{person}{Yang Song} {and} \bibinfo{person}{Stefano Ermon}.} \bibinfo{year}{2019}\natexlab{}.
\newblock \showarticletitle{Generative modeling by estimating gradients of the data distribution}.
\newblock \bibinfo{journal}{\emph{Advances in neural information processing systems}}  \bibinfo{volume}{32} (\bibinfo{year}{2019}).
\newblock


\bibitem[Song et~al\mbox{.}(2021)]%
        {song2021scorebased}
\bibfield{author}{\bibinfo{person}{Yang Song}, \bibinfo{person}{Jascha Sohl-Dickstein}, \bibinfo{person}{Diederik~P Kingma}, \bibinfo{person}{Abhishek Kumar}, \bibinfo{person}{Stefano Ermon}, {and} \bibinfo{person}{Ben Poole}.} \bibinfo{year}{2021}\natexlab{}.
\newblock \showarticletitle{Score-Based Generative Modeling through Stochastic Differential Equations}. In \bibinfo{booktitle}{\emph{International Conference on Learning Representations}}.
\newblock
\urldef\tempurl%
\url{https://openreview.net/forum?id=PxTIG12RRHS}
\showURL{%
\tempurl}


\bibitem[Vecchio and Deschaintre(2024)]%
        {vecchio2024matsynth}
\bibfield{author}{\bibinfo{person}{Giuseppe Vecchio} {and} \bibinfo{person}{Valentin Deschaintre}.} \bibinfo{year}{2024}\natexlab{}.
\newblock \showarticletitle{MatSynth: A Modern PBR Materials Dataset}. In \bibinfo{booktitle}{\emph{Proceedings of the IEEE/CVF Conference on Computer Vision and Pattern Recognition (CVPR)}}. \bibinfo{pages}{22109--22118}.
\newblock


\bibitem[Vecchio et~al\mbox{.}(2024a)]%
        {vecchio2024controlmat}
\bibfield{author}{\bibinfo{person}{Giuseppe Vecchio}, \bibinfo{person}{Rosalie Martin}, \bibinfo{person}{Arthur Roullier}, \bibinfo{person}{Adrien Kaiser}, \bibinfo{person}{Romain Rouffet}, \bibinfo{person}{Valentin Deschaintre}, {and} \bibinfo{person}{Tamy Boubekeur}.} \bibinfo{year}{2024}\natexlab{a}.
\newblock \showarticletitle{ControlMat: A Controlled Generative Approach to Material Capture}.
\newblock \bibinfo{journal}{\emph{ACM Trans. Graph.}} \bibinfo{volume}{43}, \bibinfo{number}{5}, Article \bibinfo{articleno}{164} (\bibinfo{date}{sep} \bibinfo{year}{2024}), \bibinfo{numpages}{17}~pages.
\newblock
\showISSN{0730-0301}
\urldef\tempurl%
\url{https://doi.org/10.1145/3688830}
\showDOI{\tempurl}


\bibitem[Vecchio et~al\mbox{.}(2024b)]%
        {vecchio2024matfuse}
\bibfield{author}{\bibinfo{person}{Giuseppe Vecchio}, \bibinfo{person}{Renato Sortino}, \bibinfo{person}{Simone Palazzo}, {and} \bibinfo{person}{Concetto Spampinato}.} \bibinfo{year}{2024}\natexlab{b}.
\newblock \showarticletitle{Matfuse: controllable material generation with diffusion models}. In \bibinfo{booktitle}{\emph{Proceedings of the IEEE/CVF Conference on Computer Vision and Pattern Recognition}}. \bibinfo{pages}{4429--4438}.
\newblock


\bibitem[Weijie et~al\mbox{.}(2024)]%
        {kong2024hunyuanvideo}
\bibfield{author}{\bibinfo{person}{Kong Weijie}, \bibinfo{person}{Tian Qi}, \bibinfo{person}{Zhang Zijian}, \bibinfo{person}{Min Rox}, \bibinfo{person}{Dai Zuozhuo}, \bibinfo{person}{Zhou Jin}, \bibinfo{person}{Xiong Jiangfeng}, \bibinfo{person}{Li Xin}, \bibinfo{person}{Wu Bo}, \bibinfo{person}{Zhang Jianwei}, \bibinfo{person}{Wu Kathrina}, \bibinfo{person}{Lin Qin}, \bibinfo{person}{Wang Aladdin}, \bibinfo{person}{Wang Andong}, \bibinfo{person}{Li Changlin}, \bibinfo{person}{Huang Duojun}, \bibinfo{person}{Yang Fang}, \bibinfo{person}{Tan Hao}, \bibinfo{person}{Wang Hongmei}, \bibinfo{person}{Song Jacob}, \bibinfo{person}{Bai Jiawang}, \bibinfo{person}{Wu Jianbing}, \bibinfo{person}{Xue Jinbao}, \bibinfo{person}{Wang Joey}, \bibinfo{person}{Yuan Junkun}, \bibinfo{person}{Wang Kai}, \bibinfo{person}{Liu Mengyang}, \bibinfo{person}{Li Pengyu}, \bibinfo{person}{Li Shuai}, \bibinfo{person}{Wang Weiyan}, \bibinfo{person}{Yu Wenqing}, \bibinfo{person}{Deng Xinchi}, \bibinfo{person}{Li Yang}, \bibinfo{person}{Long
  Yanxin}, \bibinfo{person}{Chen Yi}, \bibinfo{person}{Cui Yutao}, \bibinfo{person}{Peng Yuanbo}, \bibinfo{person}{Yu Zhentao}, \bibinfo{person}{He Zhiyu}, \bibinfo{person}{Xu Zhiyong}, \bibinfo{person}{Zhou Zixiang}, \bibinfo{person}{Xu Zunnan}, \bibinfo{person}{Tao Yangyu}, \bibinfo{person}{Lu Qinglin}, \bibinfo{person}{Liu Songtao}, \bibinfo{person}{Zhou Dax}, \bibinfo{person}{Wang Hongfa}, \bibinfo{person}{Yang Yong}, \bibinfo{person}{Wang Di}, \bibinfo{person}{Liu Yuhong}, \bibinfo{person}{Jiang  , Jie}, {and} \bibinfo{person}{Zhong Caesar}.} \bibinfo{year}{2024}\natexlab{}.
\newblock \bibinfo{title}{HunyuanVideo: A Systematic Framework For Large Video Generative Models}.
\newblock
\newblock
\urldef\tempurl%
\url{https://arxiv.org/abs/2412.03603}
\showURL{%
\tempurl}


\bibitem[Wu et~al\mbox{.}(2020)]%
        {wu2020deep}
\bibfield{author}{\bibinfo{person}{Huisi Wu}, \bibinfo{person}{Wei Yan}, \bibinfo{person}{Ping Li}, {and} \bibinfo{person}{Zhenkun Wen}.} \bibinfo{year}{2020}\natexlab{}.
\newblock \showarticletitle{Deep texture exemplar extraction based on trimmed T-CNN}.
\newblock \bibinfo{journal}{\emph{IEEE Transactions on Multimedia}}  \bibinfo{volume}{23} (\bibinfo{year}{2020}), \bibinfo{pages}{4502--4514}.
\newblock


\bibitem[Xue et~al\mbox{.}(2024)]%
        {xue2024reflectancefusion}
\bibfield{author}{\bibinfo{person}{Bowen Xue}, \bibinfo{person}{Claudio Guarnera}, \bibinfo{person}{Shuang Zhao}, {and} \bibinfo{person}{Zahra Montazeri}.} \bibinfo{year}{2024}\natexlab{}.
\newblock \showarticletitle{ReflectanceFusion: Diffusion-based text to SVBRDF Generation}. In \bibinfo{booktitle}{\emph{Eurographics Symposium on Rendering}}. Eurographics Association.
\newblock


\bibitem[Yan et~al\mbox{.}(2023)]%
        {psdrroom}
\bibfield{author}{\bibinfo{person}{K. Yan}, \bibinfo{person}{F. Luan}, \bibinfo{person}{M. Ha\v{s}an}, \bibinfo{person}{T. Groueix}, \bibinfo{person}{V. Deschaintre}, {and} \bibinfo{person}{S. Zhao}.} \bibinfo{year}{2023}\natexlab{}.
\newblock \showarticletitle{PSDR-Room: Single Photo to Scene using Differentiable Rendering}. In \bibinfo{booktitle}{\emph{ACM SIGGRAPH Asia 2023 Conference Proceedings}}.
\newblock


\bibitem[Yeh et~al\mbox{.}(2022)]%
        {yeh2022photoscene}
\bibfield{author}{\bibinfo{person}{Yu-Ying Yeh}, \bibinfo{person}{Zhengqin Li}, \bibinfo{person}{Yannick Hold-Geoffroy}, \bibinfo{person}{Rui Zhu}, \bibinfo{person}{Zexiang Xu}, \bibinfo{person}{Milo{\v{s}} Ha{\v{s}}an}, \bibinfo{person}{Kalyan Sunkavalli}, {and} \bibinfo{person}{Manmohan Chandraker}.} \bibinfo{year}{2022}\natexlab{}.
\newblock \showarticletitle{Photoscene: Photorealistic material and lighting transfer for indoor scenes}. In \bibinfo{booktitle}{\emph{Proceedings of the IEEE/CVF Conference on Computer Vision and Pattern Recognition}}. \bibinfo{pages}{18562--18571}.
\newblock


\bibitem[Zeng et~al\mbox{.}(2024)]%
        {zeng2024rgb}
\bibfield{author}{\bibinfo{person}{Zheng Zeng}, \bibinfo{person}{Valentin Deschaintre}, \bibinfo{person}{Iliyan Georgiev}, \bibinfo{person}{Yannick Hold-Geoffroy}, \bibinfo{person}{Yiwei Hu}, \bibinfo{person}{Fujun Luan}, \bibinfo{person}{Ling-Qi Yan}, {and} \bibinfo{person}{Milo{\v{s}} Ha{\v{s}}an}.} \bibinfo{year}{2024}\natexlab{}.
\newblock \showarticletitle{Rgb<->x: Image decomposition and synthesis using material-and lighting-aware diffusion models}. In \bibinfo{booktitle}{\emph{ACM SIGGRAPH 2024 Conference Papers}}. \bibinfo{pages}{1--11}.
\newblock


\bibitem[Zhang et~al\mbox{.}(2020)]%
        {Zhang:2020:PSDR}
\bibfield{author}{\bibinfo{person}{Cheng Zhang}, \bibinfo{person}{Bailey Miller}, \bibinfo{person}{Kai Yan}, \bibinfo{person}{Ioannis Gkioulekas}, {and} \bibinfo{person}{Shuang Zhao}.} \bibinfo{year}{2020}\natexlab{}.
\newblock \showarticletitle{Path-Space Differentiable Rendering}.
\newblock \bibinfo{journal}{\emph{ACM Trans. Graph.}} \bibinfo{volume}{39}, \bibinfo{number}{4} (\bibinfo{year}{2020}), \bibinfo{pages}{143:1--143:19}.
\newblock


\bibitem[Zhang et~al\mbox{.}(2023)]%
        {zhang2023adding}
\bibfield{author}{\bibinfo{person}{Lvmin Zhang}, \bibinfo{person}{Anyi Rao}, {and} \bibinfo{person}{Maneesh Agrawala}.} \bibinfo{year}{2023}\natexlab{}.
\newblock \bibinfo{title}{Adding Conditional Control to Text-to-Image Diffusion Models}.
\newblock
\newblock


\bibitem[Zhou et~al\mbox{.}(2022)]%
        {zhou2022tilegen}
\bibfield{author}{\bibinfo{person}{Xilong Zhou}, \bibinfo{person}{Milos Hasan}, \bibinfo{person}{Valentin Deschaintre}, \bibinfo{person}{Paul Guerrero}, \bibinfo{person}{Kalyan Sunkavalli}, {and} \bibinfo{person}{Nima~Khademi Kalantari}.} \bibinfo{year}{2022}\natexlab{}.
\newblock \showarticletitle{TileGen: Tileable, Controllable Material Generation and Capture}. In \bibinfo{booktitle}{\emph{SIGGRAPH Asia 2022 Conference Papers}} (Daegu, Republic of Korea) \emph{(\bibinfo{series}{SA '22})}. \bibinfo{publisher}{Association for Computing Machinery}, \bibinfo{address}{New York, NY, USA}, Article \bibinfo{articleno}{34}, \bibinfo{numpages}{9}~pages.
\newblock
\showISBNx{9781450394703}
\urldef\tempurl%
\url{https://doi.org/10.1145/3550469.3555403}
\showDOI{\tempurl}


\bibitem[Zhou and Kalantari(2022)]%
        {zhou2022look}
\bibfield{author}{\bibinfo{person}{Xilong Zhou} {and} \bibinfo{person}{Nima~Khademi Kalantari}.} \bibinfo{year}{2022}\natexlab{}.
\newblock \showarticletitle{Look-ahead training with learned reflectance loss for single-image svbrdf estimation}.
\newblock \bibinfo{journal}{\emph{ACM Transactions on Graphics (TOG)}} \bibinfo{volume}{41}, \bibinfo{number}{6} (\bibinfo{year}{2022}), \bibinfo{pages}{1--12}.
\newblock


\end{thebibliography}

\end{document}